\documentclass[journal]{IEEEtran}
\usepackage{amsmath,amsfonts}
\usepackage{algorithmic}
\usepackage{algorithm}
\usepackage{array}
\usepackage{textcomp}
\usepackage{stfloats}
\usepackage{url}
\usepackage{verbatim}
\usepackage{graphicx}
\usepackage[nocompress]{cite}
\usepackage[english]{babel}

\addto\captionsenglish{}
\usepackage[caption=false,font=normalsize,labelformat=simple]{subfig}

\usepackage{multirow}
\usepackage{multicol}
\usepackage{booktabs}
\usepackage{amsthm}
\usepackage{amssymb}
\usepackage{makecell}
\usepackage[dvipsnames]{xcolor}
\usepackage{hyperref}

\usepackage{bm}
\usepackage{mathtools}

\usepackage{amsthm}
\usepackage{array}   
\usepackage{amssymb}
\usepackage{arydshln}
\usepackage{bbding,pifont}
\usepackage{tabularx,ragged2e}
\usepackage[edges]{forest}
\usepackage{longtable}
\usepackage{mathrsfs}

\usepackage[framemethod=tikz]{mdframed}
\usetikzlibrary{positioning}
\usepackage[edges]{forest}
\usetikzlibrary{shadows}
\newcolumntype{C}[1]{>{\centering}p{#1}}

\definecolor{vision_color}{HTML}{85B4E8}
\definecolor{lang_color}{HTML}{FFAF48}
\definecolor{action_color}{HTML}{9CCC65}
\definecolor{mauve}{HTML}{E0B0FF}
\definecolor{mediumslateblue}{HTML}{7B68EE}
\definecolor{darkorchid}{HTML}{9932CC}
\definecolor{thistle}{HTML}{D8BFD8}
\definecolor{CustomBlue}{HTML}{1E90FF}
\definecolor{lightgray}{gray}{0.7}

\usepackage{scalerel,graphicx,xparse}

\newcommand{\cmark}{\ding{51}}%
\newcommand{\xmark}{\ding{55}}%
\newcommand{\NA}{\rule[0.5ex]{0.7em}{0.5pt}}

\begin{document}

\title{A Survey on Vision-Language-Action Models for Embodied AI}

\author{Yueen~Ma, 
     Zixing~Song, 
     Yuzheng~Zhuang, 
     Jianye~Hao, 
     Irwin~King,~\IEEEmembership{Fellow,~IEEE}
\thanks{This article has been published in IEEE Transactions on Neural Networks and Learning Systems. This is the author's version and may not fully reflect the final typeset publication. The final published version can be found via Digital Object Identifier (DOI): \href{https://doi.org/10.1109/TNNLS.2025.3650584}{10.1109/TNNLS.2025.3650584}}
\thanks{Yueen Ma and Irwin King are with the Chinese University of Hong Kong, Hong Kong, China (e-mail: \texttt{yema21@cse.cuhk.edu.hk}; \texttt{king@cse.cuhk.edu.hk}; corresponding author: Irwin King).
}
\thanks{Zixing Song is with the University of Bristol, Bristol BS8 1QU, U.K. (e-mail: \texttt{zixing.song@bristol.ac.uk}).
}
\thanks{Yuzheng Zhuang and Jianye Hao are with Huawei Noah's Ark Laboratory, Shenzhen 518000, China (e-mail: \texttt{zhuangyuzheng@huawei.com}; \texttt{haojianye@huawei.com}).
}
}

\markboth{IEEE Transactions on Neural Networks and Learning Systems}%
{Ma \MakeLowercase{\textit{et al.}}: Survey on Vision-Language-Action Models for Embodied AI}

\IEEEpubid{
\makebox[\textwidth][c]{
    \parbox{0.8\textwidth}{
        \vspace{20pt}
        \centering
        2162-237X \copyright~2025 IEEE. All rights reserved, including rights for text and data mining, and training of artificial intelligence and similar technologies. Personal use is permitted, but republication/redistribution requires IEEE permission.\\
        See \url{https://www.ieee.org/publications/rights/index.html} for more information.
    }
}
}

\maketitle
\begin{abstract}
Embodied AI is widely recognized as a cornerstone of artificial general intelligence (AGI) because it involves controlling embodied agents to perform tasks in the physical world. Building on the success of large language models (LLMs) and vision-language models (VLMs), a new category of multimodal models---referred to as vision-language-action (VLA) models---has emerged to address language-conditioned robotic tasks in embodied AI by leveraging their distinct ability to generate actions. The recent proliferation of VLAs necessitates a comprehensive survey to capture the rapidly evolving landscape. To this end, we present the first survey on VLAs for embodied AI. This work provides a detailed taxonomy of VLAs, organized into three major lines of research. The first line focuses on individual components of VLAs. The second line is dedicated to developing VLA-based control policies adept at predicting low-level actions. The third line comprises high-level task planners capable of decomposing long-horizon tasks into a sequence of subtasks, thereby guiding VLAs to follow more general user instructions. Furthermore, we provide an extensive summary of relevant resources, including datasets, simulators, and benchmarks. Finally, we discuss the challenges facing VLAs and outline promising future directions in embodied AI. A curated repository associated with this survey is available at: \url{https://github.com/yueen-ma/Awesome-VLA}
\end{abstract}


\section{Introduction}
\label{sec:intro}

Vision-language-action (VLA) models are a class of multimodal models within the field of embodied AI, designed to process information from three modalities: vision, language, and action. Unlike conversational AI, embodied AI requires controlling physical embodiments that interact with the environment, and robotics is the most prominent domain of embodied AI. In language-conditioned robotic tasks, the policy must possess the capability to understand language instructions, visually perceive the environment, and generate appropriate actions, necessitating the multimodal capabilities of VLAs. The term was recently coined by RT-2~\cite{DBLP:journals/corr/abs-2307-15818}. Compared with earlier deep reinforcement learning (RL) approaches, VLAs offer superior versatility, dexterity, and generalizability in complex environments. As a result, they are well-suited not only for controlled settings like factories but also for everyday tasks in household environments.

Early developments in deep learning primarily consisted of unimodal models. In computer vision (CV), models like AlexNet~\cite{DBLP:conf/nips/KrizhevskySH12} showcased the potential of artificial neural networks (ANNs). Recurrent neural networks laid the groundwork for numerous natural language processing (NLP) models, but NLP has seen a transition in recent years, with Transformers~\cite{DBLP:conf/nips/VaswaniSPUJGKP17} taking precedence. The deep Q-network (DQN)~\cite{DBLP:journals/nature/MnihKSRVBGRFOPB15} demonstrated that ANNs can successfully tackle RL problems. Leveraging advancements of unimodal models across diverse machine learning fields, multimodal models now address a wide range of tasks, such as visual question answering (VQA), image captioning, and text-to-video generation.

Conventional robot policies based on RL have largely focused on addressing a limited set of tasks within controlled environments, such as item grasping~\cite{DBLP:conf/iser/LevinePKQ16}. However, there is a growing need for more versatile multitask policies, akin to recent trends in large language models (LLMs) and vision-language models (VLMs). Developing a multitask policy is challenging, as it requires learning a broader set of skills and adapting to diverse environments. Furthermore, specifying tasks via language instructions offers a more intuitive user-robot interface, which necessitates the development of language-conditioned robot policies.

\begin{figure}[t]
    \centering
     \includegraphics[width=1.0\columnwidth, trim={0.37in 7.81in 5.45in 1.87in},clip]{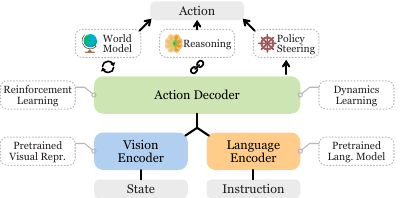}
    \caption{General architecture of VLA models. Three representative methods for action prediction are shown. Related components are presented in dashed boxes.
    }
    \label{fig:arch}
\end{figure}

\IEEEpubidadjcol

Built upon the success of large VLMs, VLA models have demonstrated their potential in addressing these challenges, as illustrated in Fig.~\ref{fig:arch}. Similar to VLMs, VLAs utilize vision foundation models as vision encoders to obtain pretrained visual representations (PVRs) of the current environmental state, such as object class, pose, and geometry. VLAs encode instructions using the token embeddings of their LLMs and employ various strategies to align vision and language embeddings, including approaches like Flamingo~\cite{DBLP:conf/nips/AlayracDLMBHLMM22}, BLIP-2~\cite{DBLP:conf/icml/0008LSH23}, and LLaVA~\cite{DBLP:journals/corr/abs-2304-08485}. By finetuning on robot data, the LLM can function as a decoder to predict actions and perform language-conditioned robotic tasks. These cross-disciplinary innovations drive the advancement of VLAs in embodied AI---a critical building block for AGI.

\begin{figure*}
    \centering
	\subfloat[Concepts \label{fig:venn}]{
	     \centering
            \includegraphics[width=0.16\textwidth, trim={0.5in 8.5in 6.5in 0.68in},clip]{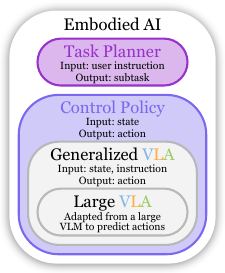}
        }
        \hfill 
	{\color{gray} \vrule width 1.0pt height 0.14\textheight} 
        \hfill 
        \subfloat[Timelines \label{fig:timeline_simple}]{
	     \centering
            \includegraphics[width=0.80\textwidth, trim={0.74in 8.745in 1.03in 0.74in},clip]{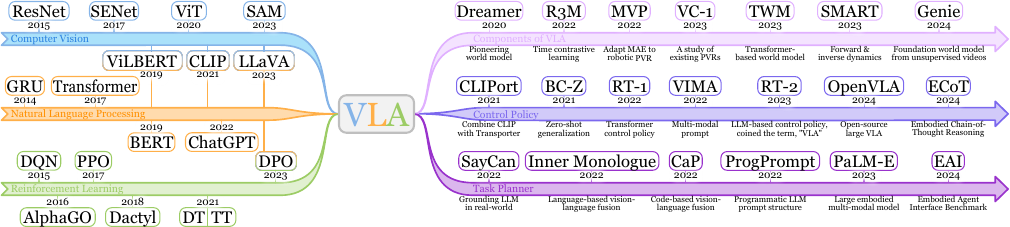}
	}
    \caption{(a) Venn diagram that outlines the main concepts in embodied AI discussed in this article. (b) Timelines that trace the evolution from unimodal models to VLA models.}
    \label{fig:intro}
\end{figure*}

VLAs are closely related to three lines of work, as depicted by the timelines in Fig.~\ref{fig:timeline_simple} and the taxonomy in Fig.~\ref{fig:taxonomy}. Some approaches focus on individual components of VLAs (see Section~\ref{sec:components}), such as PVRs, dynamics learning, world models, and reasoning. Meanwhile, a substantial body of research is dedicated to low-level control policies (see Section~\ref{sec:policy}). In this category, language instructions and visual observations are fed into the control policy, which then generates low-level actions---such as translation and rotation---thereby rendering VLAs an ideal choice for control policies. In contrast, another category of models serves as high-level task planners responsible for task decomposition (see Section~\ref{sec:planner}). These models break down long-horizon tasks into a sequence of subtasks that, in turn, guide VLAs toward the overall goal, as illustrated in Fig.~\ref{fig:illustration}. Most current robotic systems adopt such a hierarchical framework~\cite{DBLP:conf/corl/HuangXXCLFZTMCS22, DBLP:conf/corl/IchterBCFHHHIIJ22, DBLP:conf/icml/DriessXSLCIWTVY23} because the high-level task planner can leverage models with high capacity while the low-level control policy can focus on speed and precision, similar to hierarchical RL.

To provide a more comprehensive overview of current progress in embodied AI, we propose a generalized definition of ``VLA,'' as illustrated by the Venn diagram in Fig.~\ref{fig:venn}. We define a VLA as any model capable of processing multimodal inputs from vision and language to produce robot actions that accomplish embodied tasks, typically following the architecture in Fig.~\ref{fig:arch}. The original concept of VLA referred to a model that adapts VLMs to robotic tasks~\cite{DBLP:journals/corr/abs-2307-15818}. Analogous to the distinction between LLMs and more general language models, we designate the original VLAs as ``Large VLAs'' (LVLAs) because they are based on LLMs or large VLMs.

\subsection{Related Work} 
To the best of our knowledge, this survey is the first to review the recent progress of VLA models, a rapidly emerging research area. Previous surveys have investigated other facets of embodied AI. Firoozi et al.~\cite{DBLP:journals/corr/abs-2312-07843} comprehensively summarized foundation models in robotics up to 2023, while Wang et al.~\cite{DBLP:journals/corr/abs-2401-04334} focused on LLMs in robotics. Hu et al.~\cite{DBLP:journals/corr/abs-2312-08782} examined more recent vision, language, and robotic foundation models for general-purpose robots. Kawaharazuka et al.~\cite{DBLP:journals/ar/KawaharazukaMGGPZ24} concentrated on real-world robot applications. In contrast, our work emphasizes VLA models, thereby complementing and extending the existing literature on embodied AI.

\subsection{Contributions} 
To the best of our knowledge, this article is the first comprehensive survey of emerging VLA models in the field of embodied AI.
\begin{enumerate}
    \item \textit{Comprehensive Review:} We present a thorough review of emerging VLA models in embodied AI, covering various aspects including components, architectures, training objectives, and robotic tasks.
    \item \textit{Taxonomy:} We introduce a taxonomy of VLAs based on the hierarchical framework of current robotic systems, comprising two hierarchies: a low-level control policy and a high-level task planner. Control policies execute low-level actions based on specified language commands and the perceived environment. Task planners provide guidance to control policies by decomposing long-horizon tasks into subtasks. We also discuss various essential components of VLAs.
    \item \textit{Resources:} We summarize the necessary resources for training and evaluating VLAs, including recent datasets and benchmarks in real-world or simulated environments. We also discuss various approaches to address current challenges, such as data scarcity and inconsistency.
    \item \textit{Future Directions:} We outline current challenges and promising future opportunities in the field, such as safety, foundation models, and real-world deployment.
\end{enumerate}
\section{Background}

\begin{figure*}
    \centering
        \scalebox{0.75}{
          \tikzset{%
    parent/.style =          {align=center,text width=2cm,rounded corners=3pt, line width=0.5mm, fill=gray!10,draw=gray!80},
    component/.style =           {align=center,text width=5cm,rounded corners=3pt, fill=mauve!25,draw=mauve!100,line width=0.5mm},
    component_sub/.style =           {align=center,text width=4.5cm,rounded corners=3pt, fill=mauve!25,draw=none, font=\small},
    component_sub2/.style =           {align=center,text width=4cm,rounded corners=3pt, fill=mauve!25,draw=none, font=\small},
    policy/.style =           {align=center,text width=5.5cm,rounded corners=3pt, fill=mediumslateblue!25,draw=mediumslateblue!100,line width=0.5mm},
    policy_sub/.style =           {align=center,text width=5cm,rounded corners=3pt, fill=mediumslateblue!25,draw=none, font=\small},
    policy_sub2/.style =           {align=center,text width=4.5cm,rounded corners=3pt, fill=mediumslateblue!25,draw=none, font=\small},
    planner/.style =           {align=center,text width=4.5cm,rounded corners=3pt, fill=darkorchid!25,draw=darkorchid!100,line width=0.5mm},
    planner_type/.style =           {align=center,text width=4cm,rounded corners=3pt, fill=darkorchid!25,draw=none, font=\small},    
    planner_sub/.style =           {align=center,text width=3.5cm,rounded corners=3pt, fill=darkorchid!25,draw=none, font=\small},
    data/.style =           {align=center,text width=4.5cm,rounded corners=3pt, fill=thistle!25,draw=thistle!100,line width=0.5mm},
    data_sub/.style =           {align=center,text width=4cm,rounded corners=3pt, fill=thistle!25,draw=none, font=\small},    
    data_sub2/.style =           {align=center,text width=3.5cm,rounded corners=3pt, fill=thistle!25,draw=none, font=\small},    
}

	\forestset{%
	  rect/.append style={rectangle, rounded corners=2pt},
	  dir tree switch/.style args={at #1}{%
	    for tree={
	      fit=rectangle,
	    },
          where level=0{
            calign=child,
            calign child=3.2,
            xshift=-1.2cm,
          }{},
	    where level=#1{
	      for tree={
	        folder,
	        grow'=0,
	      },
	      delay={child anchor=north},
	    }{},
	    before typesetting nodes={
	      for tree={
	        content/.wrap value={\strut ##1},
	      },
	      if={isodd(n_children("!r"))}{
	        for nodewalk/.wrap pgfmath arg={{fake=r,n=##1}{calign with current edge}}{int((n_children("!r")+1)/2)},
	      }{},
	    },
	  },
	}
		\begin{forest}
            dir tree switch=at 1,
            for tree=
		  {
		    if level=1{align=center,
		            edge path={
		         \noexpand\path [draw, \forestoption{edge}] (!u.parent anchor) -- +(0,-20pt) -| (.child anchor)\forestoption{edge label};
		      },
		      }{
		        align={@{}C{25mm}@{}
		        },
		    },
		    rect,
		    draw,
		    l sep=0.5cm,
		    s sep=3pt,
		    align=center,
		    edge+={thick, rounded corners=5pt},
		    rounded corners=2pt,
		  },
		  for level=0{s sep=25pt},
	            [\textcolor{vision_color}{\Large V}\textcolor{lang_color}{\Large L}\textcolor{action_color}{\Large A} \textcolor{CustomBlue}{\S\ref{sec:vla}}, fill=gray!45, parent
	                [Components of VLA \textcolor{CustomBlue}{\S\ref{sec:components}}, for tree={component}
	                    [Reinforcement Learning \textcolor{CustomBlue}{\S\ref{sec:rl_tfm}},  component_sub]
	                    [Pretrained Visual Repr. \textcolor{CustomBlue}{\S\ref{sec:pvr}},  component_sub
		                    [Video Repr. \textcolor{CustomBlue}{\S\ref{sec:video}}, component_sub2]
                            ]
	                    [Dynamics Learning \textcolor{CustomBlue}{\S\ref{sec:dynamics}}, component_sub]
	                    [World Models \textcolor{CustomBlue}{\S\ref{sec:world-model}}, component_sub
		                    [LLM-Induced \textcolor{CustomBlue}{\S\ref{sec:llm-world-model}}, component_sub2]
		                    [Visual \textcolor{CustomBlue}{\S\ref{sec:visual-world-model}}, component_sub2]
		             	]
	                    [Reasoning \textcolor{CustomBlue}{\S\ref{sec:reasoning}}, component_sub]
	                    [Policy Steering \textcolor{CustomBlue}{\S\ref{sec:steering}}, component_sub]
	                ]
	                [Control Policies \textcolor{CustomBlue}{\S\ref{sec:policy}}, for tree={policy}
	                    [Non-Transformer \textcolor{CustomBlue}{\S\ref{sec:non-tfm}}, policy_sub]
	                    [Transformer-Based \textcolor{CustomBlue}{\S\ref{sec:tfm-based}}, policy_sub
		                    [Multimodal Instructions \textcolor{CustomBlue}{\S\ref{sec:multi-modal-instruction}},  policy_sub2]
		                    [3D Vision \textcolor{CustomBlue}{\S\ref{sec:3d_vision}}, policy_sub2, name=3d
		                        [3D Vision + Diffusion \textcolor{CustomBlue}{\S\ref{sec:3d_diffusion_policy}}, policy_sub, text width=4cm, name=3d_diff]
		                    ]
		                    [Diffusion \textcolor{CustomBlue}{\S\ref{sec:diffusion_policy}},  policy_sub2, name=diff]
		                    [Motion Planning \textcolor{CustomBlue}{\S\ref{sec:motion_planner}},  policy_sub2]
    	                    [Point-Based Actions \textcolor{CustomBlue}{\S\ref{sec:vlm-based}}, policy_sub2]
    	                    [Large VLA \textcolor{CustomBlue}{\S\ref{sec:lvla}}, policy_sub2]
                        ]
	                ]
	                [Task Planners \textcolor{CustomBlue}{\S\ref{sec:planner}}, for tree={planner, s sep=9.2pt}
                            [Monolithic \textcolor{CustomBlue}{\S\ref{sec:monolithic}}, planner_type
                                [End-to-End \textcolor{CustomBlue}{\S\ref{sec:end2end}}, planner_sub
                                    [3D Vision \textcolor{CustomBlue}{\S\ref{sec:3d_planner}}, planner_sub, text width=3cm]
                                ]
                                [Grounded \textcolor{CustomBlue}{\S\ref{sec:grounded}}, planner_sub]
                            ]
                            [Modular \textcolor{CustomBlue}{\S\ref{sec:modular}}, planner_type
                                [Language-Based \textcolor{CustomBlue}{\S\ref{sec:lang-based}}, planner_sub]
                                [Code-Based \textcolor{CustomBlue}{\S\ref{sec:code-based}}, planner_sub]
                            ]
                        ]
	                [Datasets \& Benchmarks \textcolor{CustomBlue}{\S\ref{sec:data}}, for tree={data, s sep=8.8pt}
                            [Real-World \textcolor{CustomBlue}{\S\ref{sec:real_datasets}}, data_sub]
                            [Simulation \textcolor{CustomBlue}{\S\ref{sec:simulators}}, data_sub
                                [Simulators, data_sub2]
                            ]
                            [Auto Data Collection \textcolor{CustomBlue}{\S\ref{sec:auto_data}}, data_sub]
                            [Human Data \textcolor{CustomBlue}{\S\ref{sec:human_datasets}}, data_sub]
                            [Task Planning Bench. \textcolor{CustomBlue}{\S\ref{sec:plan_benchmark}}, data_sub]
                            [Embodied QA \textcolor{CustomBlue}{\S\ref{sec:eqa}}, data_sub]
                        ]
	            ]
				\draw[thick, rounded corners=5pt]($(diff.north) - (63.5pt,0)$) -- ($(diff.north) - (63.5pt,-12pt)$) -- (3d_diff.west);
		\end{forest}


        }
    \caption{Taxonomy of VLA models. The organization of this survey follows this taxonomy.}
    \label{fig:taxonomy}
\end{figure*}

Embodied AI is a unique form of artificial intelligence that actively interacts with the physical environment. This sets it apart from other AI models, such as conversational AI, which primarily handles textual conversations, or generative AI models that focus on tasks like text-to-video generation. Embodied AI encompasses a broad spectrum of embodiments, including smart appliances, smart glasses, autonomous vehicles, and more. Among them all, robots stand out as one of the most prominent embodiments. 

Robot learning is also usually framed as an RL problem, represented as a Markov decision process (MDP) consisting of states ($s$), actions ($a$), and rewards ($r$). An MDP trajectory is denoted by $\tau = (s_{1}, a_{1}, r_{1}, \dots, s_{T}, a_{T}, r_{T})$. In certain scenarios, robotic tasks may also be viewed as partially observable Markov decision processes (POMDPs) due to incomplete observations. The primary objective of RL is to train a policy capable of generating optimal actions for the current state $\pi(a_t | s_t)$. Various methods, such as TD learning and policy gradients, can be employed to achieve this. However, in cases where defining the reward function proves challenging, imitation learning is utilized to directly model the action distribution within trajectories devoid of rewards $\tau = (s_{1}, a_{1}, \dots, s_{T}, a_{T})$. Furthermore, many multitask robot models employ language as instructions $p$ to determine which task or skill to execute, leading to the development of language-conditioned robot policies $\pi(a_t | p, s_{\leq t}, a_{< t})$.
\section{Vision-Language-Action Models}
\label{sec:vla}

\subsection{Components of VLA}
\label{sec:components}

A body of work focuses on individual components of VLA models, drawing on successes in CV, NLP, and RL. We introduce them through the lens of embodied AI.

\subsubsection{Reinforcement Learning}
\label{sec:rl_tfm}

RL laid the foundation for embodied AI and continues to help advance VLAs. DQN first demonstrated learning policies directly from high-dimensional pixel inputs, underscoring the need for greater model capacity in end-to-end RL. RL trajectories---sequences of states, actions, and rewards---naturally align with sequence modeling problems, making them well-suited to Transformer architectures. Pioneering efforts in this direction include Decision Transformer (DT)~\cite{DBLP:conf/nips/ChenLRLGLASM21} and Trajectory Transformer (TT)~\cite{DBLP:conf/nips/JannerLL21}. Gato~\cite{DBLP:journals/tmlr/ReedZPCNBGSKSEBREHCHVBF22} further extended this paradigm to a multimodal, multitask, multiembodiment setting. $\pi_{0.6}^{*}$~\cite{DBLP:journals/corr/abs-2511-14759} employs RL for VLAs to learn from experience.

Furthermore, a synergy has emerged between RL and LLMs, benefiting embodied AI in multiple ways. RL from human feedback (RLHF) aligns LLMs with human preferences and has also been applied to robot learning. SEED~\cite{DBLP:conf/iros/HiranakaHLW00Z23} utilizes RLHF alongside skill-based RL to address the sparse reward issue and to improve safety via human evaluation. Conversely, LLMs also enable novel RL methods. Reflexion~\cite{DBLP:conf/nips/ShinnCGNY23} proposes a verbal RL framework that replaces weight updates in RL models with linguistic feedback, making it naturally applicable to embodied decision-making where planning also occurs in language form. Eureka~\cite{DBLP:conf/iclr/MaLWHBJZFA24} shows that LLMs can design reward functions for embodied agents that outperform expert human-engineered ones. As LLMs evolve, this synergy accelerates progress in RL and embodied AI.

\subsubsection{Pretrained Visual Representations}
\label{sec:pvr}

The effectiveness of the vision encoder directly influences the performance of VLAs, as it provides crucial information regarding the current state $s_t$, such as object categories, positions, and affordances. Consequently, numerous methods are devoted to improving the quality of PVRs. Their technical details are compared in Table~\ref{tb:pvr}.

CLIP~\cite{DBLP:conf/icml/RadfordKHRGASAM21} has gained widespread adoption as a vision encoder in robotic models. The training objective of CLIP is to identify the correct text-image pair among all possible combinations in a given batch. CLIP is pretrained on the WIT dataset, comprising 400 million image-text pairs. This large-scale training allows CLIP to develop a rich understanding of the relationship between visual and textual information. 

R3M~\cite{DBLP:conf/corl/NairRKF022} proposes two main pretraining objectives: time contrastive learning (CL) and video-language alignment. The objective of time contrastive learning is to minimize the distance between temporally proximal video frames while simultaneously increasing the separation between temporally distant ones. This objective aims to create PVRs that capture the temporal relationships within the video sequence. On the other hand, video-language alignment aims to learn whether a video corresponds to a language instruction. This objective enriches the semantic relevance embedded in the PVRs. VIP~\cite{DBLP:conf/iclr/MaSJBK023} also capitalizes on video temporal relationships.

MVP~\cite{DBLP:conf/corl/RadosavovicXJAM22} applies a masked autoencoder (MAE) from computer vision to robotic datasets. The MAE involves masking out a portion of input patches to a ViT model and training it to reconstruct the corrupted patches. This approach closely resembles the masked language modeling technique used in BERT~\cite{DBLP:conf/naacl/DevlinCLT19} and falls within the purview of self-supervised training. RPT~\cite{DBLP:conf/corl/RadosavovicSFGD23} is pretrained with a focus not only on reconstructing visual inputs but also on robotic actions and proprioceptive states.

\begin{table*}
    \centering
    \caption{PVRs. V: Vision. L: Language. CL: contrastive learning. \textbf{Sim}/\textbf{Real}: simulated/real-world. \textbf{Mani}/\textbf{Navi}: manipulation/navigation. For simplicity, we only show the main part of the objective equation, omitting elements such as temperature, auxiliary loss. $\mathcal{S}(\cdot)$ denotes a similarity measure
    }
    \resizebox{\textwidth}{!}{%
    \begin{tabular}{>{\RaggedRight}p{1.5cm} 
                    >{\RaggedRight}p{1.5cm} 
                    >{\RaggedRight}p{1.2cm} 
                    >{\RaggedRight}p{5.3cm} 
                    >{\RaggedRight}p{5cm} 
                    >{\RaggedRight}p{5cm} 
                    @{} 
                    }
        \toprule
        \textbf{Model} & \textbf{Network} & \textbf{Type} & \textbf{Objective} & \textbf{Notations} & \textbf{Robotic Tasks} \\
        \midrule

        CLIP \cite{DBLP:conf/icml/RadfordKHRGASAM21} &
            ViT-B &
            VL-CL &
            $\sum_{i=1}^N -\log \frac{\exp \left(\mathcal{S}(x_i, y_i) \right)}{\sum_{j=1}^N \exp \left(\mathcal{S}(x_i, y_j) \right)} $ &
            $(x_i, y_i)$: image-text pair & 
            (Used by CLIPort \cite{DBLP:conf/corl/ShridharMF21}, EmbCLIP \cite{DBLP:conf/cvpr/0001WMK22}, and CoW \cite{DBLP:conf/cvpr/GadreWISS23})
            \\ \hline

        R3M \cite{DBLP:conf/corl/NairRKF022} & 
            ResNet-50 &
            Time-CL & 
            $\sum_{i,j,k} -\log \frac{\exp \left(\mathcal{S}(x_i, x_j) \right)}{\exp \left(\mathcal{S}(x_i, x_{j}) \right) + \exp \left(\mathcal{S}(x_i, x_{k}) \right)}$ &
            $x_i$: video frame at time $i$; $i<j<k$ & 
            \textbf{Sim-Mani}: Meta-World, Franka Kitchen, Adroit 
            \\ \hline

        MVP \cite{DBLP:conf/corl/RadosavovicXJAM22} &
            ViT-B/L  &
            MAE &
            $\sum_{i \in \mathcal{M}} \operatorname{MSE}\left(x_i, f_{\text{mae}}(x_{\notin \mathcal{M}}) \right)$ &
            $x_i$: image patch; $\mathcal{M}$: masked set &
            \textbf{Real-Mani} (xArm7): pick, reach block, push cube, close fridge
            \\ \hline

        VIP \cite{DBLP:conf/iclr/MaSJBK023} &
            ResNet-50  &
            Time-CL & 
            $\sum_{\tau} -\log \frac{\exp\left(\mathcal{S}(x_0, x_T)\right)}{\sum_{i, j} \exp\left(\mathcal{S}(x_i, x_T) - \gamma \mathcal{S}(x_j, x_T) \right)}$ &
            $x_i$: video frame at time $i$; $0 \le i < j < T$; $\gamma$: discount factor; $\tau$: trajectory &
            \textbf{Real-Mani} (Franka): pick and place, close drawer, push bottle, fold towel
            \\ \hline

        VC-1 \cite{DBLP:conf/nips/MajumdarYAMCSJB23} &
            ViT-L  &
            MAE, CL & 
            (A combination of previous works, including CLIP, R3M, MVP, and VIP.) &
            \NA &
            \textbf{Sim-Mani}: Meta-World, Adroit, DMC, TriFinger, Habitat 2.0; \textbf{Sim-Navi}: Habitat
            \\ \hline

        Voltron \cite{DBLP:conf/rss/KaramchetiNCKFS23} & 
            ViT-S/B &
            MAE, Lang-Gen &
            $\sum_{i \in \mathcal{M}} \operatorname{MSE}\left(x_i, f_{\text{mae}}(x_{\notin \mathcal{M}}, y) \right)$; \newline
            $\sum_{i=1}^{N} -\log P\left(y_i \mid x, y_{<i} \right)$ &
            $x_i$: image patch; $\mathcal{M}$: masked set; \newline $y$: language instruction &
            \textbf{Sim-Mani}: Franka Kitchen; \newline
            \textbf{Real-Mani} (Franka): custom study desk
            \\ \hline

        RPT \cite{DBLP:conf/corl/RadosavovicSFGD23} &
            ViT & 
            MAE & 
            $\sum_{i \in \mathcal{M}} \operatorname{MSE}\left(x_i, f_{\text{mae}}(x_{\notin \mathcal{M}}, y, z, \dots) \right)$ &
            $x, y, z$: three distinct modalities; $\mathcal{M}$: masked set &
            \textbf{Real-Mani} (Franka): stack, pick, pick from bin 
            \\ \hline

        DINOv2 \cite{DBLP:journals/tmlr/OquabDMVSKFHMEA24} & 
            ViT &
            Self-distillation &
            $\sum_{x} \sum_{x'\neq x} H(P_t(x), P_s(x'))$ &
            $x, x'$: image views; $H()$: cross-entropy; $P_t, P_s$: teacher, student &
            (Used by OpenVLA \cite{DBLP:journals/corr/abs-2406-09246}, ReKep \cite{DBLP:journals/corr/abs-2409-01652})
            \\ \hline

        I-JEPA \cite{DBLP:conf/cvpr/AssranDMBVRLB23} & 
            ViT &
            JEPA &
            $\sum_{i \in \mathcal{M}} \operatorname{MSE}\left(f'(x_i), g(f(x_{\notin \mathcal{M}})) \right)$ &
            $x_i$: image block; $\mathcal{M}$: target set; $f, f'$: encoder, EMA encoder; $g$: predictor &
            \NA 
            \\ \hline

        Theia \cite{DBLP:conf/corl/ShangSMMKWH24} & 
            ViT-T/S/B &
            Distillation &
            (Distillation of vision foundation models: ViT, CLIP, SAM, DINOv2, Depth-Anything.) &
            \NA &
            \textbf{Sim-Mani\&Navi}: CortexBench (VC-1); \textbf{Real-Mani}: pick, place, open door/drawer
        \\ \bottomrule
    \end{tabular}
    }
    \label{tb:pvr}
\end{table*}

Voltron~\cite{DBLP:conf/rss/KaramchetiNCKFS23} introduces a pretraining objective by incorporating language conditioning and language generation into the MAE objective. Employing an encoder-decoder Transformer structure, the model alternates between language-conditioned masked image reconstruction and language generation from masked images. This enhances the alignment between language and vision modalities.

VC-1~\cite{DBLP:conf/nips/MajumdarYAMCSJB23} conducts an in-depth examination of prior PVRs and introduces an enhanced PVR model by systematically exploring optimal ViT configurations across diverse datasets. In addition, they perform a comprehensive comparative analysis of their model against previous methods on various manipulation and navigation datasets, shedding light on the critical factors that contribute to the improvement of PVRs. Another study~\cite{DBLP:conf/icml/ParisiRP022} also compares previous PVRs obtained under supervised learning or self-supervised learning.

DINOv2~\cite{DBLP:journals/tmlr/OquabDMVSKFHMEA24} proposes a new self-supervised training paradigm for PVRs that achieves performance beyond that of MAE. It employs a self-distillation framework in which the teacher and student networks receive different views of the same image and match their encoded representations. The student network is updated via gradient descent, while the teacher network is maintained as the exponential moving average of the student network. 

I-JEPA~\cite{DBLP:conf/cvpr/AssranDMBVRLB23} is motivated by the joint-embedding predictive architectures proposed by LeCun~\cite{LeCun2022APT}. It constructs a ``primitive'' internal world model by comparing the embeddings of patches. Unlike DINO, which uses cropped images, I-JEPA employs masked patches. Moreover, it differs from MAE because it is a nongenerative approach.

Theia~\cite{DBLP:conf/corl/ShangSMMKWH24} proposes distilling various vision foundation models into a single model. By fusing diverse visual information, such as segmentation, depth, and semantics, it outperforms previous PVRs while requiring less data and a smaller model size.

\subsubsection{Video Representations}
\label{sec:video}

Videos are simple sequences of images and can be represented by concatenating the usual PVRs of each frame. However, their multiview nature enables a variety of unique representation techniques beyond those mentioned above, such as time CL and MAE. NeRF can be extracted from videos and contains rich 3D information for robot learning, as exemplified by F3RM~\cite{DBLP:conf/corl/ShenYYWKI23}, and 3D-LLM~\cite{DBLP:conf/nips/HongZCZDCG23}. The recent 3D Gaussian Splatting (3D-GS)~\cite{DBLP:journals/tog/KerblKLD23, DBLP:conf/cvpr/Qin0ZWP24} surpasses NeRF in visual quality and rendering speed. In addition, many videos contain audio, which can provide important cues for robot policies~\cite{DBLP:conf/corl/ThankarajP23}.


\subsubsection{Dynamics Learning}
\label{sec:dynamics}

Dynamics learning encompasses objectives aimed at endowing the model $f(\cdot)$ with an understanding of forward or inverse dynamics. Forward dynamics involves predicting the subsequent state resulting from a given action, whereas inverse dynamics entails determining the action required to transition from a previous state to a known subsequent state:
\begin{equation}
\begin{aligned}
\text{Forward dynamics: } & \hat{s}_{t+1} \leftarrow f_{\text{fwd}}(s_t, a_{t})\\
\text{Inverse dynamics: } & \hat{a}_{t} \leftarrow f_{\text{inv}}(s_t, s_{t+1}).
\end{aligned}
\end{equation}
Some approaches also frame these objectives as reordering problems for shuffled state sequences. Some forward dynamics methods closely resemble the image or video prediction pretraining used in PVRs. We compare them in Table~\ref{tb:dynamics}.

\begin{table*}
    \centering
    \caption{Dynamics learning methods for VLAs. $f(\cdot)$ is the dynamics model. Fwd, inv: forward \& inverse dynamics}
    \resizebox{\textwidth}{!}{%
    \begin{tabular}{>{\RaggedRight}p{2cm} 
                    >{\RaggedRight}p{3cm}
                    >{\RaggedRight}p{2.5cm}
                    >{\RaggedRight}p{4.5cm}
                    >{\RaggedRight}p{3cm}
                    >{\RaggedRight}p{4.5cm}
                    @{} 
                    }
        \toprule
        \textbf{Model} & \textbf{Vision Encoder} & \textbf{Type} & \textbf{Objective} & \textbf{Notations} & \textbf{Robotic Tasks} \\
        \midrule

        Vi-PRoM \cite{DBLP:conf/iros/JingZLSYFK23} &
            ResNet &
            Temporal dynamics & 
            $\sum_{t=1}^T \operatorname{CE}(t, f(s'_t))$ &
            $s'_t$: frame in shuffled seq.; $f(\cdot)$ predicts original frame index &
            \textbf{Sim-Mani}: Meta-World, Franka Kitchen; \textbf{Real-Mani}: open \& close drawer/door
            \\ \hline
        MaskDP \cite{DBLP:conf/nips/Liu0GA22} &
            ViT (MAE) & 
            MAE (implicit fwd \& inv dynamics) &
            $\sum_{t=1}^T \operatorname{MSE}\left(x_t, f_{\text{mask}}(\tau') \right)$ &
            $x_t \in \{s_t, a_t\}$; $\tau'$: masked sequence &
            \textbf{Sim}: DeepMind Control Suite
            \\ \hline
        MIDAS \cite{DBLP:conf/icml/LiGJ0HSSGW24} & 
            ViT, Mask R-CNN & 
            Inverse dynamics & 
            $\sum_{t=1}^{T-1} -\log P\left( a_t \mid f_{\text{inv}}(s_{1:T}) \right)$ &
            $s_{1:T}$: full state seq. &
            \textbf{Sim-Mani}: VIMA-Bench
            \\ \hline
        SMART \cite{DBLP:conf/iclr/SunMMBHK23} &
            CNN &
            Forward \& inverse dynamics & 
            $\sum_{t=1}^{T-1} \big( \operatorname{MSE}\left( s_{t+1}, f_{\text{fwd}}(s_t, a_t) \right) $ \newline \hspace{2.4em}
            $+ \operatorname{MSE}\left( a_t, f_{\text{inv}}(s_t, s_{t+1}) \right)\big) $ \newline
            $+ \sum \operatorname{MSE}\left( a_t, f_{\text{mask}}(\tau') \right) $
            &
            $s_t$, $a_t$: state, action; $\tau'$: masked sequence &
            \textbf{Sim}: DeepMind Control Suite
            \\ \hline
        PACT \cite{DBLP:conf/iros/BonattiVMFCK23} &
            ResNet-18, PointNet & 
            Forward dynamics & 
            $\sum_{t=1}^{T-1} \operatorname{MSE}\left( s_{t+1}, f_{\text{fwd}}(s_t, a_t) \right)$ &
            $s_t$, $a_t$: state, action &
            \textbf{Sim-Navi}: Habitat; \textbf{Real-Navi} (MuSHR vehicle)
            \\ \hline
        VPT \cite{DBLP:conf/nips/BakerAZHTEHSC22} & 
            ResNet  & 
            Inverse dynamics & 
            $\sum_{t=1}^{T} -\log P\left( a_{t} \mid f_{\text{inv}}(s_{1:T}) \right)$ &
            $s_t$, $a_t$: state, action &
            \textbf{Sim}: Minecraft
            \\ \hline
        GR-1 \cite{DBLP:conf/iclr/WuJCCXLLLK24} &
            ViT (MAE) & 
            Forward dynamics & 
            $\sum_{t=1}^{T-1} \operatorname{MSE}\left( s_{t+1}, f_{\text{fwd}}(s_t, a_{t}) \right)$ &
            $s_t$, $a_t$: state, action &
            \textbf{Sim-Mani}: CALVIN
        \\ \bottomrule
    \end{tabular}
    }
    \label{tb:dynamics}
\end{table*}

Vi-PRoM~\cite{DBLP:conf/iros/JingZLSYFK23} presents three distinct pretraining objectives. The first involves a contrastive self-supervised learning objective designed to distinguish between different videos. The remaining two objectives are centered around supervised learning tasks: temporal dynamics learning, aimed at recovering shuffled video frames, and image classification employing pseudo labels. Through a comprehensive comparison with preceding pretraining methods, Vi-PRoM demonstrates its effectiveness for behavior cloning (BC) and PPO. 

MIDAS~\cite{DBLP:conf/icml/LiGJ0HSSGW24} introduces an inverse dynamics prediction task as part of its pretraining. The objective is to train the model to predict actions from observations, formulated as a motion-following task. This approach enhances the model's understanding of the transition dynamics of the environment.

SMART~\cite{DBLP:conf/iclr/SunMMBHK23} presents a pretraining scheme encompassing three distinct objectives: forward dynamics prediction, inverse dynamics prediction, and randomly masked hindsight control. The forward dynamics prediction task involves predicting the next latent state, while the inverse dynamics prediction task entails predicting the previous action. In the case of hindsight control, the entire control sequence is provided as input, with some actions masked, and the model is trained to recover these masked actions. The incorporation of the first two dynamics prediction tasks facilitates capturing local and short-term dynamics, while the third task is designed to capture global and long-term temporal dependencies.

MaskDP~\cite{DBLP:conf/nips/Liu0GA22} features the masked decision prediction task, wherein both state and action tokens are masked for reconstruction. This masked modeling task is specifically crafted to equip the model with an understanding of both forward and inverse dynamics. Notably, in contrast to preceding masked modeling approaches like BERT or MAE, MaskDP can be applied directly to downstream tasks in a zero-shot manner.

PACT~\cite{DBLP:conf/iros/BonattiVMFCK23} introduces a pretraining objective aimed at modeling state-action transitions. It receives sequences of states and actions as input, and predicts each state and action token autoregressively. This pretrained model serves as a dynamics model, which can then be finetuned for various downstream tasks such as localization, mapping, and navigation. 

VPT~\cite{DBLP:conf/nips/BakerAZHTEHSC22} proposes a video pretraining method that harnesses unlabeled internet data to pretrain a foundation model for the game of Minecraft. The approach begins by training an inverse dynamics model using a limited amount of labeled data, which is then utilized to label internet videos. Subsequently, this newly auto-labeled data is employed to train the VPT foundation model through BC. This methodology follows semi-supervised imitation learning. As a result of this process, the model demonstrates human-level performance across a multitude of tasks. 

GR-1~\cite{DBLP:conf/iclr/WuJCCXLLLK24} introduces video prediction pretraining tailored for a GPT-style model. The ability to anticipate future frames aligns with forward dynamics learning, contributing to more accurate action prediction.

\subsubsection{World Models}
\label{sec:world-model}

A world model $P(\cdot)$ encodes commonsense knowledge about the world and predicts the future state for a given action~\cite{LeCun2022APT}:
\begin{equation}
\hat{s}_{t+1} \sim P(s_{t+1} \mid s_{t}, a_{t}).
\end{equation}
It enables model-based control and planning for embodied agents, as they can search for an optimal action sequence in imaginary space before executing any real actions. Although forward dynamics learning also attempts to predict the next state, it is usually treated as a pretraining task or an auxiliary loss to enhance the action decoder for the primary robotics tasks, rather than serving as a standalone module. In addition, new embodied experiences can be sampled from visual world models that explicitly generate images or videos of future states.

Dreamer~\cite{DBLP:conf/iclr/HafnerLB020} employs three primary modules to construct a latent dynamics model: a representation model, responsible for encoding images into latent states; a transition model, which captures transitions between latent states; and a reward model, predicting the reward associated with a given state. Under the actor-critic framework, Dreamer utilizes an action model and a value model to learn behavior through imagination by propagating analytic gradients through the learned dynamics. Building upon this foundation, DreamerV2~\cite{DBLP:conf/iclr/HafnerL0B21} introduces a discrete latent state space along with an improved objective. DreamerV3~\cite{DBLP:journals/corr/abs-2301-04104} extends its focus to a broader range of domains with fixed hyperparameters. DayDreamer~\cite{DBLP:conf/corl/WuEHAG22} applies this method to physical robots performing real-world tasks.

IRIS~\cite{DBLP:conf/iclr/MicheliAF23} employs a GPT-like autoregressive Transformer as the foundation of its world model, with a VQ-VAE serving as the vision encoder. Subsequently, a policy is trained using imagined trajectories, which are unrolled from a real observation by the world model. TWM~\cite{DBLP:conf/iclr/RobineHUH23} also investigates the application of Transformers in building world models.

\subsubsection{LLM-Induced World Models}
\label{sec:llm-world-model}
LLMs encompass a wealth of commonsense knowledge about the world, prompting many approaches to leverage that knowledge for improving VLAs.

DECKARD~\cite{DBLP:conf/icml/NottinghamAS0H023} prompts an LLM to generate abstract world models (AWMs) represented as directed acyclic graphs~\cite{DBLP:conf/nips/SongZK23, DBLP:conf/aaai/MaSHLZK23}, specifically tailored for the task of item crafting in Minecraft. DECKARD iterates between two phases: in the Dream phase, it samples a subgoal guided by AWM; in the Wake phase, DECKARD executes the subgoal and updates the AWM through interactions with the game. This guided approach enables DECKARD to achieve markedly faster item crafting compared with baselines lacking such guidance.

LLM-DM~\cite{DBLP:conf/nips/GuanVSK23} uses an LLM to construct world models in planning domain definition language (PDDL)---a capability that LLM+P~\cite{DBLP:journals/corr/abs-2304-11477} did not achieve, as its PDDL world model was hand-crafted. The LLM also acts as an interface, mediating between the generated PDDL model and corrective feedback from syntactic validators and human experts. Finally, the PDDL world model serves as a symbolic simulator, enabling plan generation.

RAP~\cite{DBLP:conf/emnlp/HaoGMHWWH23} repurposes an LLM to act as both a policy that predicts actions and a world model that provides the state transition distribution. Unlike previous chain-of-thought (CoT) prompting methods, RAP incorporates Monte Carlo tree search (MCTS) to enable structured planning, allowing the LLM to build a reasoning tree incrementally. This reasoning strategy helps RAP find a high-reward path that balances exploration and exploitation. Tree-Planner~\cite{DBLP:conf/iclr/HuMYD0SC00024} improves efficiency by prompting the LLM only once to generate diverse paths within an action tree.

LLM-MCTS~\cite{DBLP:conf/nips/ZhaoLH23} extends the problem setting to POMDPs. As a world model, the LLM generates the initial belief of the current state; as a policy, it serves as a heuristic to guide action selection. By leveraging its commonsense knowledge, the LLM reduces the search space of MCTS, thereby improving search efficiency.

\subsubsection{Visual World Models}
\label{sec:visual-world-model}

Unlike LLM-induced world models, which are in textual form, visual world models generate images, videos, or 3D scenes of future states---aligning more closely with the physical world. They can further be utilized to generate new trajectories. This direction has been gaining increasing attention since Sora demonstrated world-simulation capabilities~\cite{sora-world-model-openai}, as investigated by a dedicated survey~\cite{DBLP:journals/corr/abs-2405-03520}.

Genie~\cite{DBLP:conf/icml/BruceDEPS0LMSAA24} introduces a new class of generative models, termed generative interactive environments. It consists of three main components: a spatiotemporal video tokenizer, an autoregressive dynamics model, and a latent action model. After being trained on unlabeled videos in an unsupervised manner, Genie allows users to interact with the generative environment on a frame-by-frame basis. Consequently, Genie establishes a foundation world model.

3D-VLA~\cite{DBLP:conf/icml/ZhenQCY0DHG24} proposes a 3D world model capable of goal generation. It processes visual inputs, such as images, depth maps, and point clouds, and then generates a goal state---either as an image or a point cloud---using diffusion models in response to the user's query. The generated goal state can subsequently be used to guide robot control.

UniSim~\cite{DBLP:conf/iclr/YangDGTKSA24} builds a generative model based on real-world interaction videos. It is capable of simulating visual outcomes for both high-level and low-level actions, which can then be leveraged as new experiences for training embodied agents. E2WM~\cite{DBLP:conf/nips/XiangTGSWYH23} even treats existing simulators as world models to collect embodied experiences via MCTS.

\subsubsection{Reasoning}
\label{sec:reasoning}
Reasoning has become a key capability of LLMs, as demonstrated by chain-of-thought (CoT) methods~\cite{DBLP:conf/nips/KojimaGRMI22, DBLP:conf/nips/Wei0SBIXCLZ22}. In embodied AI, researchers are exploring how to leverage CoT reasoning to refine the decision-making process.

Reasoning is naturally compatible with high-level task planning, as both often occur in the language domain. ThinkBot~\cite{DBLP:journals/corr/abs-2312-07062} applies CoT to recover missing action descriptions in sparse human instructions, thereby enhancing instruction coherence and boosting success rates in difficult tasks. ReAct~\cite{DBLP:conf/iclr/YaoZYDSN023} interleaves reasoning traces and actions, where CoT can help create action plans, inject commonsense knowledge, and handle exceptions, ultimately improving embodied decision-making. RAT~\cite{DBLP:journals/corr/abs-2403-05313} integrates CoT with retrieval-augmented generation (RAG) to mitigate hallucinations and thus improve embodied planning. Tree-Planner~\cite{DBLP:conf/iclr/HuMYD0SC00024} employs a tree-of-thoughts approach for task planning.

Another paradigm involves equipping low-level control policies with reasoning capabilities, as pioneered by ECoT~\cite{DBLP:journals/corr/abs-2407-08693}. This method trains OpenVLA~\cite{DBLP:journals/corr/abs-2406-09246} to conduct embodied CoT reasoning about plans, subtasks, motions, and visual features, before predicting low-level actions. By relying on this multi-step reasoning rather than the ``muscle memory'' of VLAs, it improves success rates on challenging generalization tasks without requiring additional robot data. CoT-VLA~\cite{DBLP:conf/cvpr/ZhaoLKFZWLMHFHL25} introduces visual CoT reasoning for VLAs.

\subsubsection{Policy Steering}
\label{sec:steering}
Policy steering can enhance VLA performance at test-time without the need for expensive retraining. V-GPS~\cite{DBLP:conf/corl/NakamotoMKL24} re-ranks generated actions based on a learned value function, while RoboMonkey~\cite{DBLP:journals/corr/abs-2506-17811} employs a VLM-based verifier to select the optimal action from a sampled set.

\subsubsection{Strengths and Limitations}

\paragraph{PVRs} Although time CL with videos and text-guided pretraining methods, such as CLIP, provide image-level information, they lack the pixel-level details offered by MAE-based self-supervised learning methods. Pixel-level information contains rich details---including segmentation masks, object positions, and depth estimation---that are generally more useful for robot manipulation tasks requiring high precision, as demonstrated by VC-1's comparison~\cite{DBLP:conf/nips/MajumdarYAMCSJB23}. DINOv2 learns both pixel- and image-level features by combining masked image modeling with a momentum encoder and multicrop augmentation, with benefits on downstream robotic tasks evidenced by OpenVLA~\cite{DBLP:journals/corr/abs-2406-09246}. I-JEPA focuses on patches in the representation space and, as a result, captures low-level image features more effectively than view-invariance methods such as DINO. Theia distills various off-the-shelf vision foundation models into a single model that surpasses isolated individual models, as demonstrated by comprehensive evaluations against most existing PVRs in robot learning~\cite{DBLP:conf/corl/ShangSMMKWH24}.

\paragraph{Forward \& Inverse Dynamics} Forward dynamics learning is generally more challenging than inverse dynamics learning because predicting future states is more complex than predicting past actions. Consequently, the difficulty of forward dynamics often leads to greater performance improvements, as demonstrated in SMART. However, inverse dynamics models can be used to generate action labels for datasets that contain only states, such as raw robot manipulation videos.

\paragraph{World Models \& Reasoning} Although both world models and reasoning methods can be applied to low-level control policies and high-level task planners, current approaches remain distinct. World models are predominantly used to interact with control policies because they excel at generating the immediate next state given low-level actions. In contrast, CoT-based reasoning methods focus on task planning since they express thought chains in text, making them well-suited for refining text-based task plans.
\subsection{Low-Level Control Policies}
\label{sec:policy}

Through the integration of an action decoder with perception modules, such as vision encoders and language encoders, a VLA model $\pi_{\theta}$ with parameters $\theta$ is formed as a control policy to execute language instructions $p$:
\begin{equation}
\hat{a}_t \sim \pi_{\theta} (a_t \mid p, s_{\leq t}, a_{< t}).
\end{equation}
It can also be referred to as a low-level policy, low-level controller, or action primitive. The diversity among VLAs arises from individual modules and overall architectures. The general architecture is shown in Fig.~\ref{fig:arch}. This section explores various approaches to designing low-level control policies. Table~\ref{tb:policy} summarizes their technical details.

\begin{figure*}
    \centering
    \includegraphics[width=1.0\linewidth, trim={0.45in 8.45in 0.95in 0.95in},clip]{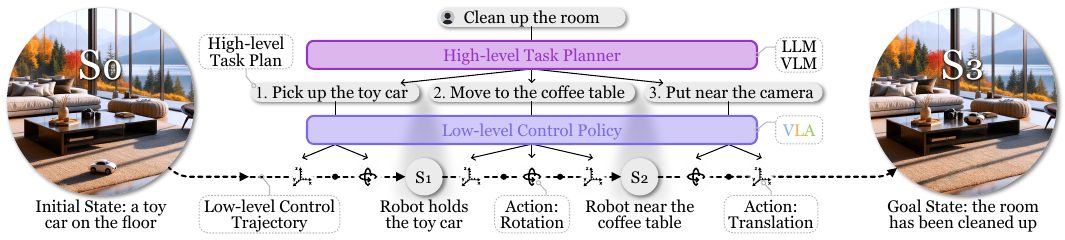}
    \caption{Illustration of a hierarchical robot policy. The high-level task planner decomposes the user instruction into subtasks, which are then executed step by step by the low-level control policy.}
    \label{fig:illustration}
\end{figure*}

\subsubsection{Non-Transformer Control Policies}
\label{sec:non-tfm}

Prior to the adoption of Transformer models, early control policies for language-conditioned robotic tasks varied significantly in architecture.

CLIPort~\cite{DBLP:conf/corl/ShridharMF21} integrates CLIP with the Transporter network, creating a two-stream architecture. The CLIP vision encoder extracts ``semantic'' information from the RGB image, while the Transporter network extracts ``spatial'' information from the RGB-D image. The CLIP sentence encoder encodes the language instruction and guides the output $SE(2)$ action, consisting of paired pick and place end-effector poses. It represents an early demonstration of language-conditioned pick-and-place capabilities.

BC-Z~\cite{DBLP:conf/corl/JangIKKELLF21} processes two types of task instructions: a language instruction or a human demonstration video. The environment is presented to the model in the form of an RGB image. Then the instruction embedding and the image embedding are combined through the FiLM layer, culminating in the generation of actions. This conditional policy is asserted to exhibit zero-shot task generalization to unseen tasks.

MCIL~\cite{DBLP:conf/rss/LynchS21} represents a pioneering robot policy that integrates free-form natural language conditioning. This is in contrast to earlier approaches that typically rely on conditions in the form of task IDs or goal images. MCIL introduces the capability to leverage unlabeled and unstructured demonstration data. This is achieved by training the policy to follow either image or language goals, with a small fraction of the training dataset consisting of paired image and language goals.

HULC~\cite{DBLP:journals/ral/MeesHB22} introduces several techniques aimed at enhancing robot learning architectures. These include a hierarchical decomposition of robot learning, a multimodal Transformer, and discrete latent plans. The Transformer learns high-level behaviors, hierarchically dividing low-level local policies and the global plan. In addition, HULC incorporates a visuo-lingual semantic alignment loss based on CL to align VL modalities. HULC++~\cite{DBLP:conf/icra/MeesBB23} further integrates a self-supervised affordance model. This model guides HULC to the actionable region specified by a language instruction, enabling it to fulfill tasks within this designated area.

UniPi~\cite{DBLP:conf/nips/DuY0DN0SA23} treats the decision-making problem as a text-conditioned video generation problem. To predict actions, UniPi generates a video based on a given text instruction, and actions are extracted from the frames of the video through inverse dynamics. This innovative policy-as-video formulation offers several advantages, including enhanced generalization across diverse robot tasks and the potential for knowledge transfer from internet videos to real robots.

\begin{table*}
    \centering
    \caption{VLAs that serve as low-level control policies. $\lozenge$ indicates large VLAs. Closely related (non-VLA models) are included in brackets. The remaining are generalized VLAs. cont/disc: continuous/discrete action. TFM: Transformer. Xattn: cross-attention. Concat: concatenation. Quant.: quantization. $p$/$s$: prompt/state vision encoder. [SC]: self-collect data
    }
    \resizebox{\textwidth}{!}{%
    \begin{tabular}{>{\RaggedRight}p{2.5cm} 
                    >{\RaggedRight}p{2.5cm} 
                    >{\RaggedRight}p{2.0cm} 
                    >{\RaggedRight}p{1.5cm} 
                    >{\RaggedRight}p{1.5cm} 
                    >{\RaggedRight}p{2.2cm} 
                    >{\RaggedRight}p{2cm} 
                    >{\RaggedRight}p{5.5cm} 
                    @{} 
                    }
        \toprule
        \textbf{Model} & \textbf{Vision\newline Encoder} & \textbf{Language Encoder / VLM} & \textbf{Action Decoder/Head} & \textbf{Architecture} & \textbf{Training Objectives} & \textbf{Training Datasets} & \textbf{Environments, Embodiments, Tasks, and Skills} \\
        \midrule

        CLIPort \cite{DBLP:conf/corl/ShridharMF21}  &
            CLIP-ResNet50, Transporter-ResNet & 
            CLIP-GPT  & 
            LingUNet & 
            Hadamard & 
            BC ($SE(2)$) & 
            [SC] & 
            \textbf{Sim}: Ravens
            \\ \hline
        MCIL \cite{DBLP:conf/rss/LynchS21}  &
            Custom-CNN &
            USE  &
            RNN  &
            LMP  & 
            MCIL & 
            Play-LMP, [SC] & 
            \textbf{Sim}: 3D Playroom environment
            \\ \hline
        HULC \cite{DBLP:journals/ral/MeesHB22, DBLP:conf/icra/MeesBB23}  &
            MCIL CNN &
            Sentence-BERT & 
            RNN  &
            LMP  & 
            MCIL & 
            CALVIN data & 
            \textbf{Sim}: CALVIN
            \\ \hline
        Language costs \cite{DBLP:conf/rss/SharmaSBPH0AF22}  &
            CLIP-ViT, UNet & 
            CLIP-GPT  & 
            STORM & 
            Concat & 
            MLE (cost map) &
            [SC] &
            \textbf{Sim} \& \textbf{Real} (Franka): pick, place
            \\ \hline
        Interactive Language \cite{DBLP:journals/corr/abs-2210-06407}  & 
            ResNet & 
            CLIP-GPT  & 
            TFM &
            Xattn & 
            BC (cont) & 
            [SC: Language-Table] &
            \textbf{Sim} \& \textbf{Real} (xArm6): rearrangement
            \\ \hline
        Hiveformer \cite{DBLP:conf/corl/GuhurCPTLS22}  &
            UNet & 
            CLIP-GPT  & 
            TFM, UNet &
            Concat & 
            BC (cont) & 
            RLBench data & 
            \textbf{Sim}: RLBench 
            \\ \hline
        PerAct \cite{DBLP:conf/corl/ShridharMF22}  &
            3D CNN & 
            CLIP-GPT  & 
            PerceiverIO & 
            Concat & 
            3D affordance & 
            RLBench data, [SC] & 
            \textbf{Sim}: RLBench; \textbf{Real} (Franka): pick, place, stack, open drawer, sweep, insert peg, etc.
            \\ \hline
        Act3D \cite{DBLP:conf/corl/GervetXGF23} &
            CLIP-ResNet50 &
            CLIP-GPT &
            TFM &
            Xattn &
            3D affordance (voxel) &
            RLBench data, [SC] &
            \textbf{Sim}: RLBench; \textbf{Real} (Franka): reach target, wipe, stack, etc.
            \\ \hline
        RVT \cite{DBLP:conf/corl/GoyalXGBCF23}, \newline RVT-2 \cite{DBLP:journals/corr/abs-2406-08545} &
            CLIP-ResNet50 &
            CLIP-GPT &
            TFM &
            Concat &
            2D affordance (project to 3D) & 
            RLBench data, [SC] & 
            \textbf{Sim}: RLBench; \textbf{Real} (Franka): stack, press, place
            \\ \hline
        RoboPoint \cite{DBLP:journals/corr/abs-2406-10721} &
            ViT-L/14 &
            Vicuna-V1.5 13B &
            \NA &
            Concat &
            2D affordance (project to 3D) &
            [SC] &
            \textbf{Real} (Franka): pick, place
            \\ \hline
        Gato \cite{DBLP:journals/tmlr/ReedZPCNBGSKSEBREHCHVBF22}  &
            ViT & 
            Sent.Piece & 
            TFM &
            Concat & 
            BC (cont \& disc) &
            [SC] &
            \textbf{Sim} \& \textbf{Real} (Sawyer): RGB-stacking 
            \\ \hline
        (RoboCat) \cite{DBLP:journals/corr/abs-2306-11706}  &
            VQ-GAN ($p, s$) & 
            \NA & 
            TFM & 
            Quant. & 
            BC, observation prediction &
            Self-improvement & 
            \textbf{Sim} \& \textbf{Real} (Sawyer, Franka, KUKA): stacking, building, lifting, insertion, removal
            \\ \hline
        VIMA \cite{DBLP:journals/corr/abs-2303-00905}  &
            ViT, Mask R-CNN & 
            T5 & 
            TFM &
            Xattn & 
            BC ($SE(2)$) & 
            [SC:VIMA-Data] &
            \textbf{Sim} (Ravens): VIMA-Bench
            \\ \hline
        BC-Z \cite{DBLP:conf/corl/JangIKKELLF21}  &
            ResNet18 ($p, s$) & 
            USE  & 
            MLP &
            FiLM & 
            BC (cont) & 
            [SC] &
            \textbf{Real} (EDR): pick-place/wipe/drag, grasp, push 
            \\ \hline
        RT-1 \cite{DBLP:conf/rss/BrohanBCCDFGHHH23}  &
            EfficientNet & 
            USE  & 
            TFM &
            FiLM  & 
            BC (disc) & 
            [SC: Fractal] &
            \textbf{Real} (EDR): pick-place, move, knock 
            \\ \hline
        MOO \cite{DBLP:journals/corr/abs-2303-00905}  &
            OWL-ViT ($p$), EfficientNet ($s$) & 
            USE  & 
            TFM &
            FiLM & 
            BC (disc) & 
            [SC] &
            \textbf{Real} (EDR): pick, move near, knock, place upright, place into 
            \\ \hline
        Q-Transformer \cite{DBLP:journals/corr/abs-2309-10150}  &
            EfficientNet & 
            USE  & 
            TFM &
            FiLM & 
            TD error &
            Fractal, Auto-collect &
            \textbf{Sim}: pick; \textbf{Real} (EDR): pick, place, open/close drawer, move near
            \\ \hline
        (RT-Trajectory) \cite{DBLP:journals/corr/abs-2311-01977}  &
            EfficientNet & 
            \NA & 
            TFM &
            \NA & 
            BC (disc) & 
            [SC] & 
            \textbf{Real} (EDR): pick, place, fold towel, swivel chair, etc.
            \\ \hline
        (ACT) \cite{DBLP:conf/rss/ZhaoKLF23} & 
            ResNet18 & 
            \NA & 
            CVAE-TFM & 
            \NA & 
            BC (cont, action chunking) & 
            [SC] with ALOHA &
            \textbf{Sim}: transfer cube, bimanual insertion; \textbf{Real} (ViperX, WidowX): slot battery, open cup, etc
            \\ \hline
        RoboAgent / MT-ACT \cite{DBLP:conf/icra/BharadhwajVSGTK24} &
            CNN & 
            \NA & 
            CVAE-TFM &
            FiLM & 
            BC (cont, action chunking) & 
            RoboSet &
            \textbf{Real} (Franka): pick, place, open/close drawers, pour, push, drag, etc.
            \\ \hline
        RoboFlamingo \cite{DBLP:journals/corr/abs-2311-01378}  &
            CLIP-ViT-L/14  &
            LLaMA, MPT, GPT-NeoX & 
            LSTM, TFM &
            Xattn &
            BC (cont) &
            CALVIN data & 
            \textbf{Sim}: CALVIN
            \\ \hline
        RoboUniView \cite{DBLP:journals/corr/abs-2406-18977} &
            UVFormer & 
            \multicolumn{3}{c}{ (Model design follows RoboFlamingo) } &
            BC (cont) &
            CALVIN data & 
            \textbf{Sim}: CALVIN
            \\ \hline
        DeeR-VLA \cite{DBLP:conf/nips/YueWKHWSF024} &
            \multicolumn{4}{c}{
              (Model design follows RoboFlamingo++)
            } &
            BC (cont) &
            CALVIN data &
            \textbf{Sim}: CALVIN
            \\ \hline
        Instruct2Act \cite{DBLP:journals/corr/abs-2305-11176} &
            CLIP, SAM &
            ChatGPT  &
            Robot APIs &
            Tool-Use &
            \NA &
            \NA &
            \textbf{Sim} (Ravens): VIMA-Bench
            \\ \hline
        VoxPoser \cite{DBLP:journals/corr/abs-2307-05973}  &
            ViLD, MDETR, OWL-ViT, SAM  & 
            GPT-4 & 
            MPC &
            Tool-Use & 
            \NA & 
            \NA & 
            \textbf{Sim}: Sapien; \textbf{Real} (Franka): move\&avoid, set table, close drawer, open bottle, sweep trash
            \\ \hline
        UniPi \cite{DBLP:conf/nips/DuY0DN0SA23}  & 
            Imagen video & 
            T5-XXL & 
            CNN & 
            Xattn &
            Inverse dynamics & 
            [SC:PF], Ravens, BridgeV1 &
            \textbf{Sim}: Painting Factory (PF), Ravens
            \\ \hline
        (Diffusion Policy) \cite{DBLP:conf/rss/ChiFDXCBS23}  &
            ResNet18 & 
            \NA & 
            UNet, TFM &
            \NA & 
            DDPM & 
            [SC] &
            \textbf{Sim}: Robomimic, Franka Kitchen, etc; 
            \textbf{Real} (UR5, Franka): push-T, flip mug, pour sauce 
            \\ \hline
        (DP3) \cite{DBLP:conf/rss/ZeZZHWX24} &
            DP3 Encoder &
            \multicolumn{3}{c}{ (Model design follows Diffusion Policy) } &
            DDIM &
            [SC] &
            \textbf{Sim}: (72 tasks); \textbf{Real} (Franka, Allegro hand): roll-up, dumpling, drill, pour
            \\ \hline
        SUDD \cite{DBLP:conf/corl/HaFS23} &
            ResNet18 & 
            CLIP-GPT  & 
            UNet, TFM &
            Concat & 
            DDPM & 
            Lang-guided data generation &
            \textbf{Sim}: MuJoCo; 
            \textbf{Real} (UR5e): pick, place
            \\ \hline
        Octo \cite{DBLP:conf/rss/GhoshWPBMDHK0LT24}  &
            CNN &
            T5-base & 
            TFM & 
            Concat & 
            DDPM &
            OXE & 
            \textbf{Real}: BridgeV2, CMU Baking, Stanford Coffee, Berkeley Peg Insert, etc.
            \\ \hline
        3D Diffuser Actor \cite{DBLP:journals/corr/abs-2402-10885} &
                \multicolumn{4}{c}{%
                  (Model design follows Act3D)%
                } &
            DDPM &
            RLBench, CALVIN, [SC] &
            \textbf{Sim}: RLBench, CALVIN; \textbf{Real} (Franka): put, open, close, stack, etc.
            \\ \hline
        MDT \cite{DBLP:journals/corr/abs-2407-05996} &
            CLIP-ViT-B/16 ($p$), Voltron/ResNet18 ($s$) &
            CLIP-GPT &
            DiT & 
            Concat &
            DDIM &
            CALVIN, LIBERO, [SC] & 
            \textbf{Sim}: CALVIN, LIBERO; 
            \textbf{Real} (Franka): pick, push, open, close, etc.
            \\ \hline
        RDT-1B \cite{liu2024rdt} &
            SigLIP &
            T5-XXL &
            DiT &
            Xattn &
            DDPM &
            (Aggregated Datasets) &
            \textbf{Real} (ALOHA dual-arm): wash, pour, fold, etc.
            \\ \hline
        RT-2 \cite{DBLP:journals/corr/abs-2307-15818} $\lozenge$ &
            ViT-4B, ViT-22B & 
            PaLI-X, PaLM-E & 
            Symbol-tuning & 
            Concat & 
            BC (disc), Co-fine-tuning &
            Fractal, VQA &
            \textbf{Sim}: Language-Table; \textbf{Real}: RT-1 evaluation tasks
            \\ \hline
        RT-H \cite{DBLP:journals/corr/abs-2403-01823} $\lozenge$ &
            \multicolumn{4}{c}{%
              (Model design follows RT-2)%
            } &
            BC (disc) &
            Diverse+Kitchen &
            \textbf{Real}: Diverse+Kitchen eval tasks
            \\ \hline
        RT-X \cite{DBLP:journals/corr/abs-2310-08864} $\lozenge$ &
            \multicolumn{4}{c}{%
              (Models from RT-1 and RT-2)%
            } &
            BC (disc) & 
            [SC: OXE] & 
            \textbf{Real}: BridgeV2, RT-1 evaluation tasks, etc.
            \\ \hline
        OpenVLA \cite{DBLP:journals/corr/abs-2406-09246} $\lozenge$ &
            DINOv2, SigLIP &
            Prismatic-7B &
            Symbol-tuning  &
            Concat &
            BC (disc) &
            OXE, DROID & 
            \textbf{Real}: BridgeV2, RT-1 evaluation tasks, Franka-Tabletop, DROID, etc.
            \\ \hline
        OpenVLA-OFT \cite{DBLP:journals/corr/abs-2502-19645} $\lozenge$ &
            \multicolumn{4}{c}{%
              (Improves OpenVLA with OFT recipe)%
            } &
            BC (cont, parallel decode w/ chunk.) &
            LIBERO, [SC] &
            \textbf{Sim}: LIBERO; \textbf{Real} (ALOHA setup): fold, scoop, put
            \\ \hline
        TraceVLA \cite{DBLP:conf/iclr/ZhengLH0DKHY25} $\lozenge$ &
            \multicolumn{4}{c}{%
              (Model design follows OpenVLA, adding visual trace prompting)%
            } &
            BC (disc) &
            BridgeV2, Fractal, [SC] & 
            \textbf{Sim}: SimplerEnv; \textbf{Real} (WidowX): pick, push, fold, swipe
            \\ \hline
        $\pi_0$ \cite{DBLP:journals/corr/abs-2410-24164} $\lozenge$ &
            SigLIP &
            PaliGemma &
            Action expert &
            Concat &
            Flow matching &
            OXE, [SC: $\pi$-cross-embod.] &
            \textbf{Real} (general robot control): sweep, open, pack, etc.
            \\ \hline
        RoboMamba \cite{DBLP:conf/nips/LiuLWALZYZGZ24} $\lozenge$ &
            CLIP-ViT &
            Mamba &
            MLP &
            Concat &
            BC (cont) &
            [SC] &
            \textbf{Sim}: Sapien; \textbf{Real} (Franka): open door/box, staple
            \\ \hline
        SpatialVLA \cite{DBLP:journals/corr/abs-2501-15830} $\lozenge$ &
            SigLIP, Ego3D Position Encoding &
            PaliGemma 2 &
            MLP &
            Concat &
            BC (Adaptive Action Grids) &
            OXE, BridgeV2, Fractal, RH20T &
            \textbf{Sim}: SimplerEnv, LIBERO; \textbf{Real} (WidowX): pick, place, close, push
            \\ \hline
        TinyVLA \cite{DBLP:journals/corr/abs-2409-12514} $\lozenge$ &
            ViT &
            Pythia &
            Diffusion Policy &
            Concat &
            DDPM &
            OXE, [SC] &
            \textbf{Sim}: MetaWorld; \textbf{Real} (Franka, UR5): place, stack, flip mug, close drawer, open box
            \\ \hline
        CogACT \cite{DBLP:journals/corr/abs-2411-19650} $\lozenge$ &
            DINOv2, SigLIP &
            LLaMA 2 &
            DiT &
            Concat &
            DDIM &
            OXE, [SC] &
            \textbf{Sim}: SimplerEnv; \textbf{Real} (Realman, Franka): pick, place, stack, open/close oven
            \\ \hline
        DexVLA \cite{DBLP:journals/corr/abs-2502-05855} $\lozenge$ &
            ViT, ResNet-50 &
            Qwen2-VL 2B, DistilBERT,  &
            ScaleDP &
            Concat &
            DDPM &
            [SC] &
            \textbf{Real} (Franka, UR5e, AgileX): pick, fold shirt, bus table, pour
            \\ \hline
        HybridVLA \cite{DBLP:journals/corr/abs-2503-10631} $\lozenge$ &
            DINOv2, SigLIP, CLIP &
            LLaMA 2, Phi-2 &
            MLP &
            Concat &
            DDIM &
            OXE, DROID &
            \textbf{Sim}: RLBench; \textbf{Real} (Franka, AgileX): pick-place, pour, open drawer, fold, etc.
            \\ \hline
        LAPA \cite{DBLP:conf/iclr/YeJJJYPMTCLLL0Z25} $\lozenge$ &
            C-ViViT, VQ-GAN &
            LWM-Chat-1M &
            MLP &
            Concat &
            LAPA &
            OXE, BridgeV2, Something V2 &
            \textbf{Sim}: Language-Table, SimplerEnv; \textbf{Real} (Franka): pick, cover with towel, knock over
            \\ \hline
        WorldVLA \cite{DBLP:journals/corr/abs-2506-21539} $\lozenge$ &
            VQ-GAN &
            Chameleon &
            TFM &
            Quant. &
            BC (disc, chunk.), world model &
            LIBERO, OpenVLA data &
            \textbf{Sim}: LIBERO
            \\ \hline
        UniVLA \cite{DBLP:journals/corr/abs-2506-19850} $\lozenge$ &
            MoVQ & 
            Emu3 &
            TFM &
            Quant. &
            BC (disc, chunk.), world model &
            (Sim benchmark data) &
            \textbf{Sim}: CALVIN, LIBERO, SimplerEnv
        \\ \bottomrule
    \end{tabular}
    }
    \label{tb:policy}
\end{table*}

\subsubsection{Transformer-Based Control Policies}
\label{sec:tfm-based}

Since the introduction of Transformers, control policies have converged to similar Transformer-based architectures.

Interactive language~\cite{DBLP:journals/corr/abs-2210-06407} presents a robotic system wherein the low-level control policy can be guided in real-time by human instructions conveyed through language, enabling the completion of long-horizon rearrangement tasks. The efficacy of such language-based guidance is primarily attributed to the utilization of a meticulously collected dataset containing diverse language instructions, which surpasses previous datasets by an order of magnitude in scale. 

Hiveformer~\cite{DBLP:conf/corl/GuhurCPTLS22} places significant emphasis on leveraging multiview scene observations and maintaining the full observation history for a language-conditioned policy. This approach represents an advancement over previous systems, such as CLIPort and BC-Z, that only use the current observation. Notably, Hiveformer stands out as one of the early adopters of Transformer architecture as its policy backbone.

Gato~\cite{DBLP:journals/tmlr/ReedZPCNBGSKSEBREHCHVBF22} proposes a model that can play Atari games, caption images, and stack blocks, all with a single set of model parameters. This achievement is facilitated by a unified tokenization scheme, harmonizing the input and output across diverse tasks and domains. Consequently, Gato enables the simultaneous training of different tasks. Astra~\cite{ma-etal-2025-astra} optimizes this architecture via trajectory attention.

RoboCat~\cite{DBLP:journals/corr/abs-2306-11706} proposes a self-improvement process designed to enable an agent to rapidly adapt to new tasks with as few as 100 demonstrations. This self-improvement process iteratively finetunes the model and self-generates new data with the finetuned model. Built upon the Gato model, RoboCat incorporates the VQ-GAN image encoder. During training, RoboCat predicts not only the next action but also future observations. The effectiveness of the self-improvement process is demonstrated through comprehensive experiments conducted in both simulated and real-world environments under multitask and multiembodiment settings.

RT-1~\cite{DBLP:conf/rss/BrohanBCCDFGHHH23}, developed by the same team as BC-Z, shares similarities with BC-Z but introduces some key distinctions. Notably, RT-1 employs a vision encoder based on the more efficient EfficientNet, departing from BC-Z's use of ResNet. However, RT-1 does not use video as a task instruction. In addition, RT-1 replaces the MLP action decoder in BC-Z with a Transformer decoder, producing discretized actions. This modification enables RT-1 to attend to past images, enhancing its performance compared with BC-Z.

Q-Transformer~\cite{DBLP:journals/corr/abs-2309-10150} extends RT-1 by introducing autoregressive Q-functions. In contrast to RT-1, which learns expert trajectories through imitation learning, Q-Transformer adopts Q-learning methods. Alongside the TD error objective of Q-learning, a conservative regularizer is incorporated to ensure that the maximum value action remains in-distribution. This approach allows Q-Transformer to leverage not only successful demonstrations but also failed trajectories for learning.

RT-Trajectory~\cite{DBLP:journals/corr/abs-2311-01977} adopts trajectory sketches as policy conditions instead of relying on language conditions or goal conditions. These trajectory sketches consist of curves that delineate the intended trajectory for the robot end-effector to follow. They can be manually specified through a graphical user interface, extracted from human demonstration videos, or generated by foundation models. RT-Trajectory's policy is built upon the backbone of RT-1 and trained to control the robot arm to accurately follow the trajectory sketches. This approach facilitates generalization to novel objects, tasks, and skills, as trajectories from various tasks are transferable.

ACT~\cite{DBLP:conf/rss/ZhaoKLF23} builds a conditional VAE policy with action chunking, requiring the policy to predict a sequence of actions rather than a single one. During inference, the action sequence is averaged using a method called temporal ensembling. RoboAgent~\cite{DBLP:conf/icra/BharadhwajVSGTK24} extends ACT via MT-ACT, demonstrating that action chunking improves temporal consistency. It also introduces an inpainting-based semantic augmentation method.

RoboFlamingo~\cite{DBLP:journals/corr/abs-2311-01378} adapts the OpenFlamingo VLM to a robot policy by attaching an LSTM-based policy head. This demonstrates that pretrained VLMs can be effectively transferred to language-conditioned robotic manipulation tasks. 

A recent trend in LLMs is equipping them with tool-use capabilities by generating code that calls tools via APIs~\cite{DBLP:journals/fcsc/QuDWCWYXW25}. Instruct2Act~\cite{DBLP:journals/corr/abs-2305-11176} follows this paradigm by integrating vision and action tools, enabling LLMs to perform robotic tasks.

\begin{figure*}
     \centering
        \subfloat[FiLM \label{type:film}]{
	     \centering
	     \includegraphics[width=0.3\columnwidth, trim={1in 5.735in 5.75in 3.62in},clip]{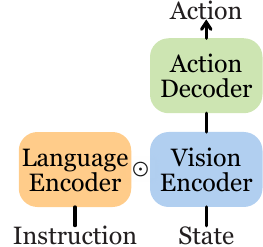}
	}
	\hfil
	\subfloat[Cross-attention \label{type:xattn}]{
		\centering
		\includegraphics[width=0.3\columnwidth, trim={1in 8.735in 5.75in 0.62in},clip]{figures/architecture.pdf}
	}
	\hfil
	\subfloat[Concatenation \label{type:concat}]{
		\centering
		\includegraphics[width=0.3\columnwidth, trim={1in 2.735in 5.75in 6.62in},clip]{figures/architecture.pdf}
	}
    \hfil
    \subfloat[Tool Use \label{type:tool_use}]{
		\centering
		\includegraphics[width=0.386\columnwidth, trim={5.75in 2.735in 0.5in 6.62in},clip]{figures/architecture.pdf}
	}
    \medskip

	\subfloat[Standard LVLA \label{type:standard}]{
		\centering
		\includegraphics[width=0.3\columnwidth, trim={3.5in 8.735in 3.25in 0.62in},clip]{figures/architecture.pdf}
	}
    \hfil
	\subfloat[Action Expert \label{type:action_expert}]{
		\centering
		\includegraphics[width=0.3\columnwidth, trim={6.0in 8.735in 0.75in 0.62in},clip]{figures/architecture.pdf}
	}
    \hfil
    \subfloat[Dual-System \label{type:dual_system}]{
		\centering
		\includegraphics[width=0.3\columnwidth, trim={6.0in 5.735in 0.75in 3.62in},clip]{figures/architecture.pdf}
	}
    \hfil
	\subfloat[VLA + World Model \label{type:world_model}]{
		\centering
		\includegraphics[width=0.386\columnwidth, trim={3.25in 5.735in 3in 3.62in},clip]{figures/architecture.pdf}
	}
    \hfil
	\subfloat[Quantization \label{type:quantization}]{
		\centering
		\includegraphics[width=0.386\columnwidth, trim={3.25in 2.735in 3in 6.62in},clip]{figures/architecture.pdf}
	}

    \caption{Representative architectures of VLA models. Some correspond to the ``Architecture'' column in Table~\ref{tb:policy}. $\odot$: Hadamard product. $\oplus$: Concatenation.
    }
    \label{fig:architectures}
\end{figure*}

\subsubsection{Control Policies for Multimodal Instructions}
\label{sec:multi-modal-instruction}

Multimodal instruction enables new ways to specify tasks, such as through demonstrations, by naming novel objects, or by pointing with a finger or mouse click.

VIMA~\cite{DBLP:journals/corr/abs-2210-03094} places a significant emphasis on multimodal prompts and the generalization capabilities of models. By incorporating multimodal prompts, more specific and intricate tasks can be formulated compared with pure text prompts. VIMA introduces six main types of tasks: object manipulation, visual goal reaching, novel concept grounding, one-shot video imitation, visual constraint satisfaction, and visual reasoning. These tasks are often challenging or even infeasible to express using only language prompts. VIMA-Bench has been developed to evaluate across four generalizability levels: placement, novel combinatorial, novel object, and novel task.

MOO~\cite{DBLP:journals/corr/abs-2303-00905} extends RT-1 to handle multimodal prompts. Leveraging the backbone of RT-1, it incorporates OWL-ViT to localize target objects specified by the prompt. By expanding the RT-1 dataset with new objects and additional prompt images, MOO enhances the generalization capabilities of RT-1. This extension also facilitates new methods of specifying target objects, such as pointing with a finger and clicking on graphical user interfaces.

\subsubsection{Control Policies with 3D Vision}
\label{sec:3d_vision}

In our 3D world, it is reasonable to assume that 3D vision provides richer information than 2D images. 

Point clouds are a common representation for 3D visual inputs due to their straightforward derivation from RGB-D inputs, as demonstrated by both DP3~\cite{DBLP:conf/rss/ZeZZHWX24} and 3D Diffuser Actor~\cite{DBLP:journals/corr/abs-2402-10885}. Voxels have likewise been extensively studied. VER~\cite{DBLP:conf/cvpr/LiuWY24} proposes voxelizing multiview images into 3D cells in a coarse-to-fine manner, enhancing performance in vision-language navigation tasks. PerAct~\cite{DBLP:conf/corl/ShridharMF22} facilitates efficient task learning with only a few demonstrations by leveraging voxel representations in both the observation and action spaces. The input comprises voxel maps reconstructed from multiview RGB-D images, while the output corresponds to the best voxel for guiding the gripper's movement. RoboUniView~\cite{DBLP:journals/corr/abs-2406-18977} improves performance by injecting 3D information from multiperspective images through a novel UVFormer vision encoder, which is pretrained on a 3D occupancy task. 

In contrast, Act3D~\cite{DBLP:conf/corl/GervetXGF23} introduces a continuous resolution 3D feature field, with adaptive resolutions based on the current task, addressing the computational cost of voxelization. RVT, RVT-2~\cite{DBLP:conf/corl/GoyalXGBCF23, DBLP:journals/corr/abs-2406-08545} propose re-rendering images from virtual views of the scene point cloud and using these images as inputs, rather than directly relying on 3D inputs.

\subsubsection{Diffusion-Based Control Policies}
\label{sec:diffusion_policy}

Diffusion-based action generation leverages the success of diffusion models in the field of CV.

Diffusion Policy~\cite{DBLP:conf/rss/ChiFDXCBS23} formulates a robot policy as a DDPM. This approach incorporates a variety of techniques, including receding horizon control, visual conditioning, and the time-series diffusion Transformer. The effectiveness of this diffusion-based visuomotor policy is underscored by its proficiency in multimodal action distributions, high-dimensional action spaces, and training stability.

SUDD~\cite{DBLP:conf/corl/HaFS23} presents a framework where an LLM guides data generation, and subsequently, the filtered dataset is distilled into a visuo-linguo-motor policy. This framework achieves language-guided data generation by composing an LLM with a suite of primitive robot utilities, such as grasp samplers and motion planners. It then extends Diffusion Policy by incorporating language-based conditioning for multitask learning and facilitates the distillation of the filtered dataset.

Octo~\cite{DBLP:conf/rss/GhoshWPBMDHK0LT24} introduces a Transformer-based diffusion policy characterized by a modular open-framework design, allowing for the flexible integration of different task definition encoders, observation encoders, and action decoders with the Octo Transformer. Being among the first to utilize the OXE dataset~\cite{DBLP:journals/corr/abs-2310-08864}, Octo demonstrates positive transfer and generalizability across diverse robots and tasks.

MDT~\cite{DBLP:journals/corr/abs-2407-05996} builds a variant of the DiT model from CV for the action prediction head. Originally proposed as a Transformer-based diffusion model, DiT replaced the classical U-Net architecture in image generation. Coupled with two auxiliary objectives---masked generative foresight and contrastive latent alignment---MDT similarly demonstrates better performance than prior U-Net-based diffusion models.

RDT-1B~\cite{liu2024rdt} is a diffusion-based foundation model for bimanual manipulation, also built on DiT. It addresses data scarcity by introducing a physically interpretable unified action format across various robots, enabling pretraining on heterogeneous multirobot datasets with over 1M trajectories. This allows RDT to scale up to 1.2B parameters and demonstrate zero-shot generalization.

\subsubsection{Diffusion-Based Control Policies with 3D Vision}
\label{sec:3d_diffusion_policy}
Several works have proposed combining 3D vision with diffusion-based policies. DP3~\cite{DBLP:conf/rss/ZeZZHWX24} introduces 3D point cloud inputs to a diffusion policy. Similarly, 3D Diffuser Actor~\cite{DBLP:journals/corr/abs-2402-10885} shares the core idea of DP3 but differs in model architecture by combining Act3D with Diffusion Policy. 3D-MoE~\cite{DBLP:journals/corr/abs-2501-16698} explores an efficient mixture-of-experts architecture with DiT-based action diffusion using a rectified flow framework.

\subsubsection{Control Policies for Motion Planning}
\label{sec:motion_planner}
Motion planning involves decomposing movement tasks into discrete waypoints while satisfying constraints such as obstacle avoidance and kinematic limits.

The \textit{language costs} framework~\cite{DBLP:conf/rss/SharmaSBPH0AF22} presents a novel approach to robot correction using natural language for human-in-the-loop control. This method leverages predicted cost maps generated from human instructions, which are then utilized by the motion planner to compute the optimal path. This framework enables users to correct goals, specify preferences, or recover from errors through intuitive language commands.

VoxPoser~\cite{DBLP:journals/corr/abs-2307-05973} employs an LLM and a VLM to create 3D voxel maps that represent affordances and constraints. It leverages the programming capability of LLMs and the perception capability of VLMs. The LLM translates language instructions into executable code, invoking the VLM to obtain object coordinates. Based on the composed affordance and constraint maps, VoxPoser employs model predictive control to generate a feasible trajectory for the robot arm's end-effector. Notably, VoxPoser does not require any training, as it directly connects LLMs and VLMs for motion planning.

RoboTAP~\cite{DBLP:conf/icra/VecerikD0DAZHAS24} breaks down demonstrations into stages marked by the opening and closing of the gripper. In each stage, RoboTAP uses the TAPIR algorithm to detect active points that track the relevant object from the source to the target pose. These tracked points are then used by a visual servoing controller to guide the robot. A motion plan is created by chaining these stages together, enabling few-shot visual imitation.

\subsubsection{Control Policies with Point-Based Actions}
\label{sec:vlm-based}

Recent research has explored leveraging the capabilities of VLMs to select or predict point-based actions---a cost-effective alternative to building full VLAs.

PIVOT~\cite{DBLP:conf/icml/NasirianyX0XL0X24} casts robotics tasks as a VQA problem, leveraging VLMs to select the best robot action from a set of visual proposals. Visual proposals are annotated in the form of keypoints on images. The VLM is iteratively prompted to refine them until the best option is identified.

RoboPoint~\cite{DBLP:journals/corr/abs-2406-10721} finetunes a VLM on the task of spatial affordance prediction, in which the model outputs 2D points indicating where to act on the image. These affordance points are subsequently projected into 3D space using depth maps, forming the predicted robot action.

ReKep~\cite{DBLP:journals/corr/abs-2409-01652} introduces constraint functions that map 3D keypoints in the scene to a numerical cost. Robotics manipulation tasks can be represented as a sequence of these constraints, which are generated by large vision models and VLMs. Consequently, robot actions can be obtained by solving the resulting constrained optimization problem.

\subsubsection{Large VLA}
\label{sec:lvla}

Large VLA is equivalent to the original VLA definition proposed by RT-2~\cite{DBLP:journals/corr/abs-2307-15818}, as illustrated in Fig.~\ref{fig:venn}. This terminology is analogous to the distinction between LLMs and general language models, or between large VLMs and general VLMs.

RT-2~\cite{DBLP:journals/corr/abs-2307-15818} leverages the capabilities of large multimodal models for robotics tasks, building upon architectures like PaLI-X and PaLM-E (see the architecture in Fig.~\ref{type:standard}). The approach introduces a co-fine-tuning strategy, jointly training the model on Internet-scale VQA data and robot data. This training scheme enhances the model's generalizability and enables emergent capabilities.

RT-H~\cite{DBLP:journals/corr/abs-2403-01823} introduces an action hierarchy that includes an intermediate prediction layer of language motions, situated between language instructions and low-level actions (translation and rotation). This additional layer facilitates improved data sharing across different tasks. For example, both the language instructions ``pick'' and ``pour'' may involve the language motion ``move the arm up''. Moreover, this action hierarchy enables users to specify corrections to recover from failures, from which the model can subsequently learn.

RT-X~\cite{DBLP:journals/corr/abs-2310-08864} builds upon the previous RT-1 and RT-2 models. These models are retrained using a newly introduced open-source dataset named open X-embodiment (OXE), which is orders of magnitude larger than previous datasets. The resulting models, RT-1-X and RT-2-X, both outperform their original versions. 

OpenVLA~\cite{DBLP:journals/corr/abs-2406-09246} was later developed as an open-source counterpart to RT-2-X. It additionally explored efficient finetuning methods, including LoRA and model quantization. OpenVLA-OFT~\cite{DBLP:journals/corr/abs-2502-19645} applies an Optimized Fine-Tuning (OFT) recipe for improved efficiency and performance. TraceVLA~\cite{DBLP:conf/iclr/ZhengLH0DKHY25} finetunes OpenVLA to enable visual trace prompting, enhancing the model's spatial-temporal awareness. 

$\pi_0$~\cite{DBLP:journals/corr/abs-2410-24164} proposes a flow-matching architecture for transforming VLMs into VLAs. By incorporating an additional action expert based on the mixture-of-experts framework (see the architecture in Fig.~\ref{type:action_expert}), it effectively inherits the Internet-scale knowledge of the base VLM while extending its capabilities to address robotic tasks.

RoboMamba~\cite{DBLP:conf/nips/LiuLWALZYZGZ24} replaces the computationally expensive Transformer architecture with a Mamba state space model featuring linear inference complexity, thereby achieving efficient robotic reasoning and action capabilities.

SpatialVLA~\cite{DBLP:journals/corr/abs-2501-15830} introduces Ego3D position encoding for injecting 3D information and represents actions with adaptive action grids for enhanced generalizability and robustness.

LAPA~\cite{DBLP:conf/iclr/YeJJJYPMTCLLL0Z25} devises the first unsupervised pretraining method for VLAs based on latent actions~\cite{DBLP:conf/icml/BruceDEPS0LMSAA24}. This approach employs a three-stage process to learn from Internet-scale unlabeled videos. First, a VQ-VAE framework is pretrained to extract quantized latent actions between image frames. Next, a VLA model is pretrained to predict these latent actions. Finally, the model is finetuned using only a small robotic dataset to map the latent actions to actual robot actions.

The integration of LVLAs with diffusion has also been gaining popularity. TinyVLA~\cite{DBLP:journals/corr/abs-2409-12514} leverages Diffusion Policy, while CogACT~\cite{DBLP:journals/corr/abs-2411-19650} utilizes a DiT action diffusion module. DexVLA~\cite{DBLP:journals/corr/abs-2502-05855} proposes embodied curriculum learning to progressively train a diffusion action expert, incorporating substep reasoning for decomposing long-horizon tasks. HybridVLA~\cite{DBLP:journals/corr/abs-2503-10631} integrates diffusion with the autoregressive paradigm to fully leverage VLMs' reasoning capabilities.

GR00T N1~\cite{DBLP:journals/corr/abs-2503-14734} introduces a dual-system architecture to build a robot foundation model for humanoid robots (see the architecture in Fig.~\ref{type:dual_system}). Its VLM-based System 2 processes image observations and language instructions at 10 Hz, while System 1 is a diffusion module that generates closed-loop motor actions in real time (120 Hz).

NORA-1.5~\cite{DBLP:journals/corr/abs-2511-14659} unifies a VLA with a world model through reward-guided post-training. Genie Envisioner~\cite{DBLP:journals/corr/abs-2508-05635} is a world foundation platform that integrates a world model and a VLA within a single video-generative framework. See the general architecture in Fig.~\ref{type:world_model}.

Visual autoregressive (VAR) modeling~\cite{DBLP:conf/nips/TianJYPW24} with quantized visual tokens demonstrates improved performance over diffusion models in image generation. This suggests that the three modalities of VLAs can be unified under the autoregressive paradigm~\cite{DBLP:journals/corr/abs-2501-12327}. WorldVLA~\cite{DBLP:journals/corr/abs-2506-21539} and UniVLA~\cite{DBLP:journals/corr/abs-2506-19850} advance this direction by integrating VLAs with world models (see the architecture in Fig.~\ref{type:quantization}). They quantize multimodal data into discrete tokens, forming a shared vocabulary of quantized multimodal tokens. Consequently, all modalities can be modeled autoregressively, enabling not only action and text generation but also image generation, thereby constituting a world model.

\subsubsection{Strengths and Limitations}

\paragraph{Architectures} Representative VLA architectures are illustrated in Fig.~\ref{fig:architectures}. FiLM is used in RT-1, and thus its follow-up models inherit this mechanism. While cross-attention may offer superior performance with smaller model sizes, concatenation is simpler to implement and can achieve comparable results with larger models~\cite{DBLP:journals/corr/abs-2210-03094}. Quantization unifies multimodal tokens into a shared vocabulary, thus enabling integration with world models. The tool-use paradigm of LLMs can also be applied to robotic tasks.

\paragraph{Action Types and Their Training Objectives} Most low-level control policies predict actions for the end-effector pose while abstracting away the motion planning module that produces more fine-grained motions. While this abstraction facilitates better generalization to different embodiments, it also imposes limitations on dexterity. 

The BC objective is used in imitation learning, with different variants for different action types. The BC objective for continuous actions can be written as
\begin{equation}
\mathcal{L}_{\text{Cont}} = \sum_t \operatorname{MSE}\left(a_{t}, \hat{a}_{t}\right),
\end{equation}
where $\operatorname{MSE}(\cdot)$ stands for mean squared error, $a_{t}$ is the action annotation from expert demonstrations, and $\hat{a}_{t}$ is the predicted action. 

Discrete actions are achieved by dividing the action value range into a fixed number of bins. Their BC objective is
\begin{equation}
\mathcal{L}_{\text{Disc}} = \sum_t \operatorname{CE}\left(a_{t}, \hat{a}_{t}\right),
\end{equation}
where $\operatorname{CE}(\cdot)$ stands for cross-entropy loss.

The BC objective for $SE(2)$ actions is
\begin{equation}
\mathcal{L}_{SE(2)} = \operatorname{CE}(a_{\text{pick}}, \hat{a}_{\text{pick}}) + \operatorname{CE}(a_{\text{place}}, \hat{a}_{\text{place}}).
\end{equation}

The DDPM objective in diffusion-based control policies is
\begin{equation}
\mathcal{L}_{\text{DDPM}} = \operatorname{MSE} \left( \varepsilon^k, \varepsilon_\theta (a_t^k, k) \right),
\end{equation}
where $a_t^k$ is the action $a_t$ corrupted by random noise $\varepsilon^k$ at diffusion iteration $k$, and $\varepsilon_\theta$ is the noise prediction network, i.e., the VLA model.


While discrete action has demonstrated superior performance in RT-1~\cite{DBLP:conf/rss/BrohanBCCDFGHHH23}, Octo~\cite{DBLP:conf/rss/GhoshWPBMDHK0LT24} argues that it leads to early grasping issues. $SE(2)$ actions require the model to predict only the pick pose and the place pose of the end-effector, which are sufficient for many tabletop manipulation tasks. However, more complex tasks---such as ``pouring water into a cup''---may require additional DoFs, thereby necessitating $SE(3)$ actions~\cite{DBLP:journals/corr/abs-2409-01652}. Although point-based actions can be coarse-grained, they are more easily obtained from VLMs in a zero-shot manner.

\begin{table*}
    \centering
    \caption{High-level task planners} 
    \resizebox{\textwidth}{!}{%
    \begin{tabular}{ 
                    >{\RaggedRight}p{3cm} 
                    >{\RaggedRight}p{3cm} 
                    >{\RaggedRight}p{2.5cm} 
                    >{\RaggedRight}p{2.7cm} 
                    >{\RaggedRight}p{1.75cm} 
                    >{\Centering}p{1.1cm} 
                    >{\RaggedRight}p{5.5cm} 
                    @{} 
                    }
        \toprule
        \textbf{Model} & \textbf{Vision \newline Model} & \textbf{Language Model / VLM} & \textbf{Low-level Control Policy} & \textbf{Architecture} & \textbf{Require Training} & \textbf{Environments, Embodiments, Tasks, and Skills} \\
        \midrule 



        PaLM-E \cite{DBLP:conf/icml/DriessXSLCIWTVY23}  &
            ViT, OSRT & 
            PaLM &  
            Interactive Language, RT-1 &
            Concat & 
            \cmark &
            \textbf{Sim}: TAMP; \textbf{Sim\&Real-Mani}: Language-Table; \textbf{Real}: SayCan setup
            \\ \hline

        EmbodiedGPT \cite{DBLP:journals/corr/abs-2305-15021}  &
            EVA ViT-G/14 & 
            LLaMA-7B & 
            MLP &
            Xattn & 
            \cmark &
            \textbf{Sim-Mani}: Franka Kitchen, Meta-World
            \\ \hline

        LEO \cite{DBLP:conf/icml/HuangYMLLW0ZJ024} &
            ConvNeXt, PointNet++ &
            Vicuna-7B & 
            MLP &
            Concat &
            \cmark &
            \textbf{Sim-Mani}: CLIPort task subset; \textbf{Sim-Navi}: Habitat-web
            \\ \hline

        3D-LLM \cite{DBLP:conf/nips/HongZCZDCG23} &
            CLIP, 3D-CLR, ConceptFusion &
            Flamingo, BLIP-2 &
            DD-PPO &
            Concat &
            \cmark &
            \textbf{Sim-Navi}: Habitat
            \\ \hline


        ShapeLLM \cite{DBLP:conf/eccv/QiDZGHGYM24} &
            ReCon++ &
            LLaMA &
            \NA &
            Concat &
            \cmark &
            3D MM-Vet Benchmark
            \\ \hline

        SayCan \cite{DBLP:conf/corl/IchterBCFHHHIIJ22}  &
            \NA &
            PaLM, FLAN & 
            BC-Z, MT-Opt & 
            \NA & 
            \xmark &
            \textbf{Real-Navi\&Mani} (EDR): office kitchen
            \\ \hline

        Translated $\langle$LM$\rangle$ \cite{DBLP:conf/icml/HuangAPM22}  &
            \NA & 
            GPT-3, Codex & 
            \NA & 
            \NA & 
            \xmark & 
            \textbf{Sim}: VirtualHome
            \\ \hline

        (SL)$^3$ \cite{DBLP:conf/acl/Sharma0A22}  &
            \NA & 
            T5-small & 
            seq2seq & 
            \NA &
            \cmark &
            \textbf{Sim-Navi}: ALFRED
            \\ \hline

        Inner Monologue \cite{DBLP:conf/corl/HuangXXCLFZTMCS22}  &
            MDETR, CLIP & 
            InstructGPT, PaLM &
            CLIPort, heuristics, SayCan policies & 
            Language & 
            \xmark & 
            \textbf{Sim\&Real-Mani}: Tabletop Rearrangement; \textbf{Real-Navi\&Mani}: Kitchen Mobile Mani.
            \\ \hline

        LLM-Planner \cite{DBLP:conf/iccv/SongSWCW023}  &
            HLSM object detector & 
            GPT-3 & 
            HLSM & 
            Language & 
            \cmark &
            \textbf{Sim-Navi}: ALFRED
            \\ \hline

        LID \cite{DBLP:conf/nips/LiPPDWF0HAAAM0Z22}  &
            CNN &
            GPT-2 & 
            \NA &
            Lang., Embed &
            \cmark &
            \textbf{Sim}: VirtualHome
            \\ \hline

        SMs \cite{DBLP:conf/iclr/ZengAICWWTPRSLV23}  & 
            CLIP, ViLD & 
            GPT-3, RoBERTa &
            CLIPort & 
            Lang., Code &
            \xmark &
            \textbf{Sim-Mani}: pick, place
            \\ \hline

        ProgPrompt \cite{DBLP:conf/icra/SinghBMGXTFTG23}  & 
            ViLD & 
            GPT-3 & 
            API & 
            Code & 
            \xmark &
            \textbf{Sim}: VirtualHome
            \\ \hline
        ChatGPT for Robotics \cite{DBLP:journals/corr/abs-2306-17582}  &
            YOLOv8 & 
            ChatGPT & 
            API & 
            Code & 
            \xmark &
            \textbf{Real-Navi}: drone flight; \textbf{Sim-Navi}: AirSim, Habitat
            \\ \hline
        CaP \cite{DBLP:conf/icra/LiangHXXHIFZ23}  & 
            ViLD, MDETR & 
            GPT-3, Codex &
            API &
            Code & 
            \xmark &
            \textbf{Sim-Mani} \& \textbf{Sim-Navi}: pick, place, etc.
            \\ \hline            
        DEPS \cite{DBLP:journals/corr/abs-2302-01560}  &
            CLIP  & 
            ChatGPT & 
            MC-Controller, Steve-1 &
            Code & 
            \xmark &
            \textbf{Sim}: Minecraft
            \\ \hline
        ConceptGraphs \cite{DBLP:journals/corr/abs-2309-16650}  &
            SAM, CLIP, LLaVA & 
            GPT-4 & 
            API &
            Code (JSON) & 
            \xmark &
            \textbf{Sim}: AI2-THOR; \textbf{Real} (Spot Arm): pick-place
        \\ \bottomrule
    \end{tabular}
    }
    \label{tb:planner}
\end{table*}

\paragraph{RT Series} RT-1~\cite{DBLP:conf/rss/BrohanBCCDFGHHH23} inspired a series of ``Robotic Transformer'' models. The Transformer backbone surpasses previous RNN backbones by harnessing the higher capacity of Transformers to absorb larger robot datasets. Preceding RT-1 was BC-Z, which solely utilized MLP layers for action prediction. Subsequent to RT-1, several works emerged to introduce new capabilities. MOO adapted RT-1 to accommodate multimodal prompts. RT-Trajectory enabled RT-1 to process trajectory sketches as prompts. Q-Transformer utilized Q-learning to train RT-1. RT-2, based on ViT and LLM, introduced a completely different architecture from RT-1. RT-X retrained RT-1 and RT-2 with a significantly larger dataset, resulting in improved performance. Based on RT-2, RT-H~\cite{DBLP:journals/corr/abs-2403-01823} introduced action hierarchies for better data sharing.

\paragraph{LVLA vs. Generalized VLA} While LVLAs can greatly enhance instruction-following abilities because they can better parse user intentions, concerns arise regarding their training cost and deployment speed. Slow inference speed, in particular, can significantly impact performance in dynamic environments, as changes in the environment may occur during inference. Therefore, several methods have been proposed to improve efficiency. TinyVLA~\cite{DBLP:journals/corr/abs-2409-12514} focuses on inference speed and data efficiency through a smaller VLM and diffusion head for robot action. DeeR-VLA~\cite{DBLP:conf/nips/YueWKHWSF024} proposes only partially activating the model via early-exit dynamic inference.

\paragraph{Scaling Laws} As in LLMs, scaling laws have also been observed in robotics~\cite{DBLP:conf/iclr/LinHSWY025, DBLP:conf/icml/PearceRBGDH25}, revealing the importance of model size, dataset size, and the diversity of environments and objects. Further research in this area can guide the development of VLAs with robust in-the-wild generalization capabilities.
\section{Task Planners}
\label{sec:planner}

A high-level task planner $\pi_{\phi}$ aims to decompose a complex task $\ell$ into a sequence of subtasks $\hat{\mathbf{p}} = [p_1, p_2, \dots, p_N]$ (i.e., a task plan), where each $p_i$ serves as an instruction to low-level control policies $\pi_{\theta}$:
\begin{equation}
\hat{\mathbf{p}} \sim \pi_{\phi}(\mathbf{p} \mid \ell, s_t).
\end{equation}
This process is sometimes referred to as task or subgoal decomposition and is closely related to task and motion planning (TAMP) and embodied decision making. When equipped with task planners, VLAs can complete more complex, long-horizon tasks, as illustrated in Fig.~\ref{fig:illustration}. Details are summarized in Table~\ref{tb:planner}. Ideally, task plans should also incorporate optimal scheduling for these subtasks.

\subsection{Monolithic Task Planners}
\label{sec:monolithic}

A single LLM or multimodal LLM (MLLM) can typically generate task plans by employing a tailored framework or through finetuning on embodied datasets. We refer to these as \textit{monolithic} models.

\subsubsection{End-to-End Task Planners}
\label{sec:end2end}

Similar to LVLAs, task planners can be implemented as end-to-end MLLMs, leveraging their Internet-scale knowledge for task planning.

PaLM-E~\cite{DBLP:conf/icml/DriessXSLCIWTVY23} integrates ViT and PaLM to create an embodied MLLM capable of performing high-level embodied reasoning tasks. Based on perceived images and high-level language instructions, PaLM-E generates a text plan that serves as instructions for low-level robotic policies. In a mobile manipulation environment, the generated plan is mapped to executable low-level instructions using SayCan~\cite{DBLP:conf/corl/IchterBCFHHHIIJ22}. As the low-level policy executes actions, PaLM-E can also replan based on changes in the environment. With PaLM as its backbone, PaLM-E can handle standard VQA tasks along with additional embodied VQA tasks.

EmbodiedGPT~\cite{DBLP:journals/corr/abs-2305-15021} introduces the embodied-former, which outputs task-relevant instance-level features. This is achieved by incorporating vision encoder embeddings and embodied planning information provided by an LLM. These instance features serve to inform the low-level policy about the immediate next action to take.

\subsubsection{End-to-End Task Planners with 3D Vision}
\label{sec:3d_planner}

Some task planners also explore the use of 3D vision. Because the majority of current MLLMs deal with images as visual inputs, they require architectural changes to incorporate 3D visual inputs. Consequently, they are usually end-to-end models.

LEO~\cite{DBLP:conf/icml/HuangYMLLW0ZJ024} uses a two-stage training strategy to integrate a point cloud encoder with an LLM: the first stage focuses on 3D vision-language alignment, followed by the second stage of 3D vision-language-action instruction tuning. LEO performs well not only in 3D question-answering tasks but also in manipulation, navigation, and task planning.

3D-LLM~\cite{DBLP:conf/nips/HongZCZDCG23} injects 3D information into LLMs and empowers them to perform 3D tasks, such as 3D-assisted dialog and navigation, using features from point clouds, gradSLAM, and neural voxel fields. MultiPLY~\cite{DBLP:conf/cvpr/HongZCWLG24} is an object-centric embodied LLM that incorporates even more modalities, including audio, tactile, and thermal.

ShapeLLM~\cite{DBLP:conf/eccv/QiDZGHGYM24} is built on the novel 3D vision encoder ReCon++, which distills knowledge from multiview image and text teacher models, along with a point cloud MAE. By integrating ReCon++ with LLaMA, ShapeLLM improves embodied interaction performance on their newly proposed 3D benchmark, 3D MM-Vet.

\subsubsection{Grounded Task Planners}
\label{sec:grounded}

Grounded task planning involves generating high-level actions while considering whether they can be executed by low-level control policies.

SayCan~\cite{DBLP:conf/corl/IchterBCFHHHIIJ22} is a framework designed to integrate high-level LLM planners with low-level control policies. In this framework, the LLM planner accepts a user's high-level instruction and ``says'' what the most probable next low-level skill is, a concept referred to as \textit{task-grounding}. The low-level policy provides the value function as the affordance function, determining the probability that the policy ``can'' complete the skill, known as \textit{world-grounding}. By considering both the LLM's plan and the affordances, the framework selects the optimal skill for the current state.

Translated $\langle$LM$\rangle$~\cite{DBLP:conf/icml/HuangAPM22} employs a two-step process to translate high-level instructions into executable actions. Initially, a pretrained causal LLM is utilized for \textit{plan generation}, breaking down the high-level instruction into the next action, expressed as a free-form language phrase. Then, as this phrase may not directly map to VirtualHome actions, a pretrained masked LLM is employed for \textit{action translation}. This step involves calculating the similarity between the generated action phrase and the admissible VirtualHome actions. The translated action is appended to the plan, and the updated plan is fed back into the LLM to generate the next action phrase. This two-step process is repeated until a complete plan is formed. A ``re-prompting'' strategy~\cite{DBLP:journals/corr/abs-2211-09935} is further proposed to generate corrective actions when the agent encounters precondition errors.

(SL)$^3$~\cite{DBLP:conf/acl/Sharma0A22} is a learning algorithm that alternates between three steps: segmentation, labeling, and parameter update. In the segmentation step, high-level subtasks are aligned with low-level actions; subtask descriptions are then inferred in the labeling step; and finally, the network parameters are updated. This approach enables a hierarchical policy to discover reusable skills with sparse natural language annotations.

\subsection{Modular Task Planners}
\label{sec:modular}

Finetuning end-to-end models on embodied data can be expensive; therefore, some approaches adopt a modular design by assembling off-the-shelf LLMs and VLMs into task planners. These approaches can also be viewed as following the tool-use architecture~\cite{DBLP:journals/fcsc/QuDWCWYXW25}.

\begin{figure}[t]
     \subfloat[Language-based \label{fig:lang_based}]{
         \centering
         \includegraphics[width=0.45\columnwidth, trim={1in 8in 4.5in 0.99in},clip]{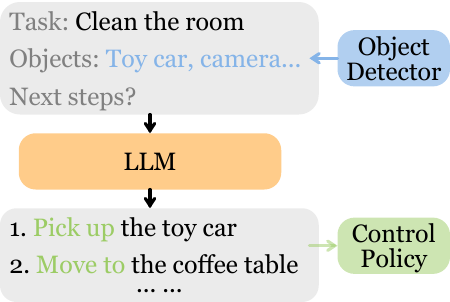}
     }
     \hfill
     \subfloat[Code-based \label{fig:code_based}]{
         \centering
         \includegraphics[width=0.45\columnwidth, trim={1in 5in 4.5in 4in},clip]{figures/planner.pdf}
    }
    \caption{Different approaches to connect LLM to multimodal modules in modular task planners.}
    \label{fig:planner_types}
\end{figure}

\subsubsection{Language-Based Task Planners}
\label{sec:lang-based}

Language-based approaches use natural language descriptions as the medium for exchanging multimodal information, as shown in Fig.~\ref{fig:lang_based}.

Inner monologue~\cite{DBLP:conf/corl/HuangXXCLFZTMCS22} sits between high-level commands and low-level policies to enable closed-loop control planning. It employs an LLM to generate language instructions for low-level policies and dynamically updates these instructions based on the received feedback. The feedback encompasses various sources: success feedback, object and scene feedback, and human feedback. As the feedback is communicated to the LLM in a textual format, no additional training is required for the LLM. A similar approach is used in ReAct~\cite{DBLP:conf/iclr/YaoZYDSN023}.

LLM-Planner~\cite{DBLP:conf/iccv/SongSWCW023} introduces a novel approach to constructing a hierarchical policy comprising a high-level planner and a low-level planner. The high-level planner harnesses the capabilities of an LLM to generate natural language plans, while the low-level planner translates each subgoal within the plan into primitive actions. While sharing similarities with previous methods in its overall architecture, LLM-Planner distinguishes itself by incorporating a re-planning mechanism, enabling the robot to ``get unstuck''.

LID~\cite{DBLP:conf/nips/LiPPDWF0HAAAM0Z22} introduces a novel data collection procedure termed active data gathering (ADG). A key aspect of ADG is hindsight relabeling, which reassigns labels to unsuccessful trajectories, effectively maximizing the utilization of data irrespective of trajectory success. By converting all environmental inputs into textual descriptions, the proposed language-model-based policy gains enhanced combinatorial generalization.

Socratic models (SMs)~\cite{DBLP:conf/iclr/ZengAICWWTPRSLV23} present a unique framework wherein diverse pretrained models are effectively composed without the need for finetuning. The framework relies on a key mechanism called multimodal-informed prompting, which facilitates information exchange among models with varied multimodal capabilities. The idea is to utilize multimodal models to convert non-language inputs into language descriptions, effectively unifying different modalities within the language space. Beyond excelling in conventional multimodal tasks, SMs showcase their versatility in robot perception and planning. In addition to natural language plans, task plans can also be represented in the form of pseudocode.

\subsubsection{Code-Based Task Planners}
\label{sec:code-based}

Code-based task planners leverage the coding ability of LLMs to generate task plans in the form of executable programs. Object detectors, VLMs, and control policies can be invoked via APIs, as shown in Fig.~\ref{fig:code_based}.

ProgPrompt~\cite{DBLP:conf/icra/SinghBMGXTFTG23} introduces a novel task-planning approach by prompting LLMs with program-like specifications detailing available actions and objects. This enables LLMs to generate high-level plans for household tasks in a few-shot manner. Environmental feedback can be incorporated through assertions within the program. 

ChatGPT for robotics~\cite{DBLP:journals/corr/abs-2306-17582} takes advantage of the programming ability of ChatGPT to facilitate ``user on the loop'' control, a departure from the conventional ``engineer in the loop'' methodology. The procedure includes several steps. First, a list of APIs is defined, such as an object-detection API, a grasp API, and a move API. Second, a prompt is constructed for ChatGPT, specifying the environment, API functionality, and task goal. Third, ChatGPT is iteratively prompted to write code utilizing the defined APIs to execute the task, provided access to simulation and user feedback for evaluating code quality and safety. Finally, the ChatGPT-generated code is executed. In this procedure, ChatGPT serves as a high-level task planner, and actions are generated through function calls to corresponding low-level APIs.

Code as policies (CaP)~\cite{DBLP:conf/icra/LiangHXXHIFZ23} also leverages the code-writing capability of LLMs. It employs GPT-3 or Codex to generate policy code, which, in turn, invokes perception modules and control APIs. CaP exhibits proficiency in spatial-geometric reasoning, generalization to new instructions, and parameterization for low-level control primitives. By leveraging the multimodal capabilities of GPT-4V, COME-robot~\cite{DBLP:journals/corr/abs-2404-10220} eliminates the need for perception APIs required by CaP. This opens up possibilities for open-ended reasoning and adaptive planning within a closed-loop framework, enabling capabilities such as failure recovery and free-form instruction following.

DEPS~\cite{DBLP:journals/corr/abs-2302-01560} stands for ``describe, explain, plan, and select.'' This approach employs an LLM to generate plans and explain failures based on feedback descriptions collected from the environment---a process referred to as ``self-explanation,'' which aids in replanning. In addition, DEPS introduces a trainable goal selector to choose among parallel candidate subgoals based on how easily they can be achieved, a crucial aspect often overlooked by other high-level task planners.

ConceptGraphs~\cite{DBLP:journals/corr/abs-2309-16650} introduces a method to convert observation sequences into open-vocabulary 3D scene graphs. Objects are extracted from RGB images using 2D segmentation models, and VLMs are employed to caption objects and establish inter-object relations, resulting in the construction of a 3D scene graph. This graph can then be translated into a text description (JSON), providing LLMs with rich semantic and spatial relationships between entities for task planning.

\begin{table}[t]
    \caption{Embodied Datasets. Per RT-X, skills correspond to verbs, and tasks are different combinations of verbs and objects. $\ast$ Dataset collected in simulation rather than the real world. Some datasets are continually updated, and we include only the original version. Table adapted from \cite{DBLP:conf/icra/FangFTLW0ZL24, DBLP:journals/corr/abs-2403-12945}}
    \resizebox{\columnwidth}{!}{%
        \begin{tabular}{
            >{\RaggedRight}p{2.1cm} 
            >{\RaggedRight}p{0.4cm} 
            >{\RaggedRight}p{0.4cm} 
            >{\RaggedRight}p{0.5cm} 
            >{\RaggedRight}p{0.85cm} 
            >{\RaggedRight}p{1.8cm} 
            >{\RaggedRight}p{0.75cm} 
            >{\RaggedRight}p{1.4cm} 
            >{\RaggedRight}p{1.3cm} 
        }
        \toprule
        \textbf{Dataset}           & \textbf{Skills}        & \textbf{Tasks}  & \textbf{Scenes} & \textbf{Episodes}     & \textbf{Collection}    & \textbf{Obs.}   & \textbf{Instruction}          & \textbf{Robots}     
        	\\ \hline

        MIME \cite{DBLP:conf/corl/SharmaMPG18} & 
        	12            & 20      & 1  & 8.3K     &
        	Human teleop. & RGBD  & Demo                & Baxter  
        	\\ \hline
			
		$\ast$RoboTurk \cite{DBLP:conf/corl/MandlekarZGBSTG18} &
        	2            & 3      & 1  & 2.1K     &
        	Human teleop. & RGB    & \NA               & Sawyer  
        	\\ \hline

        RoboNet \cite{DBLP:conf/corl/DasariETNBSSLF19}      & 
        	\NA             & \NA      & 10  & 162K     &
        	Script    & RGB      & Goal image          & (7 robots)
        	\\ \hline

        MT-Opt  \cite{DBLP:journals/corr/abs-2104-08212}  & 
        	2             & 12     & 1   & 800K     &
        	Script, RL     & RGB     & Task ID            & (7 robots)
        	\\ \hline

        BC-Z   \cite{DBLP:conf/corl/JangIKKELLF21}  & 
        	3             & 100    & 1   & 25.9K   &
        	Human teleop.  & RGB       & Lang, demo           & EDR
        	\\ \hline

        Fractal  \cite{DBLP:conf/rss/BrohanBCCDFGHHH23} & 
        	12            & 700+   & 2   & 130K     &
        	Human teleop.    & RGB       & Lang               & EDR    
        	\\ \hline

        MOO  \cite{DBLP:journals/corr/abs-2303-00905} & 
        	5             & \NA      & \NA   & 59.1K   &
        	Human teleop.  & RGB     & Multimodal     & EDR
        	\\ \hline

        $\ast$VIMA \cite{DBLP:journals/corr/abs-2210-03094} & 
        	17            & \NA      & 1     & 650K     &
        	Script   & RGB     & Multimodal      & UR5
        	\\ \hline

        RoboSet \cite{DBLP:conf/nips/KumarSZMC0R23}    & 
        	12            & 38     & 11     & 98.5K    &
        	Human, script    & RGBD   & Lang          & Franka
        	\\ \hline

        BridgeV2 \cite{DBLP:conf/rss/EbertYSBGDFL22} & 
        	13            & \NA      & 24     & 60.1K    &
        	Human, script      & RGB(D)  & Lang          & WidowX
	        \\ \hline

        RH20T \cite{DBLP:conf/icra/FangFTLW0ZL24} & 
        	42            & 147   & 7      & 110K+  &
        	Human teleop. & RGBD   & Lang   & (4 robots)
        	\\ \hline

        DROID \cite{DBLP:journals/corr/abs-2403-12945} & 
        	86             & \NA      & 564   & 76K &
        	Human teleop.      & RGBD  & Lang	  & Franka
        	\\ \hline

        OXE  \cite{DBLP:journals/corr/abs-2310-08864} & 
        	527           & 160K & 311   & 1M+      &
        	Aggregate data      & RGBD   & Lang        & (22 robots)
        	\\ \bottomrule
        \end{tabular}
    }
    \label{tb:datasets}
\end{table}

\subsection{Strengths and Limitations}

\textit{Monolithic} task planners that utilize grounded task planning focus on generating executable plans. End-to-end models share an architecture similar to most LVLAs and can be finetuned on specialized embodied data to achieve better performance. However, the training costs of such large models are often substantial. In contrast, \textit{modular} task planners are more readily deployable because they leverage off-the-shelf LLMs and VLMs. Language-based task planners offer the advantage of seamless integration of LLMs and VLMs, as they are designed to operate in the natural language space. However, they often require extra steps to align the generated task plans with language instructions that are admissible to low-level control policies. Conversely, while code-based task planners may require manually wrapping VLMs and control policies within APIs and preparing clear documentation in advance, they enable code debugging and provide greater controllability. Ultimately, their performance remains constrained by the programming capabilities of the underlying models.
\section{Datasets and Benchmarks}
\label{sec:data}

\begin{table*}
    \centering
    \caption{Simulators and simulated benchmarks. Control: continuous control tasks. D, S, A, N: depth, segmentation, audio, normal. Force: simulated contact force between end-effector and item. PD: pre-defined. Table adapted from \cite{DBLP:conf/corl/Srivastava0LMXV21, DBLP:conf/corl/0002XMLSSVGDJKL21}
    }
    \resizebox{\textwidth}{!}{%
        \begin{tabular}{
            >{\RaggedRight}p{2.5cm} 
            >{\RaggedRight}p{0.75cm} 
            >{\RaggedRight}p{0.75cm} 
            >{\RaggedRight}p{0.6cm} 
            >{\RaggedRight}p{0.9cm} 
            >{\RaggedRight}p{1.5cm} 
            >{\RaggedRight}p{2cm} 
            >{\RaggedRight}p{1.2cm} 
            >{\RaggedRight}p{2cm} 
            >{\RaggedRight}p{4cm} 
            >{\RaggedRight}p{4cm}
            @{} 
        }
        \toprule
        \textbf{Name}                & \textbf{Scenes\newline/Rooms}        & \textbf{Objects\newline/Cat} & \textbf{UI}        & \textbf{Physics Engine}      & \textbf{Task}       & \textbf{Observation}              & \textbf{Action}            & \textbf{Agent}                              & \textbf{Description}                             &  \textbf{Related}                                       \\ \hline
        Gibson \cite{DBLP:conf/cvpr/XiaZHSMS18} & 572/-               & \NA                  & \NA         & Pybullet    & Navi       & RGB, D, N, S           & \NA                 & \NA                                  & Navi only                               &  \NA                                              \\ \hline
        
        iGibson \cite{DBLP:journals/ral/XiaSLKTTMS20, DBLP:conf/iros/ShenX0MFWPBSTTV21, DBLP:conf/corl/0002XMLSSVGDJKL21} & 
            15/108              & 
            152/5              & 
            Mouse, VR & 
            Pybullet    & 
            Navi, Mani & 
            RGB, D, S, N, Flow, LiDAR & 
            Force             & 
            TurtleBot v2, LoCoBot, etc. & 
            VR, Continuous Extended States. Versions: 0.5, 1.0, 2.0          & 
            Benchmarks: BEHAVIOR-100 \cite{DBLP:conf/corl/Srivastava0LMXV21}, BEHAVIOR-1K \cite{DBLP:journals/corr/abs-2403-09227} 
            \\ \hline

        SAPIEN \cite{DBLP:conf/cvpr/XiangQMXZLLJYWY20} & 
            \NA                   & 
            2346/-             & 
            Code      & 
            PhysX       & 
            Navi, Mani & 
            RGB, D, S              & 
            Force             & 
            Franka                             & 
            Articulation, Ray Tracing               & 
            VoxPoser. Benchmark: SIMPLER \cite{DBLP:journals/corr/abs-2405-05941}
            \\ \hline

        AI2-THOR \cite{DBLP:journals/corr/abs-1712-05474} & 
            -/120               & 
            118/118            & 
            Mouse     & 
            Unity       & 
            Navi, Mani & 
            RGB, D, S, A           & 
            Force, PD & 
            ManipulaTHOR, LoCoBot, etc.        & 
            Object States, Task Planning. Versions: \cite{DBLP:conf/cvpr/EhsaniHHVWKKM21, DBLP:conf/cvpr/WeihsDKM21}        &  
            Benchmarks: ALFRED, RoPOR \cite{DBLP:conf/iclr/MirakhorG0B24}   
            \\ \hline  
        
        VirtualHome \cite{DBLP:conf/cvpr/PuigRBLWF018} & 
            7/-                 & 
            -/509              & 
            Lang  & 
            Unity       & 
            Navi, Mani & 
            RGB, D, S           & 
            Force, PD & 
            Human                              & 
            Object States, Task Planning            & 
            LID, Translated〈LM〉, ProgPrompt                 
            \\ \hline
        
        TDW  \cite{DBLP:conf/nips/GanSAMSTFKBHSKW21} & 
            15/120           & 
            112/50             & 
            VR        & 
            Unity, Flex & 
            Navi, Mani & 
            RGB, D, S, A           & 
            Force             & 
            Fetch, Sawyer, Baxter              & 
            Audio, Fluids                           &  
            \NA                                              
            \\ \hline
        
        RLBench  \cite{DBLP:journals/ral/JamesMAD20} & 
            1/-                 & 
            28/28              & 
            Code      &  
            V-REP      & 
            Mani       & 
            RGB, D, S           & 
            Force             & 
            Franka                             & 
            Tiered Task Difficulty                   & 
            Hiveformer, PerAct                              
            \\ \hline
        
        Meta-World \cite{DBLP:conf/corl/YuQHJHFL19} & 
            1/-                 & 
            80/7               & 
            Code      & 
            MuJoCo      & 
            Mani       & 
            Pose                        & 
            Force             & 
            Sawyer                             & 
            Meta-RL             & 
            R3M, VC-1, Vi-PRoM, EmbodiedGPT                 
            \\ \hline
        
        CALVIN \cite{DBLP:journals/ral/MeesHRB22} & 
            4/-                 & 
            7/5                & 
            \NA         & 
            Pybullet    & 
            Mani       & 
            RGB, D                   & 
            Force             & 
            Franka                             & 
            Long-horizon Lang-cond tasks & 
            GR-1, HULC, RoboFlamingo                        
            \\ \hline
        
        Franka Kitchen \cite{DBLP:conf/corl/0004KLLH19} & 
            1/-                 & 
            10/6               & 
            VR        & 
            MuJoCo      & 
            Mani       & 
            Pose                     & 
            Force             & 
            Franka                             & 
            Extended by R3M with RGB                & 
            R3M, Voltron, Vi-PRoM, Diffusion Policy, EmbodiedGPT 
            \\ \hline
        
        Habitat \cite{DBLP:conf/iccv/SavvaMPBKMZWJSL19, DBLP:conf/nips/SzotCUWZTMMCMGV21} & 
            \multicolumn{2}{l}{(Matterport + Gibson)}            & 
            Mouse     & 
            Bullet      & 
            Navi       & 
            RGB, D, S, A           & 
            Force        & 
            Fetch, Franka, AlienGO           & 
            Fast, Navi only. Versions: 1.0, 2.0, Rearrangement \cite{DBLP:journals/corr/abs-2011-01975}   & 
            VC-1, PACT; Benchmark: OVMM \cite{DBLP:conf/corl/YenamandraRYWKG23} 
            \\ \hline
        
        ALFRED \cite{DBLP:conf/cvpr/ShridharTGBHMZF20} & 
            -/120               & 
            84/84              &  
            \NA   &   
            Unity    & 
            Navi, Mani &   
            RGB, D, S    & 
            PD        & 
            Human             & 
            Diverse long-horizon tasks &  
            (SL)$^3$, LLM-Planner     
            \\ \hline
        
        DMC \cite{DBLP:journals/corr/abs-1801-00690} & 
            1/-                 & 
            4/4                & 
            Code       & 
            MuJoCo      & 
            Control    & 
            RGB, D                   & 
            Force             & 
            \NA                                  & 
            Continuous RL                           & 
            VC-1, SMART                                     
            \\ \hline
        
        OpenAI Gym \cite{DBLP:journals/corr/BrockmanCPSSTZ16} & 
            1/-                 & 
            4/4                & 
            Code      & 
            MuJoCo      & 
            Control    & 
            RGB                      & 
            Force             & 
            \NA                                  & 
            Single agent RL environments            &  
            \NA                                              
            \\ \hline
        Genesis \cite{Genesis} & 
            \multicolumn{2}{p{2.3cm}}{(Rigid, deformable, liquid, etc.)}                 & 
            Code      & 
            (Custom)     & 
            Navi, Mani     & 
            RGB, D, S, N                      & 
            Force             & 
            Franka, Unitree, etc.                                 & 
            High-speed comprehensive physics simulation   &  
            \NA                                              
            \\
        \bottomrule
        \end{tabular}
    }
    \label{tb:simulator}
\end{table*}

\subsection{Real-World Robot Datasets \& Benchmarks}
\label{sec:real_datasets}

Embodied AI faces significant data scarcity issues because real-world robot data are not as readily available as language data. Collecting real-world robot datasets poses multiple challenges. First, the process is impeded by the cost and time required to procure robotic equipment, set up environments, and gather expert data through dedicated policies or human teleoperation. Second, the diverse types and configurations of robots introduce inconsistencies in factors such as sensory data, control modes, and gripper types. Finally, accurately capturing 6D object poses and reproducing setups remain elusive. We summarize recent embodied datasets in Table~\ref{tb:datasets}. In addition, real-world benchmarks are further complicated by the need for human evaluation.

\subsection{Simulators, Simulated Robot Datasets \& Benchmarks}
\label{sec:simulators}

Many researchers resort to simulated environments to circumvent real-world obstacles and scale the data collection process. We compare simulators and simulated benchmarks in Table~\ref{tb:simulator}. Nevertheless, this strategy presents its own challenges, chief among which is the sim-to-real gap. This discrepancy arises when models trained on simulated data exhibit poor performance during real-world deployment. The causes of this gap are multifaceted, encompassing unrealistic rendering quality, inaccuracies in physics simulations, and domain shifts characterized by object properties and robot motion planners. For instance, simulating non-rigid entities such as deformable objects or liquids presents significant difficulties. Moreover, importing new objects into simulators requires considerable effort, often involving techniques such as 3D scanning and mesh editing. Despite these hurdles, simulated environments provide automated evaluation metrics that aid researchers in consistently evaluating robotic models. Many benchmarks are based on simulators because they can precisely reproduce the experimental setup and yield fair comparisons of different models. Another technique, known as real-to-sim, can improve simulation fidelity, recreate failure cases, or facilitate digital twins.

\subsection{Automated Dataset Collection}
\label{sec:auto_data}

Several approaches advocate for automated dataset collection. RoboGen~\cite{DBLP:conf/icml/WangXCWWFEHG24} employs a generative simulation paradigm that proposes interesting skills, simulates corresponding environments, and selects optimal learning approaches to train policies for acquiring those skills. AutoRT~\cite{DBLP:journals/corr/abs-2401-12963} functions as a robot orchestrator driven by LLMs, generating tasks, filtering them by affordance, and utilizing either autonomous policies or human teleoperators to collect and evaluate data. DIAL~\cite{DBLP:conf/rss/XiaoCSWBHLT23} focuses on augmenting language instructions in existing datasets using VLMs. RoboPoint~\cite{DBLP:journals/corr/abs-2406-10721} generates scenes procedurally with randomized 3D layouts, objects, and camera viewpoints.

\subsection{Human Datasets}
\label{sec:human_datasets}

An alternative strategy to address data scarcity in real-world settings is to leverage human data. Human behavior offers plentiful guidance for robot policies due to its dexterity and diversity~\cite{DBLP:conf/naacl/MaCLSZK24}. However, this strategy also comes with inherent limitations. Capturing and transferring human hand and body motions to robot embodiments is difficult. Moreover, the inconsistency in human data poses a hurdle, as some data are egocentric while others are captured from third-person perspectives. In addition, filtering human data to extract useful information can be labor-intensive. These obstacles underscore the complexities involved in incorporating human data into robot learning processes. UMI~\cite{DBLP:conf/rss/ChiXPCB0TS24} proposes a method to mitigate these issues using hand-held grippers. For a more comprehensive comparison of human datasets, we refer interested readers to~\cite{DBLP:conf/nips/MoonSXJBBRRMBSP23}.

\subsection{Task Planning Benchmarks}
\label{sec:plan_benchmark}

EgoPlan-Bench~\cite{DBLP:journals/corr/abs-2312-06722} focuses on benchmarking real-world task planning with human annotations. PlanBench~\cite{DBLP:conf/nips/ValmeekamMHSK23, DBLP:conf/nips/ValmeekamMSK23} comprehensively assesses various aspects of task planning ability, such as cost optimality, plan verification, and replanning. LoTa-Bench~\cite{DBLP:journals/corr/abs-2402-08178} directly evaluates task planning by executing the generated plans in simulators and calculating success rates. Embodied agent interface (EAI)~\cite{DBLP:conf/nips/LiZWWZSGLLZLL0M24} argues that this approach fails to pinpoint issues in LLMs. By formalizing the input-output of LLM-based modules for decision-making tasks, EAI enables more fine-grained metrics beyond success rates.

\subsection{Embodied Question Answering (EQA) Benchmarks}
\label{sec:eqa}

EQA benchmarks, as summarized in Table~\ref{tb:eqa}, do not directly evaluate robotic tasks like manipulation and navigation, but instead target other relevant abilities for embodied AI, such as spatial reasoning, physics understanding, and world knowledge. EQA is akin to traditional VQA benchmarks, but differs in that the agent can actively explore the environment before providing an answer. EmbodiedQA~\cite{DBLP:conf/cvpr/DasDGLPB18} and IQUAD~\cite{DBLP:conf/cvpr/GordonKRRFF18} were among the first works to introduce this type of benchmark. MT-EQA~\cite{DBLP:conf/cvpr/YuCGBBB19} focuses on complex questions involving multiple targets. MP3D-EQA~\cite{DBLP:conf/cvpr/WijmansDMDGLEPB19} converts previous RGB inputs to point clouds, testing 3D perception capabilities. 

Active exploration requires access to a simulator, thereby restricting the ability to utilize real-world data. EgoVQA~\cite{DBLP:conf/iccvw/Fan19} shifts the focus of VQA to egocentric videos. EgoTaskQA~\cite{DBLP:conf/nips/JiaLZH22} emphasizes spatial, temporal, and causal relationship reasoning. EQA-MX~\cite{DBLP:conf/iclr/IslamGII24} investigates multimodal expressions (MXs), including regular verbal utterances and nonverbal gestures like eye gaze and pointing. OpenEQA~\cite{majumdar2023openeqa} evaluates seven main categories, including functional reasoning and world knowledge.

\begin{table*}
    \centering
    \caption{Embodied question answering benchmarks. Explore: active exploration} 
    \resizebox{\textwidth}{!}{%
        \begin{tabular}{
            >{\RaggedRight}p{2.5cm} 
            >{\RaggedRight}p{1.5cm} 
            >{\RaggedRight}p{1.75cm} 
            >{\RaggedRight}p{3cm} 
            >{\RaggedRight}p{3.85cm} 
            >{\RaggedRight}p{4cm} 
            >{\Centering}p{1cm}
            >{\RaggedRight}p{1.5cm}
            @{} 
        }
        \toprule
           \textbf{Benchmark} & \textbf{QA Pairs} & \textbf{Scenes} & \textbf{Source} & \textbf{Answer Type} & \textbf{Collection} & \textbf{Explore} & \textbf{Metrics}    
           \\ \midrule
            EQA \cite{DBLP:conf/cvpr/DasDGLPB18} & 
                5K & 750 envs &
                House3D simulator     & Answer set (172 answers)  & Template &
                \cmark   & Accuracy
                \\ \hline

            IQUAD \cite{DBLP:conf/cvpr/GordonKRRFF18} & 
                75K & 30 rooms & 
                AI2-THOR & Multiple choice & Template &
                \cmark     & Accuracy
                \\ \hline

            MT-EQA \cite{DBLP:conf/cvpr/YuCGBBB19} & 
                19.3K & 588 envs &
                House3D simulator     & Binary answer  & Template &
                \cmark     & Accuracy
                \\ \hline

            MP3D-EQA \cite{DBLP:conf/cvpr/WijmansDMDGLEPB19}  & 
                1,136 & 83 envs &
                MatterPort3D          & Answer set (53 answers)     & Template &
                \cmark     & Accuracy
                \\ \hline

            EgoVQA \cite{DBLP:conf/iccvw/Fan19} & 
                600 &    16 videos &
                IU Multi-view         & Multiple choice (1 out of 5) & Human annotators &
                \xmark      & Accuracy
                \\ \hline

            EgoTaskQA \cite{DBLP:conf/nips/JiaLZH22} & 
                40K & 2K videos & 
                LEMMA dataset & Open answer, binary verification & Human annotators \& template &
                \xmark      & Accuracy
                \\ \hline

            EQA-MX  \cite{DBLP:conf/iclr/IslamGII24} & 
                8.2M & 750K images & 
                CAESAR simulator      & Answer set & Question templates, answer set  & 
                \xmark & Accuracy
                \\ \hline

            OpenEQA \cite{majumdar2023openeqa}  & 
                557 + 1079 & 180 envs & 
                HM3D + ScanNet & Open answer & Human annotators & 
                \xmark & LLM Score      
                \\ 

            \bottomrule
    \end{tabular}
    }
    \label{tb:eqa}
\end{table*}
\section{Challenges and Future Directions}
\label{sec:future}

\paragraph{Safety First} Safety is paramount in robotics, as robots interact directly with the physical world. Ensuring the safety of robotic systems requires the integration of real-world commonsense. This involves the incorporation of robust safety guardrails, risk assessment frameworks, and human-robot interaction protocols. RLHF and ``evaluation without execution'' can also significantly lower safety risks~\cite{DBLP:conf/iros/HiranakaHLW00Z23}. Interpretability and explainability of VLA decision-making processes are also crucial for enhancing robot safety through error diagnosis and troubleshooting. 

\paragraph{Datasets \& Benchmarks} In addition to the issues discussed in Section~\ref{sec:data}, comprehensive benchmarks that cover a wide range of skills, objects, embodiments, and environments remain to be developed. Moreover, metrics beyond the success rate are needed for a fine-grained diagnosis of issues in VLA models, as highlighted by EAI~\cite{DBLP:conf/nips/LiZWWZSGLLZLL0M24} for LLMs. 

\paragraph{Foundation Models \& Generalization} VLA foundation models or robotic foundation models (RFMs) for embodied AI remain an open research topic primarily due to the diversity in embodiments, environments, and tasks. Many have made significant progress~\cite{liu2024rdt, DBLP:journals/corr/abs-2410-24164}, but still lack generalization capability on par with LLMs in NLP. Attaining such a level of generalization is very challenging, as it requires the development of many core AGI capabilities.

\paragraph{Multimodality} VLAs inherit many challenges associated with multimodal models, such as obtaining useful embeddings and aligning different modalities. Current approaches, like ImageBind~\cite{DBLP:conf/cvpr/GirdharELSAJM23} and LanguageBind~\cite{DBLP:conf/iclr/ZhuLNYCWPJZLZ0024}, align different modalities to the image or language embedding space, respectively. Within a unified embedding space, MLLMs can accommodate diverse modalities and become more general. However, whether focusing on embeddings alone is sufficient remains under debate. Although beyond the scope of VLAs, other modalities---such as audio~\cite{DBLP:conf/corl/ThankarajP23}, haptics~\cite{DBLP:conf/corl/GuzeyECP23}, and gaze~\cite{ijcai2020p689}---have proven useful for certain embodied AI applications. For instance, modeling human gaze data using an additional gaze network~\cite{DBLP:conf/eccv/ZhangLZWMHB18} or incorporating it via an auxiliary loss~\cite{DBLP:conf/atal/SaranZSN21} has been shown to enhance the performance of the primary policy network. These approaches demonstrate that RL policies can benefit from human-like visual attention, as human gaze often reveals the most salient locations in the environment~\cite{DBLP:conf/nips/GuoZLZBHS21}. While incorporating additional modalities is often advantageous, it inevitably introduces extra complexity into the model design.

\paragraph{Framework for Long-Horizon Tasks} The hierarchical framework is currently the most practical approach for long-horizon tasks. However, it increases system complexity and potential points of failure. Frequent task execution failures can trigger replanning, which can cause significant latency. Moreover, monolithic task planners share similar architectures with LVLAs, so employing two large models can be redundant and may hinder scalability. In addition, although modular task planners typically do not require training, they are not plug-and-play: language-based models may generate subtasks that control policies cannot execute, whereas code-based models require modules to be prewrapped in APIs manually. Therefore, developing a unified framework that directly translates long-horizon tasks into low-level control signals in an end-to-end fashion is worth exploring.

\paragraph{Real-Time Responsiveness} Unlike conversational AI, many robotic applications require real-time decision-making to respond to dynamic environments. If inference time cannot keep pace with environmental changes, the model may generate obsolete actions repeatedly. However, current VLA models---especially LVLAs and task planners---face a tradeoff between speed and capacity. Novel mechanisms are thus needed to strike an optimal balance.

\paragraph{Multiagent Systems} Cooperative multiagent systems offer benefits such as distributed perception and collaborative fault recovery. However, they also face challenges, including effective communication, coordinated dispatching, and fleet heterogeneity. In certain scenarios, individual agents may have conflicting goals, which further increases complexity.

\paragraph{Ethical and Societal Implications} Robotics has always raised various ethical, societal, and legal concerns. These include risks related to privacy, job displacement, decision-making bias, and the impact on social norms and human relationships. 

\paragraph{Applications} Most current VLAs focus on household or industrial settings, but a wider range of applications is possible, such as virtual assistants, autonomous vehicles, and agricultural robots. Various embodiments may also call for specialized VLAs, including dexterous hands, drones, quadruped robots, and humanoid robots. One particularly important field is healthcare, encompassing surgical robots~\cite{Le2016ASO} and care robots~\cite{DBLP:conf/ichi/LaursenPJSBAV22}. Healthcare demands higher safety and privacy standards and may necessitate novel techniques such as human-in-the-loop (HITL) control and federated learning. Moreover, due to the significant domain gap, specialized vision models might be needed for medical images.

\section{Conclusion}
VLA models hold immense promise for enabling embodied agents to interact with the physical world and fulfill user instructions. This article is the first survey to review LVLAs alongside generalized VLAs. Our taxonomy provides a high-level overview of three main lines of research: key components, control policies, and task planners. We meticulously analyze and compare their technical details, including model architectures, training strategies, and individual modules. In addition, we highlight essential resources for training and evaluating VLAs, such as datasets, simulators, and benchmarks. We hope this survey captures the rapidly evolving landscape of embodied AI and inspires future research.

\section*{Acknowledgments}
This work was supported by the Research Grants Council of the Hong Kong Special Administrative Region, China, CUHK 2300246 (RGC C1043-24G), CUHK 7010870, and CUHK 6906061.


\begin{thebibliography}{100}
\providecommand{\url}[1]{#1}
\csname url@samestyle\endcsname
\providecommand{\newblock}{\relax}
\providecommand{\bibinfo}[2]{#2}
\providecommand{\BIBentrySTDinterwordspacing}{\spaceskip=0pt\relax}
\providecommand{\BIBentryALTinterwordstretchfactor}{4}
\providecommand{\BIBentryALTinterwordspacing}{\spaceskip=\fontdimen2\font plus
\BIBentryALTinterwordstretchfactor\fontdimen3\font minus
  \fontdimen4\font\relax}
\providecommand{\BIBforeignlanguage}[2]{{%
\expandafter\ifx\csname l@#1\endcsname\relax
\typeout{** WARNING: IEEEtran.bst: No hyphenation pattern has been}%
\typeout{** loaded for the language `#1'. Using the pattern for}%
\typeout{** the default language instead.}%
\else
\language=\csname l@#1\endcsname
\fi
#2}}
\providecommand{\BIBdecl}{\relax}
\BIBdecl

\bibitem{DBLP:journals/corr/abs-2307-15818}
A.~Brohan, N.~Brown, J.~Carbajal, Y.~Chebotar, X.~Chen, K.~Choromanski,
  T.~Ding, D.~Driess, A.~Dubey, C.~Finn, P.~Florence, C.~Fu, M.~G. Arenas,
  K.~Gopalakrishnan, K.~Han, K.~Hausman, A.~Herzog, J.~Hsu, B.~Ichter,
  A.~Irpan, N.~J. Joshi, R.~Julian, D.~Kalashnikov, Y.~Kuang, I.~Leal, L.~Lee,
  T.~E. Lee, S.~Levine, Y.~Lu, H.~Michalewski, I.~Mordatch, K.~Pertsch, K.~Rao,
  K.~Reymann, M.~S. Ryoo, G.~Salazar, P.~Sanketi, P.~Sermanet, J.~Singh,
  A.~Singh, R.~Soricut, H.~T. Tran, V.~Vanhoucke, Q.~Vuong, A.~Wahid,
  S.~Welker, P.~Wohlhart, J.~Wu, F.~Xia, T.~Xiao, P.~Xu, S.~Xu, T.~Yu, and
  B.~Zitkovich, ``{RT-2:} vision-language-action models transfer web knowledge
  to robotic control,'' \emph{CoRR}, vol. abs/2307.15818, 2023.

\bibitem{DBLP:conf/nips/KrizhevskySH12}
A.~Krizhevsky, I.~Sutskever, and G.~E. Hinton, ``Imagenet classification with
  deep convolutional neural networks,'' in \emph{{NIPS}}, 2012, pp. 1106--1114.

\bibitem{DBLP:conf/nips/VaswaniSPUJGKP17}
A.~Vaswani, N.~Shazeer, N.~Parmar, J.~Uszkoreit, L.~Jones, A.~N. Gomez,
  L.~Kaiser, and I.~Polosukhin, ``Attention is all you need,'' in
  \emph{{NIPS}}, 2017, pp. 5998--6008.

\bibitem{DBLP:journals/nature/MnihKSRVBGRFOPB15}
V.~Mnih, K.~Kavukcuoglu, D.~Silver, A.~A. Rusu, J.~Veness, M.~G. Bellemare,
  A.~Graves, M.~A. Riedmiller, A.~Fidjeland, G.~Ostrovski, S.~Petersen,
  C.~Beattie, A.~Sadik, I.~Antonoglou, H.~King, D.~Kumaran, D.~Wierstra,
  S.~Legg, and D.~Hassabis, ``Human-level control through deep reinforcement
  learning,'' \emph{Nat.}, vol. 518, no. 7540, pp. 529--533, 2015.

\bibitem{DBLP:conf/iser/LevinePKQ16}
S.~Levine, P.~Pastor, A.~Krizhevsky, and D.~Quillen, ``Learning hand-eye
  coordination for robotic grasping with large-scale data collection,'' in
  \emph{{ISER}}, ser. Springer Proceedings in Advanced Robotics, vol.~1.\hskip
  1em plus 0.5em minus 0.4em\relax Springer, 2016, pp. 173--184.

\bibitem{DBLP:conf/nips/AlayracDLMBHLMM22}
J.~Alayrac, J.~Donahue, P.~Luc, A.~Miech, I.~Barr, Y.~Hasson, K.~Lenc,
  A.~Mensch, K.~Millican, M.~Reynolds, R.~Ring, E.~Rutherford, S.~Cabi, T.~Han,
  Z.~Gong, S.~Samangooei, M.~Monteiro, J.~L. Menick, S.~Borgeaud, A.~Brock,
  A.~Nematzadeh, S.~Sharifzadeh, M.~Binkowski, R.~Barreira, O.~Vinyals,
  A.~Zisserman, and K.~Simonyan, ``Flamingo: a visual language model for
  few-shot learning,'' in \emph{NeurIPS}, 2022.

\bibitem{DBLP:conf/icml/0008LSH23}
J.~Li, D.~Li, S.~Savarese, and S.~C.~H. Hoi, ``{BLIP-2:} bootstrapping
  language-image pre-training with frozen image encoders and large language
  models,'' in \emph{{ICML}}, vol. 202.\hskip 1em plus 0.5em minus 0.4em\relax
  {PMLR}, 2023, pp. 19\,730--19\,742.

\bibitem{DBLP:journals/corr/abs-2304-08485}
H.~Liu, C.~Li, Q.~Wu, and Y.~J. Lee, ``Visual instruction tuning,''
  \emph{CoRR}, vol. abs/2304.08485, 2023.

\bibitem{DBLP:conf/corl/HuangXXCLFZTMCS22}
W.~Huang, F.~Xia, T.~Xiao, H.~Chan, J.~Liang, P.~Florence, A.~Zeng, J.~Tompson,
  I.~Mordatch, Y.~Chebotar, P.~Sermanet, T.~Jackson, N.~Brown, L.~Luu,
  S.~Levine, K.~Hausman, and B.~Ichter, ``Inner monologue: Embodied reasoning
  through planning with language models,'' in \emph{CoRL}, vol. 205.\hskip 1em
  plus 0.5em minus 0.4em\relax {PMLR}, 2022, pp. 1769--1782.

\bibitem{DBLP:conf/corl/IchterBCFHHHIIJ22}
B.~Ichter, A.~Brohan, Y.~Chebotar, C.~Finn, K.~Hausman, A.~Herzog, D.~Ho,
  J.~Ibarz, A.~Irpan, E.~Jang, R.~Julian, D.~Kalashnikov, S.~Levine, Y.~Lu,
  C.~Parada, K.~Rao, P.~Sermanet, A.~Toshev, V.~Vanhoucke, F.~Xia, T.~Xiao,
  P.~Xu, M.~Yan, N.~Brown, M.~Ahn, O.~Cortes, N.~Sievers, C.~Tan, S.~Xu,
  D.~Reyes, J.~Rettinghouse, J.~Quiambao, P.~Pastor, L.~Luu, K.~Lee, Y.~Kuang,
  S.~Jesmonth, N.~J. Joshi, K.~Jeffrey, R.~J. Ruano, J.~Hsu, K.~Gopalakrishnan,
  B.~David, A.~Zeng, and C.~K. Fu, ``Do as {I} can, not as {I} say: Grounding
  language in robotic affordances,'' in \emph{CoRL}, vol. 205.\hskip 1em plus
  0.5em minus 0.4em\relax {PMLR}, 2022, pp. 287--318.

\bibitem{DBLP:conf/icml/DriessXSLCIWTVY23}
D.~Driess, F.~Xia, M.~S.~M. Sajjadi, C.~Lynch, A.~Chowdhery, B.~Ichter,
  A.~Wahid, J.~Tompson, Q.~Vuong, T.~Yu, W.~Huang, Y.~Chebotar, P.~Sermanet,
  D.~Duckworth, S.~Levine, V.~Vanhoucke, K.~Hausman, M.~Toussaint, K.~Greff,
  A.~Zeng, I.~Mordatch, and P.~Florence, ``Palm-e: An embodied multimodal
  language model,'' in \emph{{ICML}}, vol. 202.\hskip 1em plus 0.5em minus
  0.4em\relax {PMLR}, 2023, pp. 8469--8488.

\bibitem{DBLP:journals/corr/abs-2312-07843}
R.~Firoozi, J.~Tucker, S.~Tian, A.~Majumdar, J.~Sun, W.~Liu, Y.~Zhu, S.~Song,
  A.~Kapoor, K.~Hausman, B.~Ichter, D.~Driess, J.~Wu, C.~Lu, and M.~Schwager,
  ``Foundation models in robotics: Applications, challenges, and the future,''
  \emph{CoRR}, vol. abs/2312.07843, 2023.

\bibitem{DBLP:journals/corr/abs-2401-04334}
J.~Wang, Z.~Wu, Y.~Li, H.~Jiang, P.~Shu, E.~Shi, H.~Hu, C.~Ma, Y.~Liu, X.~Wang,
  Y.~Yao, X.~Liu, H.~Zhao, Z.~Liu, H.~Dai, L.~Zhao, B.~Ge, X.~Li, T.~Liu, and
  S.~Zhang, ``Large language models for robotics: Opportunities, challenges,
  and perspectives,'' \emph{CoRR}, vol. abs/2401.04334, 2024.

\bibitem{DBLP:journals/corr/abs-2312-08782}
Y.~Hu, Q.~Xie, V.~Jain, J.~Francis, J.~Patrikar, N.~V. Keetha, S.~Kim, Y.~Xie,
  T.~Zhang, S.~Zhao, Y.~Q. Chong, C.~Wang, K.~P. Sycara, M.~Johnson{-}Roberson,
  D.~Batra, X.~Wang, S.~A. Scherer, Z.~Kira, F.~Xia, and Y.~Bisk, ``Toward
  general-purpose robots via foundation models: {A} survey and meta-analysis,''
  \emph{CoRR}, vol. abs/2312.08782, 2023.

\bibitem{DBLP:journals/ar/KawaharazukaMGGPZ24}
K.~Kawaharazuka, T.~Matsushima, A.~Gambardella, J.~Guo, C.~Paxton, and A.~Zeng,
  ``Real-world robot applications of foundation models: a review,'' \emph{Adv.
  Robotics}, vol.~38, no.~18, pp. 1232--1254, 2024.

\bibitem{DBLP:conf/nips/ChenLRLGLASM21}
L.~Chen, K.~Lu, A.~Rajeswaran, K.~Lee, A.~Grover, M.~Laskin, P.~Abbeel,
  A.~Srinivas, and I.~Mordatch, ``Decision transformer: Reinforcement learning
  via sequence modeling,'' in \emph{NeurIPS}, 2021, pp. 15\,084--15\,097.

\bibitem{DBLP:conf/nips/JannerLL21}
M.~Janner, Q.~Li, and S.~Levine, ``Offline reinforcement learning as one big
  sequence modeling problem,'' in \emph{NeurIPS}, 2021, pp. 1273--1286.

\bibitem{DBLP:journals/tmlr/ReedZPCNBGSKSEBREHCHVBF22}
S.~E. Reed, K.~Zolna, E.~Parisotto, S.~G. Colmenarejo, A.~Novikov,
  G.~Barth{-}Maron, M.~Gimenez, Y.~Sulsky, J.~Kay, J.~T. Springenberg,
  T.~Eccles, J.~Bruce, A.~Razavi, A.~Edwards, N.~Heess, Y.~Chen, R.~Hadsell,
  O.~Vinyals, M.~Bordbar, and N.~de~Freitas, ``A generalist agent,''
  \emph{Trans. Mach. Learn. Res.}, vol. 2022, 2022.

\bibitem{DBLP:journals/corr/abs-2511-14759}
A.~A. Physical Intelligence~and, R.~Aniceto, A.~Balakrishna, K.~Black,
  K.~Conley, G.~Connors, J.~Darpinian, K.~Dhabalia, J.~DiCarlo, D.~Driess,
  M.~Equi, A.~Esmail, Y.~Fang, C.~Finn, C.~Glossop, T.~Godden, I.~Goryachev,
  L.~Groom, H.~Hancock, K.~Hausman, G.~Hussein, B.~Ichter, S.~Jakubczak,
  R.~Jen, T.~Jones, B.~Katz, L.~Ke, C.~Kuchi, M.~Lamb, D.~LeBlanc, S.~Levine,
  A.~Li{-}Bell, Y.~Lu, V.~Mano, M.~Mothukuri, S.~Nair, K.~Pertsch, A.~Z. Ren,
  C.~Sharma, L.~X. Shi, L.~Smith, J.~T. Springenberg, K.~Stachowicz,
  W.~Stoeckle, A.~Swerdlow, J.~Tanner, M.~Torne, Q.~Vuong, A.~Walling, H.~Wang,
  B.~Williams, S.~Yoo, L.~Yu, U.~Zhilinsky, and Z.~Zhou,
  ``{\(\pi\)}\({}^{\mbox{*}}\)\({}_{\mbox{0.6}}\): a {VLA} that learns, from
  experience,'' \emph{CoRR}, vol. abs/2511.14759, 2025.

\bibitem{DBLP:conf/iros/HiranakaHLW00Z23}
A.~Hiranaka, M.~Hwang, S.~Lee, C.~Wang, L.~Fei{-}Fei, J.~Wu, and R.~Zhang,
  ``Primitive skill-based robot learning from human evaluative feedback,'' in
  \emph{{IROS}}, 2023, pp. 7817--7824.

\bibitem{DBLP:conf/nips/ShinnCGNY23}
N.~Shinn, F.~Cassano, A.~Gopinath, K.~Narasimhan, and S.~Yao, ``Reflexion:
  language agents with verbal reinforcement learning,'' in \emph{NeurIPS},
  2023.

\bibitem{DBLP:conf/iclr/MaLWHBJZFA24}
Y.~J. Ma, W.~Liang, G.~Wang, D.~Huang, O.~Bastani, D.~Jayaraman, Y.~Zhu,
  L.~Fan, and A.~Anandkumar, ``Eureka: Human-level reward design via coding
  large language models,'' in \emph{{ICLR}}, 2024.

\bibitem{DBLP:conf/icml/RadfordKHRGASAM21}
A.~Radford, J.~W. Kim, C.~Hallacy, A.~Ramesh, G.~Goh, S.~Agarwal, G.~Sastry,
  A.~Askell, P.~Mishkin, J.~Clark, G.~Krueger, and I.~Sutskever, ``Learning
  transferable visual models from natural language supervision,'' in
  \emph{{ICML}}, vol. 139.\hskip 1em plus 0.5em minus 0.4em\relax {PMLR}, 2021,
  pp. 8748--8763.

\bibitem{DBLP:conf/corl/NairRKF022}
S.~Nair, A.~Rajeswaran, V.~Kumar, C.~Finn, and A.~Gupta, ``{R3M:} {A} universal
  visual representation for robot manipulation,'' in \emph{CoRL}, vol.
  205.\hskip 1em plus 0.5em minus 0.4em\relax {PMLR}, 2022, pp. 892--909.

\bibitem{DBLP:conf/iclr/MaSJBK023}
Y.~J. Ma, S.~Sodhani, D.~Jayaraman, O.~Bastani, V.~Kumar, and A.~Zhang,
  ``{VIP:} towards universal visual reward and representation via
  value-implicit pre-training,'' in \emph{{ICLR}}, 2023.

\bibitem{DBLP:conf/corl/RadosavovicXJAM22}
I.~Radosavovic, T.~Xiao, S.~James, P.~Abbeel, J.~Malik, and T.~Darrell,
  ``Real-world robot learning with masked visual pre-training,'' in
  \emph{CoRL}, vol. 205.\hskip 1em plus 0.5em minus 0.4em\relax {PMLR}, 2022,
  pp. 416--426.

\bibitem{DBLP:conf/naacl/DevlinCLT19}
J.~Devlin, M.~Chang, K.~Lee, and K.~Toutanova, ``{BERT:} pre-training of deep
  bidirectional transformers for language understanding,'' in \emph{{NAACL-HLT}
  {(1)}}.\hskip 1em plus 0.5em minus 0.4em\relax Association for Computational
  Linguistics, 2019, pp. 4171--4186.

\bibitem{DBLP:conf/corl/RadosavovicSFGD23}
I.~Radosavovic, B.~Shi, L.~Fu, K.~Goldberg, T.~Darrell, and J.~Malik, ``Robot
  learning with sensorimotor pre-training,'' in \emph{CoRL}, vol. 229.\hskip
  1em plus 0.5em minus 0.4em\relax {PMLR}, 2023, pp. 683--693.

\bibitem{DBLP:conf/corl/ShridharMF21}
M.~Shridhar, L.~Manuelli, and D.~Fox, ``Cliport: What and where pathways for
  robotic manipulation,'' in \emph{CoRL}, vol. 164.\hskip 1em plus 0.5em minus
  0.4em\relax {PMLR}, 2021, pp. 894--906.

\bibitem{DBLP:conf/cvpr/0001WMK22}
A.~Khandelwal, L.~Weihs, R.~Mottaghi, and A.~Kembhavi, ``Simple but effective:
  {CLIP} embeddings for embodied {AI},'' in \emph{{CVPR}}.\hskip 1em plus 0.5em
  minus 0.4em\relax {IEEE}, 2022, pp. 14\,809--14\,818.

\bibitem{DBLP:conf/cvpr/GadreWISS23}
S.~Y. Gadre, M.~Wortsman, G.~Ilharco, L.~Schmidt, and S.~Song, ``Cows on
  pasture: Baselines and benchmarks for language-driven zero-shot object
  navigation,'' in \emph{{CVPR}}.\hskip 1em plus 0.5em minus 0.4em\relax
  {IEEE}, 2023, pp. 23\,171--23\,181.

\bibitem{DBLP:conf/nips/MajumdarYAMCSJB23}
A.~Majumdar, K.~Yadav, S.~Arnaud, Y.~J. Ma, C.~Chen, S.~Silwal, A.~Jain,
  V.~Berges, T.~Wu, J.~Vakil, P.~Abbeel, J.~Malik, D.~Batra, Y.~Lin,
  O.~Maksymets, A.~Rajeswaran, and F.~Meier, ``Where are we in the search for
  an artificial visual cortex for embodied intelligence?'' in \emph{NeurIPS},
  2023.

\bibitem{DBLP:conf/rss/KaramchetiNCKFS23}
S.~Karamcheti, S.~Nair, A.~S. Chen, T.~Kollar, C.~Finn, D.~Sadigh, and
  P.~Liang, ``Language-driven representation learning for robotics,'' in
  \emph{RSS}, 2023.

\bibitem{DBLP:journals/tmlr/OquabDMVSKFHMEA24}
M.~Oquab, T.~Darcet, T.~Moutakanni, H.~V. Vo, M.~Szafraniec, V.~Khalidov,
  P.~Fernandez, D.~Haziza, F.~Massa, A.~El{-}Nouby, M.~Assran, N.~Ballas,
  W.~Galuba, R.~Howes, P.~Huang, S.~Li, I.~Misra, M.~Rabbat, V.~Sharma,
  G.~Synnaeve, H.~Xu, H.~J{\'{e}}gou, J.~Mairal, P.~Labatut, A.~Joulin, and
  P.~Bojanowski, ``Dinov2: Learning robust visual features without
  supervision,'' \emph{Trans. Mach. Learn. Res.}, vol. 2024, 2024.

\bibitem{DBLP:journals/corr/abs-2406-09246}
M.~J. Kim, K.~Pertsch, S.~Karamcheti, T.~Xiao, A.~Balakrishna, S.~Nair,
  R.~Rafailov, E.~P. Foster, G.~Lam, P.~Sanketi, Q.~Vuong, T.~Kollar,
  B.~Burchfiel, R.~Tedrake, D.~Sadigh, S.~Levine, P.~Liang, and C.~Finn,
  ``Openvla: An open-source vision-language-action model,'' \emph{CoRR}, vol.
  abs/2406.09246, 2024.

\bibitem{DBLP:journals/corr/abs-2409-01652}
W.~Huang, C.~Wang, Y.~Li, R.~Zhang, and L.~Fei{-}Fei, ``Rekep: Spatio-temporal
  reasoning of relational keypoint constraints for robotic manipulation,''
  \emph{CoRR}, vol. abs/2409.01652, 2024.

\bibitem{DBLP:conf/cvpr/AssranDMBVRLB23}
M.~Assran, Q.~Duval, I.~Misra, P.~Bojanowski, P.~Vincent, M.~G. Rabbat,
  Y.~LeCun, and N.~Ballas, ``Self-supervised learning from images with a
  joint-embedding predictive architecture,'' in \emph{{CVPR}}.\hskip 1em plus
  0.5em minus 0.4em\relax {IEEE}, 2023, pp. 15\,619--15\,629.

\bibitem{DBLP:conf/corl/ShangSMMKWH24}
J.~Shang, K.~Schmeckpeper, B.~B. May, M.~V. Minniti, T.~Kelestemur, D.~Watkins,
  and L.~Herlant, ``Theia: Distilling diverse vision foundation models for
  robot learning,'' in \emph{CoRL}, ser. Proceedings of Machine Learning
  Research, vol. 270.\hskip 1em plus 0.5em minus 0.4em\relax {PMLR}, 2024, pp.
  724--748.

\bibitem{DBLP:conf/icml/ParisiRP022}
S.~Parisi, A.~Rajeswaran, S.~Purushwalkam, and A.~Gupta, ``The unsurprising
  effectiveness of pre-trained vision models for control,'' in \emph{{ICML}},
  vol. 162.\hskip 1em plus 0.5em minus 0.4em\relax {PMLR}, 2022, pp.
  17\,359--17\,371.

\bibitem{LeCun2022APT}
\BIBentryALTinterwordspacing
Y.~LeCun, ``A path towards autonomous machine intelligence,'' 2022. [Online].
  Available: \url{https://openreview.net/pdf?id=BZ5a1r-kVsf}
\BIBentrySTDinterwordspacing

\bibitem{DBLP:conf/corl/ShenYYWKI23}
W.~Shen, G.~Yang, A.~Yu, J.~Wong, L.~P. Kaelbling, and P.~Isola, ``Distilled
  feature fields enable few-shot language-guided manipulation,'' in
  \emph{CoRL}, vol. 229.\hskip 1em plus 0.5em minus 0.4em\relax {PMLR}, 2023,
  pp. 405--424.

\bibitem{DBLP:conf/nips/HongZCZDCG23}
Y.~Hong, H.~Zhen, P.~Chen, S.~Zheng, Y.~Du, Z.~Chen, and C.~Gan, ``3d-llm:
  Injecting the 3d world into large language models,'' in \emph{NeurIPS}, 2023.

\bibitem{DBLP:journals/tog/KerblKLD23}
B.~Kerbl, G.~Kopanas, T.~Leimk{\"{u}}hler, and G.~Drettakis, ``3d gaussian
  splatting for real-time radiance field rendering,'' \emph{{ACM} Trans.
  Graph.}, vol.~42, no.~4, pp. 139:1--139:14, 2023.

\bibitem{DBLP:conf/cvpr/Qin0ZWP24}
M.~Qin, W.~Li, J.~Zhou, H.~Wang, and H.~Pfister, ``Langsplat: 3d language
  gaussian splatting,'' in \emph{{CVPR}}.\hskip 1em plus 0.5em minus
  0.4em\relax {IEEE}, 2024, pp. 20\,051--20\,060.

\bibitem{DBLP:conf/corl/ThankarajP23}
A.~Thankaraj and L.~Pinto, ``That sounds right: Auditory self-supervision for
  dynamic robot manipulation,'' in \emph{CoRL}, vol. 229.\hskip 1em plus 0.5em
  minus 0.4em\relax {PMLR}, 2023, pp. 1036--1049.

\bibitem{DBLP:conf/iros/JingZLSYFK23}
Y.~Jing, X.~Zhu, X.~Liu, Q.~Sima, T.~Yang, Y.~Feng, and T.~Kong, ``Exploring
  visual pre-training for robot manipulation: Datasets, models and methods,''
  in \emph{{IROS}}, 2023, pp. 11\,390--11\,395.

\bibitem{DBLP:conf/nips/Liu0GA22}
F.~Liu, H.~Liu, A.~Grover, and P.~Abbeel, ``Masked autoencoding for scalable
  and generalizable decision making,'' in \emph{NeurIPS}, 2022.

\bibitem{DBLP:conf/icml/LiGJ0HSSGW24}
J.~Li, Q.~Gao, M.~Johnston, X.~Gao, X.~He, H.~Shi, S.~Shakiah, R.~Ghanadan, and
  W.~Y. Wang, ``Mastering robot manipulation with multimodal prompts through
  pretraining and multi-task fine-tuning,'' in \emph{{ICML}}, 2024.

\bibitem{DBLP:conf/iclr/SunMMBHK23}
Y.~Sun, S.~Ma, R.~Madaan, R.~Bonatti, F.~Huang, and A.~Kapoor, ``{SMART:}
  self-supervised multi-task pretraining with control transformers,'' in
  \emph{{ICLR}}, 2023.

\bibitem{DBLP:conf/iros/BonattiVMFCK23}
R.~Bonatti, S.~Vemprala, S.~Ma, F.~Frujeri, S.~Chen, and A.~Kapoor, ``{PACT:}
  perception-action causal transformer for autoregressive robotics
  pre-training,'' in \emph{{IROS}}, 2023, pp. 3621--3627.

\bibitem{DBLP:conf/nips/BakerAZHTEHSC22}
B.~Baker, I.~Akkaya, P.~Zhokhov, J.~Huizinga, J.~Tang, A.~Ecoffet, B.~Houghton,
  R.~Sampedro, and J.~Clune, ``Video pretraining {(VPT):} learning to act by
  watching unlabeled online videos,'' in \emph{NeurIPS}, 2022.

\bibitem{DBLP:conf/iclr/WuJCCXLLLK24}
H.~Wu, Y.~Jing, C.~Cheang, G.~Chen, J.~Xu, X.~Li, M.~Liu, H.~Li, and T.~Kong,
  ``Unleashing large-scale video generative pre-training for visual robot
  manipulation,'' in \emph{{ICLR}}, 2024.

\bibitem{DBLP:conf/iclr/HafnerLB020}
D.~Hafner, T.~P. Lillicrap, J.~Ba, and M.~Norouzi, ``Dream to control: Learning
  behaviors by latent imagination,'' in \emph{{ICLR}}, 2020.

\bibitem{DBLP:conf/iclr/HafnerL0B21}
D.~Hafner, T.~P. Lillicrap, M.~Norouzi, and J.~Ba, ``Mastering atari with
  discrete world models,'' in \emph{{ICLR}}, 2021.

\bibitem{DBLP:journals/corr/abs-2301-04104}
D.~Hafner, J.~Pasukonis, J.~Ba, and T.~P. Lillicrap, ``Mastering diverse
  domains through world models,'' \emph{CoRR}, vol. abs/2301.04104, 2023.

\bibitem{DBLP:conf/corl/WuEHAG22}
P.~Wu, A.~Escontrela, D.~Hafner, P.~Abbeel, and K.~Goldberg, ``Daydreamer:
  World models for physical robot learning,'' in \emph{CoRL}, vol. 205.\hskip
  1em plus 0.5em minus 0.4em\relax {PMLR}, 2022, pp. 2226--2240.

\bibitem{DBLP:conf/iclr/MicheliAF23}
V.~Micheli, E.~Alonso, and F.~Fleuret, ``Transformers are sample-efficient
  world models,'' in \emph{{ICLR}}, 2023.

\bibitem{DBLP:conf/iclr/RobineHUH23}
J.~Robine, M.~H{\"{o}}ftmann, T.~Uelwer, and S.~Harmeling, ``Transformer-based
  world models are happy with 100k interactions,'' in \emph{{ICLR}}, 2023.

\bibitem{DBLP:conf/icml/NottinghamAS0H023}
K.~Nottingham, P.~Ammanabrolu, A.~Suhr, Y.~Choi, H.~Hajishirzi, S.~Singh, and
  R.~Fox, ``Do embodied agents dream of pixelated sheep: Embodied decision
  making using language guided world modelling,'' in \emph{{ICML}}, vol.
  202.\hskip 1em plus 0.5em minus 0.4em\relax {PMLR}, 2023, pp.
  26\,311--26\,325.

\bibitem{DBLP:conf/nips/SongZK23}
Z.~Song, Y.~Zhang, and I.~King, ``No change, no gain: Empowering graph neural
  networks with expected model change maximization for active learning,'' in
  \emph{NeurIPS}, 2023.

\bibitem{DBLP:conf/aaai/MaSHLZK23}
Y.~Ma, Z.~Song, X.~Hu, J.~Li, Y.~Zhang, and I.~King, ``Graph component
  contrastive learning for concept relatedness estimation,'' in
  \emph{{AAAI}}.\hskip 1em plus 0.5em minus 0.4em\relax {AAAI} Press, 2023, pp.
  13\,362--13\,370.

\bibitem{DBLP:conf/nips/GuanVSK23}
L.~Guan, K.~Valmeekam, S.~Sreedharan, and S.~Kambhampati, ``Leveraging
  pre-trained large language models to construct and utilize world models for
  model-based task planning,'' in \emph{NeurIPS}, 2023.

\bibitem{DBLP:journals/corr/abs-2304-11477}
B.~Liu, Y.~Jiang, X.~Zhang, Q.~Liu, S.~Zhang, J.~Biswas, and P.~Stone,
  ``{LLM+P:} empowering large language models with optimal planning
  proficiency,'' \emph{CoRR}, vol. abs/2304.11477, 2023.

\bibitem{DBLP:conf/emnlp/HaoGMHWWH23}
S.~Hao, Y.~Gu, H.~Ma, J.~J. Hong, Z.~Wang, D.~Z. Wang, and Z.~Hu, ``Reasoning
  with language model is planning with world model,'' in \emph{{EMNLP}}, 2023,
  pp. 8154--8173.

\bibitem{DBLP:conf/iclr/HuMYD0SC00024}
M.~Hu, Y.~Mu, X.~Yu, M.~Ding, S.~Wu, W.~Shao, Q.~Chen, B.~Wang, Y.~Qiao, and
  P.~Luo, ``Tree-planner: Efficient close-loop task planning with large
  language models,'' in \emph{{ICLR}}, 2024.

\bibitem{DBLP:conf/nips/ZhaoLH23}
Z.~Zhao, W.~S. Lee, and D.~Hsu, ``Large language models as commonsense
  knowledge for large-scale task planning,'' in \emph{NeurIPS}, 2023.

\bibitem{sora-world-model-openai}
\BIBentryALTinterwordspacing
OpenAI. (2024) Video generation models as world simulators. [Online].
  Available:
  \url{https://openai.com/index/video-generation-models-as-world-simulators/}
\BIBentrySTDinterwordspacing

\bibitem{DBLP:journals/corr/abs-2405-03520}
Z.~Zhu, X.~Wang, W.~Zhao, C.~Min, N.~Deng, M.~Dou, Y.~Wang, B.~Shi, K.~Wang,
  C.~Zhang, Y.~You, Z.~Zhang, D.~Zhao, L.~Xiao, J.~Zhao, J.~Lu, and G.~Huang,
  ``Is sora a world simulator? {A} comprehensive survey on general world models
  and beyond,'' \emph{CoRR}, vol. abs/2405.03520, 2024.

\bibitem{DBLP:conf/icml/BruceDEPS0LMSAA24}
J.~Bruce, M.~D. Dennis, A.~Edwards, J.~Parker{-}Holder, Y.~Shi, E.~Hughes,
  M.~Lai, A.~Mavalankar, R.~Steigerwald, C.~Apps, Y.~Aytar, S.~Bechtle,
  F.~M.~P. Behbahani, S.~C.~Y. Chan, N.~Heess, L.~Gonzalez, S.~Osindero,
  S.~Ozair, S.~E. Reed, J.~Zhang, K.~Zolna, J.~Clune, N.~de~Freitas, S.~Singh,
  and T.~Rockt{\"{a}}schel, ``Genie: Generative interactive environments,'' in
  \emph{{ICML}}, 2024.

\bibitem{DBLP:conf/icml/ZhenQCY0DHG24}
H.~Zhen, X.~Qiu, P.~Chen, J.~Yang, X.~Yan, Y.~Du, Y.~Hong, and C.~Gan,
  ``3d-vla: {A} 3d vision-language-action generative world model,'' in
  \emph{{ICML}}, 2024.

\bibitem{DBLP:conf/iclr/YangDGTKSA24}
S.~Yang, Y.~Du, S.~K.~S. Ghasemipour, J.~Tompson, L.~P. Kaelbling,
  D.~Schuurmans, and P.~Abbeel, ``Learning interactive real-world simulators,''
  in \emph{{ICLR}}, 2024.

\bibitem{DBLP:conf/nips/XiangTGSWYH23}
J.~Xiang, T.~Tao, Y.~Gu, T.~Shu, Z.~Wang, Z.~Yang, and Z.~Hu, ``Language models
  meet world models: Embodied experiences enhance language models,'' in
  \emph{NeurIPS}, 2023.

\bibitem{DBLP:conf/nips/KojimaGRMI22}
T.~Kojima, S.~S. Gu, M.~Reid, Y.~Matsuo, and Y.~Iwasawa, ``Large language
  models are zero-shot reasoners,'' in \emph{NeurIPS}, 2022.

\bibitem{DBLP:conf/nips/Wei0SBIXCLZ22}
J.~Wei, X.~Wang, D.~Schuurmans, M.~Bosma, B.~Ichter, F.~Xia, E.~H. Chi, Q.~V.
  Le, and D.~Zhou, ``Chain-of-thought prompting elicits reasoning in large
  language models,'' in \emph{NeurIPS}, 2022.

\bibitem{DBLP:journals/corr/abs-2312-07062}
G.~Lu, Z.~Wang, C.~Liu, J.~Lu, and Y.~Tang, ``Thinkbot: Embodied instruction
  following with thought chain reasoning,'' \emph{CoRR}, vol. abs/2312.07062,
  2023.

\bibitem{DBLP:conf/iclr/YaoZYDSN023}
S.~Yao, J.~Zhao, D.~Yu, N.~Du, I.~Shafran, K.~R. Narasimhan, and Y.~Cao,
  ``React: Synergizing reasoning and acting in language models,'' in
  \emph{{ICLR}}, 2023.

\bibitem{DBLP:journals/corr/abs-2403-05313}
Z.~Wang, A.~Liu, H.~Lin, J.~Li, X.~Ma, and Y.~Liang, ``{RAT:} retrieval
  augmented thoughts elicit context-aware reasoning in long-horizon
  generation,'' \emph{CoRR}, vol. abs/2403.05313, 2024.

\bibitem{DBLP:journals/corr/abs-2407-08693}
M.~Zawalski, W.~Chen, K.~Pertsch, O.~Mees, C.~Finn, and S.~Levine, ``Robotic
  control via embodied chain-of-thought reasoning,'' \emph{CoRR}, vol.
  abs/2407.08693, 2024.

\bibitem{DBLP:conf/cvpr/ZhaoLKFZWLMHFHL25}
Q.~Zhao, Y.~Lu, M.~J. Kim, Z.~Fu, Z.~Zhang, Y.~Wu, Z.~Li, Q.~Ma, S.~Han,
  C.~Finn, A.~Handa, T.~Lin, G.~Wetzstein, M.~Liu, and D.~Xiang, ``Cot-vla:
  Visual chain-of-thought reasoning for vision-language-action models,'' in
  \emph{{CVPR}}, 2025, pp. 1702--1713.

\bibitem{DBLP:conf/corl/NakamotoMKL24}
M.~Nakamoto, O.~Mees, A.~Kumar, and S.~Levine, ``Steering your generalists:
  Improving robotic foundation models via value guidance,'' in \emph{CoRL},
  ser. Proceedings of Machine Learning Research, vol. 270.\hskip 1em plus 0.5em
  minus 0.4em\relax {PMLR}, 2024, pp. 4996--5013.

\bibitem{DBLP:journals/corr/abs-2506-17811}
J.~Kwok, C.~Agia, R.~Sinha, M.~Foutter, S.~Li, I.~Stoica, A.~Mirhoseini, and
  M.~Pavone, ``Robomonkey: Scaling test-time sampling and verification for
  vision-language-action models,'' \emph{CoRR}, vol. abs/2506.17811, 2025.

\bibitem{DBLP:conf/corl/JangIKKELLF21}
E.~Jang, A.~Irpan, M.~Khansari, D.~Kappler, F.~Ebert, C.~Lynch, S.~Levine, and
  C.~Finn, ``{BC-Z:} zero-shot task generalization with robotic imitation
  learning,'' in \emph{CoRL}, vol. 164.\hskip 1em plus 0.5em minus 0.4em\relax
  {PMLR}, 2021, pp. 991--1002.

\bibitem{DBLP:conf/rss/LynchS21}
C.~Lynch and P.~Sermanet, ``Language conditioned imitation learning over
  unstructured data,'' in \emph{RSS}, 2021.

\bibitem{DBLP:journals/ral/MeesHB22}
O.~Mees, L.~Hermann, and W.~Burgard, ``What matters in language conditioned
  robotic imitation learning over unstructured data,'' \emph{{IEEE} Robotics
  Autom. Lett.}, vol.~7, no.~4, pp. 11\,205--11\,212, 2022.

\bibitem{DBLP:conf/icra/MeesBB23}
O.~Mees, J.~Borja{-}Diaz, and W.~Burgard, ``Grounding language with visual
  affordances over unstructured data,'' in \emph{{ICRA}}.\hskip 1em plus 0.5em
  minus 0.4em\relax {IEEE}, 2023, pp. 11\,576--11\,582.

\bibitem{DBLP:conf/nips/DuY0DN0SA23}
Y.~Du, S.~Yang, B.~Dai, H.~Dai, O.~Nachum, J.~Tenenbaum, D.~Schuurmans, and
  P.~Abbeel, ``Learning universal policies via text-guided video generation,''
  in \emph{NeurIPS}, 2023.

\bibitem{DBLP:conf/rss/SharmaSBPH0AF22}
P.~Sharma, B.~Sundaralingam, V.~Blukis, C.~Paxton, T.~Hermans, A.~Torralba,
  J.~Andreas, and D.~Fox, ``Correcting robot plans with natural language
  feedback,'' in \emph{RSS}, 2022.

\bibitem{DBLP:journals/corr/abs-2210-06407}
C.~Lynch, A.~Wahid, J.~Tompson, T.~Ding, J.~Betker, R.~Baruch, T.~Armstrong,
  and P.~Florence, ``Interactive language: Talking to robots in real time,''
  \emph{CoRR}, vol. abs/2210.06407, 2022.

\bibitem{DBLP:conf/corl/GuhurCPTLS22}
P.~Guhur, S.~Chen, R.~G. Pinel, M.~Tapaswi, I.~Laptev, and C.~Schmid,
  ``Instruction-driven history-aware policies for robotic manipulations,'' in
  \emph{CoRL}, vol. 205.\hskip 1em plus 0.5em minus 0.4em\relax {PMLR}, 2022,
  pp. 175--187.

\bibitem{DBLP:conf/corl/ShridharMF22}
M.~Shridhar, L.~Manuelli, and D.~Fox, ``Perceiver-actor: {A} multi-task
  transformer for robotic manipulation,'' in \emph{CoRL}, vol. 205.\hskip 1em
  plus 0.5em minus 0.4em\relax {PMLR}, 2022, pp. 785--799.

\bibitem{DBLP:conf/corl/GervetXGF23}
T.~Gervet, Z.~Xian, N.~Gkanatsios, and K.~Fragkiadaki, ``Act3d: 3d feature
  field transformers for multi-task robotic manipulation,'' in \emph{CoRL},
  vol. 229.\hskip 1em plus 0.5em minus 0.4em\relax {PMLR}, 2023, pp.
  3949--3965.

\bibitem{DBLP:conf/corl/GoyalXGBCF23}
A.~Goyal, J.~Xu, Y.~Guo, V.~Blukis, Y.~Chao, and D.~Fox, ``{RVT:} robotic view
  transformer for 3d object manipulation,'' in \emph{CoRL}, vol. 229.\hskip 1em
  plus 0.5em minus 0.4em\relax {PMLR}, 2023, pp. 694--710.

\bibitem{DBLP:journals/corr/abs-2406-08545}
A.~Goyal, V.~Blukis, J.~Xu, Y.~Guo, Y.~Chao, and D.~Fox, ``{RVT-2:} learning
  precise manipulation from few demonstrations,'' \emph{CoRR}, vol.
  abs/2406.08545, 2024.

\bibitem{DBLP:journals/corr/abs-2406-10721}
W.~Yuan, J.~Duan, V.~Blukis, W.~Pumacay, R.~Krishna, A.~Murali, A.~Mousavian,
  and D.~Fox, ``Robopoint: {A} vision-language model for spatial affordance
  prediction for robotics,'' \emph{CoRR}, vol. abs/2406.10721, 2024.

\bibitem{DBLP:journals/corr/abs-2306-11706}
K.~Bousmalis, G.~Vezzani, D.~Rao, C.~Devin, A.~X. Lee, M.~Bauz{\'{a}},
  T.~Davchev, Y.~Zhou, A.~Gupta, A.~Raju, A.~Laurens, C.~Fantacci, V.~Dalibard,
  M.~Zambelli, M.~F. Martins, R.~Pevceviciute, M.~Blokzijl, M.~Denil,
  N.~Batchelor, T.~Lampe, E.~Parisotto, K.~Zolna, S.~E. Reed, S.~G.
  Colmenarejo, J.~Scholz, A.~Abdolmaleki, O.~Groth, J.~Regli, O.~Sushkov,
  T.~Roth{\"{o}}rl, J.~E. Chen, Y.~Aytar, D.~Barker, J.~Ortiz, M.~A.
  Riedmiller, J.~T. Springenberg, R.~Hadsell, F.~Nori, and N.~Heess, ``Robocat:
  {A} self-improving foundation agent for robotic manipulation,'' \emph{CoRR},
  vol. abs/2306.11706, 2023.

\bibitem{DBLP:journals/corr/abs-2303-00905}
A.~Stone, T.~Xiao, Y.~Lu, K.~Gopalakrishnan, K.~Lee, Q.~Vuong, P.~Wohlhart,
  B.~Zitkovich, F.~Xia, C.~Finn, and K.~Hausman, ``Open-world object
  manipulation using pre-trained vision-language models,'' \emph{CoRR}, vol.
  abs/2303.00905, 2023.

\bibitem{DBLP:conf/rss/BrohanBCCDFGHHH23}
A.~Brohan, N.~Brown, J.~Carbajal, Y.~Chebotar, J.~Dabis, C.~Finn,
  K.~Gopalakrishnan, K.~Hausman, A.~Herzog, J.~Hsu, J.~Ibarz, B.~Ichter,
  A.~Irpan, T.~Jackson, S.~Jesmonth, N.~J. Joshi, R.~Julian, D.~Kalashnikov,
  Y.~Kuang, I.~Leal, K.~Lee, S.~Levine, Y.~Lu, U.~Malla, D.~Manjunath,
  I.~Mordatch, O.~Nachum, C.~Parada, J.~Peralta, E.~Perez, K.~Pertsch,
  J.~Quiambao, K.~Rao, M.~S. Ryoo, G.~Salazar, P.~R. Sanketi, K.~Sayed,
  J.~Singh, S.~Sontakke, A.~Stone, C.~Tan, H.~T. Tran, V.~Vanhoucke, S.~Vega,
  Q.~Vuong, F.~Xia, T.~Xiao, P.~Xu, S.~Xu, T.~Yu, and B.~Zitkovich, ``{RT-1:}
  robotics transformer for real-world control at scale,'' in \emph{RSS}, 2023.

\bibitem{DBLP:journals/corr/abs-2309-10150}
Y.~Chebotar, Q.~Vuong, A.~Irpan, K.~Hausman, F.~Xia, Y.~Lu, A.~Kumar, T.~Yu,
  A.~Herzog, K.~Pertsch, K.~Gopalakrishnan, J.~Ibarz, O.~Nachum, S.~Sontakke,
  G.~Salazar, H.~T. Tran, J.~Peralta, C.~Tan, D.~Manjunath, J.~Singh,
  B.~Zitkovich, T.~Jackson, K.~Rao, C.~Finn, and S.~Levine, ``Q-transformer:
  Scalable offline reinforcement learning via autoregressive q-functions,''
  \emph{CoRR}, vol. abs/2309.10150, 2023.

\bibitem{DBLP:journals/corr/abs-2311-01977}
J.~Gu, S.~Kirmani, P.~Wohlhart, Y.~Lu, M.~G. Arenas, K.~Rao, W.~Yu, C.~Fu,
  K.~Gopalakrishnan, Z.~Xu, P.~Sundaresan, P.~Xu, H.~Su, K.~Hausman, C.~Finn,
  Q.~Vuong, and T.~Xiao, ``Rt-trajectory: Robotic task generalization via
  hindsight trajectory sketches,'' \emph{CoRR}, vol. abs/2311.01977, 2023.

\bibitem{DBLP:conf/rss/ZhaoKLF23}
T.~Z. Zhao, V.~Kumar, S.~Levine, and C.~Finn, ``Learning fine-grained bimanual
  manipulation with low-cost hardware,'' in \emph{RSS}, 2023.

\bibitem{DBLP:conf/icra/BharadhwajVSGTK24}
H.~Bharadhwaj, J.~Vakil, M.~Sharma, A.~Gupta, S.~Tulsiani, and V.~Kumar,
  ``Roboagent: Generalization and efficiency in robot manipulation via semantic
  augmentations and action chunking,'' in \emph{{ICRA}}.\hskip 1em plus 0.5em
  minus 0.4em\relax {IEEE}, 2024, pp. 4788--4795.

\bibitem{DBLP:journals/corr/abs-2311-01378}
X.~Li, M.~Liu, H.~Zhang, C.~Yu, J.~Xu, H.~Wu, C.~Cheang, Y.~Jing, W.~Zhang,
  H.~Liu, H.~Li, and T.~Kong, ``Vision-language foundation models as effective
  robot imitators,'' \emph{CoRR}, vol. abs/2311.01378, 2023.

\bibitem{DBLP:journals/corr/abs-2406-18977}
F.~Liu, F.~Yan, L.~Zheng, C.~Feng, Y.~Huang, and L.~Ma, ``Robouniview:
  Visual-language model with unified view representation for robotic
  manipulaiton,'' \emph{CoRR}, vol. abs/2406.18977, 2024.

\bibitem{DBLP:conf/nips/YueWKHWSF024}
Y.~Yue, Y.~Wang, B.~Kang, Y.~Han, S.~Wang, S.~Song, J.~Feng, and G.~Huang,
  ``Deer-vla: Dynamic inference of multimodal large language models for
  efficient robot execution,'' in \emph{NeurIPS}, 2024.

\bibitem{DBLP:journals/corr/abs-2305-11176}
S.~Huang, Z.~Jiang, H.~Dong, Y.~Qiao, P.~Gao, and H.~Li, ``Instruct2act:
  Mapping multi-modality instructions to robotic actions with large language
  model,'' \emph{CoRR}, vol. abs/2305.11176, 2023.

\bibitem{DBLP:journals/corr/abs-2307-05973}
W.~Huang, C.~Wang, R.~Zhang, Y.~Li, J.~Wu, and L.~Fei{-}Fei, ``Voxposer:
  Composable 3d value maps for robotic manipulation with language models,''
  \emph{CoRR}, vol. abs/2307.05973, 2023.

\bibitem{DBLP:conf/rss/ChiFDXCBS23}
C.~Chi, S.~Feng, Y.~Du, Z.~Xu, E.~Cousineau, B.~Burchfiel, and S.~Song,
  ``Diffusion policy: Visuomotor policy learning via action diffusion,'' in
  \emph{RSS}, 2023.

\bibitem{DBLP:conf/rss/ZeZZHWX24}
Y.~Ze, G.~Zhang, K.~Zhang, C.~Hu, M.~Wang, and H.~Xu, ``3d diffusion policy:
  Generalizable visuomotor policy learning via simple 3d representations,'' in
  \emph{RSS}, 2024.

\bibitem{DBLP:conf/corl/HaFS23}
H.~Ha, P.~Florence, and S.~Song, ``Scaling up and distilling down:
  Language-guided robot skill acquisition,'' in \emph{CoRL}, vol. 229.\hskip
  1em plus 0.5em minus 0.4em\relax {PMLR}, 2023, pp. 3766--3777.

\bibitem{DBLP:conf/rss/GhoshWPBMDHK0LT24}
D.~Ghosh, H.~R. Walke, K.~Pertsch, K.~Black, O.~Mees, S.~Dasari, J.~Hejna,
  T.~Kreiman, C.~Xu, J.~Luo, Y.~L. Tan, L.~Y. Chen, Q.~Vuong, T.~Xiao, P.~R.
  Sanketi, D.~Sadigh, C.~Finn, and S.~Levine, ``Octo: An open-source generalist
  robot policy,'' in \emph{RSS}, 2024.

\bibitem{DBLP:journals/corr/abs-2402-10885}
T.~Ke, N.~Gkanatsios, and K.~Fragkiadaki, ``3d diffuser actor: Policy diffusion
  with 3d scene representations,'' \emph{CoRR}, vol. abs/2402.10885, 2024.

\bibitem{DBLP:journals/corr/abs-2407-05996}
M.~Reuss, {\"{O}}.~E. Yagmurlu, F.~Wenzel, and R.~Lioutikov, ``Multimodal
  diffusion transformer: Learning versatile behavior from multimodal goals,''
  \emph{CoRR}, vol. abs/2407.05996, 2024.

\bibitem{liu2024rdt}
S.~Liu, L.~Wu, B.~Li, H.~Tan, H.~Chen, Z.~Wang, K.~Xu, H.~Su, and J.~Zhu,
  ``Rdt-1b: a diffusion foundation model for bimanual manipulation,''
  \emph{arXiv preprint arXiv:2410.07864}, 2024.

\bibitem{DBLP:journals/corr/abs-2403-01823}
S.~Belkhale, T.~Ding, T.~Xiao, P.~Sermanet, Q.~Vuong, J.~Tompson, Y.~Chebotar,
  D.~Dwibedi, and D.~Sadigh, ``{RT-H:} action hierarchies using language,''
  \emph{CoRR}, vol. abs/2403.01823, 2024.

\bibitem{DBLP:journals/corr/abs-2310-08864}
O.~X. Collaboration, A.~Padalkar, A.~Pooley, A.~Jain, A.~Bewley, A.~Herzog,
  A.~Irpan, A.~Khazatsky, A.~Raj, A.~Singh, A.~Brohan, A.~Raffin, A.~Wahid,
  B.~Burgess{-}Limerick, B.~Kim, B.~Sch{\"{o}}lkopf, B.~Ichter, C.~Lu, C.~Xu,
  C.~Finn, C.~Xu, C.~Chi, C.~Huang, C.~Chan, C.~Pan, C.~Fu, C.~Devin,
  D.~Driess, D.~Pathak, D.~Shah, D.~B{\"{u}}chler, D.~Kalashnikov, D.~Sadigh,
  E.~Johns, F.~Ceola, F.~Xia, F.~Stulp, G.~Zhou, G.~S. Sukhatme, G.~Salhotra,
  G.~Yan, G.~Schiavi, G.~Kahn, H.~Su, H.~Fang, H.~Shi, H.~B. Amor, H.~I.
  Christensen, H.~Furuta, H.~Walke, H.~Fang, I.~Mordatch, I.~Radosavovic, and
  et~al., ``Open x-embodiment: Robotic learning datasets and {RT-X} models,''
  \emph{CoRR}, vol. abs/2310.08864, 2023.

\bibitem{DBLP:journals/corr/abs-2502-19645}
M.~J. Kim, C.~Finn, and P.~Liang, ``Fine-tuning vision-language-action models:
  Optimizing speed and success,'' \emph{CoRR}, vol. abs/2502.19645, 2025.

\bibitem{DBLP:conf/iclr/ZhengLH0DKHY25}
R.~Zheng, Y.~Liang, S.~Huang, J.~Gao, H.~D. III, A.~Kolobov, F.~Huang, and
  J.~Yang, ``Tracevla: Visual trace prompting enhances spatial-temporal
  awareness for generalist robotic policies,'' in \emph{{ICLR}}, 2025.

\bibitem{DBLP:journals/corr/abs-2410-24164}
K.~Black, N.~Brown, D.~Driess, A.~Esmail, M.~Equi, C.~Finn, N.~Fusai, L.~Groom,
  K.~Hausman, B.~Ichter, S.~Jakubczak, T.~Jones, L.~Ke, S.~Levine,
  A.~Li{-}Bell, M.~Mothukuri, S.~Nair, K.~Pertsch, L.~X. Shi, J.~Tanner,
  Q.~Vuong, A.~Walling, H.~Wang, and U.~Zhilinsky,
  ``{\(\pi\)}\({}_{\mbox{0}}\): {A} vision-language-action flow model for
  general robot control,'' \emph{CoRR}, vol. abs/2410.24164, 2024.

\bibitem{DBLP:conf/nips/LiuLWALZYZGZ24}
J.~Liu, M.~Liu, Z.~Wang, P.~An, X.~Li, K.~Zhou, S.~Yang, R.~Zhang, Y.~Guo, and
  S.~Zhang, ``Robomamba: Efficient vision-language-action model for robotic
  reasoning and manipulation,'' in \emph{NeurIPS}, 2024.

\bibitem{DBLP:journals/corr/abs-2501-15830}
D.~Qu, H.~Song, Q.~Chen, Y.~Yao, X.~Ye, Y.~Ding, Z.~Wang, J.~Gu, B.~Zhao,
  D.~Wang, and X.~Li, ``Spatialvla: Exploring spatial representations for
  visual-language-action model,'' \emph{CoRR}, vol. abs/2501.15830, 2025.

\bibitem{DBLP:journals/corr/abs-2409-12514}
J.~Wen, Y.~Zhu, J.~Li, M.~Zhu, K.~Wu, Z.~Xu, N.~Liu, R.~Cheng, C.~Shen,
  Y.~Peng, F.~Feng, and J.~Tang, ``Tinyvla: Towards fast, data-efficient
  vision-language-action models for robotic manipulation,'' \emph{CoRR}, vol.
  abs/2409.12514, 2024.

\bibitem{DBLP:journals/corr/abs-2411-19650}
Q.~Li, Y.~Liang, Z.~Wang, L.~Luo, X.~Chen, M.~Liao, F.~Wei, Y.~Deng, S.~Xu,
  Y.~Zhang, X.~Wang, B.~Liu, J.~Fu, J.~Bao, D.~Chen, Y.~Shi, J.~Yang, and
  B.~Guo, ``Cogact: {A} foundational vision-language-action model for
  synergizing cognition and action in robotic manipulation,'' \emph{CoRR}, vol.
  abs/2411.19650, 2024.

\bibitem{DBLP:journals/corr/abs-2502-05855}
J.~Wen, Y.~Zhu, J.~Li, Z.~Tang, C.~Shen, and F.~Feng, ``Dexvla: Vision-language
  model with plug-in diffusion expert for general robot control,'' \emph{CoRR},
  vol. abs/2502.05855, 2025.

\bibitem{DBLP:journals/corr/abs-2503-10631}
J.~Liu, H.~Chen, P.~An, Z.~Liu, R.~Zhang, C.~Gu, X.~Li, Z.~Guo, S.~Chen,
  M.~Liu, C.~Hou, M.~Zhao, K.~alex Zhou, P.~Heng, and S.~Zhang, ``Hybridvla:
  Collaborative diffusion and autoregression in a unified
  vision-language-action model,'' \emph{CoRR}, vol. abs/2503.10631, 2025.

\bibitem{DBLP:conf/iclr/YeJJJYPMTCLLL0Z25}
S.~Ye, J.~Jang, B.~Jeon, S.~J. Joo, J.~Yang, B.~Peng, A.~Mandlekar, R.~Tan,
  Y.~Chao, B.~Y. Lin, L.~Liden, K.~Lee, J.~Gao, L.~Zettlemoyer, D.~Fox, and
  M.~Seo, ``Latent action pretraining from videos,'' in \emph{{ICLR}}, 2025.

\bibitem{DBLP:journals/corr/abs-2506-21539}
J.~Cen, C.~Yu, H.~Yuan, Y.~Jiang, S.~Huang, J.~Guo, X.~Li, Y.~Song, H.~Luo,
  F.~Wang, D.~Zhao, and H.~Chen, ``Worldvla: Towards autoregressive action
  world model,'' \emph{CoRR}, vol. abs/2506.21539, 2025.

\bibitem{DBLP:journals/corr/abs-2506-19850}
Y.~Wang, X.~Li, W.~Wang, J.~Zhang, Y.~Li, Y.~Chen, X.~Wang, and Z.~Zhang,
  ``Unified vision-language-action model,'' \emph{CoRR}, vol. abs/2506.19850,
  2025.

\bibitem{ma-etal-2025-astra}
Y.~Ma, D.~Chi, S.~Wu, Y.~Liu, Y.~Zhuang, and I.~King, ``Astra: Efficient
  transformer architecture and contrastive dynamics learning for embodied
  instruction following,'' in \emph{EMNLP}, 2025.

\bibitem{DBLP:journals/fcsc/QuDWCWYXW25}
C.~Qu, S.~Dai, X.~Wei, H.~Cai, S.~Wang, D.~Yin, J.~Xu, and J.~Wen, ``Tool
  learning with large language models: a survey,'' \emph{Frontiers Comput.
  Sci.}, vol.~19, no.~8, p. 198343, 2025.

\bibitem{DBLP:journals/corr/abs-2210-03094}
Y.~Jiang, A.~Gupta, Z.~Zhang, G.~Wang, Y.~Dou, Y.~Chen, L.~Fei{-}Fei,
  A.~Anandkumar, Y.~Zhu, and L.~Fan, ``{VIMA:} general robot manipulation with
  multimodal prompts,'' \emph{CoRR}, vol. abs/2210.03094, 2022.

\bibitem{DBLP:conf/cvpr/LiuWY24}
R.~Liu, W.~Wang, and Y.~Yang, ``Volumetric environment representation for
  vision-language navigation,'' in \emph{{CVPR}}.\hskip 1em plus 0.5em minus
  0.4em\relax {IEEE}, 2024, pp. 16\,317--16\,328.

\bibitem{DBLP:journals/corr/abs-2501-16698}
Y.~Ma and I.~King, ``3d-moe: {A} mixture-of-experts multi-modal {LLM} for 3d
  vision and pose diffusion via rectified flow,'' \emph{CoRR}, vol.
  abs/2501.16698, 2025.

\bibitem{DBLP:conf/icra/VecerikD0DAZHAS24}
M.~Vecer{\'{\i}}k, C.~Doersch, Y.~Yang, T.~Davchev, Y.~Aytar, G.~Zhou,
  R.~Hadsell, L.~Agapito, and J.~Scholz, ``Robotap: Tracking arbitrary points
  for few-shot visual imitation,'' in \emph{{ICRA}}.\hskip 1em plus 0.5em minus
  0.4em\relax {IEEE}, 2024, pp. 5397--5403.

\bibitem{DBLP:conf/icml/NasirianyX0XL0X24}
S.~Nasiriany, F.~Xia, W.~Yu, T.~Xiao, J.~Liang, I.~Dasgupta, A.~Xie, D.~Driess,
  A.~Wahid, Z.~Xu, Q.~Vuong, T.~Zhang, T.~E. Lee, K.~Lee, P.~Xu, S.~Kirmani,
  Y.~Zhu, A.~Zeng, K.~Hausman, N.~Heess, C.~Finn, S.~Levine, and B.~Ichter,
  ``{PIVOT:} iterative visual prompting elicits actionable knowledge for
  vlms,'' in \emph{{ICML}}, 2024.

\bibitem{DBLP:journals/corr/abs-2503-14734}
J.~Bjorck, F.~Casta{\~{n}}eda, N.~Cherniadev, X.~Da, R.~Ding, Linxi, Y.~Fang,
  D.~Fox, F.~Hu, S.~Huang, J.~Jang, Z.~Jiang, J.~Kautz, K.~Kundalia, L.~Lao,
  Z.~Li, Z.~Lin, K.~Lin, G.~Liu, E.~LLontop, L.~Magne, A.~Mandlekar,
  A.~Narayan, S.~Nasiriany, S.~Reed, Y.~L. Tan, G.~Wang, Z.~Wang, J.~Wang,
  Q.~Wang, J.~Xiang, Y.~Xie, Y.~Xu, Z.~Xu, S.~Ye, Z.~Yu, A.~Zhang, H.~Zhang,
  Y.~Zhao, R.~Zheng, and Y.~Zhu, ``{GR00T} {N1:} an open foundation model for
  generalist humanoid robots,'' \emph{CoRR}, vol. abs/2503.14734, 2025.

\bibitem{DBLP:journals/corr/abs-2511-14659}
N.~M. Chia{-}Yu Hung~and, H.~Deng, L.~Renhang, Y.~Ang, A.~Zadeh, C.~Li,
  D.~Herremans, Z.~Wang, and S.~Poria, ``{NORA-1.5:} {A} vision-language-action
  model trained using world model-, and action-based preference rewards,''
  \emph{CoRR}, vol. abs/2511.14659, 2025.

\bibitem{DBLP:journals/corr/abs-2508-05635}
P.~Z. Yue Liao~and, S.~Huang, D.~Yang, S.~Chen, Y.~Jiang, H.~Yue, J.~Cai,
  S.~Liu, J.~Luo, L.~Chen, S.~Yan, M.~Yao, and G.~Ren, ``Genie envisioner: {A}
  unified world foundation platform for robotic, manipulation,'' \emph{CoRR},
  vol. abs/2508.05635, 2025.

\bibitem{DBLP:conf/nips/TianJYPW24}
K.~Tian, Y.~Jiang, Z.~Yuan, B.~Peng, and L.~Wang, ``Visual autoregressive
  modeling: Scalable image generation via next-scale prediction,'' in
  \emph{NeurIPS}, 2024.

\bibitem{DBLP:journals/corr/abs-2501-12327}
X.~Zhuang, Y.~Xie, Y.~Deng, L.~Liang, J.~Ru, Y.~Yin, and Y.~Zou, ``{VARGPT:}
  unified understanding and generation in a visual autoregressive multimodal
  large language model,'' \emph{CoRR}, vol. abs/2501.12327, 2025.

\bibitem{DBLP:journals/corr/abs-2305-15021}
Y.~Mu, Q.~Zhang, M.~Hu, W.~Wang, M.~Ding, J.~Jin, B.~Wang, J.~Dai, Y.~Qiao, and
  P.~Luo, ``Embodiedgpt: Vision-language pre-training via embodied chain of
  thought,'' \emph{CoRR}, vol. abs/2305.15021, 2023.

\bibitem{DBLP:conf/icml/HuangYMLLW0ZJ024}
J.~Huang, S.~Yong, X.~Ma, X.~Linghu, P.~Li, Y.~Wang, Q.~Li, S.~Zhu, B.~Jia, and
  S.~Huang, ``An embodied generalist agent in 3d world,'' in \emph{{ICML}},
  2024.

\bibitem{DBLP:conf/eccv/QiDZGHGYM24}
Z.~Qi, R.~Dong, S.~Zhang, H.~Geng, C.~Han, Z.~Ge, L.~Yi, and K.~Ma, ``Shapellm:
  Universal 3d object understanding for embodied interaction,'' in \emph{{ECCV}
  {(43)}}, ser. Lecture Notes in Computer Science, vol. 15101.\hskip 1em plus
  0.5em minus 0.4em\relax Springer, 2024, pp. 214--238.

\bibitem{DBLP:conf/icml/HuangAPM22}
W.~Huang, P.~Abbeel, D.~Pathak, and I.~Mordatch, ``Language models as zero-shot
  planners: Extracting actionable knowledge for embodied agents,'' in
  \emph{{ICML}}, vol. 162.\hskip 1em plus 0.5em minus 0.4em\relax {PMLR}, 2022,
  pp. 9118--9147.

\bibitem{DBLP:conf/acl/Sharma0A22}
P.~Sharma, A.~Torralba, and J.~Andreas, ``Skill induction and planning with
  latent language,'' in \emph{{ACL} {(1)}}, 2022, pp. 1713--1726.

\bibitem{DBLP:conf/iccv/SongSWCW023}
C.~H. Song, B.~M. Sadler, J.~Wu, W.~Chao, C.~Washington, and Y.~Su,
  ``Llm-planner: Few-shot grounded planning for embodied agents with large
  language models,'' in \emph{{ICCV}}.\hskip 1em plus 0.5em minus 0.4em\relax
  {IEEE}, 2023, pp. 2986--2997.

\bibitem{DBLP:conf/nips/LiPPDWF0HAAAM0Z22}
S.~Li, X.~Puig, C.~Paxton, Y.~Du, C.~Wang, L.~Fan, T.~Chen, D.~Huang,
  E.~Aky{\"{u}}rek, A.~Anandkumar, J.~Andreas, I.~Mordatch, A.~Torralba, and
  Y.~Zhu, ``Pre-trained language models for interactive decision-making,'' in
  \emph{NeurIPS}, 2022.

\bibitem{DBLP:conf/iclr/ZengAICWWTPRSLV23}
A.~Zeng, M.~Attarian, B.~Ichter, K.~M. Choromanski, A.~Wong, S.~Welker,
  F.~Tombari, A.~Purohit, M.~S. Ryoo, V.~Sindhwani, J.~Lee, V.~Vanhoucke, and
  P.~Florence, ``Socratic models: Composing zero-shot multimodal reasoning with
  language,'' in \emph{{ICLR}}, 2023.

\bibitem{DBLP:conf/icra/SinghBMGXTFTG23}
I.~Singh, V.~Blukis, A.~Mousavian, A.~Goyal, D.~Xu, J.~Tremblay, D.~Fox,
  J.~Thomason, and A.~Garg, ``Progprompt: Generating situated robot task plans
  using large language models,'' in \emph{{ICRA}}.\hskip 1em plus 0.5em minus
  0.4em\relax {IEEE}, 2023, pp. 11\,523--11\,530.

\bibitem{DBLP:journals/corr/abs-2306-17582}
S.~Vemprala, R.~Bonatti, A.~Bucker, and A.~Kapoor, ``Chatgpt for robotics:
  Design principles and model abilities,'' \emph{CoRR}, vol. abs/2306.17582,
  2023.

\bibitem{DBLP:conf/icra/LiangHXXHIFZ23}
J.~Liang, W.~Huang, F.~Xia, P.~Xu, K.~Hausman, B.~Ichter, P.~Florence, and
  A.~Zeng, ``Code as policies: Language model programs for embodied control,''
  in \emph{{ICRA}}.\hskip 1em plus 0.5em minus 0.4em\relax {IEEE}, 2023, pp.
  9493--9500.

\bibitem{DBLP:journals/corr/abs-2302-01560}
Z.~Wang, S.~Cai, A.~Liu, X.~Ma, and Y.~Liang, ``Describe, explain, plan and
  select: Interactive planning with large language models enables open-world
  multi-task agents,'' \emph{CoRR}, vol. abs/2302.01560, 2023.

\bibitem{DBLP:journals/corr/abs-2309-16650}
Q.~Gu, A.~Kuwajerwala, S.~Morin, K.~M. Jatavallabhula, B.~Sen, A.~Agarwal,
  C.~Rivera, W.~Paul, K.~Ellis, R.~Chellappa, C.~Gan, C.~M. de~Melo, J.~B.
  Tenenbaum, A.~Torralba, F.~Shkurti, and L.~Paull, ``Conceptgraphs:
  Open-vocabulary 3d scene graphs for perception and planning,'' \emph{CoRR},
  vol. abs/2309.16650, 2023.

\bibitem{DBLP:conf/iclr/LinHSWY025}
F.~Lin, Y.~Hu, P.~Sheng, C.~Wen, J.~You, and Y.~Gao, ``Data scaling laws in
  imitation learning for robotic manipulation,'' in \emph{{ICLR}}, 2025.

\bibitem{DBLP:conf/icml/PearceRBGDH25}
T.~Pearce, T.~Rashid, D.~Bignell, R.~Georgescu, S.~Devlin, and K.~Hofmann,
  ``Scaling laws for pre-training agents and world models,'' in \emph{{ICML}},
  2025.

\bibitem{DBLP:conf/cvpr/HongZCWLG24}
Y.~Hong, Z.~Zheng, P.~Chen, Y.~Wang, J.~Li, and C.~Gan, ``Multiply: {A}
  multisensory object-centric embodied large language model in 3d world,'' in
  \emph{{CVPR}}.\hskip 1em plus 0.5em minus 0.4em\relax {IEEE}, 2024, pp.
  26\,396--26\,406.

\bibitem{DBLP:journals/corr/abs-2211-09935}
S.~S. Raman, V.~Cohen, E.~Rosen, I.~Idrees, D.~Paulius, and S.~Tellex,
  ``Planning with large language models via corrective re-prompting,''
  \emph{CoRR}, vol. abs/2211.09935, 2022.

\bibitem{DBLP:journals/corr/abs-2404-10220}
P.~Zhi, Z.~Zhang, M.~Han, Z.~Zhang, Z.~Li, Z.~Jiao, B.~Jia, and S.~Huang,
  ``Closed-loop open-vocabulary mobile manipulation with {GPT-4V},''
  \emph{CoRR}, vol. abs/2404.10220, 2024.

\bibitem{DBLP:conf/icra/FangFTLW0ZL24}
H.~Fang, H.~Fang, Z.~Tang, J.~Liu, C.~Wang, J.~Wang, H.~Zhu, and C.~Lu,
  ``{RH20T:} {A} comprehensive robotic dataset for learning diverse skills in
  one-shot,'' in \emph{{ICRA}}.\hskip 1em plus 0.5em minus 0.4em\relax {IEEE},
  2024, pp. 653--660.

\bibitem{DBLP:journals/corr/abs-2403-12945}
A.~Khazatsky, K.~Pertsch, S.~Nair, A.~Balakrishna, S.~Dasari, S.~Karamcheti,
  S.~Nasiriany, M.~K. Srirama, L.~Y. Chen, K.~Ellis, P.~D. Fagan, J.~Hejna,
  M.~Itkina, M.~Lepert, Y.~J. Ma, P.~T. Miller, J.~Wu, S.~Belkhale, S.~Dass,
  H.~Ha, A.~Jain, A.~Lee, Y.~Lee, M.~Memmel, S.~Park, I.~Radosavovic, K.~Wang,
  A.~Zhan, K.~Black, C.~Chi, K.~B. Hatch, S.~Lin, J.~Lu, J.~Mercat, A.~Rehman,
  P.~R. Sanketi, A.~Sharma, C.~Simpson, Q.~Vuong, H.~R. Walke, B.~Wulfe,
  T.~Xiao, J.~H. Yang, A.~Yavary, T.~Z. Zhao, C.~Agia, R.~Baijal, M.~G. Castro,
  D.~Chen, Q.~Chen, T.~Chung, J.~Drake, E.~P. Foster, and et~al., ``{DROID:}
  {A} large-scale in-the-wild robot manipulation dataset,'' \emph{CoRR}, vol.
  abs/2403.12945, 2024.

\bibitem{DBLP:conf/corl/SharmaMPG18}
P.~Sharma, L.~Mohan, L.~Pinto, and A.~Gupta, ``Multiple interactions made easy
  {(MIME):} large scale demonstrations data for imitation,'' in \emph{CoRL},
  vol.~87.\hskip 1em plus 0.5em minus 0.4em\relax {PMLR}, 2018, pp. 906--915.

\bibitem{DBLP:conf/corl/MandlekarZGBSTG18}
A.~Mandlekar, Y.~Zhu, A.~Garg, J.~Booher, M.~Spero, A.~Tung, J.~Gao, J.~Emmons,
  A.~Gupta, E.~Orbay, S.~Savarese, and L.~Fei{-}Fei, ``{ROBOTURK:} {A}
  crowdsourcing platform for robotic skill learning through imitation,'' in
  \emph{CoRL}, vol.~87.\hskip 1em plus 0.5em minus 0.4em\relax {PMLR}, 2018,
  pp. 879--893.

\bibitem{DBLP:conf/corl/DasariETNBSSLF19}
S.~Dasari, F.~Ebert, S.~Tian, S.~Nair, B.~Bucher, K.~Schmeckpeper, S.~Singh,
  S.~Levine, and C.~Finn, ``Robonet: Large-scale multi-robot learning,'' in
  \emph{CoRL}, vol. 100.\hskip 1em plus 0.5em minus 0.4em\relax {PMLR}, 2019,
  pp. 885--897.

\bibitem{DBLP:journals/corr/abs-2104-08212}
D.~Kalashnikov, J.~Varley, Y.~Chebotar, B.~Swanson, R.~Jonschkowski, C.~Finn,
  S.~Levine, and K.~Hausman, ``Mt-opt: Continuous multi-task robotic
  reinforcement learning at scale,'' \emph{CoRR}, vol. abs/2104.08212, 2021.

\bibitem{DBLP:conf/nips/KumarSZMC0R23}
V.~Kumar, R.~M. Shah, G.~Zhou, V.~Moens, V.~Caggiano, A.~Gupta, and
  A.~Rajeswaran, ``Robohive: {A} unified framework for robot learning,'' in
  \emph{NeurIPS}, 2023.

\bibitem{DBLP:conf/rss/EbertYSBGDFL22}
F.~Ebert, Y.~Yang, K.~Schmeckpeper, B.~Bucher, G.~Georgakis, K.~Daniilidis,
  C.~Finn, and S.~Levine, ``Bridge data: Boosting generalization of robotic
  skills with cross-domain datasets,'' in \emph{RSS}, 2022.

\bibitem{DBLP:conf/corl/Srivastava0LMXV21}
S.~Srivastava, C.~Li, M.~Lingelbach, R.~Mart{\'{\i}}n{-}Mart{\'{\i}}n, F.~Xia,
  K.~E. Vainio, Z.~Lian, C.~Gokmen, S.~Buch, C.~K. Liu, S.~Savarese, H.~Gweon,
  J.~Wu, and L.~Fei{-}Fei, ``{BEHAVIOR:} benchmark for everyday household
  activities in virtual, interactive, and ecological environments,'' in
  \emph{CoRL}, vol. 164.\hskip 1em plus 0.5em minus 0.4em\relax {PMLR}, 2021,
  pp. 477--490.

\bibitem{DBLP:conf/corl/0002XMLSSVGDJKL21}
C.~Li, F.~Xia, R.~Mart{\'{\i}}n{-}Mart{\'{\i}}n, M.~Lingelbach, S.~Srivastava,
  B.~Shen, K.~E. Vainio, C.~Gokmen, G.~Dharan, T.~Jain, A.~Kurenkov, C.~K. Liu,
  H.~Gweon, J.~Wu, L.~Fei{-}Fei, and S.~Savarese, ``igibson 2.0: Object-centric
  simulation for robot learning of everyday household tasks,'' in \emph{CoRL},
  vol. 164.\hskip 1em plus 0.5em minus 0.4em\relax {PMLR}, 2021, pp. 455--465.

\bibitem{DBLP:conf/cvpr/XiaZHSMS18}
F.~Xia, A.~R. Zamir, Z.~He, A.~Sax, J.~Malik, and S.~Savarese, ``Gibson env:
  Real-world perception for embodied agents,'' in \emph{{CVPR}}.\hskip 1em plus
  0.5em minus 0.4em\relax Computer Vision Foundation / {IEEE} Computer Society,
  2018, pp. 9068--9079.

\bibitem{DBLP:journals/ral/XiaSLKTTMS20}
F.~Xia, W.~B. Shen, C.~Li, P.~Kasimbeg, M.~Tchapmi, A.~Toshev,
  R.~Mart{\'{\i}}n{-}Mart{\'{\i}}n, and S.~Savarese, ``Interactive gibson
  benchmark: {A} benchmark for interactive navigation in cluttered
  environments,'' \emph{{IEEE} Robotics Autom. Lett.}, vol.~5, no.~2, pp.
  713--720, 2020.

\bibitem{DBLP:conf/iros/ShenX0MFWPBSTTV21}
B.~Shen, F.~Xia, C.~Li, R.~Mart{\'{\i}}n{-}Mart{\'{\i}}n, L.~Fan, G.~Wang,
  C.~P{\'{e}}rez{-}D'Arpino, S.~Buch, S.~Srivastava, L.~Tchapmi, M.~Tchapmi,
  K.~Vainio, J.~Wong, L.~Fei{-}Fei, and S.~Savarese, ``igibson 1.0: {A}
  simulation environment for interactive tasks in large realistic scenes,'' in
  \emph{{IROS}}.\hskip 1em plus 0.5em minus 0.4em\relax {IEEE}, 2021, pp.
  7520--7527.

\bibitem{DBLP:journals/corr/abs-2403-09227}
C.~Li, R.~Zhang, J.~Wong, C.~Gokmen, S.~Srivastava,
  R.~Mart{\'{\i}}n{-}Mart{\'{\i}}n, C.~Wang, G.~Levine, W.~Ai, B.~Martinez,
  H.~Yin, M.~Lingelbach, M.~Hwang, A.~Hiranaka, S.~Garlanka, A.~Aydin, S.~Lee,
  J.~Sun, M.~Anvari, M.~Sharma, D.~Bansal, S.~Hunter, K.~Kim, A.~Lou, C.~R.
  Matthews, I.~Villa{-}Renteria, J.~H. Tang, C.~Tang, F.~Xia, Y.~Li,
  S.~Savarese, H.~Gweon, C.~K. Liu, J.~Wu, and L.~Fei{-}Fei, ``{BEHAVIOR-1K:}
  {A} human-centered, embodied {AI} benchmark with 1, 000 everyday activities
  and realistic simulation,'' \emph{CoRR}, vol. abs/2403.09227, 2024.

\bibitem{DBLP:conf/cvpr/XiangQMXZLLJYWY20}
F.~Xiang, Y.~Qin, K.~Mo, Y.~Xia, H.~Zhu, F.~Liu, M.~Liu, H.~Jiang, Y.~Yuan,
  H.~Wang, L.~Yi, A.~X. Chang, L.~J. Guibas, and H.~Su, ``{SAPIEN:} {A}
  simulated part-based interactive environment,'' in \emph{{CVPR}}.\hskip 1em
  plus 0.5em minus 0.4em\relax Computer Vision Foundation / {IEEE}, 2020, pp.
  11\,094--11\,104.

\bibitem{DBLP:journals/corr/abs-2405-05941}
X.~Li, K.~Hsu, J.~Gu, K.~Pertsch, O.~Mees, H.~R. Walke, C.~Fu, I.~Lunawat,
  I.~Sieh, S.~Kirmani, S.~Levine, J.~Wu, C.~Finn, H.~Su, Q.~Vuong, and T.~Xiao,
  ``Evaluating real-world robot manipulation policies in simulation,''
  \emph{CoRR}, vol. abs/2405.05941, 2024.

\bibitem{DBLP:journals/corr/abs-1712-05474}
E.~Kolve, R.~Mottaghi, D.~Gordon, Y.~Zhu, A.~Gupta, and A.~Farhadi,
  ``{AI2-THOR:} an interactive 3d environment for visual {AI},'' \emph{CoRR},
  vol. abs/1712.05474, 2017.

\bibitem{DBLP:conf/cvpr/EhsaniHHVWKKM21}
K.~Ehsani, W.~Han, A.~Herrasti, E.~VanderBilt, L.~Weihs, E.~Kolve, A.~Kembhavi,
  and R.~Mottaghi, ``Manipulathor: {A} framework for visual object
  manipulation,'' in \emph{{CVPR}}.\hskip 1em plus 0.5em minus 0.4em\relax
  Computer Vision Foundation / {IEEE}, 2021, pp. 4497--4506.

\bibitem{DBLP:conf/cvpr/WeihsDKM21}
L.~Weihs, M.~Deitke, A.~Kembhavi, and R.~Mottaghi, ``Visual room
  rearrangement,'' in \emph{{CVPR}}.\hskip 1em plus 0.5em minus 0.4em\relax
  Computer Vision Foundation / {IEEE}, 2021, pp. 5922--5931.

\bibitem{DBLP:conf/iclr/MirakhorG0B24}
K.~Mirakhor, S.~Ghosh, D.~Das, and B.~Bhowmick, ``Task planning for visual room
  rearrangement under partial observability,'' in \emph{{ICLR}}, 2024.

\bibitem{DBLP:conf/cvpr/PuigRBLWF018}
X.~Puig, K.~Ra, M.~Boben, J.~Li, T.~Wang, S.~Fidler, and A.~Torralba,
  ``Virtualhome: Simulating household activities via programs,'' in \emph{2018
  {IEEE} Conference on Computer Vision and Pattern Recognition, {CVPR} 2018,
  Salt Lake City, UT, USA, June 18-22, 2018}.\hskip 1em plus 0.5em minus
  0.4em\relax Computer Vision Foundation / {IEEE} Computer Society, 2018, pp.
  8494--8502.

\bibitem{DBLP:conf/nips/GanSAMSTFKBHSKW21}
C.~Gan, J.~Schwartz, S.~Alter, D.~Mrowca, M.~Schrimpf, J.~Traer, J.~D. Freitas,
  J.~Kubilius, A.~Bhandwaldar, N.~Haber, M.~Sano, K.~Kim, E.~Wang,
  M.~Lingelbach, A.~Curtis, K.~T. Feigelis, D.~Bear, D.~Gutfreund, D.~D. Cox,
  A.~Torralba, J.~J. DiCarlo, J.~Tenenbaum, J.~H. McDermott, and D.~Yamins,
  ``Threedworld: {A} platform for interactive multi-modal physical
  simulation,'' in \emph{NeurIPS Datasets and Benchmarks}, 2021.

\bibitem{DBLP:journals/ral/JamesMAD20}
S.~James, Z.~Ma, D.~R. Arrojo, and A.~J. Davison, ``Rlbench: The robot learning
  benchmark {\&} learning environment,'' \emph{{IEEE} Robotics Autom. Lett.},
  vol.~5, no.~2, pp. 3019--3026, 2020.

\bibitem{DBLP:conf/corl/YuQHJHFL19}
T.~Yu, D.~Quillen, Z.~He, R.~Julian, K.~Hausman, C.~Finn, and S.~Levine,
  ``Meta-world: {A} benchmark and evaluation for multi-task and meta
  reinforcement learning,'' in \emph{CoRL}, vol. 100.\hskip 1em plus 0.5em
  minus 0.4em\relax {PMLR}, 2019, pp. 1094--1100.

\bibitem{DBLP:journals/ral/MeesHRB22}
O.~Mees, L.~Hermann, E.~Rosete{-}Beas, and W.~Burgard, ``{CALVIN:} {A}
  benchmark for language-conditioned policy learning for long-horizon robot
  manipulation tasks,'' \emph{{IEEE} Robotics Autom. Lett.}, vol.~7, no.~3, pp.
  7327--7334, 2022.

\bibitem{DBLP:conf/corl/0004KLLH19}
A.~Gupta, V.~Kumar, C.~Lynch, S.~Levine, and K.~Hausman, ``Relay policy
  learning: Solving long-horizon tasks via imitation and reinforcement
  learning,'' in \emph{CoRL}, vol. 100.\hskip 1em plus 0.5em minus 0.4em\relax
  {PMLR}, 2019, pp. 1025--1037.

\bibitem{DBLP:conf/iccv/SavvaMPBKMZWJSL19}
M.~Savva, J.~Malik, D.~Parikh, D.~Batra, A.~Kadian, O.~Maksymets, Y.~Zhao,
  E.~Wijmans, B.~Jain, J.~Straub, J.~Liu, and V.~Koltun, ``Habitat: {A}
  platform for embodied {AI} research,'' in \emph{{ICCV}}.\hskip 1em plus 0.5em
  minus 0.4em\relax {IEEE}, 2019, pp. 9338--9346.

\bibitem{DBLP:conf/nips/SzotCUWZTMMCMGV21}
A.~Szot, A.~Clegg, E.~Undersander, E.~Wijmans, Y.~Zhao, J.~M. Turner,
  N.~Maestre, M.~Mukadam, D.~S. Chaplot, O.~Maksymets, A.~Gokaslan, V.~Vondrus,
  S.~Dharur, F.~Meier, W.~Galuba, A.~X. Chang, Z.~Kira, V.~Koltun, J.~Malik,
  M.~Savva, and D.~Batra, ``Habitat 2.0: Training home assistants to rearrange
  their habitat,'' in \emph{NeurIPS}, 2021, pp. 251--266.

\bibitem{DBLP:journals/corr/abs-2011-01975}
D.~Batra, A.~X. Chang, S.~Chernova, A.~J. Davison, J.~Deng, V.~Koltun,
  S.~Levine, J.~Malik, I.~Mordatch, R.~Mottaghi, M.~Savva, and H.~Su,
  ``Rearrangement: {A} challenge for embodied {AI},'' \emph{CoRR}, vol.
  abs/2011.01975, 2020.

\bibitem{DBLP:conf/corl/YenamandraRYWKG23}
S.~Yenamandra, A.~Ramachandran, K.~Yadav, A.~S. Wang, M.~Khanna, T.~Gervet,
  T.~Yang, V.~Jain, A.~Clegg, J.~M. Turner, Z.~Kira, M.~Savva, A.~X. Chang,
  D.~S. Chaplot, D.~Batra, R.~Mottaghi, Y.~Bisk, and C.~Paxton, ``Homerobot:
  Open-vocabulary mobile manipulation,'' in \emph{CoRL}, vol. 229.\hskip 1em
  plus 0.5em minus 0.4em\relax {PMLR}, 2023, pp. 1975--2011.

\bibitem{DBLP:conf/cvpr/ShridharTGBHMZF20}
M.~Shridhar, J.~Thomason, D.~Gordon, Y.~Bisk, W.~Han, R.~Mottaghi,
  L.~Zettlemoyer, and D.~Fox, ``{ALFRED:} {A} benchmark for interpreting
  grounded instructions for everyday tasks,'' in \emph{{CVPR}}.\hskip 1em plus
  0.5em minus 0.4em\relax Computer Vision Foundation / {IEEE}, 2020, pp.
  10\,737--10\,746.

\bibitem{DBLP:journals/corr/abs-1801-00690}
Y.~Tassa, Y.~Doron, A.~Muldal, T.~Erez, Y.~Li, D.~de~Las~Casas, D.~Budden,
  A.~Abdolmaleki, J.~Merel, A.~Lefrancq, T.~P. Lillicrap, and M.~A. Riedmiller,
  ``Deepmind control suite,'' \emph{CoRR}, vol. abs/1801.00690, 2018.

\bibitem{DBLP:journals/corr/BrockmanCPSSTZ16}
G.~Brockman, V.~Cheung, L.~Pettersson, J.~Schneider, J.~Schulman, J.~Tang, and
  W.~Zaremba, ``Openai gym,'' \emph{CoRR}, vol. abs/1606.01540, 2016.

\bibitem{Genesis}
\BIBentryALTinterwordspacing
Genesis-Team, ``Genesis: A universal and generative physics engine for robotics
  and beyond,'' December 2024. [Online]. Available:
  \url{https://github.com/Genesis-Embodied-AI/Genesis}
\BIBentrySTDinterwordspacing

\bibitem{DBLP:conf/icml/WangXCWWFEHG24}
Y.~Wang, Z.~Xian, F.~Chen, T.~Wang, Y.~Wang, K.~Fragkiadaki, Z.~Erickson,
  D.~Held, and C.~Gan, ``Robogen: Towards unleashing infinite data for
  automated robot learning via generative simulation,'' in \emph{{ICML}}, 2024.

\bibitem{DBLP:journals/corr/abs-2401-12963}
M.~Ahn, D.~Dwibedi, C.~Finn, M.~G. Arenas, K.~Gopalakrishnan, K.~Hausman,
  B.~Ichter, A.~Irpan, N.~J. Joshi, R.~Julian, S.~Kirmani, I.~Leal, T.~E. Lee,
  S.~Levine, Y.~Lu, S.~Maddineni, K.~Rao, D.~Sadigh, P.~Sanketi, P.~Sermanet,
  Q.~Vuong, S.~Welker, F.~Xia, T.~Xiao, P.~Xu, S.~Xu, and Z.~Xu, ``Autort:
  Embodied foundation models for large scale orchestration of robotic agents,''
  \emph{CoRR}, vol. abs/2401.12963, 2024.

\bibitem{DBLP:conf/rss/XiaoCSWBHLT23}
T.~Xiao, H.~Chan, P.~Sermanet, A.~Wahid, A.~Brohan, K.~Hausman, S.~Levine, and
  J.~Tompson, ``Robotic skill acquisition via instruction augmentation with
  vision-language models,'' in \emph{RSS}, 2023.

\bibitem{DBLP:conf/naacl/MaCLSZK24}
Y.~Ma, D.~Chi, J.~Li, K.~Song, Y.~Zhuang, and I.~King, ``{VOLTA:} improving
  generative diversity by variational mutual information maximizing
  autoencoder,'' in \emph{{NAACL-HLT} (Findings)}, 2024.

\bibitem{DBLP:conf/rss/ChiXPCB0TS24}
C.~Chi, Z.~Xu, C.~Pan, E.~Cousineau, B.~Burchfiel, S.~Feng, R.~Tedrake, and
  S.~Song, ``Universal manipulation interface: In-the-wild robot teaching
  without in-the-wild robots,'' in \emph{RSS}, 2024.

\bibitem{DBLP:conf/nips/MoonSXJBBRRMBSP23}
G.~Moon, S.~Saito, W.~Xu, R.~Joshi, J.~Buffalini, H.~Bellan, N.~Rosen,
  J.~Richardson, M.~Mize, P.~de~Bree, T.~Simon, B.~Peng, S.~Garg, K.~McPhail,
  and T.~Shiratori, ``A dataset of relighted 3d interacting hands,'' in
  \emph{NeurIPS}, 2023.

\bibitem{DBLP:journals/corr/abs-2312-06722}
Y.~Chen, Y.~Ge, Y.~Ge, M.~Ding, B.~Li, R.~Wang, R.~Xu, Y.~Shan, and X.~Liu,
  ``Egoplan-bench: Benchmarking egocentric embodied planning with multimodal
  large language models,'' \emph{CoRR}, vol. abs/2312.06722, 2023.

\bibitem{DBLP:conf/nips/ValmeekamMHSK23}
K.~Valmeekam, M.~Marquez, A.~O. Hernandez, S.~Sreedharan, and S.~Kambhampati,
  ``Planbench: An extensible benchmark for evaluating large language models on
  planning and reasoning about change,'' in \emph{NeurIPS}, 2023.

\bibitem{DBLP:conf/nips/ValmeekamMSK23}
K.~Valmeekam, M.~Marquez, S.~Sreedharan, and S.~Kambhampati, ``On the planning
  abilities of large language models - {A} critical investigation,'' in
  \emph{NeurIPS}, 2023.

\bibitem{DBLP:journals/corr/abs-2402-08178}
J.~Choi, Y.~Yoon, H.~Ong, J.~Kim, and M.~Jang, ``Lota-bench: Benchmarking
  language-oriented task planners for embodied agents,'' \emph{CoRR}, vol.
  abs/2402.08178, 2024.

\bibitem{DBLP:conf/nips/LiZWWZSGLLZLL0M24}
M.~Li, S.~Zhao, Q.~Wang, K.~Wang, Y.~Zhou, S.~Srivastava, C.~Gokmen, T.~Lee,
  L.~E. Li, R.~Zhang, W.~Liu, P.~Liang, L.~Fei{-}Fei, J.~Mao, and J.~Wu,
  ``Embodied agent interface: Benchmarking llms for embodied decision making,''
  in \emph{NeurIPS}, 2024.

\bibitem{DBLP:conf/cvpr/DasDGLPB18}
A.~Das, S.~Datta, G.~Gkioxari, S.~Lee, D.~Parikh, and D.~Batra, ``Embodied
  question answering,'' in \emph{{CVPR}}.\hskip 1em plus 0.5em minus
  0.4em\relax Computer Vision Foundation / {IEEE} Computer Society, 2018, pp.
  1--10.

\bibitem{DBLP:conf/cvpr/GordonKRRFF18}
D.~Gordon, A.~Kembhavi, M.~Rastegari, J.~Redmon, D.~Fox, and A.~Farhadi,
  ``{IQA:} visual question answering in interactive environments,'' in
  \emph{{CVPR}}.\hskip 1em plus 0.5em minus 0.4em\relax Computer Vision
  Foundation / {IEEE} Computer Society, 2018, pp. 4089--4098.

\bibitem{DBLP:conf/cvpr/YuCGBBB19}
L.~Yu, X.~Chen, G.~Gkioxari, M.~Bansal, T.~L. Berg, and D.~Batra,
  ``Multi-target embodied question answering,'' in \emph{{CVPR}}.\hskip 1em
  plus 0.5em minus 0.4em\relax Computer Vision Foundation / {IEEE}, 2019, pp.
  6309--6318.

\bibitem{DBLP:conf/cvpr/WijmansDMDGLEPB19}
E.~Wijmans, S.~Datta, O.~Maksymets, A.~Das, G.~Gkioxari, S.~Lee, I.~Essa,
  D.~Parikh, and D.~Batra, ``Embodied question answering in photorealistic
  environments with point cloud perception,'' in \emph{{CVPR}}.\hskip 1em plus
  0.5em minus 0.4em\relax Computer Vision Foundation / {IEEE}, 2019, pp.
  6659--6668.

\bibitem{DBLP:conf/iccvw/Fan19}
C.~Fan, ``Egovqa - an egocentric video question answering benchmark dataset,''
  in \emph{{ICCV} Workshops}.\hskip 1em plus 0.5em minus 0.4em\relax {IEEE},
  2019, pp. 4359--4366.

\bibitem{DBLP:conf/nips/JiaLZH22}
B.~Jia, T.~Lei, S.~Zhu, and S.~Huang, ``Egotaskqa: Understanding human tasks in
  egocentric videos,'' in \emph{NeurIPS}, 2022.

\bibitem{DBLP:conf/iclr/IslamGII24}
M.~M. Islam, A.~Gladstone, R.~Islam, and T.~Iqbal, ``{EQA-MX:} embodied
  question answering using multimodal expression,'' in \emph{{ICLR}}, 2024.

\bibitem{majumdar2023openeqa}
A.~Majumdar, A.~Ajay, X.~Zhang, P.~Putta, S.~Yenamandra, M.~Henaff, S.~Silwal,
  P.~Mcvay, O.~Maksymets, S.~Arnaud, K.~Yadav, Q.~Li, B.~Newman, M.~Sharma,
  V.~Berges, S.~Zhang, P.~Agrawal, Y.~Bisk, D.~Batra, M.~Kalakrishnan,
  F.~Meier, C.~Paxton, S.~Sax, and A.~Rajeswaran, ``{OpenEQA: Embodied Question
  Answering in the Era of Foundation Models},'' in \emph{{CVPR}}, 2024.

\bibitem{DBLP:conf/cvpr/GirdharELSAJM23}
R.~Girdhar, A.~El{-}Nouby, Z.~Liu, M.~Singh, K.~V. Alwala, A.~Joulin, and
  I.~Misra, ``Imagebind one embedding space to bind them all,'' in
  \emph{{CVPR}}.\hskip 1em plus 0.5em minus 0.4em\relax {IEEE}, 2023, pp.
  15\,180--15\,190.

\bibitem{DBLP:conf/iclr/ZhuLNYCWPJZLZ0024}
B.~Zhu, B.~Lin, M.~Ning, Y.~Yan, J.~Cui, H.~Wang, Y.~Pang, W.~Jiang, J.~Zhang,
  Z.~Li, C.~Zhang, Z.~Li, W.~Liu, and L.~Yuan, ``Languagebind: Extending
  video-language pretraining to n-modality by language-based semantic
  alignment,'' in \emph{{ICLR}}, 2024.

\bibitem{DBLP:conf/corl/GuzeyECP23}
I.~G{\"{u}}zey, B.~Evans, S.~Chintala, and L.~Pinto, ``Dexterity from touch:
  Self-supervised pre-training of tactile representations with robotic play,''
  in \emph{CoRL}, vol. 229.\hskip 1em plus 0.5em minus 0.4em\relax {PMLR},
  2023, pp. 3142--3166.

\bibitem{ijcai2020p689}
R.~Zhang, A.~Saran, B.~Liu, Y.~Zhu, S.~Guo, S.~Niekum, D.~Ballard, and
  M.~Hayhoe, ``Human gaze assisted artificial intelligence: A review,'' in
  \emph{{IJCAI-20}}, 7 2020, pp. 4951--4958, survey track.

\bibitem{DBLP:conf/eccv/ZhangLZWMHB18}
R.~Zhang, Z.~Liu, L.~Zhang, J.~A. Whritner, K.~S. Muller, M.~M. Hayhoe, and
  D.~H. Ballard, ``{AGIL:} learning attention from human for visuomotor
  tasks,'' in \emph{{ECCV} {(11)}}, vol. 11215.\hskip 1em plus 0.5em minus
  0.4em\relax Springer, 2018, pp. 692--707.

\bibitem{DBLP:conf/atal/SaranZSN21}
A.~Saran, R.~Zhang, E.~S. Short, and S.~Niekum, ``Efficiently guiding imitation
  learning agents with human gaze,'' in \emph{{AAMAS}}.\hskip 1em plus 0.5em
  minus 0.4em\relax {ACM}, 2021, pp. 1109--1117.

\bibitem{DBLP:conf/nips/GuoZLZBHS21}
S.~Guo, R.~Zhang, B.~Liu, Y.~Zhu, D.~H. Ballard, M.~M. Hayhoe, and P.~Stone,
  ``Machine versus human attention in deep reinforcement learning tasks,'' in
  \emph{NeurIPS}, 2021, pp. 25\,370--25\,385.

\bibitem{Le2016ASO}
H.~M. Le, T.~N. Do, and S.~J. Phee, ``A survey on actuators-driven surgical
  robots,'' \emph{Sensors and Actuators A-physical}, vol. 247, pp. 323--354,
  2016.

\bibitem{DBLP:conf/ichi/LaursenPJSBAV22}
M.~S. Laursen, J.~S. Pedersen, S.~A. Just, T.~R. Savarimuthu, B.~Blomholt,
  J.~K.~H. Andersen, and P.~J. Vinholt, ``Factors facilitating the acceptance
  of diagnostic robots in healthcare: {A} survey,'' in \emph{{ICHI}}.\hskip 1em
  plus 0.5em minus 0.4em\relax {IEEE}, 2022, pp. 442--448.

\bibitem{DBLP:conf/cvpr/HeZRS16}
K.~He, X.~Zhang, S.~Ren, and J.~Sun, ``Deep residual learning for image
  recognition,'' in \emph{{CVPR}}.\hskip 1em plus 0.5em minus 0.4em\relax
  {IEEE} Computer Society, 2016, pp. 770--778.

\bibitem{DBLP:conf/iclr/DosovitskiyB0WZ21}
A.~Dosovitskiy, L.~Beyer, A.~Kolesnikov, D.~Weissenborn, X.~Zhai,
  T.~Unterthiner, M.~Dehghani, M.~Minderer, G.~Heigold, S.~Gelly, J.~Uszkoreit,
  and N.~Houlsby, ``An image is worth 16x16 words: Transformers for image
  recognition at scale,'' in \emph{{ICLR}}, 2021.

\bibitem{DBLP:journals/corr/abs-2304-02643}
A.~Kirillov, E.~Mintun, N.~Ravi, H.~Mao, C.~Rolland, L.~Gustafson, T.~Xiao,
  S.~Whitehead, A.~C. Berg, W.~Lo, P.~Doll{\'{a}}r, and R.~B. Girshick,
  ``Segment anything,'' \emph{CoRR}, vol. abs/2304.02643, 2023.

\bibitem{DBLP:journals/neco/HochreiterS97}
S.~Hochreiter and J.~Schmidhuber, ``Long short-term memory,'' \emph{Neural
  Comput.}, vol.~9, no.~8, pp. 1735--1780, 1997.

\bibitem{DBLP:conf/emnlp/ChoMGBBSB14}
K.~Cho, B.~van Merrienboer, {\c{C}}.~G{\"{u}}l{\c{c}}ehre, D.~Bahdanau,
  F.~Bougares, H.~Schwenk, and Y.~Bengio, ``Learning phrase representations
  using {RNN} encoder-decoder for statistical machine translation,'' in
  \emph{{EMNLP}}.\hskip 1em plus 0.5em minus 0.4em\relax {ACL}, 2014, pp.
  1724--1734.

\bibitem{chatgpt-openai}
\BIBentryALTinterwordspacing
OpenAI. (2023) Introducing chatgpt. [Online]. Available:
  \url{https://openai.com/blog/chatgpt}
\BIBentrySTDinterwordspacing

\bibitem{DBLP:journals/nature/SilverHMGSDSAPL16}
D.~Silver, A.~Huang, C.~J. Maddison, A.~Guez, L.~Sifre, G.~van~den Driessche,
  J.~Schrittwieser, I.~Antonoglou, V.~Panneershelvam, M.~Lanctot, S.~Dieleman,
  D.~Grewe, J.~Nham, N.~Kalchbrenner, I.~Sutskever, T.~P. Lillicrap, M.~Leach,
  K.~Kavukcuoglu, T.~Graepel, and D.~Hassabis, ``Mastering the game of go with
  deep neural networks and tree search,'' \emph{Nat.}, vol. 529, no. 7587, pp.
  484--489, 2016.

\bibitem{DBLP:journals/corr/SchulmanWDRK17}
J.~Schulman, F.~Wolski, P.~Dhariwal, A.~Radford, and O.~Klimov, ``Proximal
  policy optimization algorithms,'' \emph{CoRR}, vol. abs/1707.06347, 2017.

\bibitem{DBLP:journals/corr/abs-1808-00177}
OpenAI, M.~Andrychowicz, B.~Baker, M.~Chociej, R.~J{\'{o}}zefowicz, B.~McGrew,
  J.~Pachocki, A.~Petron, M.~Plappert, G.~Powell, A.~Ray, J.~Schneider,
  S.~Sidor, J.~Tobin, P.~Welinder, L.~Weng, and W.~Zaremba, ``Learning
  dexterous in-hand manipulation,'' \emph{CoRR}, vol. abs/1808.00177, 2018.

\bibitem{DBLP:conf/nips/RafailovSMMEF23}
R.~Rafailov, A.~Sharma, E.~Mitchell, C.~D. Manning, S.~Ermon, and C.~Finn,
  ``Direct preference optimization: Your language model is secretly a reward
  model,'' in \emph{NeurIPS}, 2023.

\bibitem{DBLP:journals/corr/ChenFLVGDZ15}
X.~Chen, H.~Fang, T.~Lin, R.~Vedantam, S.~Gupta, P.~Doll{\'{a}}r, and C.~L.
  Zitnick, ``Microsoft {COCO} captions: Data collection and evaluation
  server,'' \emph{CoRR}, vol. abs/1504.00325, 2015.

\bibitem{DBLP:conf/iccv/AntolALMBZP15}
S.~Antol, A.~Agrawal, J.~Lu, M.~Mitchell, D.~Batra, C.~L. Zitnick, and
  D.~Parikh, ``{VQA:} visual question answering,'' in \emph{{ICCV}}.\hskip 1em
  plus 0.5em minus 0.4em\relax {IEEE} Computer Society, 2015, pp. 2425--2433.

\bibitem{DBLP:conf/eccv/YuPYBB16}
L.~Yu, P.~Poirson, S.~Yang, A.~C. Berg, and T.~L. Berg, ``Modeling context in
  referring expressions,'' in \emph{{ECCV} {(2)}}, ser. Lecture Notes in
  Computer Science, vol. 9906.\hskip 1em plus 0.5em minus 0.4em\relax Springer,
  2016, pp. 69--85.

\bibitem{DBLP:conf/cvpr/VinyalsTBE15}
O.~Vinyals, A.~Toshev, S.~Bengio, and D.~Erhan, ``Show and tell: {A} neural
  image caption generator,'' in \emph{{CVPR}}.\hskip 1em plus 0.5em minus
  0.4em\relax {IEEE} Computer Society, 2015, pp. 3156--3164.

\bibitem{radford2018improving}
A.~Radford, K.~Narasimhan, T.~Salimans, and I.~Sutskever, ``Improving language
  understanding by generative pre-training,'' \emph{OpenAI blog}, 2018.

\bibitem{DBLP:conf/nips/LuBPL19}
J.~Lu, D.~Batra, D.~Parikh, and S.~Lee, ``Vilbert: Pretraining task-agnostic
  visiolinguistic representations for vision-and-language tasks,'' in
  \emph{NeurIPS}, 2019, pp. 13--23.

\bibitem{DBLP:journals/air/KhanSZQ20}
A.~Khan, A.~Sohail, U.~Zahoora, and A.~S. Qureshi, ``A survey of the recent
  architectures of deep convolutional neural networks,'' \emph{Artif. Intell.
  Rev.}, vol.~53, no.~8, pp. 5455--5516, 2020.

\bibitem{DBLP:journals/csur/KhanNHZKS22}
S.~H. Khan, M.~Naseer, M.~Hayat, S.~W. Zamir, F.~S. Khan, and M.~Shah,
  ``Transformers in vision: {A} survey,'' \emph{{ACM} Comput. Surv.}, vol.~54,
  no. 10s, pp. 200:1--200:41, 2022.

\bibitem{DBLP:journals/neco/LeCunBDHHHJ89}
Y.~LeCun, B.~E. Boser, J.~S. Denker, D.~Henderson, R.~E. Howard, W.~E. Hubbard,
  and L.~D. Jackel, ``Backpropagation applied to handwritten zip code
  recognition,'' \emph{Neural Comput.}, vol.~1, no.~4, pp. 541--551, 1989.

\bibitem{DBLP:journals/corr/SimonyanZ14a}
K.~Simonyan and A.~Zisserman, ``Very deep convolutional networks for
  large-scale image recognition,'' in \emph{{ICLR}}, 2015.

\bibitem{DBLP:conf/cvpr/SzegedyLJSRAEVR15}
C.~Szegedy, W.~Liu, Y.~Jia, P.~Sermanet, S.~E. Reed, D.~Anguelov, D.~Erhan,
  V.~Vanhoucke, and A.~Rabinovich, ``Going deeper with convolutions,'' in
  \emph{{CVPR}}.\hskip 1em plus 0.5em minus 0.4em\relax {IEEE} Computer
  Society, 2015, pp. 1--9.

\bibitem{DBLP:conf/aaai/SzegedyIVA17}
C.~Szegedy, S.~Ioffe, V.~Vanhoucke, and A.~A. Alemi, ``Inception-v4,
  inception-resnet and the impact of residual connections on learning,'' in
  \emph{{AAAI}}.\hskip 1em plus 0.5em minus 0.4em\relax {AAAI} Press, 2017, pp.
  4278--4284.

\bibitem{DBLP:conf/cvpr/XieGDTH17}
S.~Xie, R.~B. Girshick, P.~Doll{\'{a}}r, Z.~Tu, and K.~He, ``Aggregated
  residual transformations for deep neural networks,'' in \emph{{CVPR}}.\hskip
  1em plus 0.5em minus 0.4em\relax {IEEE} Computer Society, 2017, pp.
  5987--5995.

\bibitem{DBLP:conf/cvpr/HuSS18}
J.~Hu, L.~Shen, and G.~Sun, ``Squeeze-and-excitation networks,'' in
  \emph{{CVPR}}.\hskip 1em plus 0.5em minus 0.4em\relax Computer Vision
  Foundation / {IEEE} Computer Society, 2018, pp. 7132--7141.

\bibitem{DBLP:conf/icml/TanL19}
M.~Tan and Q.~V. Le, ``Efficientnet: Rethinking model scaling for convolutional
  neural networks,'' in \emph{{ICML}}, vol.~97.\hskip 1em plus 0.5em minus
  0.4em\relax {PMLR}, 2019, pp. 6105--6114.

\bibitem{DBLP:conf/cvpr/GirshickDDM14}
R.~B. Girshick, J.~Donahue, T.~Darrell, and J.~Malik, ``Rich feature
  hierarchies for accurate object detection and semantic segmentation,'' in
  \emph{{CVPR}}.\hskip 1em plus 0.5em minus 0.4em\relax {IEEE} Computer
  Society, 2014, pp. 580--587.

\bibitem{DBLP:conf/iccv/Girshick15}
R.~B. Girshick, ``Fast {R-CNN},'' in \emph{{ICCV}}.\hskip 1em plus 0.5em minus
  0.4em\relax {IEEE} Computer Society, 2015, pp. 1440--1448.

\bibitem{DBLP:conf/nips/RenHGS15}
S.~Ren, K.~He, R.~B. Girshick, and J.~Sun, ``Faster {R-CNN:} towards real-time
  object detection with region proposal networks,'' in \emph{{NIPS}}, 2015, pp.
  91--99.

\bibitem{DBLP:conf/iccv/HeGDG17}
K.~He, G.~Gkioxari, P.~Doll{\'{a}}r, and R.~B. Girshick, ``Mask {R-CNN},'' in
  \emph{{ICCV}}.\hskip 1em plus 0.5em minus 0.4em\relax {IEEE} Computer
  Society, 2017, pp. 2980--2988.

\bibitem{DBLP:conf/cvpr/RedmonDGF16}
J.~Redmon, S.~K. Divvala, R.~B. Girshick, and A.~Farhadi, ``You only look once:
  Unified, real-time object detection,'' in \emph{{CVPR}}.\hskip 1em plus 0.5em
  minus 0.4em\relax {IEEE} Computer Society, 2016, pp. 779--788.

\bibitem{DBLP:conf/cvpr/LinDGHHB17}
T.~Lin, P.~Doll{\'{a}}r, R.~B. Girshick, K.~He, B.~Hariharan, and S.~J.
  Belongie, ``Feature pyramid networks for object detection,'' in
  \emph{{CVPR}}.\hskip 1em plus 0.5em minus 0.4em\relax {IEEE} Computer
  Society, 2017, pp. 936--944.

\bibitem{DBLP:conf/iccv/LinGGHD17}
T.~Lin, P.~Goyal, R.~B. Girshick, K.~He, and P.~Doll{\'{a}}r, ``Focal loss for
  dense object detection,'' in \emph{{ICCV}}.\hskip 1em plus 0.5em minus
  0.4em\relax {IEEE} Computer Society, 2017, pp. 2999--3007.

\bibitem{DBLP:conf/cvpr/00010BT0GZ18}
P.~Anderson, X.~He, C.~Buehler, D.~Teney, M.~Johnson, S.~Gould, and L.~Zhang,
  ``Bottom-up and top-down attention for image captioning and visual question
  answering,'' in \emph{{CVPR}}.\hskip 1em plus 0.5em minus 0.4em\relax
  Computer Vision Foundation / {IEEE} Computer Society, 2018, pp. 6077--6086.

\bibitem{DBLP:conf/cvpr/LongSD15}
J.~Long, E.~Shelhamer, and T.~Darrell, ``Fully convolutional networks for
  semantic segmentation,'' in \emph{{CVPR}}.\hskip 1em plus 0.5em minus
  0.4em\relax {IEEE} Computer Society, 2015, pp. 3431--3440.

\bibitem{DBLP:journals/pami/BadrinarayananK17}
V.~Badrinarayanan, A.~Kendall, and R.~Cipolla, ``Segnet: {A} deep convolutional
  encoder-decoder architecture for image segmentation,'' \emph{{IEEE} Trans.
  Pattern Anal. Mach. Intell.}, vol.~39, no.~12, pp. 2481--2495, 2017.

\bibitem{DBLP:conf/miccai/RonnebergerFB15}
O.~Ronneberger, P.~Fischer, and T.~Brox, ``U-net: Convolutional networks for
  biomedical image segmentation,'' in \emph{{MICCAI} {(3)}}, ser. Lecture Notes
  in Computer Science, vol. 9351.\hskip 1em plus 0.5em minus 0.4em\relax
  Springer, 2015, pp. 234--241.

\bibitem{DBLP:conf/eccv/CarionMSUKZ20}
N.~Carion, F.~Massa, G.~Synnaeve, N.~Usunier, A.~Kirillov, and S.~Zagoruyko,
  ``End-to-end object detection with transformers,'' in \emph{{ECCV} {(1)}},
  ser. Lecture Notes in Computer Science, vol. 12346.\hskip 1em plus 0.5em
  minus 0.4em\relax Springer, 2020, pp. 213--229.

\bibitem{DBLP:conf/iccv/StrudelPLS21}
R.~Strudel, R.~G. Pinel, I.~Laptev, and C.~Schmid, ``Segmenter: Transformer for
  semantic segmentation,'' in \emph{{ICCV}}.\hskip 1em plus 0.5em minus
  0.4em\relax {IEEE}, 2021, pp. 7242--7252.

\bibitem{DBLP:journals/csur/IoannidouCNK17}
A.~Ioannidou, E.~Chatzilari, S.~Nikolopoulos, and I.~Kompatsiaris, ``Deep
  learning advances in computer vision with 3d data: {A} survey,'' \emph{{ACM}
  Comput. Surv.}, vol.~50, no.~2, pp. 20:1--20:38, 2017.

\bibitem{DBLP:journals/corr/abs-1808-01462}
E.~Ahmed, A.~Saint, A.~E.~R. Shabayek, K.~Cherenkova, R.~Das, G.~Gusev,
  D.~Aouada, and B.~E. Ottersten, ``Deep learning advances on different 3d data
  representations: {A} survey,'' \emph{CoRR}, vol. abs/1808.01462, 2018.

\bibitem{DBLP:journals/pami/GuoWHLLB21}
Y.~Guo, H.~Wang, Q.~Hu, H.~Liu, L.~Liu, and M.~Bennamoun, ``Deep learning for
  3d point clouds: {A} survey,'' \emph{{IEEE} Trans. Pattern Anal. Mach.
  Intell.}, vol.~43, no.~12, pp. 4338--4364, 2021.

\bibitem{DBLP:conf/cvpr/QiSNDYG16}
C.~R. Qi, H.~Su, M.~Nie{\ss}ner, A.~Dai, M.~Yan, and L.~J. Guibas, ``Volumetric
  and multi-view cnns for object classification on 3d data,'' in
  \emph{{CVPR}}.\hskip 1em plus 0.5em minus 0.4em\relax {IEEE} Computer
  Society, 2016, pp. 5648--5656.

\bibitem{DBLP:conf/aaai/FengFYZG19}
Y.~Feng, Y.~Feng, H.~You, X.~Zhao, and Y.~Gao, ``Meshnet: Mesh neural network
  for 3d shape representation,'' in \emph{{AAAI}}.\hskip 1em plus 0.5em minus
  0.4em\relax {AAAI} Press, 2019, pp. 8279--8286.

\bibitem{DBLP:journals/tnn/OtterMK21}
D.~W. Otter, J.~R. Medina, and J.~K. Kalita, ``A survey of the usages of deep
  learning for natural language processing,'' \emph{{IEEE} Trans. Neural
  Networks Learn. Syst.}, vol.~32, no.~2, pp. 604--624, 2021.

\bibitem{DBLP:journals/csur/LiuYFJHN23}
P.~Liu, W.~Yuan, J.~Fu, Z.~Jiang, H.~Hayashi, and G.~Neubig, ``Pre-train,
  prompt, and predict: {A} systematic survey of prompting methods in natural
  language processing,'' \emph{{ACM} Comput. Surv.}, vol.~55, no.~9, pp.
  195:1--195:35, 2023.

\bibitem{DBLP:conf/nips/BengioDV00}
Y.~Bengio, R.~Ducharme, and P.~Vincent, ``A neural probabilistic language
  model,'' in \emph{{NIPS}}.\hskip 1em plus 0.5em minus 0.4em\relax {MIT}
  Press, 2000, pp. 932--938.

\bibitem{DBLP:conf/nips/MikolovSCCD13}
T.~Mikolov, I.~Sutskever, K.~Chen, G.~S. Corrado, and J.~Dean, ``Distributed
  representations of words and phrases and their compositionality,'' in
  \emph{{NIPS}}, 2013, pp. 3111--3119.

\bibitem{DBLP:journals/corr/abs-1301-3781}
T.~Mikolov, K.~Chen, G.~Corrado, and J.~Dean, ``Efficient estimation of word
  representations in vector space,'' in \emph{{ICLR} (Workshop Poster)}, 2013.

\bibitem{DBLP:conf/emnlp/PenningtonSM14}
J.~Pennington, R.~Socher, and C.~D. Manning, ``Glove: Global vectors for word
  representation,'' in \emph{{EMNLP}}.\hskip 1em plus 0.5em minus 0.4em\relax
  {ACL}, 2014, pp. 1532--1543.

\bibitem{DBLP:journals/cogsci/Elman90}
J.~L. Elman, ``Finding structure in time,'' \emph{Cogn. Sci.}, vol.~14, no.~2,
  pp. 179--211, 1990.

\bibitem{DBLP:journals/corr/BahdanauCB14}
D.~Bahdanau, K.~Cho, and Y.~Bengio, ``Neural machine translation by jointly
  learning to align and translate,'' in \emph{{ICLR}}, 2015.

\bibitem{DBLP:journals/corr/HuangXY15}
Z.~Huang, W.~Xu, and K.~Yu, ``Bidirectional {LSTM-CRF} models for sequence
  tagging,'' \emph{CoRR}, vol. abs/1508.01991, 2015.

\bibitem{DBLP:conf/naacl/PetersNIGCLZ18}
M.~E. Peters, M.~Neumann, M.~Iyyer, M.~Gardner, C.~Clark, K.~Lee, and
  L.~Zettlemoyer, ``Deep contextualized word representations,'' in
  \emph{{NAACL-HLT}}.\hskip 1em plus 0.5em minus 0.4em\relax Association for
  Computational Linguistics, 2018, pp. 2227--2237.

\bibitem{DBLP:conf/acl/RuderH18}
J.~Howard and S.~Ruder, ``Universal language model fine-tuning for text
  classification,'' in \emph{{ACL} {(1)}}.\hskip 1em plus 0.5em minus
  0.4em\relax Association for Computational Linguistics, 2018, pp. 328--339.

\bibitem{DBLP:conf/emnlp/Kim14}
Y.~Kim, ``Convolutional neural networks for sentence classification,'' in
  \emph{{EMNLP}}.\hskip 1em plus 0.5em minus 0.4em\relax {ACL}, 2014, pp.
  1746--1751.

\bibitem{DBLP:conf/nips/ZhangZL15}
X.~Zhang, J.~J. Zhao, and Y.~LeCun, ``Character-level convolutional networks
  for text classification,'' in \emph{{NIPS}}, 2015, pp. 649--657.

\bibitem{DBLP:conf/acl/MaH16}
X.~Ma and E.~H. Hovy, ``End-to-end sequence labeling via bi-directional
  lstm-cnns-crf,'' in \emph{{ACL} {(1)}}.\hskip 1em plus 0.5em minus
  0.4em\relax The Association for Computer Linguistics, 2016.

\bibitem{radford2019language}
A.~Radford, J.~Wu, R.~Child, D.~Luan, D.~Amodei, I.~Sutskever \emph{et~al.},
  ``Language models are unsupervised multitask learners,'' \emph{OpenAI blog},
  vol.~1, no.~8, p.~9, 2019.

\bibitem{brown2020language}
T.~B. Brown, B.~Mann, N.~Ryder, M.~Subbiah, J.~Kaplan, P.~Dhariwal,
  A.~Neelakantan, P.~Shyam, G.~Sastry, A.~Askell \emph{et~al.}, ``Language
  models are few-shot learners,'' \emph{arXiv preprint arXiv:2005.14165}, 2020.

\bibitem{DBLP:journals/corr/abs-2303-08774}
OpenAI, ``{GPT-4} technical report,'' \emph{CoRR}, vol. abs/2303.08774, 2023.

\bibitem{DBLP:journals/corr/abs-1907-11692}
Y.~Liu, M.~Ott, N.~Goyal, J.~Du, M.~Joshi, D.~Chen, O.~Levy, M.~Lewis,
  L.~Zettlemoyer, and V.~Stoyanov, ``Roberta: {A} robustly optimized {BERT}
  pretraining approach,'' \emph{CoRR}, vol. abs/1907.11692, 2019.

\bibitem{DBLP:conf/iclr/LanCGGSS20}
Z.~Lan, M.~Chen, S.~Goodman, K.~Gimpel, P.~Sharma, and R.~Soricut, ``{ALBERT:}
  {A} lite {BERT} for self-supervised learning of language representations,''
  in \emph{{ICLR}}, 2020.

\bibitem{DBLP:conf/iclr/ClarkLLM20}
K.~Clark, M.~Luong, Q.~V. Le, and C.~D. Manning, ``{ELECTRA:} pre-training text
  encoders as discriminators rather than generators,'' in \emph{{ICLR}}, 2020.

\bibitem{DBLP:conf/nips/YangDYCSL19}
Z.~Yang, Z.~Dai, Y.~Yang, J.~G. Carbonell, R.~Salakhutdinov, and Q.~V. Le,
  ``Xlnet: Generalized autoregressive pretraining for language understanding,''
  in \emph{NeurIPS}, 2019, pp. 5754--5764.

\bibitem{DBLP:journals/corr/abs-2205-01068}
S.~Zhang, S.~Roller, N.~Goyal, M.~Artetxe, M.~Chen, S.~Chen, C.~Dewan, M.~T.
  Diab, X.~Li, X.~V. Lin, T.~Mihaylov, M.~Ott, S.~Shleifer, K.~Shuster,
  D.~Simig, P.~S. Koura, A.~Sridhar, T.~Wang, and L.~Zettlemoyer, ``{OPT:} open
  pre-trained transformer language models,'' \emph{CoRR}, vol. abs/2205.01068,
  2022.

\bibitem{DBLP:conf/acl/LewisLGGMLSZ20}
M.~Lewis, Y.~Liu, N.~Goyal, M.~Ghazvininejad, A.~Mohamed, O.~Levy, V.~Stoyanov,
  and L.~Zettlemoyer, ``{BART:} denoising sequence-to-sequence pre-training for
  natural language generation, translation, and comprehension,'' in
  \emph{{ACL}}.\hskip 1em plus 0.5em minus 0.4em\relax Association for
  Computational Linguistics, 2020, pp. 7871--7880.

\bibitem{DBLP:journals/jmlr/RaffelSRLNMZLL20}
C.~Raffel, N.~Shazeer, A.~Roberts, K.~Lee, S.~Narang, M.~Matena, Y.~Zhou,
  W.~Li, and P.~J. Liu, ``Exploring the limits of transfer learning with a
  unified text-to-text transformer,'' \emph{J. Mach. Learn. Res.}, vol.~21, pp.
  140:1--140:67, 2020.

\bibitem{DBLP:journals/jmlr/ChowdheryNDBMRBCSGSSTMRBTSPRDHPBAI23}
A.~Chowdhery, S.~Narang, J.~Devlin, M.~Bosma, G.~Mishra, A.~Roberts, P.~Barham,
  H.~W. Chung, C.~Sutton, S.~Gehrmann, P.~Schuh, K.~Shi, S.~Tsvyashchenko,
  J.~Maynez, A.~Rao, P.~Barnes, Y.~Tay, N.~Shazeer, V.~Prabhakaran, E.~Reif,
  N.~Du, B.~Hutchinson, R.~Pope, J.~Bradbury, J.~Austin, M.~Isard,
  G.~Gur{-}Ari, P.~Yin, T.~Duke, A.~Levskaya, S.~Ghemawat, S.~Dev,
  H.~Michalewski, X.~Garcia, V.~Misra, K.~Robinson, L.~Fedus, D.~Zhou,
  D.~Ippolito, D.~Luan, H.~Lim, B.~Zoph, A.~Spiridonov, R.~Sepassi, D.~Dohan,
  S.~Agrawal, M.~Omernick, A.~M. Dai, T.~S. Pillai, M.~Pellat, A.~Lewkowycz,
  E.~Moreira, R.~Child, O.~Polozov, K.~Lee, Z.~Zhou, X.~Wang, B.~Saeta,
  M.~Diaz, O.~Firat, M.~Catasta, J.~Wei, K.~Meier{-}Hellstern, D.~Eck, J.~Dean,
  S.~Petrov, and N.~Fiedel, ``Palm: Scaling language modeling with pathways,''
  \emph{J. Mach. Learn. Res.}, vol.~24, pp. 240:1--240:113, 2023.

\bibitem{DBLP:journals/corr/abs-2305-10403}
R.~Anil, A.~M. Dai, O.~Firat, M.~Johnson, D.~Lepikhin, A.~Passos, S.~Shakeri,
  E.~Taropa, P.~Bailey, Z.~Chen, E.~Chu, J.~H. Clark, L.~E. Shafey, Y.~Huang,
  K.~Meier{-}Hellstern, G.~Mishra, E.~Moreira, M.~Omernick, K.~Robinson,
  S.~Ruder, Y.~Tay, K.~Xiao, Y.~Xu, Y.~Zhang, G.~H. {\'{A}}brego, J.~Ahn,
  J.~Austin, P.~Barham, J.~A. Botha, J.~Bradbury, S.~Brahma, K.~Brooks,
  M.~Catasta, Y.~Cheng, C.~Cherry, C.~A. Choquette{-}Choo, A.~Chowdhery,
  C.~Crepy, S.~Dave, M.~Dehghani, S.~Dev, J.~Devlin, M.~D{\'{\i}}az, N.~Du,
  E.~Dyer, V.~Feinberg, F.~Feng, V.~Fienber, M.~Freitag, X.~Garcia,
  S.~Gehrmann, L.~Gonzalez, and et~al., ``Palm 2 technical report,''
  \emph{CoRR}, vol. abs/2305.10403, 2023.

\bibitem{DBLP:journals/corr/abs-2302-13971}
H.~Touvron, T.~Lavril, G.~Izacard, X.~Martinet, M.~Lachaux, T.~Lacroix,
  B.~Rozi{\`{e}}re, N.~Goyal, E.~Hambro, F.~Azhar, A.~Rodriguez, A.~Joulin,
  E.~Grave, and G.~Lample, ``Llama: Open and efficient foundation language
  models,'' \emph{CoRR}, vol. abs/2302.13971, 2023.

\bibitem{DBLP:journals/corr/abs-2307-09288}
H.~Touvron, L.~Martin, K.~Stone, P.~Albert, A.~Almahairi, Y.~Babaei,
  N.~Bashlykov, S.~Batra, P.~Bhargava, S.~Bhosale, D.~Bikel, L.~Blecher,
  C.~Canton{-}Ferrer, M.~Chen, G.~Cucurull, D.~Esiobu, J.~Fernandes, J.~Fu,
  W.~Fu, B.~Fuller, C.~Gao, V.~Goswami, N.~Goyal, A.~Hartshorn, S.~Hosseini,
  R.~Hou, H.~Inan, M.~Kardas, V.~Kerkez, M.~Khabsa, I.~Kloumann, A.~Korenev,
  P.~S. Koura, M.~Lachaux, T.~Lavril, J.~Lee, D.~Liskovich, Y.~Lu, Y.~Mao,
  X.~Martinet, T.~Mihaylov, P.~Mishra, I.~Molybog, Y.~Nie, A.~Poulton,
  J.~Reizenstein, R.~Rungta, K.~Saladi, A.~Schelten, R.~Silva, E.~M. Smith,
  R.~Subramanian, X.~E. Tan, B.~Tang, R.~Taylor, A.~Williams, J.~X. Kuan,
  P.~Xu, Z.~Yan, I.~Zarov, Y.~Zhang, A.~Fan, M.~Kambadur, S.~Narang,
  A.~Rodriguez, R.~Stojnic, S.~Edunov, and T.~Scialom, ``Llama 2: Open
  foundation and fine-tuned chat models,'' \emph{CoRR}, vol. abs/2307.09288,
  2023.

\bibitem{ernie35-baidu}
\BIBentryALTinterwordspacing
Baidu. (2023) Introducing ernie 3.5: Baidu’s knowledge-enhanced foundation
  model takes a giant leap forward. [Online]. Available:
  \url{http://research.baidu.com/Blog/index-view?id=185}
\BIBentrySTDinterwordspacing

\bibitem{DBLP:conf/nips/Ouyang0JAWMZASR22}
L.~Ouyang, J.~Wu, X.~Jiang, D.~Almeida, C.~L. Wainwright, P.~Mishkin, C.~Zhang,
  S.~Agarwal, K.~Slama, A.~Ray, J.~Schulman, J.~Hilton, F.~Kelton, L.~Miller,
  M.~Simens, A.~Askell, P.~Welinder, P.~F. Christiano, J.~Leike, and R.~Lowe,
  ``Training language models to follow instructions with human feedback,'' in
  \emph{NeurIPS}, 2022.

\bibitem{DBLP:conf/iclr/WeiBZGYLDDL22}
J.~Wei, M.~Bosma, V.~Y. Zhao, K.~Guu, A.~W. Yu, B.~Lester, N.~Du, A.~M. Dai,
  and Q.~V. Le, ``Finetuned language models are zero-shot learners,'' in
  \emph{{ICLR}}, 2022.

\bibitem{DBLP:journals/corr/abs-2210-11416}
H.~W. Chung, L.~Hou, S.~Longpre, B.~Zoph, Y.~Tay, W.~Fedus, E.~Li, X.~Wang,
  M.~Dehghani, S.~Brahma, A.~Webson, S.~S. Gu, Z.~Dai, M.~Suzgun, X.~Chen,
  A.~Chowdhery, S.~Narang, G.~Mishra, A.~Yu, V.~Y. Zhao, Y.~Huang, A.~M. Dai,
  H.~Yu, S.~Petrov, E.~H. Chi, J.~Dean, J.~Devlin, A.~Roberts, D.~Zhou, Q.~V.
  Le, and J.~Wei, ``Scaling instruction-finetuned language models,''
  \emph{CoRR}, vol. abs/2210.11416, 2022.

\bibitem{alpaca}
R.~Taori, I.~Gulrajani, T.~Zhang, Y.~Dubois, X.~Li, C.~Guestrin, P.~Liang, and
  T.~B. Hashimoto, ``Stanford alpaca: An instruction-following llama model,''
  \url{https://github.com/tatsu-lab/stanford_alpaca}, 2023.

\bibitem{vicuna2023}
\BIBentryALTinterwordspacing
W.-L. Chiang, Z.~Li, Z.~Lin, Y.~Sheng, Z.~Wu, H.~Zhang, L.~Zheng, S.~Zhuang,
  Y.~Zhuang, J.~E. Gonzalez, I.~Stoica, and E.~P. Xing, ``Vicuna: An
  open-source chatbot impressing gpt-4 with 90\%* chatgpt quality,'' March
  2023. [Online]. Available: \url{https://lmsys.org/blog/2023-03-30-vicuna/}
\BIBentrySTDinterwordspacing

\bibitem{DBLP:journals/corr/Li17b}
Y.~Li, ``Deep reinforcement learning: An overview,'' \emph{CoRR}, vol.
  abs/1701.07274, 2017.

\bibitem{DBLP:journals/corr/abs-1708-05866}
K.~Arulkumaran, M.~P. Deisenroth, M.~Brundage, and A.~A. Bharath, ``A brief
  survey of deep reinforcement learning,'' \emph{CoRR}, vol. abs/1708.05866,
  2017.

\bibitem{DBLP:journals/corr/abs-2301-03044}
W.~Li, H.~Luo, Z.~Lin, C.~Zhang, Z.~Lu, and D.~Ye, ``A survey on transformers
  in reinforcement learning,'' \emph{CoRR}, vol. abs/2301.03044, 2023.

\bibitem{DBLP:conf/aaai/HasseltGS16}
H.~van Hasselt, A.~Guez, and D.~Silver, ``Deep reinforcement learning with
  double q-learning,'' in \emph{{AAAI}}.\hskip 1em plus 0.5em minus 0.4em\relax
  {AAAI} Press, 2016, pp. 2094--2100.

\bibitem{DBLP:conf/nips/AndrychowiczCRS17}
M.~Andrychowicz, D.~Crow, A.~Ray, J.~Schneider, R.~Fong, P.~Welinder,
  B.~McGrew, J.~Tobin, P.~Abbeel, and W.~Zaremba, ``Hindsight experience
  replay,'' in \emph{{NIPS}}, 2017, pp. 5048--5058.

\bibitem{DBLP:conf/icml/FujimotoMP19}
S.~Fujimoto, D.~Meger, and D.~Precup, ``Off-policy deep reinforcement learning
  without exploration,'' in \emph{{ICML}}, vol.~97.\hskip 1em plus 0.5em minus
  0.4em\relax {PMLR}, 2019, pp. 2052--2062.

\bibitem{DBLP:conf/nips/KumarFSTL19}
A.~Kumar, J.~Fu, M.~Soh, G.~Tucker, and S.~Levine, ``Stabilizing off-policy
  q-learning via bootstrapping error reduction,'' in \emph{NeurIPS}, 2019, pp.
  11\,761--11\,771.

\bibitem{DBLP:conf/nips/KumarZTL20}
A.~Kumar, A.~Zhou, G.~Tucker, and S.~Levine, ``Conservative q-learning for
  offline reinforcement learning,'' in \emph{NeurIPS}, 2020.

\bibitem{DBLP:conf/icml/LevineK13}
S.~Levine and V.~Koltun, ``Guided policy search,'' in \emph{{ICML} {(3)}}, ser.
  {JMLR} Workshop and Conference Proceedings, vol.~28.\hskip 1em plus 0.5em
  minus 0.4em\relax JMLR.org, 2013, pp. 1--9.

\bibitem{DBLP:conf/icml/SilverLHDWR14}
D.~Silver, G.~Lever, N.~Heess, T.~Degris, D.~Wierstra, and M.~A. Riedmiller,
  ``Deterministic policy gradient algorithms,'' in \emph{{ICML}}, ser. {JMLR}
  Workshop and Conference Proceedings, vol.~32.\hskip 1em plus 0.5em minus
  0.4em\relax JMLR.org, 2014, pp. 387--395.

\bibitem{DBLP:journals/corr/LillicrapHPHETS15}
T.~P. Lillicrap, J.~J. Hunt, A.~Pritzel, N.~Heess, T.~Erez, Y.~Tassa,
  D.~Silver, and D.~Wierstra, ``Continuous control with deep reinforcement
  learning,'' in \emph{{ICLR} (Poster)}, 2016.

\bibitem{DBLP:conf/icml/MnihBMGLHSK16}
V.~Mnih, A.~P. Badia, M.~Mirza, A.~Graves, T.~P. Lillicrap, T.~Harley,
  D.~Silver, and K.~Kavukcuoglu, ``Asynchronous methods for deep reinforcement
  learning,'' in \emph{{ICML}}, ser. {JMLR} Workshop and Conference
  Proceedings, vol.~48.\hskip 1em plus 0.5em minus 0.4em\relax JMLR.org, 2016,
  pp. 1928--1937.

\bibitem{DBLP:conf/icml/GuLSL16}
S.~Gu, T.~P. Lillicrap, I.~Sutskever, and S.~Levine, ``Continuous deep
  q-learning with model-based acceleration,'' in \emph{{ICML}}, ser. {JMLR}
  Workshop and Conference Proceedings, vol.~48.\hskip 1em plus 0.5em minus
  0.4em\relax JMLR.org, 2016, pp. 2829--2838.

\bibitem{DBLP:conf/icml/HaarnojaZAL18}
T.~Haarnoja, A.~Zhou, P.~Abbeel, and S.~Levine, ``Soft actor-critic: Off-policy
  maximum entropy deep reinforcement learning with a stochastic actor,'' in
  \emph{{ICML}}, vol.~80.\hskip 1em plus 0.5em minus 0.4em\relax {PMLR}, 2018,
  pp. 1856--1865.

\bibitem{DBLP:conf/icml/SchulmanLAJM15}
J.~Schulman, S.~Levine, P.~Abbeel, M.~I. Jordan, and P.~Moritz, ``Trust region
  policy optimization,'' in \emph{{ICML}}, ser. {JMLR} Workshop and Conference
  Proceedings, vol.~37.\hskip 1em plus 0.5em minus 0.4em\relax JMLR.org, 2015,
  pp. 1889--1897.

\bibitem{DBLP:journals/corr/SchulmanMLJA15}
J.~Schulman, P.~Moritz, S.~Levine, M.~I. Jordan, and P.~Abbeel,
  ``High-dimensional continuous control using generalized advantage
  estimation,'' in \emph{{ICLR} (Poster)}, 2016.

\bibitem{DBLP:conf/nips/HoE16}
J.~Ho and S.~Ermon, ``Generative adversarial imitation learning,'' in
  \emph{{NIPS}}, 2016, pp. 4565--4573.

\bibitem{DBLP:conf/icml/PintoDSG17}
L.~Pinto, J.~Davidson, R.~Sukthankar, and A.~Gupta, ``Robust adversarial
  reinforcement learning,'' in \emph{{ICML}}, vol.~70.\hskip 1em plus 0.5em
  minus 0.4em\relax {PMLR}, 2017, pp. 2817--2826.

\bibitem{DBLP:conf/nips/ChristianoLBMLA17}
P.~F. Christiano, J.~Leike, T.~B. Brown, M.~Martic, S.~Legg, and D.~Amodei,
  ``Deep reinforcement learning from human preferences,'' in \emph{{NIPS}},
  2017, pp. 4299--4307.

\bibitem{DBLP:conf/icml/VezhnevetsOSHJS17}
A.~S. Vezhnevets, S.~Osindero, T.~Schaul, N.~Heess, M.~Jaderberg, D.~Silver,
  and K.~Kavukcuoglu, ``Feudal networks for hierarchical reinforcement
  learning,'' in \emph{{ICML}}, vol.~70.\hskip 1em plus 0.5em minus 0.4em\relax
  {PMLR}, 2017, pp. 3540--3549.

\bibitem{DBLP:journals/jmlr/LevineFDA16}
S.~Levine, C.~Finn, T.~Darrell, and P.~Abbeel, ``End-to-end training of deep
  visuomotor policies,'' \emph{J. Mach. Learn. Res.}, vol.~17, pp. 39:1--39:40,
  2016.

\bibitem{DBLP:journals/corr/abs-1806-10293}
D.~Kalashnikov, A.~Irpan, P.~Pastor, J.~Ibarz, A.~Herzog, E.~Jang, D.~Quillen,
  E.~Holly, M.~Kalakrishnan, V.~Vanhoucke, and S.~Levine, ``Qt-opt: Scalable
  deep reinforcement learning for vision-based robotic manipulation,''
  \emph{CoRR}, vol. abs/1806.10293, 2018.

\bibitem{DBLP:journals/corr/abs-1910-07113}
OpenAI, I.~Akkaya, M.~Andrychowicz, M.~Chociej, M.~Litwin, B.~McGrew,
  A.~Petron, A.~Paino, M.~Plappert, G.~Powell, R.~Ribas, J.~Schneider,
  N.~Tezak, J.~Tworek, P.~Welinder, L.~Weng, Q.~Yuan, W.~Zaremba, and L.~Zhang,
  ``Solving rubik's cube with a robot hand,'' \emph{CoRR}, vol. abs/1910.07113,
  2019.

\bibitem{DBLP:journals/tnn/ScarselliGTHM09}
F.~Scarselli, M.~Gori, A.~C. Tsoi, M.~Hagenbuchner, and G.~Monfardini, ``The
  graph neural network model,'' \emph{{IEEE} Trans. Neural Networks}, vol.~20,
  no.~1, pp. 61--80, 2009.

\bibitem{DBLP:journals/tnn/WuPCLZY21}
Z.~Wu, S.~Pan, F.~Chen, G.~Long, C.~Zhang, and P.~S. Yu, ``A comprehensive
  survey on graph neural networks,'' \emph{{IEEE} Trans. Neural Networks Learn.
  Syst.}, vol.~32, no.~1, pp. 4--24, 2021.

\bibitem{DBLP:journals/corr/BrunaZSL13}
J.~Bruna, W.~Zaremba, A.~Szlam, and Y.~LeCun, ``Spectral networks and locally
  connected networks on graphs,'' in \emph{{ICLR}}, 2014.

\bibitem{DBLP:conf/nips/DefferrardBV16}
M.~Defferrard, X.~Bresson, and P.~Vandergheynst, ``Convolutional neural
  networks on graphs with fast localized spectral filtering,'' in
  \emph{{NIPS}}, 2016, pp. 3837--3845.

\bibitem{DBLP:conf/iclr/KipfW17}
T.~N. Kipf and M.~Welling, ``Semi-supervised classification with graph
  convolutional networks,'' in \emph{{ICLR} (Poster)}, 2017.

\bibitem{DBLP:journals/tnn/Micheli09}
A.~Micheli, ``Neural network for graphs: {A} contextual constructive
  approach,'' \emph{{IEEE} Trans. Neural Networks}, vol.~20, no.~3, pp.
  498--511, 2009.

\bibitem{DBLP:conf/icml/GilmerSRVD17}
J.~Gilmer, S.~S. Schoenholz, P.~F. Riley, O.~Vinyals, and G.~E. Dahl, ``Neural
  message passing for quantum chemistry,'' in \emph{{ICML}}, vol.~70.\hskip 1em
  plus 0.5em minus 0.4em\relax {PMLR}, 2017, pp. 1263--1272.

\bibitem{DBLP:conf/iclr/XuHLJ19}
K.~Xu, W.~Hu, J.~Leskovec, and S.~Jegelka, ``How powerful are graph neural
  networks?'' in \emph{{ICLR}}, 2019.

\bibitem{DBLP:conf/nips/HamiltonYL17}
W.~L. Hamilton, Z.~Ying, and J.~Leskovec, ``Inductive representation learning
  on large graphs,'' in \emph{{NIPS}}, 2017, pp. 1024--1034.

\bibitem{DBLP:conf/iclr/VelickovicCCRLB18}
P.~Velickovic, G.~Cucurull, A.~Casanova, A.~Romero, P.~Li{\`{o}}, and
  Y.~Bengio, ``Graph attention networks,'' in \emph{{ICLR} (Poster)}, 2018.

\bibitem{DBLP:conf/aaai/CaoLX16}
S.~Cao, W.~Lu, and Q.~Xu, ``Deep neural networks for learning graph
  representations,'' in \emph{{AAAI}}.\hskip 1em plus 0.5em minus 0.4em\relax
  {AAAI} Press, 2016, pp. 1145--1152.

\bibitem{DBLP:journals/corr/KipfW16a}
T.~N. Kipf and M.~Welling, ``Variational graph auto-encoders,'' \emph{CoRR},
  vol. abs/1611.07308, 2016.

\bibitem{DBLP:conf/iconip/SeoDVB18}
Y.~Seo, M.~Defferrard, P.~Vandergheynst, and X.~Bresson, ``Structured sequence
  modeling with graph convolutional recurrent networks,'' in \emph{{ICONIP}
  {(1)}}, ser. Lecture Notes in Computer Science, vol. 11301.\hskip 1em plus
  0.5em minus 0.4em\relax Springer, 2018, pp. 362--373.

\bibitem{DBLP:conf/icml/DuZDMC0SL22}
W.~Du, H.~Zhang, Y.~Du, Q.~Meng, W.~Chen, N.~Zheng, B.~Shao, and T.~Liu,
  ``{SE(3)} equivariant graph neural networks with complete local frames,'' in
  \emph{{ICML}}, vol. 162.\hskip 1em plus 0.5em minus 0.4em\relax {PMLR}, 2022,
  pp. 5583--5608.

\bibitem{DBLP:conf/icml/SatorrasHW21}
V.~G. Satorras, E.~Hoogeboom, and M.~Welling, ``E(n) equivariant graph neural
  networks,'' in \emph{{ICML}}, vol. 139.\hskip 1em plus 0.5em minus
  0.4em\relax {PMLR}, 2021, pp. 9323--9332.

\bibitem{DBLP:conf/nips/RongBXX0HH20}
Y.~Rong, Y.~Bian, T.~Xu, W.~Xie, Y.~Wei, W.~Huang, and J.~Huang,
  ``Self-supervised graph transformer on large-scale molecular data,'' in
  \emph{NeurIPS}, 2020.

\bibitem{DBLP:conf/nips/FuchsW0W20}
F.~Fuchs, D.~E. Worrall, V.~Fischer, and M.~Welling, ``Se(3)-transformers: 3d
  roto-translation equivariant attention networks,'' in \emph{NeurIPS}, 2020.

\bibitem{DBLP:journals/ijcv/KrishnaZGJHKCKL17}
R.~Krishna, Y.~Zhu, O.~Groth, J.~Johnson, K.~Hata, J.~Kravitz, S.~Chen,
  Y.~Kalantidis, L.~Li, D.~A. Shamma, M.~S. Bernstein, and L.~Fei{-}Fei,
  ``Visual genome: Connecting language and vision using crowdsourced dense
  image annotations,'' \emph{Int. J. Comput. Vis.}, vol. 123, no.~1, pp.
  32--73, 2017.

\bibitem{DBLP:journals/ftml/WuCSGGLPL23}
L.~Wu, Y.~Chen, K.~Shen, X.~Guo, H.~Gao, S.~Li, J.~Pei, and B.~Long, ``Graph
  neural networks for natural language processing: {A} survey,'' \emph{Found.
  Trends Mach. Learn.}, vol.~16, no.~2, pp. 119--328, 2023.

\bibitem{DBLP:conf/emnlp/GhosalMPCG19}
D.~Ghosal, N.~Majumder, S.~Poria, N.~Chhaya, and A.~F. Gelbukh, ``Dialoguegcn:
  {A} graph convolutional neural network for emotion recognition in
  conversation,'' in \emph{{EMNLP/IJCNLP} {(1)}}.\hskip 1em plus 0.5em minus
  0.4em\relax Association for Computational Linguistics, 2019, pp. 154--164.

\bibitem{DBLP:conf/acl/ZhongXTXDZWY20}
W.~Zhong, J.~Xu, D.~Tang, Z.~Xu, N.~Duan, M.~Zhou, J.~Wang, and J.~Yin,
  ``Reasoning over semantic-level graph for fact checking,'' in
  \emph{{ACL}}.\hskip 1em plus 0.5em minus 0.4em\relax Association for
  Computational Linguistics, 2020, pp. 6170--6180.

\bibitem{DBLP:journals/corr/abs-2107-09556}
L.~Wang, Y.~Li, {\"{O}}.~Aslan, and O.~Vinyals, ``Wikigraphs: {A} wikipedia
  text - knowledge graph paired dataset,'' \emph{CoRR}, vol. abs/2107.09556,
  2021.

\bibitem{DBLP:conf/kdd/TangZYLZS08}
J.~Tang, J.~Zhang, L.~Yao, J.~Li, L.~Zhang, and Z.~Su, ``Arnetminer: extraction
  and mining of academic social networks,'' in \emph{{KDD}}.\hskip 1em plus
  0.5em minus 0.4em\relax {ACM}, 2008, pp. 990--998.

\bibitem{DBLP:conf/emnlp/TanB19}
H.~Tan and M.~Bansal, ``{LXMERT:} learning cross-modality encoder
  representations from transformers,'' in \emph{{EMNLP/IJCNLP} {(1)}}.\hskip
  1em plus 0.5em minus 0.4em\relax Association for Computational Linguistics,
  2019, pp. 5099--5110.

\bibitem{DBLP:journals/corr/abs-1908-03557}
L.~H. Li, M.~Yatskar, D.~Yin, C.~Hsieh, and K.~Chang, ``Visualbert: {A} simple
  and performant baseline for vision and language,'' \emph{CoRR}, vol.
  abs/1908.03557, 2019.

\bibitem{DBLP:conf/iclr/SuZCLLWD20}
W.~Su, X.~Zhu, Y.~Cao, B.~Li, L.~Lu, F.~Wei, and J.~Dai, ``{VL-BERT:}
  pre-training of generic visual-linguistic representations,'' in
  \emph{{ICLR}}, 2020.

\bibitem{DBLP:conf/eccv/ChenLYK0G0020}
Y.~Chen, L.~Li, L.~Yu, A.~E. Kholy, F.~Ahmed, Z.~Gan, Y.~Cheng, and J.~Liu,
  ``{UNITER:} universal image-text representation learning,'' in \emph{{ECCV}
  {(30)}}, ser. Lecture Notes in Computer Science, vol. 12375.\hskip 1em plus
  0.5em minus 0.4em\relax Springer, 2020, pp. 104--120.

\bibitem{DBLP:conf/icml/KimSK21}
W.~Kim, B.~Son, and I.~Kim, ``Vilt: Vision-and-language transformer without
  convolution or region supervision,'' in \emph{{ICML}}, vol. 139.\hskip 1em
  plus 0.5em minus 0.4em\relax {PMLR}, 2021, pp. 5583--5594.

\bibitem{DBLP:conf/iclr/WangYYDT022}
Z.~Wang, J.~Yu, A.~W. Yu, Z.~Dai, Y.~Tsvetkov, and Y.~Cao, ``Simvlm: Simple
  visual language model pretraining with weak supervision,'' in \emph{{ICLR}},
  2022.

\bibitem{DBLP:conf/nips/DaiLLT21}
Z.~Dai, H.~Liu, Q.~V. Le, and M.~Tan, ``Coatnet: Marrying convolution and
  attention for all data sizes,'' in \emph{NeurIPS}, 2021, pp. 3965--3977.

\bibitem{DBLP:journals/tmlr/WangYHLLGLLW22}
J.~Wang, Z.~Yang, X.~Hu, L.~Li, K.~Lin, Z.~Gan, Z.~Liu, C.~Liu, and L.~Wang,
  ``{GIT:} {A} generative image-to-text transformer for vision and language,''
  \emph{Trans. Mach. Learn. Res.}, vol. 2022, 2022.

\bibitem{DBLP:journals/corr/abs-2111-11432}
L.~Yuan, D.~Chen, Y.~Chen, N.~Codella, X.~Dai, J.~Gao, H.~Hu, X.~Huang, B.~Li,
  C.~Li, C.~Liu, M.~Liu, Z.~Liu, Y.~Lu, Y.~Shi, L.~Wang, J.~Wang, B.~Xiao,
  Z.~Xiao, J.~Yang, M.~Zeng, L.~Zhou, and P.~Zhang, ``Florence: {A} new
  foundation model for computer vision,'' \emph{CoRR}, vol. abs/2111.11432,
  2021.

\bibitem{DBLP:conf/cvpr/WangBDBPLAMSSW23}
W.~Wang, H.~Bao, L.~Dong, J.~Bjorck, Z.~Peng, Q.~Liu, K.~Aggarwal, O.~K.
  Mohammed, S.~Singhal, S.~Som, and F.~Wei, ``Image as a foreign language:
  {BEIT} pretraining for vision and vision-language tasks,'' in
  \emph{{CVPR}}.\hskip 1em plus 0.5em minus 0.4em\relax {IEEE}, 2023, pp.
  19\,175--19\,186.

\bibitem{DBLP:conf/iclr/YaoHHLNXLLJX22}
L.~Yao, R.~Huang, L.~Hou, G.~Lu, M.~Niu, H.~Xu, X.~Liang, Z.~Li, X.~Jiang, and
  C.~Xu, ``{FILIP:} fine-grained interactive language-image pre-training,'' in
  \emph{{ICLR}}, 2022.

\bibitem{DBLP:conf/icml/JiaYXCPPLSLD21}
C.~Jia, Y.~Yang, Y.~Xia, Y.~Chen, Z.~Parekh, H.~Pham, Q.~V. Le, Y.~Sung, Z.~Li,
  and T.~Duerig, ``Scaling up visual and vision-language representation
  learning with noisy text supervision,'' in \emph{{ICML}}, vol. 139.\hskip 1em
  plus 0.5em minus 0.4em\relax {PMLR}, 2021, pp. 4904--4916.

\bibitem{DBLP:conf/cvpr/XieLHL20}
Q.~Xie, M.~Luong, E.~H. Hovy, and Q.~V. Le, ``Self-training with noisy student
  improves imagenet classification,'' in \emph{{CVPR}}.\hskip 1em plus 0.5em
  minus 0.4em\relax Computer Vision Foundation / {IEEE}, 2020, pp.
  10\,684--10\,695.

\bibitem{DBLP:conf/nips/LiSGJXH21}
J.~Li, R.~R. Selvaraju, A.~Gotmare, S.~R. Joty, C.~Xiong, and S.~C. Hoi,
  ``Align before fuse: Vision and language representation learning with
  momentum distillation,'' in \emph{NeurIPS}, 2021, pp. 9694--9705.

\bibitem{DBLP:conf/cvpr/SinghHGCGRK22}
A.~Singh, R.~Hu, V.~Goswami, G.~Couairon, W.~Galuba, M.~Rohrbach, and D.~Kiela,
  ``{FLAVA:} {A} foundational language and vision alignment model,'' in
  \emph{{CVPR}}.\hskip 1em plus 0.5em minus 0.4em\relax {IEEE}, 2022, pp.
  15\,617--15\,629.

\bibitem{DBLP:conf/iccv/LiuL00W0LG21}
Z.~Liu, Y.~Lin, Y.~Cao, H.~Hu, Y.~Wei, Z.~Zhang, S.~Lin, and B.~Guo, ``Swin
  transformer: Hierarchical vision transformer using shifted windows,'' in
  \emph{{ICCV}}.\hskip 1em plus 0.5em minus 0.4em\relax {IEEE}, 2021, pp.
  9992--10\,002.

\bibitem{DBLP:conf/iccv/WuXCLDY021}
H.~Wu, B.~Xiao, N.~Codella, M.~Liu, X.~Dai, L.~Yuan, and L.~Zhang, ``Cvt:
  Introducing convolutions to vision transformers,'' in \emph{{ICCV}}.\hskip
  1em plus 0.5em minus 0.4em\relax {IEEE}, 2021, pp. 22--31.

\bibitem{DBLP:conf/icml/BrockDSS21}
A.~Brock, S.~De, S.~L. Smith, and K.~Simonyan, ``High-performance large-scale
  image recognition without normalization,'' in \emph{{ICML}}, vol. 139.\hskip
  1em plus 0.5em minus 0.4em\relax {PMLR}, 2021, pp. 1059--1071.

\bibitem{DBLP:journals/corr/abs-2203-15556}
J.~Hoffmann, S.~Borgeaud, A.~Mensch, E.~Buchatskaya, T.~Cai, E.~Rutherford,
  D.~de~Las~Casas, L.~A. Hendricks, J.~Welbl, A.~Clark, T.~Hennigan, E.~Noland,
  K.~Millican, G.~van~den Driessche, B.~Damoc, A.~Guy, S.~Osindero,
  K.~Simonyan, E.~Elsen, J.~W. Rae, O.~Vinyals, and L.~Sifre, ``Training
  compute-optimal large language models,'' \emph{CoRR}, vol. abs/2203.15556,
  2022.

\bibitem{DBLP:conf/cvpr/FangWXSWW0WC23}
Y.~Fang, W.~Wang, B.~Xie, Q.~Sun, L.~Wu, X.~Wang, T.~Huang, X.~Wang, and
  Y.~Cao, ``{EVA:} exploring the limits of masked visual representation
  learning at scale,'' in \emph{{CVPR}}.\hskip 1em plus 0.5em minus 0.4em\relax
  {IEEE}, 2023, pp. 19\,358--19\,369.

\bibitem{DBLP:conf/iclr/Chen0CPPSGGMB0P23}
X.~Chen, X.~Wang, S.~Changpinyo, A.~J. Piergiovanni, P.~Padlewski, D.~Salz,
  S.~Goodman, A.~Grycner, B.~Mustafa, L.~Beyer, A.~Kolesnikov, J.~Puigcerver,
  N.~Ding, K.~Rong, H.~Akbari, G.~Mishra, L.~Xue, A.~V. Thapliyal, J.~Bradbury,
  and W.~Kuo, ``Pali: {A} jointly-scaled multilingual language-image model,''
  in \emph{{ICLR}}, 2023.

\bibitem{DBLP:conf/naacl/XueCRKASBR21}
L.~Xue, N.~Constant, A.~Roberts, M.~Kale, R.~Al{-}Rfou, A.~Siddhant, A.~Barua,
  and C.~Raffel, ``mt5: {A} massively multilingual pre-trained text-to-text
  transformer,'' in \emph{{NAACL-HLT}}.\hskip 1em plus 0.5em minus 0.4em\relax
  Association for Computational Linguistics, 2021, pp. 483--498.

\bibitem{DBLP:journals/corr/abs-2305-18565}
X.~Chen, J.~Djolonga, P.~Padlewski, B.~Mustafa, S.~Changpinyo, J.~Wu, C.~R.
  Ruiz, S.~Goodman, X.~Wang, Y.~Tay, S.~Shakeri, M.~Dehghani, D.~Salz,
  M.~Lucic, M.~Tschannen, A.~Nagrani, H.~Hu, M.~Joshi, B.~Pang, C.~Montgomery,
  P.~Pietrzyk, M.~Ritter, A.~J. Piergiovanni, M.~Minderer, F.~Pavetic,
  A.~Waters, G.~Li, I.~Alabdulmohsin, L.~Beyer, J.~Amelot, K.~Lee, A.~P.
  Steiner, Y.~Li, D.~Keysers, A.~Arnab, Y.~Xu, K.~Rong, A.~Kolesnikov,
  M.~Seyedhosseini, A.~Angelova, X.~Zhai, N.~Houlsby, and R.~Soricut, ``Pali-x:
  On scaling up a multilingual vision and language model,'' \emph{CoRR}, vol.
  abs/2305.18565, 2023.

\bibitem{DBLP:journals/corr/abs-2302-05442}
M.~Dehghani, J.~Djolonga, B.~Mustafa, P.~Padlewski, J.~Heek, J.~Gilmer,
  A.~Steiner, M.~Caron, R.~Geirhos, I.~Alabdulmohsin, R.~Jenatton, L.~Beyer,
  M.~Tschannen, A.~Arnab, X.~Wang, C.~Riquelme, M.~Minderer, J.~Puigcerver,
  U.~Evci, M.~Kumar, S.~van Steenkiste, G.~F. Elsayed, A.~Mahendran, F.~Yu,
  A.~Oliver, F.~Huot, J.~Bastings, M.~P. Collier, A.~A. Gritsenko, V.~Birodkar,
  C.~Vasconcelos, Y.~Tay, T.~Mensink, A.~Kolesnikov, F.~Pavetic, D.~Tran,
  T.~Kipf, M.~Lucic, X.~Zhai, D.~Keysers, J.~Harmsen, and N.~Houlsby, ``Scaling
  vision transformers to 22 billion parameters,'' \emph{CoRR}, vol.
  abs/2302.05442, 2023.

\bibitem{DBLP:conf/iclr/Tay00GW0CBSZZHM23}
Y.~Tay, M.~Dehghani, V.~Q. Tran, X.~Garcia, J.~Wei, X.~Wang, H.~W. Chung,
  D.~Bahri, T.~Schuster, H.~S. Zheng, D.~Zhou, N.~Houlsby, and D.~Metzler,
  ``{UL2:} unifying language learning paradigms,'' in \emph{{ICLR}}, 2023.

\bibitem{DBLP:journals/corr/abs-2303-16199}
R.~Zhang, J.~Han, A.~Zhou, X.~Hu, S.~Yan, P.~Lu, H.~Li, P.~Gao, and Y.~Qiao,
  ``Llama-adapter: Efficient fine-tuning of language models with zero-init
  attention,'' \emph{CoRR}, vol. abs/2303.16199, 2023.

\bibitem{DBLP:journals/corr/abs-2304-15010}
P.~Gao, J.~Han, R.~Zhang, Z.~Lin, S.~Geng, A.~Zhou, W.~Zhang, P.~Lu, C.~He,
  X.~Yue, H.~Li, and Y.~Qiao, ``Llama-adapter {V2:} parameter-efficient visual
  instruction model,'' \emph{CoRR}, vol. abs/2304.15010, 2023.

\bibitem{DBLP:journals/corr/abs-2302-14045}
S.~Huang, L.~Dong, W.~Wang, Y.~Hao, S.~Singhal, S.~Ma, T.~Lv, L.~Cui, O.~K.
  Mohammed, B.~Patra, Q.~Liu, K.~Aggarwal, Z.~Chi, J.~Bjorck, V.~Chaudhary,
  S.~Som, X.~Song, and F.~Wei, ``Language is not all you need: Aligning
  perception with language models,'' \emph{CoRR}, vol. abs/2302.14045, 2023.

\bibitem{DBLP:journals/corr/abs-2306-14824}
Z.~Peng, W.~Wang, L.~Dong, Y.~Hao, S.~Huang, S.~Ma, and F.~Wei, ``Kosmos-2:
  Grounding multimodal large language models to the world,'' \emph{CoRR}, vol.
  abs/2306.14824, 2023.

\bibitem{DBLP:journals/corr/abs-2210-06423}
H.~Wang, S.~Ma, S.~Huang, L.~Dong, W.~Wang, Z.~Peng, Y.~Wu, P.~Bajaj,
  S.~Singhal, A.~Benhaim, B.~Patra, Z.~Liu, V.~Chaudhary, X.~Song, and F.~Wei,
  ``Foundation transformers,'' \emph{CoRR}, vol. abs/2210.06423, 2022.

\bibitem{DBLP:journals/corr/abs-2305-06500}
W.~Dai, J.~Li, D.~Li, A.~M.~H. Tiong, J.~Zhao, W.~Wang, B.~Li, P.~Fung, and
  S.~C.~H. Hoi, ``Instructblip: Towards general-purpose vision-language models
  with instruction tuning,'' \emph{CoRR}, vol. abs/2305.06500, 2023.

\bibitem{DBLP:journals/corr/abs-2304-10592}
D.~Zhu, J.~Chen, X.~Shen, X.~Li, and M.~Elhoseiny, ``Minigpt-4: Enhancing
  vision-language understanding with advanced large language models,''
  \emph{CoRR}, vol. abs/2304.10592, 2023.

\bibitem{DBLP:journals/corr/abs-2306-02858}
H.~Zhang, X.~Li, and L.~Bing, ``Video-llama: An instruction-tuned audio-visual
  language model for video understanding,'' \emph{CoRR}, vol. abs/2306.02858,
  2023.

\bibitem{DBLP:journals/corr/abs-2305-16355}
Y.~Su, T.~Lan, H.~Li, J.~Xu, Y.~Wang, and D.~Cai, ``Pandagpt: One model to
  instruction-follow them all,'' \emph{CoRR}, vol. abs/2305.16355, 2023.

\bibitem{DBLP:journals/corr/abs-2305-06355}
K.~Li, Y.~He, Y.~Wang, Y.~Li, W.~Wang, P.~Luo, Y.~Wang, L.~Wang, and Y.~Qiao,
  ``Videochat: Chat-centric video understanding,'' \emph{CoRR}, vol.
  abs/2305.06355, 2023.

\bibitem{stablelm}
\BIBentryALTinterwordspacing
S.~contributors. (2023) Stablelm: Stability ai language models. [Online].
  Available: \url{https://github.com/stability-AI/stableLM}
\BIBentrySTDinterwordspacing

\bibitem{DBLP:journals/corr/abs-2307-09474}
L.~Zhao, E.~Yu, Z.~Ge, J.~Yang, H.~Wei, H.~Zhou, J.~Sun, Y.~Peng, R.~Dong,
  C.~Han, and X.~Zhang, ``Chatspot: Bootstrapping multimodal llms via precise
  referring instruction tuning,'' \emph{CoRR}, vol. abs/2307.09474, 2023.

\bibitem{DBLP:journals/corr/abs-2304-14178}
Q.~Ye, H.~Xu, G.~Xu, J.~Ye, M.~Yan, Y.~Zhou, J.~Wang, A.~Hu, P.~Shi, Y.~Shi,
  C.~Li, Y.~Xu, H.~Chen, J.~Tian, Q.~Qi, J.~Zhang, and F.~Huang, ``mplug-owl:
  Modularization empowers large language models with multimodality,''
  \emph{CoRR}, vol. abs/2304.14178, 2023.

\bibitem{DBLP:journals/corr/abs-2311-04257}
Q.~Ye, H.~Xu, J.~Ye, M.~Yan, A.~Hu, H.~Liu, Q.~Qian, J.~Zhang, F.~Huang, and
  J.~Zhou, ``mplug-owl2: Revolutionizing multi-modal large language model with
  modality collaboration,'' \emph{CoRR}, vol. abs/2311.04257, 2023.

\bibitem{DBLP:conf/emnlp/XuGDM23}
C.~Xu, D.~Guo, N.~Duan, and J.~J. McAuley, ``Baize: An open-source chat model
  with parameter-efficient tuning on self-chat data,'' in \emph{{EMNLP}}.\hskip
  1em plus 0.5em minus 0.4em\relax Association for Computational Linguistics,
  2023, pp. 6268--6278.

\bibitem{DBLP:journals/corr/abs-2303-04671}
C.~Wu, S.~Yin, W.~Qi, X.~Wang, Z.~Tang, and N.~Duan, ``Visual chatgpt: Talking,
  drawing and editing with visual foundation models,'' \emph{CoRR}, vol.
  abs/2303.04671, 2023.

\bibitem{DBLP:journals/corr/abs-2305-04160}
F.~Chen, M.~Han, H.~Zhao, Q.~Zhang, J.~Shi, S.~Xu, and B.~Xu, ``{X-LLM:}
  bootstrapping advanced large language models by treating multi-modalities as
  foreign languages,'' \emph{CoRR}, vol. abs/2305.04160, 2023.

\bibitem{DBLP:conf/cvpr/Zhai0HB22}
X.~Zhai, A.~Kolesnikov, N.~Houlsby, and L.~Beyer, ``Scaling vision
  transformers,'' in \emph{{CVPR}}.\hskip 1em plus 0.5em minus 0.4em\relax
  {IEEE}, 2022, pp. 1204--1213.

\bibitem{DBLP:conf/acl/DuQLDQY022}
Z.~Du, Y.~Qian, X.~Liu, M.~Ding, J.~Qiu, Z.~Yang, and J.~Tang, ``{GLM:} general
  language model pretraining with autoregressive blank infilling,'' in
  \emph{{ACL} {(1)}}.\hskip 1em plus 0.5em minus 0.4em\relax Association for
  Computational Linguistics, 2022, pp. 320--335.

\bibitem{DBLP:journals/ijautcomp/ChenZHCSXX23}
F.~Chen, D.~Zhang, M.~Han, X.~Chen, J.~Shi, S.~Xu, and B.~Xu, ``{VLP:} {A}
  survey on vision-language pre-training,'' \emph{Int. J. Autom. Comput.},
  vol.~20, no.~1, pp. 38--56, 2023.

\bibitem{DBLP:journals/pami/XuZC23}
P.~Xu, X.~Zhu, and D.~A. Clifton, ``Multimodal learning with transformers: {A}
  survey,'' \emph{{IEEE} Trans. Pattern Anal. Mach. Intell.}, vol.~45, no.~10,
  pp. 12\,113--12\,132, 2023.

\bibitem{DBLP:journals/ijautcomp/WangCQGWWTG23}
X.~Wang, G.~Chen, G.~Qian, P.~Gao, X.~Wei, Y.~Wang, Y.~Tian, and W.~Gao,
  ``Large-scale multi-modal pre-trained models: {A} comprehensive survey,''
  \emph{Mach. Intell. Res.}, vol.~20, no.~4, pp. 447--482, 2023.

\bibitem{DBLP:journals/corr/abs-2304-00685}
J.~Zhang, J.~Huang, S.~Jin, and S.~Lu, ``Vision-language models for vision
  tasks: {A} survey,'' \emph{CoRR}, vol. abs/2304.00685, 2023.

\bibitem{DBLP:conf/iccv/SunMV0S19}
C.~Sun, A.~Myers, C.~Vondrick, K.~Murphy, and C.~Schmid, ``Videobert: {A} joint
  model for video and language representation learning,'' in
  \emph{{ICCV}}.\hskip 1em plus 0.5em minus 0.4em\relax {IEEE}, 2019, pp.
  7463--7472.

\bibitem{DBLP:conf/nips/BaoW0LMASPW22}
H.~Bao, W.~Wang, L.~Dong, Q.~Liu, O.~K. Mohammed, K.~Aggarwal, S.~Som, S.~Piao,
  and F.~Wei, ``Vlmo: Unified vision-language pre-training with
  mixture-of-modality-experts,'' in \emph{NeurIPS}, 2022.

\bibitem{DBLP:conf/cvpr/ZhaiWMSK0B22}
X.~Zhai, X.~Wang, B.~Mustafa, A.~Steiner, D.~Keysers, A.~Kolesnikov, and
  L.~Beyer, ``Lit: Zero-shot transfer with locked-image text tuning,'' in
  \emph{{CVPR}}.\hskip 1em plus 0.5em minus 0.4em\relax {IEEE}, 2022, pp.
  18\,102--18\,112.

\bibitem{DBLP:conf/nips/TsimpoukelliMCE21}
M.~Tsimpoukelli, J.~Menick, S.~Cabi, S.~M.~A. Eslami, O.~Vinyals, and F.~Hill,
  ``Multimodal few-shot learning with frozen language models,'' in
  \emph{NeurIPS}, 2021, pp. 200--212.

\bibitem{DBLP:journals/tmlr/YuWVYSW22}
J.~Yu, Z.~Wang, V.~Vasudevan, L.~Yeung, M.~Seyedhosseini, and Y.~Wu, ``Coca:
  Contrastive captioners are image-text foundation models,'' \emph{Trans. Mach.
  Learn. Res.}, vol. 2022, 2022.

\bibitem{DBLP:conf/icml/WangYMLBLMZZY22}
P.~Wang, A.~Yang, R.~Men, J.~Lin, S.~Bai, Z.~Li, J.~Ma, C.~Zhou, J.~Zhou, and
  H.~Yang, ``{OFA:} unifying architectures, tasks, and modalities through a
  simple sequence-to-sequence learning framework,'' in \emph{{ICML}}, vol.
  162.\hskip 1em plus 0.5em minus 0.4em\relax {PMLR}, 2022, pp.
  23\,318--23\,340.

\bibitem{DBLP:conf/iclr/HuSWALWWC22}
E.~J. Hu, Y.~Shen, P.~Wallis, Z.~Allen{-}Zhu, Y.~Li, S.~Wang, L.~Wang, and
  W.~Chen, ``Lora: Low-rank adaptation of large language models,'' in
  \emph{{ICLR}}, 2022.

\bibitem{DBLP:conf/cvpr/EsserRO21}
P.~Esser, R.~Rombach, and B.~Ommer, ``Taming transformers for high-resolution
  image synthesis,'' in \emph{{CVPR}}.\hskip 1em plus 0.5em minus 0.4em\relax
  Computer Vision Foundation / {IEEE}, 2021, pp. 12\,873--12\,883.

\bibitem{DBLP:conf/nips/ZhengVCP22}
C.~Zheng, T.~Vuong, J.~Cai, and D.~Phung, ``Movq: Modulating quantized vectors
  for high-fidelity image generation,'' in \emph{NeurIPS}, 2022.

\bibitem{DBLP:conf/iccv/ZhaiM0B23}
X.~Zhai, B.~Mustafa, A.~Kolesnikov, and L.~Beyer, ``Sigmoid loss for language
  image pre-training,'' in \emph{{ICCV}}.\hskip 1em plus 0.5em minus
  0.4em\relax {IEEE}, 2023, pp. 11\,941--11\,952.

\bibitem{DBLP:conf/iclr/GuLKC22}
X.~Gu, T.~Lin, W.~Kuo, and Y.~Cui, ``Open-vocabulary object detection via
  vision and language knowledge distillation,'' in \emph{{ICLR}}, 2022.

\bibitem{DBLP:conf/iccv/KamathSLSMC21}
A.~Kamath, M.~Singh, Y.~LeCun, G.~Synnaeve, I.~Misra, and N.~Carion, ``{MDETR}
  - modulated detection for end-to-end multi-modal understanding,'' in
  \emph{{ICCV}}.\hskip 1em plus 0.5em minus 0.4em\relax {IEEE}, 2021, pp.
  1760--1770.

\bibitem{DBLP:conf/corl/BlukisPFGA21}
V.~Blukis, C.~Paxton, D.~Fox, A.~Garg, and Y.~Artzi, ``A persistent spatial
  semantic representation for high-level natural language instruction
  execution,'' in \emph{CoRL}, vol. 164.\hskip 1em plus 0.5em minus 0.4em\relax
  {PMLR}, 2021, pp. 706--717.

\bibitem{DBLP:conf/cvpr/HeCXLDG22}
K.~He, X.~Chen, S.~Xie, Y.~Li, P.~Doll{\'{a}}r, and R.~B. Girshick, ``Masked
  autoencoders are scalable vision learners,'' in \emph{{CVPR}}.\hskip 1em plus
  0.5em minus 0.4em\relax {IEEE}, 2022, pp. 15\,979--15\,988.

\bibitem{DBLP:journals/corr/abs-2205-06230}
M.~Minderer, A.~A. Gritsenko, A.~Stone, M.~Neumann, D.~Weissenborn,
  A.~Dosovitskiy, A.~Mahendran, A.~Arnab, M.~Dehghani, Z.~Shen, X.~Wang,
  X.~Zhai, T.~Kipf, and N.~Houlsby, ``Simple open-vocabulary object detection
  with vision transformers,'' \emph{CoRR}, vol. abs/2205.06230, 2022.

\bibitem{DBLP:conf/cvpr/0003MWFDX22}
Z.~Liu, H.~Mao, C.~Wu, C.~Feichtenhofer, T.~Darrell, and S.~Xie, ``A convnet
  for the 2020s,'' in \emph{{CVPR}}.\hskip 1em plus 0.5em minus 0.4em\relax
  {IEEE}, 2022, pp. 11\,966--11\,976.

\bibitem{DBLP:journals/corr/abs-2210-02303}
J.~Ho, W.~Chan, C.~Saharia, J.~Whang, R.~Gao, A.~A. Gritsenko, D.~P. Kingma,
  B.~Poole, M.~Norouzi, D.~J. Fleet, and T.~Salimans, ``Imagen video: High
  definition video generation with diffusion models,'' \emph{CoRR}, vol.
  abs/2210.02303, 2022.

\bibitem{DBLP:conf/iclr/VillegasBKM0SCK23}
R.~Villegas, M.~Babaeizadeh, P.~Kindermans, H.~Moraldo, H.~Zhang, M.~T. Saffar,
  S.~Castro, J.~Kunze, and D.~Erhan, ``Phenaki: Variable length video
  generation from open domain textual descriptions,'' in \emph{{ICLR}}.\hskip
  1em plus 0.5em minus 0.4em\relax OpenReview.net, 2023.

\bibitem{DBLP:conf/nips/SajjadiDMSPLGGK22}
M.~S.~M. Sajjadi, D.~Duckworth, A.~Mahendran, S.~van Steenkiste, F.~Pavetic,
  M.~Lucic, L.~J. Guibas, K.~Greff, and T.~Kipf, ``Object scene representation
  transformer,'' in \emph{NeurIPS}, 2022.

\bibitem{DBLP:conf/cvpr/QiSMG17}
C.~R. Qi, H.~Su, K.~Mo, and L.~J. Guibas, ``Pointnet: Deep learning on point
  sets for 3d classification and segmentation,'' in \emph{{CVPR}}.\hskip 1em
  plus 0.5em minus 0.4em\relax {IEEE} Computer Society, 2017, pp. 77--85.

\bibitem{DBLP:conf/nips/QiYSG17}
C.~R. Qi, L.~Yi, H.~Su, and L.~J. Guibas, ``Pointnet++: Deep hierarchical
  feature learning on point sets in a metric space,'' in \emph{{NIPS}}, 2017,
  pp. 5099--5108.

\bibitem{DBLP:conf/cvpr/HongLDCTG23}
Y.~Hong, C.~Lin, Y.~Du, Z.~Chen, J.~B. Tenenbaum, and C.~Gan, ``3d concept
  learning and reasoning from multi-view images,'' in \emph{{CVPR}}.\hskip 1em
  plus 0.5em minus 0.4em\relax {IEEE}, 2023, pp. 9202--9212.

\bibitem{DBLP:conf/rss/JatavallabhulaK23}
K.~M. Jatavallabhula, A.~Kuwajerwala, Q.~Gu, M.~Omama, G.~Iyer, S.~Saryazdi,
  T.~Chen, A.~Maalouf, S.~Li, N.~V. Keetha, A.~Tewari, J.~B. Tenenbaum, C.~M.
  de~Melo, K.~M. Krishna, L.~Paull, F.~Shkurti, and A.~Torralba,
  ``Conceptfusion: Open-set multimodal 3d mapping,'' in \emph{RSS}, 2023.

\bibitem{DBLP:conf/emnlp/ReimersG19}
N.~Reimers and I.~Gurevych, ``Sentence-bert: Sentence embeddings using siamese
  bert-networks,'' in \emph{{EMNLP/IJCNLP} {(1)}}.\hskip 1em plus 0.5em minus
  0.4em\relax Association for Computational Linguistics, 2019, pp. 3980--3990.

\bibitem{DBLP:journals/corr/abs-1910-01108}
V.~Sanh, L.~Debut, J.~Chaumond, and T.~Wolf, ``Distilbert, a distilled version
  of {BERT:} smaller, faster, cheaper and lighter,'' \emph{CoRR}, vol.
  abs/1910.01108, 2019.

\bibitem{MosaicML2023Introducing}
\BIBentryALTinterwordspacing
M.~N. Team. (2023) Introducing mpt-7b: A new standard for open-source,
  commercially usable llms. Accessed: 2023-05-05. [Online]. Available:
  \url{www.mosaicml.com/blog/mpt-7b}
\BIBentrySTDinterwordspacing

\bibitem{DBLP:journals/corr/abs-2308-01390}
A.~Awadalla, I.~Gao, J.~Gardner, J.~Hessel, Y.~Hanafy, W.~Zhu, K.~Marathe,
  Y.~Bitton, S.~Y. Gadre, S.~Sagawa, J.~Jitsev, S.~Kornblith, P.~W. Koh,
  G.~Ilharco, M.~Wortsman, and L.~Schmidt, ``Openflamingo: An open-source
  framework for training large autoregressive vision-language models,''
  \emph{CoRR}, vol. abs/2308.01390, 2023.

\bibitem{DBLP:conf/icml/BidermanSABOHKP23}
S.~Biderman, H.~Schoelkopf, Q.~G. Anthony, H.~Bradley, K.~O'Brien, E.~Hallahan,
  M.~A. Khan, S.~Purohit, U.~S. Prashanth, E.~Raff, A.~Skowron, L.~Sutawika,
  and O.~van~der Wal, ``Pythia: {A} suite for analyzing large language models
  across training and scaling,'' in \emph{{ICML}}, vol. 202.\hskip 1em plus
  0.5em minus 0.4em\relax {PMLR}, 2023, pp. 2397--2430.

\bibitem{javaheripi2023phi}
M.~Javaheripi, S.~Bubeck, M.~Abdin, J.~Aneja, S.~Bubeck, C.~C.~T. Mendes,
  W.~Chen, A.~Del~Giorno, R.~Eldan, S.~Gopi \emph{et~al.}, ``Phi-2: The
  surprising power of small language models,'' \emph{Microsoft Research Blog},
  2023.

\bibitem{DBLP:journals/corr/abs-2204-06745}
S.~Black, S.~Biderman, E.~Hallahan, Q.~Anthony, L.~Gao, L.~Golding, H.~He,
  C.~Leahy, K.~McDonell, J.~Phang, M.~Pieler, U.~S. Prashanth, S.~Purohit,
  L.~Reynolds, J.~Tow, B.~Wang, and S.~Weinbach, ``Gpt-neox-20b: An open-source
  autoregressive language model,'' \emph{CoRR}, vol. abs/2204.06745, 2022.

\bibitem{DBLP:journals/corr/abs-2107-03374}
M.~Chen, J.~Tworek, H.~Jun, Q.~Yuan, H.~P. de~Oliveira~Pinto, J.~Kaplan,
  H.~Edwards, Y.~Burda, N.~Joseph, G.~Brockman, A.~Ray, R.~Puri, G.~Krueger,
  M.~Petrov, H.~Khlaaf, G.~Sastry, P.~Mishkin, B.~Chan, S.~Gray, N.~Ryder,
  M.~Pavlov, A.~Power, L.~Kaiser, M.~Bavarian, C.~Winter, P.~Tillet, F.~P.
  Such, D.~Cummings, M.~Plappert, F.~Chantzis, E.~Barnes, A.~Herbert{-}Voss,
  W.~H. Guss, A.~Nichol, A.~Paino, N.~Tezak, J.~Tang, I.~Babuschkin, S.~Balaji,
  S.~Jain, W.~Saunders, C.~Hesse, A.~N. Carr, J.~Leike, J.~Achiam, V.~Misra,
  E.~Morikawa, A.~Radford, M.~Knight, M.~Brundage, M.~Murati, K.~Mayer,
  P.~Welinder, B.~McGrew, D.~Amodei, S.~McCandlish, I.~Sutskever, and
  W.~Zaremba, ``Evaluating large language models trained on code,''
  \emph{CoRR}, vol. abs/2107.03374, 2021.

\bibitem{DBLP:conf/icml/Karamcheti0BLKS24}
S.~Karamcheti, S.~Nair, A.~Balakrishna, P.~Liang, T.~Kollar, and D.~Sadigh,
  ``Prismatic vlms: Investigating the design space of visually-conditioned
  language models,'' in \emph{{ICML}}, 2024.

\bibitem{DBLP:journals/corr/abs-2407-07726}
L.~Beyer, A.~Steiner, A.~S. Pinto, A.~Kolesnikov, X.~Wang, D.~Salz, M.~Neumann,
  I.~Alabdulmohsin, M.~Tschannen, E.~Bugliarello, T.~Unterthiner, D.~Keysers,
  S.~Koppula, F.~Liu, A.~Grycner, A.~A. Gritsenko, N.~Houlsby, M.~Kumar,
  K.~Rong, J.~Eisenschlos, R.~Kabra, M.~Bauer, M.~Bosnjak, X.~Chen,
  M.~Minderer, P.~Voigtlaender, I.~Bica, I.~Balazevic, J.~Puigcerver,
  P.~Papalampidi, O.~J. H{\'{e}}naff, X.~Xiong, R.~Soricut, J.~Harmsen, and
  X.~Zhai, ``Paligemma: {A} versatile 3b {VLM} for transfer,'' \emph{CoRR},
  vol. abs/2407.07726, 2024.

\bibitem{DBLP:journals/corr/abs-2412-03555}
A.~Steiner, A.~S. Pinto, M.~Tschannen, D.~Keysers, X.~Wang, Y.~Bitton, A.~A.
  Gritsenko, M.~Minderer, A.~Sherbondy, S.~Long, S.~Qin, R.~R. Ingle,
  E.~Bugliarello, S.~Kazemzadeh, T.~Mesnard, I.~Alabdulmohsin, L.~Beyer, and
  X.~Zhai, ``Paligemma 2: {A} family of versatile vlms for transfer,''
  \emph{CoRR}, vol. abs/2412.03555, 2024.

\bibitem{DBLP:journals/corr/abs-2405-09818}
Chameleon-Team, ``Chameleon: Mixed-modal early-fusion foundation models,''
  \emph{CoRR}, vol. abs/2405.09818, 2024.

\bibitem{DBLP:journals/corr/abs-2409-18869}
X.~Wang, X.~Zhang, Z.~Luo, Q.~Sun, Y.~Cui, J.~Wang, F.~Zhang, Y.~Wang, Z.~Li,
  Q.~Yu, Y.~Zhao, Y.~Ao, X.~Min, T.~Li, B.~Wu, B.~Zhao, B.~Zhang, L.~Wang,
  G.~Liu, Z.~He, X.~Yang, J.~Liu, Y.~Lin, T.~Huang, and Z.~Wang, ``Emu3:
  Next-token prediction is all you need,'' \emph{CoRR}, vol. abs/2409.18869,
  2024.

\bibitem{DBLP:conf/iclr/0055YZA25}
H.~Liu, W.~Yan, M.~Zaharia, and P.~Abbeel, ``World model on million-length
  video and language with blockwise ringattention,'' in \emph{{ICLR}}.\hskip
  1em plus 0.5em minus 0.4em\relax OpenReview.net, 2025.

\bibitem{DBLP:conf/emnlp/MisraBBNSA18}
D.~K. Misra, A.~Bennett, V.~Blukis, E.~Niklasson, M.~Shatkhin, and Y.~Artzi,
  ``Mapping instructions to actions in 3d environments with visual goal
  prediction,'' in \emph{{EMNLP}}, 2018, pp. 2667--2678.

\bibitem{DBLP:conf/corl/LynchKXKTLS19}
C.~Lynch, M.~Khansari, T.~Xiao, V.~Kumar, J.~Tompson, S.~Levine, and
  P.~Sermanet, ``Learning latent plans from play,'' in \emph{CoRL}, vol.
  100.\hskip 1em plus 0.5em minus 0.4em\relax {PMLR}, 2019, pp. 1113--1132.

\bibitem{DBLP:conf/corl/BhardwajSMRFRB21}
M.~Bhardwaj, B.~Sundaralingam, A.~Mousavian, N.~D. Ratliff, D.~Fox, F.~Ramos,
  and B.~Boots, ``{STORM:} an integrated framework for fast joint-space
  model-predictive control for reactive manipulation,'' in \emph{CoRL}, vol.
  164.\hskip 1em plus 0.5em minus 0.4em\relax {PMLR}, 2021, pp. 750--759.

\bibitem{DBLP:conf/iclr/JaegleBADIDKZBS22}
A.~Jaegle, S.~Borgeaud, J.~Alayrac, C.~Doersch, C.~Ionescu, D.~Ding,
  S.~Koppula, D.~Zoran, A.~Brock, E.~Shelhamer, O.~J. H{\'{e}}naff, M.~M.
  Botvinick, A.~Zisserman, O.~Vinyals, and J.~Carreira, ``Perceiver {IO:} {A}
  general architecture for structured inputs {\&} outputs,'' in \emph{{ICLR}},
  2022.

\bibitem{DBLP:conf/iclr/WijmansKMLEPSB20}
E.~Wijmans, A.~Kadian, A.~Morcos, S.~Lee, I.~Essa, D.~Parikh, M.~Savva, and
  D.~Batra, ``{DD-PPO:} learning near-perfect pointgoal navigators from 2.5
  billion frames,'' in \emph{{ICLR}}, 2020.

\bibitem{DBLP:journals/corr/abs-2409-14411}
M.~Zhu, Y.~Zhu, J.~Li, J.~Wen, Z.~Xu, N.~Liu, R.~Cheng, C.~Shen, Y.~Peng,
  F.~Feng, and J.~Tang, ``Scaling diffusion policy in transformer to 1 billion
  parameters for robotic manipulation,'' \emph{CoRR}, vol. abs/2409.14411,
  2024.

\bibitem{DBLP:journals/corr/abs-2305-08298}
J.~W. Wei, L.~Hou, A.~K. Lampinen, X.~Chen, D.~Huang, Y.~Tay, X.~Chen, Y.~Lu,
  D.~Zhou, T.~Ma, and Q.~V. Le, ``Symbol tuning improves in-context learning in
  language models,'' \emph{CoRR}, vol. abs/2305.08298, 2023.

\bibitem{DBLP:conf/iccv/PeeblesX23}
W.~Peebles and S.~Xie, ``Scalable diffusion models with transformers,'' in
  \emph{{ICCV}}.\hskip 1em plus 0.5em minus 0.4em\relax {IEEE}, 2023, pp.
  4172--4182.

\bibitem{DBLP:conf/nips/HoJA20}
J.~Ho, A.~Jain, and P.~Abbeel, ``Denoising diffusion probabilistic models,'' in
  \emph{NeurIPS}, 2020.

\bibitem{DBLP:conf/iclr/0011SKKEP21}
Y.~Song, J.~Sohl{-}Dickstein, D.~P. Kingma, A.~Kumar, S.~Ermon, and B.~Poole,
  ``Score-based generative modeling through stochastic differential
  equations,'' in \emph{{ICLR}}, 2021.

\bibitem{DBLP:conf/corl/ZengFTWCAAKDSL20}
A.~Zeng, P.~Florence, J.~Tompson, S.~Welker, J.~Chien, M.~Attarian,
  T.~Armstrong, I.~Krasin, D.~Duong, V.~Sindhwani, and J.~Lee, ``Transporter
  networks: Rearranging the visual world for robotic manipulation,'' in
  \emph{CoRL}, vol. 155.\hskip 1em plus 0.5em minus 0.4em\relax {PMLR}, 2020,
  pp. 726--747.

\bibitem{DBLP:conf/iccv/GoyalKMMWKHFYMH17}
R.~Goyal, S.~E. Kahou, V.~Michalski, J.~Materzynska, S.~Westphal, H.~Kim,
  V.~Haenel, I.~Fr{\"{u}}nd, P.~Yianilos, M.~Mueller{-}Freitag, F.~Hoppe,
  C.~Thurau, I.~Bax, and R.~Memisevic, ``The "something something" video
  database for learning and evaluating visual common sense,'' in
  \emph{{ICCV}}.\hskip 1em plus 0.5em minus 0.4em\relax {IEEE} Computer
  Society, 2017, pp. 5843--5851.

\bibitem{DBLP:journals/corr/abs-2307-03567}
X.~Lin, J.~So, S.~Mahalingam, F.~Liu, and P.~Abbeel, ``Spawnnet: Learning
  generalizable visuomotor skills from pre-trained networks,'' \emph{CoRR},
  vol. abs/2307.03567, 2023.

\bibitem{DBLP:conf/icra/ArunachalamGCP23}
S.~P. Arunachalam, I.~G{\"{u}}zey, S.~Chintala, and L.~Pinto, ``Holo-dex:
  Teaching dexterity with immersive mixed reality,'' in \emph{{ICRA}}.\hskip
  1em plus 0.5em minus 0.4em\relax {IEEE}, 2023, pp. 5962--5969.

\bibitem{DBLP:journals/corr/abs-2311-16098}
N.~M.~M. Shafiullah, A.~Rai, H.~Etukuru, Y.~Liu, I.~Misra, S.~Chintala, and
  L.~Pinto, ``On bringing robots home,'' \emph{CoRR}, vol. abs/2311.16098,
  2023.

\bibitem{DBLP:conf/corl/SeoHLLJLA22}
Y.~Seo, D.~Hafner, H.~Liu, F.~Liu, S.~James, K.~Lee, and P.~Abbeel, ``Masked
  world models for visual control,'' in \emph{CoRL}, vol. 205.\hskip 1em plus
  0.5em minus 0.4em\relax {PMLR}, 2022, pp. 1332--1344.

\bibitem{DBLP:conf/rss/MendoncaBP23}
R.~Mendonca, S.~Bahl, and D.~Pathak, ``Structured world models from human
  videos,'' in \emph{RSS}, 2023.

\bibitem{DBLP:conf/nips/Pan0WY22}
M.~Pan, X.~Zhu, Y.~Wang, and X.~Yang, ``Iso-dream: Isolating and leveraging
  noncontrollable visual dynamics in world models,'' in \emph{NeurIPS}, 2022.

\bibitem{DBLP:conf/acl/DaiYYCLS19}
Z.~Dai, Z.~Yang, Y.~Yang, J.~G. Carbonell, Q.~V. Le, and R.~Salakhutdinov,
  ``Transformer-xl: Attentive language models beyond a fixed-length context,''
  in \emph{{ACL} {(1)}}.\hskip 1em plus 0.5em minus 0.4em\relax Association for
  Computational Linguistics, 2019, pp. 2978--2988.

\bibitem{DBLP:journals/corr/abs-2403-13064}
A.~Avetisyan, C.~Xie, H.~Howard{-}Jenkins, T.~Yang, S.~Aroudj, S.~Patra,
  F.~Zhang, D.~P. Frost, L.~Holland, C.~Orme, J.~Engel, E.~Miller, R.~A.
  Newcombe, and V.~Balntas, ``Scenescript: Reconstructing scenes with an
  autoregressive structured language model,'' \emph{CoRR}, vol. abs/2403.13064,
  2024.

\bibitem{DBLP:conf/nips/ShafiullahCAP22}
N.~M. Shafiullah, Z.~J. Cui, A.~Altanzaya, and L.~Pinto, ``Behavior
  transformers: Cloning {\textdollar}k{\textdollar} modes with one stone,'' in
  \emph{NeurIPS}, 2022.

\bibitem{DBLP:conf/iclr/CuiWSP23}
Z.~J. Cui, Y.~Wang, N.~M.~M. Shafiullah, and L.~Pinto, ``From play to policy:
  Conditional behavior generation from uncurated robot data,'' in
  \emph{{ICLR}}, 2023.

\bibitem{DBLP:conf/icml/00010EKSP24}
S.~Lee, Y.~Wang, H.~Etukuru, H.~J. Kim, N.~M.~M. Shafiullah, and L.~Pinto,
  ``Behavior generation with latent actions,'' in \emph{{ICML}}, 2024.

\bibitem{DBLP:journals/corr/abs-2505-04769}
R.~Sapkota, Y.~Cao, K.~I. Roumeliotis, and M.~Karkee, ``Vision-language-action
  models: Concepts, progress, applications and challenges,'' \emph{CoRR}, vol.
  abs/2505.04769, 2025.

\bibitem{DBLP:journals/corr/abs-2507-01925}
Y.~Zhong, F.~Bai, S.~Cai, X.~Huang, Z.~Chen, X.~Zhang, Y.~Wang, S.~Guo,
  T.~Guan, K.~N. Lui, Z.~Qi, Y.~Liang, Y.~Chen, and Y.~Yang, ``A survey on
  vision-language-action models: An action tokenization perspective,''
  \emph{CoRR}, vol. abs/2507.01925, 2025.

\bibitem{DBLP:journals/corr/abs-2507-10672}
M.~U. Din, W.~Akram, L.~S. Saoud, J.~Rosell, and I.~Hussain, ``Vision language
  action models in robotic manipulation: {A} systematic review,'' \emph{CoRR},
  vol. abs/2507.10672, 2025.

\bibitem{DBLP:journals/corr/abs-2509-19012}
D.~Zhang, J.~Sun, C.~Hu, X.~Wu, Z.~Yuan, R.~Zhou, F.~Shen, and Q.~Zhou, ``Pure
  vision language action {(VLA)} models: {A} comprehensive survey,''
  \emph{CoRR}, vol. abs/2509.19012, 2025.

\bibitem{DBLP:journals/corr/abs-2509-23121}
S.~Li, Y.~Chen, L.~Dong, S.~Liu, D.~Lan, L.~Yu, and Z.~Pang, ``Transferring
  vision-language-action models to industry applications: Architectures,
  performance, and challenges,'' \emph{CoRR}, vol. abs/2509.23121, 2025.

\bibitem{DBLP:journals/access/KawaharazukaOYPZ25}
K.~Kawaharazuka, J.~Oh, J.~Yamada, I.~Posner, and Y.~Zhu,
  ``Vision-language-action models for robotics: {A} review towards real-world
  applications,'' \emph{{IEEE} Access}, vol.~13, pp. 162\,467--162\,504, 2025.

\bibitem{DBLP:journals/corr/abs-2510-24795}
Z.~Yu, B.~Wang, P.~Zeng, H.~Zhang, J.~Zhang, L.~Gao, J.~Song, N.~Sebe, and
  H.~T. Shen, ``A survey on efficient vision-language-action models,''
  \emph{CoRR}, vol. abs/2510.24795, 2025.

\bibitem{DBLP:journals/corr/abs-2511-05936}
S.~Poria, N.~Majumder, C.~Hung, A.~A. Bagherzadeh, C.~Li, K.~Kwok, Z.~Wang,
  C.~Tan, J.~Wu, and D.~Hsu, ``10 open challenges steering the future of
  vision-language-action models,'' \emph{CoRR}, vol. abs/2511.05936, 2025.

\bibitem{DBLP:journals/corr/abs-2512-11362}
C.~Xu, S.~Zhang, Y.~Liu, B.~Sun, W.~Chen, B.~Xu, Q.~Liu, J.~Wang, S.~Wang,
  S.~Luo, J.~Peters, A.~V. Vasilakos, S.~Zafeiriou, and J.~Deng, ``An anatomy
  of vision-language-action models: From modules to milestones and
  challenges,'' \emph{CoRR}, vol. abs/2512.11362, 2025.

\bibitem{pine2025rlvla}
H.~Deng, Z.~Wu, H.~Liu, W.~Guo, Y.~Xue, Z.~Shan, C.~Zhang, B.~Jia, Y.~Ling,
  G.~Lu, and Z.~Wang, ``A survey on reinforcement learning of
  vision-language-action models for robotic manipulation,'' \emph{TechRxiv},
  2025, preprint.

\bibitem{DBLP:journals/corr/abs-2512-16760}
T.~Hu, X.~Liu, S.~Wang, Y.~Zhu, A.~Liang, L.~Kong, G.~Zhao, Z.~Gong, J.~Cen,
  Z.~Huang, X.~Hao, L.~Li, H.~Song, X.~Li, J.~Ma, S.~Shen, J.~Zhu, D.~Tao,
  Z.~Liu, and J.~Liang, ``Vision-language-action models for autonomous driving:
  Past, present, and future,'' \emph{CoRR}, vol. abs/2512.16760, 2025.

\bibitem{DBLP:conf/iclr/0109M0KJSRBCLR25}
X.~Li, C.~Mata, J.~Park, K.~Kahatapitiya, Y.~S. Jang, J.~Shang, K.~Ranasinghe,
  R.~D. Burgert, M.~Cai, Y.~J. Lee, and M.~S. Ryoo, ``Llara: Supercharging
  robot learning data for vision-language policy,'' in \emph{{ICLR}}.\hskip 1em
  plus 0.5em minus 0.4em\relax OpenReview.net, 2025.

\bibitem{DBLP:journals/corr/abs-2501-09747}
K.~Pertsch, K.~Stachowicz, B.~Ichter, D.~Driess, S.~Nair, Q.~Vuong, O.~Mees,
  C.~Finn, and S.~Levine, ``{FAST:} efficient action tokenization for
  vision-language-action models,'' \emph{CoRR}, vol. abs/2501.09747, 2025.

\bibitem{DBLP:journals/corr/abs-2504-16054}
P.~Intelligence, K.~Black, N.~Brown, J.~Darpinian, K.~Dhabalia, D.~Driess,
  A.~Esmail, M.~Equi, C.~Finn, N.~Fusai, M.~Y. Galliker, D.~Ghosh, L.~Groom,
  K.~Hausman, B.~Ichter, S.~Jakubczak, T.~Jones, L.~Ke, D.~LeBlanc, S.~Levine,
  A.~Li{-}Bell, M.~Mothukuri, S.~Nair, K.~Pertsch, A.~Z. Ren, L.~X. Shi, L.~M.
  Smith, J.~T. Springenberg, K.~Stachowicz, J.~Tanner, Q.~Vuong, H.~Walke,
  A.~Walling, H.~Wang, L.~Yu, and U.~Zhilinsky, ``{\(\pi\)}\({}_{\mbox{0.5}}\):
  a vision-language-action model with open-world generalization,'' \emph{CoRR},
  vol. abs/2504.16054, 2025.

\bibitem{DBLP:conf/corl/XuCFJZL0SRS0HHF24}
Z.~Xu, H.~L. Chiang, Z.~Fu, M.~G. Jacob, T.~Zhang, T.~E. Lee, W.~Yu,
  C.~Schenck, D.~Rendleman, D.~Shah, F.~Xia, J.~Hsu, J.~Hoech, P.~Florence,
  S.~Kirmani, S.~Singh, V.~Sindhwani, C.~Parada, C.~Finn, P.~Xu, S.~Levine, and
  J.~Tan, ``Mobility {VLA:} multimodal instruction navigation with long-context
  vlms and topological graphs,'' in \emph{CoRL}, ser. Proceedings of Machine
  Learning Research, vol. 270.\hskip 1em plus 0.5em minus 0.4em\relax {PMLR},
  2024, pp. 3866--3887.

\bibitem{DBLP:journals/corr/abs-2502-14795}
P.~Ding, J.~Ma, X.~Tong, B.~Zou, X.~Luo, Y.~Fan, T.~Wang, H.~Lu, P.~Mo, J.~Liu,
  Y.~Wang, H.~Zhou, W.~Feng, J.~Liu, S.~Huang, and D.~Wang, ``Humanoid-vla:
  Towards universal humanoid control with visual integration,'' \emph{CoRR},
  vol. abs/2502.14795, 2025.

\bibitem{DBLP:conf/eccv/DingZZSZHYW24}
P.~Ding, H.~Zhao, W.~Zhang, W.~Song, M.~Zhang, S.~Huang, N.~Yang, and D.~Wang,
  ``{QUAR-VLA:} vision-language-action model for quadruped robots,'' in
  \emph{{ECCV} {(5)}}, ser. Lecture Notes in Computer Science, vol.
  15063.\hskip 1em plus 0.5em minus 0.4em\relax Springer, 2024, pp. 352--367.

\bibitem{DBLP:journals/corr/abs-2412-15576}
X.~Tong, P.~Ding, D.~Wang, W.~Zhang, C.~Cui, M.~Sun, Y.~Fan, H.~Zhao, H.~Zhang,
  Y.~Dang, S.~Huang, and S.~Lyu, ``Quart-online: Latency-free large multimodal
  language model for quadruped robot learning,'' \emph{CoRR}, vol.
  abs/2412.15576, 2024.

\bibitem{DBLP:journals/corr/abs-2503-08007}
H.~Zhao, W.~Song, D.~Wang, X.~Tong, P.~Ding, X.~Cheng, and Z.~Ge, ``More:
  Unlocking scalability in reinforcement learning for quadruped
  vision-language-action models,'' \emph{CoRR}, vol. abs/2503.08007, 2025.

\bibitem{DBLP:journals/corr/abs-2502-20900}
Y.~Zhong, X.~Huang, R.~Li, C.~Zhang, Y.~Liang, Y.~Yang, and Y.~Chen,
  ``Dexgraspvla: {A} vision-language-action framework towards general dexterous
  grasping,'' \emph{CoRR}, vol. abs/2502.20900, 2025.

\bibitem{DBLP:journals/corr/abs-2511-00139}
Y.~Cui, Y.~Zhang, L.~Tao, Y.~Li, X.~Yi, and Z.~Li, ``End-to-end dexterous
  arm-hand {VLA} policies via shared autonomy: {VR} teleoperation augmented by
  autonomous hand {VLA} policy for efficient data collection,'' \emph{CoRR},
  vol. abs/2511.00139, 2025.

\bibitem{DBLP:journals/corr/abs-2511-00088}
NVIDIA, ``Alpamayo-r1: Bridging reasoning and action prediction for
  generalizable autonomous driving in the long tail,'' \emph{CoRR}, vol.
  abs/2511.00088, 2025.

\bibitem{DBLP:conf/wacv/AraiMSWY0Y25}
H.~Arai, K.~Miwa, K.~Sasaki, K.~Watanabe, Y.~Yamaguchi, S.~Aoki, and
  I.~Yamamoto, ``Covla: Comprehensive vision-language-action dataset for
  autonomous driving,'' in \emph{{WACV}}.\hskip 1em plus 0.5em minus
  0.4em\relax {IEEE}, 2025, pp. 1933--1943.

\bibitem{DBLP:journals/corr/abs-2510-12796}
Y.~Li, S.~Shang, W.~Liu, B.~Zhan, H.~Wang, Y.~Wang, Y.~Chen, X.~Wang, Y.~An,
  C.~Tang, L.~Hou, L.~Fan, and Z.~Zhang, ``Drivevla-w0: World models amplify
  data scaling law in autonomous driving,'' \emph{CoRR}, vol. abs/2510.12796,
  2025.

\bibitem{DBLP:journals/tmlr/HwangXLHJCHHCSZGAT25}
J.~Hwang, R.~Xu, H.~Lin, W.~Hung, J.~Ji, K.~Choi, D.~Huang, T.~He,
  P.~Covington, B.~Sapp, Y.~Zhou, J.~Guo, D.~Anguelov, and M.~Tan, ``{EMMA:}
  end-to-end multimodal model for autonomous driving,'' \emph{Trans. Mach.
  Learn. Res.}, vol. 2025, 2025.

\bibitem{DBLP:journals/corr/abs-2503-09527}
P.~Chen, P.~Bu, Y.~Wang, X.~Wang, Z.~Wang, J.~Guo, Y.~Zhao, Q.~Zhu, J.~Song,
  S.~Yang, J.~Wang, and B.~Zheng, ``Combatvla: An efficient
  vision-language-action model for combat tasks in 3d action role-playing
  games,'' \emph{CoRR}, vol. abs/2503.09527, 2025.

\bibitem{DBLP:conf/acl/LiWH0L25}
M.~Li, Z.~Wang, K.~He, X.~Ma, and Y.~Liang, ``{JARVIS-VLA:} post-training
  large-scale vision language models to play visual games with keyboards and
  mouse,'' in \emph{{ACL} (Findings)}.\hskip 1em plus 0.5em minus 0.4em\relax
  Association for Computational Linguistics, 2025, pp. 17\,878--17\,899.

\bibitem{DBLP:conf/corl/KeGF24}
N.~G. Tsung{-}Wei Ke~and and K.~Fragkiadaki, ``3d diffuser actor: Policy
  diffusion with 3d scene representations,'' in \emph{CoRL}, ser. Proceedings
  of Machine Learning Research, vol. 270.\hskip 1em plus 0.5em minus
  0.4em\relax {PMLR}, 2024, pp. 1949--1974.

\bibitem{vla_3dsvla_2025}
X.~Li, L.~Heng, J.~Liu, Y.~Shen, C.~Gu, Z.~Liu, H.~Chen, N.~Han, R.~Zhang,
  H.~Tang, S.~Zhang, and H.~Dong, ``3ds-vla,'' \emph{arXiv preprint}, 2025.

\bibitem{DBLP:journals/corr/abs-2506-22242}
Y.~C. Jiahui Zhang~and, Y.~Xu, Z.~Huang, Y.~Zhou, Y.~Yuan, X.~Cai, G.~Huang,
  X.~Quan, H.~Xu, and L.~Zhang, ``4d-vla: Spatiotemporal vision-language-action
  pretraining with cross-scene, calibration,'' \emph{CoRR}, vol.
  abs/2506.22242, 2025.

\bibitem{DBLP:journals/corr/abs-2509-23224}
M.~A. Kohei Sendai~and, T.~Matsushima, Y.~Matsuo, and Y.~Iwasawa, ``Leave no
  observation behind: Real-time correction for {VLA} action, chunks,''
  \emph{CoRR}, vol. abs/2509.23224, 2025.

\bibitem{DBLP:conf/corl/0004CLZ0B0024}
H.~C. Siyuan Huang~and, Y.~Liu, Y.~Zhu, H.~Dong, A.~Boularias, P.~Gao, and
  H.~Li, ``{A3VLM:} actionable articulation-aware vision language model,'' in
  \emph{CoRL}, ser. Proceedings of Machine Learning Research, vol. 270.\hskip
  1em plus 0.5em minus 0.4em\relax {PMLR}, 2024, pp. 1675--1690.

\bibitem{DBLP:journals/corr/abs-2510-22201}
K.~K. Minho Park~and, J.~Hyung, H.~Jang, H.~Jin, J.~Yun, H.~Lee, and J.~Choo,
  ``{ACG:} action coherence guidance for flow-based {VLA} models,''
  \emph{CoRR}, vol. abs/2510.22201, 2025.

\bibitem{DBLP:journals/corr/abs-2511-18082}
T.~W. Wencheng Ye~and, L.~Zhu, F.~Li, and G.~Yang, ``Actdistill: General
  action-guided self-derived distillation for efficient, vision-language-action
  models,'' \emph{CoRR}, vol. abs/2511.18082, 2025.

\bibitem{DBLP:journals/corr/abs-2512-20276}
H.~G. Yuntao Dai~and, T.~Wang, Q.~Cheng, Y.~Zheng, Z.~Qiu, L.~Gong, W.~Lou, and
  X.~Zhou, ``Actionflow: {A} pipelined action acceleration for vision language,
  models on edge,'' \emph{CoRR}, vol. abs/2512.20276, 2025.

\bibitem{DBLP:journals/corr/abs-2510-14300}
Y.~L. Weijie Shen~and, Y.~Wu, Z.~Liang, S.~Gu, D.~Wang, T.~Nian, L.~Xu, Y.~Qin,
  J.~Pang, X.~Guan, X.~Yang, and Y.~Mu, ``Expertise need not monopolize:
  Action-specialized mixture of experts, for vision-language-action learning,''
  \emph{CoRR}, vol. abs/2510.14300, 2025.

\bibitem{DBLP:journals/corr/abs-2509-22093}
Y.~C. Xiaohuan Pei~and, S.~Xu, Y.~Wang, Y.~Shi, and C.~Xu, ``Action-aware
  dynamic pruning for efficient vision-language-action, manipulation,''
  \emph{CoRR}, vol. abs/2509.22093, 2025.

\bibitem{DBLP:journals/corr/abs-2512-07472}
Z.~W. Siyu Xu~and, Y.~Wang, C.~Xia, T.~Huang, and C.~Xu, ``Affordance field
  intervention: Enabling vlas to escape memory traps, in robotic
  manipulation,'' \emph{CoRR}, vol. abs/2512.07472, 2025.

\bibitem{DBLP:journals/corr/abs-2508-07770}
Z.~Y. Yizheng Zhang~and, J.~Lai, C.~Lu, and L.~Han, ``Agentworld: An
  interactive simulation platform for scene construction, and mobile robotic
  manipulation,'' \emph{CoRR}, vol. abs/2508.07770, 2025.

\bibitem{DBLP:journals/corr/abs-2511-16661}
H.~Q. Irmak Guzey~and, J.~Urain, C.~Wang, J.~Yin, K.~Bodduluri, M.~Lambeta,
  L.~Pinto, A.~Rai, J.~Malik, T.~Wu, A.~Sharma, and H.~Bharadhwaj, ``Dexterity
  from smart lenses: Multi-fingered robot manipulation with, in-the-wild human
  demonstrations,'' \emph{CoRR}, vol. abs/2511.16661, 2025.

\bibitem{DBLP:journals/corr/abs-2508-10259}
B.~L. Wenxin Zheng~and, B.~Xu, E.~Feng, J.~Gu, and H.~Chen, ``Leveraging
  os-level primitives for robotic action management,'' \emph{CoRR}, vol.
  abs/2508.10259, 2025.

\bibitem{DBLP:journals/corr/abs-2509-03383}
Z.~W. Yiyang Huang~and, Z.~Wan, Y.~Tian, H.~Xu, Y.~Han, and Y.~Gan, ``{ANNIE:}
  be careful of your robots,'' \emph{CoRR}, vol. abs/2509.03383, 2025.

\bibitem{DBLP:journals/corr/abs-2511-14148}
S.~C. Yuhua Jiang~and, Y.~Ding, F.~Gao, and B.~Qi, ``Asyncvla: Asynchronous
  flow matching for vision-language-action models,'' \emph{CoRR}, vol.
  abs/2511.14148, 2025.

\bibitem{DBLP:journals/corr/abs-2509-02055}
\BIBentryALTinterwordspacing
C.~W. Yang Zhang~and, O.~Lu, Y.~Zhao, Y.~Ge, Z.~S. 0001, X.~L. 0001, C.~Z.
  0012, C.~Bai, and X.~L. 0001, ``Align-then-steer: Adapting the
  vision-language action models through unified latent guidance,'' 2025.
  [Online]. Available: \url{https://dblp.org/rec/journals/corr/abs-2509-02055}
\BIBentrySTDinterwordspacing

\bibitem{DBLP:journals/corr/abs-2509-01944}
J.~T. Zhenlong Yuan~and, J.~Luo, R.~Chen, C.~Qian, L.~Sun, X.~Chu, Y.~Cai,
  D.~Zhang, and S.~Li, ``Autodrive-r\({}^{\mbox{2}}\): Incentivizing reasoning
  and self-reflection, capacity for {VLA} model in autonomous driving,''
  \emph{CoRR}, vol. abs/2509.01944, 2025.

\bibitem{DBLP:journals/corr/abs-2509-23931}
Y.~X. Hanshi Wang~and, Z.~Xu, J.~Gao, Y.~Liu, W.~Hu, K.~Wang, and Z.~Zhang,
  ``Autoprune: Each complexity deserves a pruning policy,'' \emph{CoRR}, vol.
  abs/2509.23931, 2025.

\bibitem{DBLP:journals/corr/abs-2506-13757}
T.~C. Zewei Zhou~and, S.~Z. Zhao, Y.~Zhang, Z.~Huang, B.~Zhou, and J.~Ma,
  ``Autovla: {A} vision-language-action model for end-to-end autonomous,
  driving with adaptive reasoning and reinforcement fine-tuning,'' \emph{CoRR},
  vol. abs/2506.13757, 2025.

\bibitem{DBLP:journals/corr/abs-2511-18960}
\BIBentryALTinterwordspacing
J.~L. Lei Xiao~and, J.~Gao, F.~Ye, Y.~Jin, J.~Qian, J.~Zhang, Y.~Wu, and X.~Yu,
  ``Ava-vla: Improving vision-language-action models with active visual
  attention,'' 2025. [Online]. Available:
  \url{https://dblp.org/rec/journals/corr/abs-2511-18960}
\BIBentrySTDinterwordspacing

\bibitem{DBLP:journals/corr/abs-2510-21746}
L.~L. Harris Song~and, ``Avi: Action from volumetric inference,'' \emph{CoRR},
  vol. abs/2510.21746, 2025.

\bibitem{DBLP:journals/corr/abs-2505-16640}
G.~T. Xueyang Zhou~and, G.~Zhang, H.~Wang, P.~Zhou, and L.~Sun, ``Badvla:
  Towards backdoor attacks on vision-language-action models, via
  objective-decoupled optimization,'' \emph{CoRR}, vol. abs/2505.16640, 2025.

\bibitem{DBLP:journals/corr/abs-2512-11218}
Z.~Z. Kechun Xu~and, A.~Chen, S.~Zhao, Q.~Huang, Y.~Yang, H.~Lu, R.~Xiong,
  M.~Tomizuka, and Y.~Wang, ``Seeing to act, prompting to specify: {A} bayesian
  factorization of, vision language action policy,'' \emph{CoRR}, vol.
  abs/2512.11218, 2025.

\bibitem{DBLP:journals/corr/abs-2507-15597}
Y.~F. Hao Luo~and, W.~Zhang, S.~Zheng, Y.~Wang, H.~Yuan, J.~Liu, C.~Xu, Q.~Jin,
  and Z.~Lu, ``Being-h0: Vision-language-action pretraining from large-scale
  human, videos,'' \emph{CoRR}, vol. abs/2507.15597, 2025.

\bibitem{DBLP:journals/corr/abs-2511-22555}
J.~F. Yanbo Mao~and, R.~Zhang, H.~Xie, and M.~Yao, ``Beyond success: Refining
  elegant robot manipulation from mixed-quality, data via just-in-time
  intervention,'' \emph{CoRR}, vol. abs/2511.22555, 2025.

\bibitem{DBLP:conf/smc/GbagbeCAALT24}
M.~A.~C. Koffivi Fid{\`{e}}le Gbagbe~and, A.~Alabbas, O.~Alyunes, A.~Lykov, and
  D.~Tsetserukou, ``Bi-vla: Vision-language-action model-based system for
  bimanual robotic, dexterous manipulations,'' in \emph{{SMC}}.\hskip 1em plus
  0.5em minus 0.4em\relax {IEEE}, 2024, pp. 2864--2869.

\bibitem{DBLP:journals/corr/abs-2509-18865}
T.~B. Masato Kobayashi~and, ``Bi-vla: Bilateral control-based imitation
  learning via vision-language, fusion for action generation,'' \emph{CoRR},
  vol. abs/2509.18865, 2025.

\bibitem{DBLP:journals/corr/abs-2506-07961}
Y.~C. Peiyan Li~and, H.~Wu, X.~Ma, X.~Wu, Y.~Huang, L.~Wang, T.~Kong, and
  T.~Tan, ``Bridgevla: Input-output alignment for efficient 3d manipulation
  learning, with vision-language models,'' \emph{CoRR}, vol. abs/2506.07961,
  2025.

\bibitem{DBLP:conf/icra/HancockRM25}
A.~Z.~R. Asher J. Hancock~and and A.~Majumdar, ``Run-time observation
  interventions make vision-language-action models, more visually robust,'' in
  \emph{{ICRA}}.\hskip 1em plus 0.5em minus 0.4em\relax {IEEE}, 2025, pp.
  9499--9506.

\bibitem{DBLP:journals/corr/abs-2511-14396}
Y.~Y. Xiuxiu Qi~and, J.~Cao, L.~Bai, C.~Fan, C.~Cao, and H.~Wang, ``Continuous
  vision-language-action co-learning with semantic-physical, alignment for
  behavioral cloning,'' \emph{CoRR}, vol. abs/2511.14396, 2025.

\bibitem{DBLP:journals/corr/abs-2505-21906}
Y.~Z. Zhongyi Zhou~and, J.~Wen, C.~Shen, and Y.~Xu, ``Chatvla-2:
  Vision-language-action model with open-world embodied reasoning, from
  pretrained knowledge,'' \emph{CoRR}, vol. abs/2505.21906, 2025.

\bibitem{DBLP:journals/corr/abs-2509-14143}
R.~Y. Zijian An~and, Y.~Feng, and L.~Zhou, ``{CLAW:} {A} vision-language-action
  framework for weight-aware robotic, grasping,'' \emph{CoRR}, vol.
  abs/2509.14143, 2025.

\bibitem{DBLP:journals/corr/abs-2411-00508}
J.~K. Gi{-}Cheon Kang~and, K.~Shim, J.~K. Lee, and B.~Zhang, ``{CLIP-RT:}
  learning language-conditioned robotic policies from natural, language
  supervision,'' \emph{CoRR}, vol. abs/2411.00508, 2024.

\bibitem{DBLP:journals/corr/abs-2506-14317}
Q.~Y. Zeyuan Chen~and, Y.~Chen, T.~Wu, J.~Zhang, Z.~Ding, J.~Li, Y.~Yang, and
  H.~Dong, ``Clutterdexgrasp: {A} sim-to-real system for general dexterous
  grasping, in cluttered scenes,'' \emph{CoRR}, vol. abs/2506.14317, 2025.

\bibitem{DBLP:journals/corr/abs-2511-19914}
F.~S. Dapeng Zhang~and, R.~Zhao, Y.~Chen, P.~Zhi, C.~Li, R.~Zhou, and Q.~Zhou,
  ``Coc-vla: Delving into adversarial domain transfer for explainable,
  autonomous driving via chain-of-causality visual-language-action model,''
  \emph{CoRR}, vol. abs/2511.19914, 2025.

\bibitem{DBLP:journals/corr/abs-2508-21046}
R.~Z. Wei Li~and, R.~Shao, J.~He, and L.~Nie, ``Cogvla: Cognition-aligned
  vision-language-action model via instruction-driven, routing {\&}
  sparsification,'' \emph{CoRR}, vol. abs/2508.21046, 2025.

\bibitem{DBLP:journals/corr/abs-2512-22939}
X.~C. Qihang Peng~and, C.~Yang, S.~Shi, and H.~Li, ``Colavla: Leveraging
  cognitive latent reasoning for hierarchical parallel, trajectory planning in
  autonomous driving,'' \emph{CoRR}, vol. abs/2512.22939, 2025.

\bibitem{DBLP:journals/corr/abs-2502-05450}
S.~T. Yuhui Chen~and, S.~Liu, Y.~Zhou, H.~Li, and D.~Zhao, ``Conrft: {A}
  reinforced fine-tuning method for {VLA} models via consistency, policy,''
  \emph{CoRR}, vol. abs/2502.05450, 2025.

\bibitem{DBLP:journals/corr/abs-2506-16211}
Y.~W. Puhao Li~and, Z.~Xi, W.~Li, Y.~Huang, Z.~Zhang, Y.~Chen, J.~Wang, S.~Zhu,
  T.~Liu, and S.~Huang, ``Controlvla: Few-shot object-centric adaptation for
  pre-trained vision-language-action, models,'' \emph{CoRR}, vol.
  abs/2506.16211, 2025.

\bibitem{DBLP:journals/corr/abs-2509-23823}
W.~K. Tian Nian~and, Y.~Mu, T.~Chen, S.~Zhu, and B.~Hu, ``Control your robot:
  {A} unified system for robot control and policy, deployment,'' \emph{CoRR},
  vol. abs/2509.23823, 2025.

\bibitem{DBLP:journals/corr/abs-2509-15968}
Y.~C. Shiyu Fang~and, H.~Liang, C.~Lv, P.~Hang, and J.~Sun, ``Corevla: {A}
  dual-stage end-to-end autonomous driving framework for, long-tail scenarios
  via collect-and-refine,'' \emph{CoRR}, vol. abs/2509.15968, 2025.

\bibitem{DBLP:journals/corr/abs-2511-22532}
T.~Y. Zhaohui Wang~and and H.~Tang, ``Cot4ad: {A} vision-language-action model
  with explicit chain-of-thought, reasoning for autonomous driving,''
  \emph{CoRR}, vol. abs/2511.22532, 2025.

\bibitem{DBLP:journals/corr/abs-2512-24426}
W.~D. Zhenghao "Mark" Peng~and, Y.~You, Y.~Chen, W.~Luo, T.~Tian, Y.~Cao,
  A.~Sharma, D.~Xu, B.~Ivanovic, B.~Li, B.~Zhou, Y.~Wang, and M.~Pavone,
  ``Counterfactual {VLA:} self-reflective vision-language-action model, with
  adaptive reasoning,'' \emph{CoRR}, vol. abs/2512.24426, 2025.

\bibitem{DBLP:journals/corr/abs-2509-06819}
O.~H. Daniel San Jos{\'{e}} Pro~and, R.~R{\"{o}}mer, M.~D{\"{o}}sch, M.~Schuck,
  and A.~P. Schoellig, ``{CRISP} - compliant {ROS2} controllers for
  learning-based manipulation, policies and teleoperation,'' \emph{CoRR}, vol.
  abs/2509.06819, 2025.

\bibitem{DBLP:journals/corr/abs-2512-01022}
H.~L. Yi{-}Lin Wei~and, Y.~Lin, P.~Wang, Z.~Liang, G.~Liu, and W.~Zheng,
  ``Cyclemanip: Enabling cyclic task manipulation via effective historical,
  perception and understanding,'' \emph{CoRR}, vol. abs/2512.01022, 2025.

\bibitem{DBLP:journals/corr/abs-2510-07730}
H.~L. Changyeon Kim~and, Y.~Seo, K.~Lee, and Y.~Zhu, ``{DEAS:} detached value
  learning with action sequence for scalable, offline {RL},'' \emph{CoRR}, vol.
  abs/2510.07730, 2025.

\bibitem{DBLP:journals/corr/abs-2511-20720}
L.~H. Haibo Hu~and, N.~Guan, and C.~J. Xue, ``Deead: Dynamic early exit of
  vision-language action for efficient, autonomous driving,'' \emph{CoRR}, vol.
  abs/2511.20720, 2025.

\bibitem{DBLP:journals/corr/abs-2511-15669}
Y.~L. Cheng Yin~and, W.~Xu, S.~Tam, X.~Zeng, Z.~Liu, and Z.~Yin,
  ``Deepthinkvla: Enhancing reasoning capability of vision-language-action,
  models,'' \emph{CoRR}, vol. abs/2511.15669, 2025.

\bibitem{vla_dejavu_2025}
S.~Wu, Y.~Ji, Q.~Li, Z.~Zhang, Q.~He, W.~Xie, G.~Zhang, B.~Bayramli, Y.~Ding,
  and H.~Lu, ``Dejavu: Towards experience feedback learning for embodied
  intelligence,'' \emph{arXiv preprint}, 2025.

\bibitem{DBLP:journals/corr/abs-2510-13375}
Y.~L. Tianyuan Yuan~and, C.~Lu, Z.~Chen, T.~Jiang, and H.~Zhao, ``Depthvla:
  Enhancing vision-language-action models with depth-aware, spatial
  reasoning,'' \emph{CoRR}, vol. abs/2510.13375, 2025.

\bibitem{DBLP:journals/corr/abs-2510-23511}
E.~Z. Bin Xie~and, F.~Jia, H.~Shi, H.~Fan, H.~Zhang, H.~Li, J.~Sun, J.~Bin,
  J.~Huang, K.~Liu, K.~Liu, K.~Gu, L.~Sun, M.~Zhang, P.~Han, R.~Hao, R.~Zhang,
  S.~Huang, S.~Xie, T.~Wang, T.~Liu, W.~Tang, W.~Zhu, Y.~Chen, Y.~Liu, Y.~Zhou,
  Y.~Liu, Y.~Zhao, Y.~Ma, Y.~Wei, Y.~Chen, Z.~Chen, Z.~Li, Z.~Wu, Z.~Zhang,
  Z.~Liu, Z.~Yan, and Z.~Zhang, ``Dexbotic: Open-source vision-language-action
  toolbox,'' \emph{CoRR}, vol. abs/2510.23511, 2025.

\bibitem{DBLP:journals/ijrr/ChiXFCDBTS25}
Z.~X. Cheng Chi~and, S.~Feng, E.~Cousineau, Y.~Du, B.~Burchfiel, R.~Tedrake,
  and S.~Song, ``Diffusion policy: Visuomotor policy learning via action
  diffusion,'' \emph{Int. J. Robotics Res.}, vol.~44, no. 10-11, pp.
  1684--1704, 2025.

\bibitem{DBLP:journals/corr/abs-2412-03293}
M.~Z. Junjie Wen~and, Y.~Zhu, Z.~Tang, J.~Li, Z.~Zhou, C.~Li, X.~Liu, Y.~Peng,
  C.~Shen, and F.~Feng, ``Diffusion-vla: Scaling robot foundation models via
  unified diffusion, and autoregression,'' \emph{CoRR}, vol. abs/2412.03293,
  2024.

\bibitem{DBLP:journals/corr/abs-2510-17148}
A.~J. Yu~Gao~and, Y.~Wang, J.~Wang, H.~Jiang, Z.~Sun, Y.~Heng, W.~Shuo,
  H.~Zhao, and H.~Sun, ``Diffvla++: Bridging cognitive reasoning and end-to-end
  driving through, metric-guided alignment,'' \emph{CoRR}, vol. abs/2510.17148,
  2025.

\bibitem{vla_dipole_2025}
R.~Liang, Y.~Zheng, K.~Zheng, T.~Tan, J.~Li, L.~Mao, Z.~Wang, G.~Chen, H.~Ye,
  J.~Liu, J.~Wang, and X.~Zhan, ``Dichotomous diffusion policy optimization,''
  \emph{arXiv preprint}, 2025.

\bibitem{DBLP:journals/corr/abs-2511-19528}
Z.~F. Rushuai Yang~and, T.~Zhang, K.~Wang, C.~Zhang, L.~Zhao, X.~Su, Y.~Chen,
  and J.~Bian, ``Discover, learn, and reinforce: Scaling vision-language-action
  pretraining, with diverse rl-generated trajectories,'' \emph{CoRR}, vol.
  abs/2511.19528, 2025.

\bibitem{DBLP:journals/corr/abs-2510-25616}
M.~K. Nikita Kachaev~and, D.~Zelezetsky, A.~K. Kovalev, and A.~I. Panov,
  ``Don't blind your {VLA:} aligning visual representations for {OOD},
  generalization,'' \emph{CoRR}, vol. abs/2510.25616, 2025.

\bibitem{DBLP:journals/corr/abs-2410-15549}
J.~K. ByungOk Han~and and J.~Jang, ``A dual process {VLA:} efficient robotic
  manipulation leveraging {VLM},'' \emph{CoRR}, vol. abs/2410.15549, 2024.

\bibitem{DBLP:journals/corr/abs-2512-23864}
Z.~Z. Guo Ye~and, X.~Zhao, S.~Wu, H.~Lu, S.~Lu, and H.~Liu, ``Learning to feel
  the future: Dreamtacvla for contact-rich manipulation,'' \emph{CoRR}, vol.
  abs/2512.23864, 2025.

\bibitem{DBLP:journals/corr/abs-2507-04447}
H.~L. Wenyao Zhang~and, Z.~Qi, Y.~Wang, X.~Yu, J.~Zhang, R.~Dong, J.~He,
  H.~Wang, Z.~Zhang, L.~Yi, W.~Zeng, and X.~Jin, ``Dreamvla: {A}
  vision-language-action model dreamed with comprehensive, world knowledge,''
  \emph{CoRR}, vol. abs/2507.04447, 2025.

\bibitem{DBLP:journals/corr/abs-2512-22615}
S.~G. Jiacheng Ye~and, J.~Gao, J.~Fan, S.~Wu, W.~Bi, H.~Bai, L.~Shang, and
  L.~Kong, ``Dream-vl {\&} dream-vla: Open vision-language and
  vision-language-action, models with diffusion language model backbone,''
  \emph{CoRR}, vol. abs/2512.22615, 2025.

\bibitem{DBLP:journals/corr/abs-2511-22134}
Z.~L. Zhen Fang~and, J.~Liu, H.~Chen, Y.~Zeng, S.~Huang, Z.~Chen, L.~Chen,
  S.~Zhang, and F.~Zhao, ``Dualvla: Building a generalizable embodied agent via
  partial decoupling, of reasoning and action,'' \emph{CoRR}, vol.
  abs/2511.22134, 2025.

\bibitem{DBLP:journals/corr/abs-2512-20188}
H.~Z. Teqiang Zou~and, Y.~Nong, Y.~Li, K.~Liu, H.~Yang, X.~Ling, X.~Li, and
  L.~Ma, ``Asynchronous fast-slow vision-language-action policies for
  whole-body, robotic manipulation,'' \emph{CoRR}, vol. abs/2512.20188, 2025.

\bibitem{DBLP:journals/corr/abs-2509-25681}
M.~Z. Junjie Wen~and, J.~Liu, Z.~Liu, Y.~Yang, L.~Zhang, S.~Zhang, Y.~Zhu, and
  Y.~Xu, ``dvla: Diffusion vision-language-action model with multimodal
  chain-of-thought,'' \emph{CoRR}, vol. abs/2509.25681, 2025.

\bibitem{DBLP:journals/corr/abs-2510-27545}
Y.~H. Travis Davies~and, A.~Gladstone, Y.~Liu, X.~Chen, H.~Ji, H.~Liu, and
  L.~Hu, ``Ebt-policy: Energy unlocks emergent physical reasoning
  capabilities,'' \emph{CoRR}, vol. abs/2510.27545, 2025.

\bibitem{DBLP:conf/corl/ZawalskiCPMFL24}
W.~C. Michal Zawalski~and, K.~Pertsch, O.~Mees, C.~Finn, and S.~Levine,
  ``Robotic control via embodied chain-of-thought reasoning,'' in \emph{CoRL},
  ser. Proceedings of Machine Learning Research, vol. 270.\hskip 1em plus 0.5em
  minus 0.4em\relax {PMLR}, 2024, pp. 3157--3181.

\bibitem{DBLP:journals/corr/abs-2505-08243}
S.~B. William Chen~and, S.~Mirchandani, O.~Mees, D.~Driess, K.~Pertsch, and
  S.~Levine, ``Training strategies for efficient embodied reasoning,''
  \emph{CoRR}, vol. abs/2505.08243, 2025.

\bibitem{DBLP:journals/corr/abs-2506-10100}
Y.~W. Yantai Yang~and, Z.~Wen, L.~Zhongwei, C.~Zou, Z.~Zhang, C.~Wen, and
  L.~Zhang, ``Efficientvla: Training-free acceleration and compression for
  vision-language-action, models,'' \emph{CoRR}, vol. abs/2506.10100, 2025.

\bibitem{DBLP:journals/corr/abs-2508-19852}
M.~Z.~S. Binjie Zhang~and, ``Ego-centric predictive model conditioned on hand
  trajectories,'' \emph{CoRR}, vol. abs/2508.19852, 2025.

\bibitem{DBLP:journals/corr/abs-2510-06207}
R.~C. Zefu Lin~and, C.~Hanning, X.~Wang, J.~Xu, X.~Jin, C.~Wenbo, H.~Zhou,
  L.~Fan, W.~Li, and Z.~Zhang, ``Embodiedcoder: Parameterized embodied mobile
  manipulation via modern, coding model,'' \emph{CoRR}, vol. abs/2510.06207,
  2025.

\bibitem{DBLP:journals/corr/abs-2511-11478}
T.~H. Nhat Chung~and, T.~Nguyen, H.~Le, F.~Bumgarner, D.~M.~H. Nguyen, K.~Vo,
  K.~Yamazaki, C.~Rainwater, T.~Kieu, A.~Nguyen, and N.~Le, ``Rethinking
  progression of memory state in robotic manipulation: An, object-centric
  perspective,'' \emph{CoRR}, vol. abs/2511.11478, 2025.

\bibitem{DBLP:journals/corr/abs-2509-22407}
X.~W. Zhehao Dong~and, Z.~Zhu, Y.~Wang, Y.~Wang, Y.~Zhou, B.~Wang, C.~Ni,
  R.~Ouyang, W.~Qin, X.~Chen, Y.~Ye, and G.~Huang, ``{EMMA:} generalizing
  real-world robot manipulation via generative, visual transfer,'' \emph{CoRR},
  vol. abs/2509.22407, 2025.

\bibitem{DBLP:conf/acl/SunHDTTGP25}
P.~H. Qi~Sun~and, P.~T. Deep, V.~Toh, U.~Tan, D.~Ghosal, and S.~Poria,
  ``Emma-x: An embodied multimodal action model with grounded chain of, thought
  and look-ahead spatial reasoning,'' in \emph{{ACL} {(1)}}.\hskip 1em plus
  0.5em minus 0.4em\relax Association for Computational Linguistics, 2025, pp.
  14\,199--14\,214.

\bibitem{DBLP:journals/corr/abs-2505-15206}
L.~B. Chikit Ng~and, G.~Wang, Y.~Wang, H.~Gao, K.~Yuan, C.~Jin, T.~Zeng, and
  H.~Ren, ``Endovla: Dual-phase vision-language-action model for autonomous
  tracking, in endoscopy,'' \emph{CoRR}, vol. abs/2505.15206, 2025.

\bibitem{DBLP:journals/corr/abs-2508-21112}
H.~S. Delin Qu~and, Q.~Chen, Z.~Chen, X.~Gao, X.~Ye, Q.~Lv, M.~Shi, G.~Ren,
  C.~Ruan, M.~Yao, H.~Yang, J.~Bao, B.~Zhao, and D.~Wang, ``Embodiedonevision:
  Interleaved vision-text-action pretraining for, general robot control,''
  \emph{CoRR}, vol. abs/2508.21112, 2025.

\bibitem{DBLP:journals/corr/abs-2512-24125}
S.~W. Yi~Liu~and, D.~Wei, X.~Cai, L.~Zhong, J.~Yang, G.~Ren, J.~Zhang, M.~Yao,
  C.~Li, X.~He, L.~Chen, and J.~Luo, ``Unified embodied {VLM} reasoning with
  robotic action via autoregressive, discretized pre-training,'' \emph{CoRR},
  vol. abs/2512.24125, 2025.

\bibitem{DBLP:journals/corr/abs-2507-17462}
G.~W. Chang Nie~and, Z.~Lie, and H.~Wang, ``{ERMV:} editing 4d robotic
  multi-view images to enhance embodied agents,'' \emph{CoRR}, vol.
  abs/2507.17462, 2025.

\bibitem{DBLP:journals/corr/abs-2511-01224}
Y.~P. Chengmeng Li~and, ``Embodiment transfer learning for
  vision-language-action models,'' \emph{CoRR}, vol. abs/2511.01224, 2025.

\bibitem{DBLP:journals/corr/abs-2511-05397}
A.~M. Samarth Chopra~and, B.~Carnovale, E.~Sokolson, R.~Kubendran, and
  S.~Dickerson, ``Everydayvla: {A} vision-language-action model for affordable
  robotic, manipulation,'' \emph{CoRR}, vol. abs/2511.05397, 2025.

\bibitem{DBLP:journals/corr/abs-2511-04555}
Y.~Z. Tao Lin~and, Y.~Du, J.~Zhang, J.~Liu, Y.~Chen, E.~Gu, Z.~Liu, H.~Cai,
  Y.~Zou, L.~Zou, Z.~Zhou, G.~Li, and B.~Zhao, ``Evo-1: Lightweight
  vision-language-action model with preserved semantic, alignment,''
  \emph{CoRR}, vol. abs/2511.04555, 2025.

\bibitem{DBLP:journals/corr/abs-2511-06202}
Y.~A. Shahram Najam Syed~and, A.~Jakobsson, and J.~Ichnowski, ``Expres-vla:
  Specializing vision-language-action models through experience, replay and
  retrieval,'' \emph{CoRR}, vol. abs/2511.06202, 2025.

\bibitem{DBLP:journals/corr/abs-2511-15279}
Y.~H. Jiashu Yang~and, Y.~Xie, N.~Guo, and W.~Lian, ``Look, zoom, understand:
  The robotic eyeball for embodied perception,'' \emph{CoRR}, vol.
  abs/2511.15279, 2025.

\bibitem{DBLP:journals/corr/abs-2509-06951}
W.~K. Qi~Lv~and, H.~Li, J.~Zeng, Z.~Qiu, D.~Qu, H.~Song, Q.~Chen, X.~Deng, and
  J.~Pang, ``{F1:} {A} vision-language-action model bridging understanding and,
  generation to actions,'' \emph{CoRR}, vol. abs/2509.06951, 2025.

\bibitem{DBLP:journals/corr/abs-2510-01642}
J.~D. Zijun Lin~and, H.~Fang, D.~Fox, R.~Krishna, C.~Tan, and B.~Wen,
  ``Failsafe: Reasoning and recovery from failures in vision-language-action,
  models,'' \emph{CoRR}, vol. abs/2510.01642, 2025.

\bibitem{DBLP:journals/corr/abs-2507-23318}
Q.~Z. Jiajun Cao~and, P.~Jia, X.~Zhao, B.~Lan, X.~Zhang, X.~Wei, S.~Chen,
  Z.~Li, Y.~Wang, L.~Li, X.~Liu, M.~Lu, and S.~Zhang, ``Fastdrivevla: Efficient
  end-to-end driving via plug-and-play reconstruction-based, token pruning,''
  \emph{CoRR}, vol. abs/2507.23318, 2025.

\bibitem{vla_fastinslow_2025}
H.~Chen, J.~Liu, C.~Gu, Z.~Liu, R.~Zhang, X.~Li, and X.~He, ``Fast-in-slow: A
  dual-system vla model unifying fast manipulation within slow reasoning,''
  \emph{arXiv preprint}, 2025.

\bibitem{DBLP:journals/corr/abs-2505-15659}
J.~W. Ruijie Zheng~and, S.~Reed, J.~Bjorck, Y.~Fang, F.~Hu, J.~Jang,
  K.~Kundalia, Z.~Lin, L.~Magne, A.~Narayan, Y.~L. Tan, G.~Wang, Q.~Wang,
  J.~Xiang, Y.~Xu, S.~Ye, J.~Kautz, F.~Huang, Y.~Zhu, and L.~Fan, ``{FLARE:}
  robot learning with implicit world modeling,'' \emph{CoRR}, vol.
  abs/2505.15659, 2025.

\bibitem{DBLP:journals/corr/abs-2509-04996}
H.~Z. Moritz Reuss~and, M.~R{\"{u}}hle, {\"{O}}.~E. Yagmurlu, F.~Otto, and
  R.~Lioutikov, ``{FLOWER:} democratizing generalist robot policies with
  efficient vision-language-action, flow policies,'' \emph{CoRR}, vol.
  abs/2509.04996, 2025.

\bibitem{vla_flowvla_2025}
Z.~Zhong, H.~Yan, J.~Li, X.~Liu, X.~Gong, T.~Zhang, W.~Song, J.~Chen, X.~Zheng,
  H.~Wang, and H.~Li, ``Flowvla: Visual chain of thought-based motion reasoning
  for vision-language-action models,'' \emph{arXiv preprint}, 2025.

\bibitem{DBLP:journals/corr/abs-2505-22159}
H.~L. Jiawen Yu~and, Q.~Yu, J.~Ren, C.~Hao, H.~Ding, G.~Huang, G.~Huang,
  Y.~Song, P.~Cai, C.~Lu, and W.~Zhang, ``Forcevla: Enhancing {VLA} models with
  a force-aware moe for contact-rich, manipulation,'' \emph{CoRR}, vol.
  abs/2505.22159, 2025.

\bibitem{DBLP:journals/corr/abs-2510-21744}
S.~L. Yanjia Huang~and, S.~Liu, Q.~Xu, M.~Wu, X.~Gao, and Z.~Tu, ``Forge-tree:
  Diffusion-forcing tree search for long-horizon robot manipulation,''
  \emph{CoRR}, vol. abs/2510.21744, 2025.

\bibitem{DBLP:journals/corr/abs-2509-04018}
Z.~D. Yifan Yang~and, T.~Xie, F.~Cao, P.~Shen, P.~Song, P.~Jin, G.~Sun, S.~Xu,
  Y.~You, and J.~Liu, ``{FPC-VLA:} {A} vision-language-action framework with a
  supervisor, for failure prediction and correction,'' \emph{CoRR}, vol.
  abs/2509.04018, 2025.

\bibitem{DBLP:journals/corr/abs-2509-19870}
J.~L. Xin Wang~and, Z.~Weng, Y.~Wang, Y.~Gao, T.~Pang, C.~Du, Y.~Teng, Y.~Wang,
  Z.~Wu, X.~Ma, and Y.~Jiang, ``Freezevla: Action-freezing attacks against
  vision-language-action, models,'' \emph{CoRR}, vol. abs/2509.19870, 2025.

\bibitem{DBLP:journals/corr/abs-2512-02902}
Q.~Z. Weiqi Li~and, R.~Zhai, L.~Lin, and G.~Wang, ``{VLA} models are more
  generalizable than you think: Revisiting physical, and spatial modeling,''
  \emph{CoRR}, vol. abs/2512.02902, 2025.

\bibitem{DBLP:journals/corr/abs-2509-00576}
T.~Y. Tao Jiang~and, Y.~Liu, C.~Lu, J.~Cui, X.~Liu, S.~Cheng, J.~Gao, H.~Xu,
  and H.~Zhao, ``Galaxea open-world dataset and {G0} dual-system {VLA} model,''
  \emph{CoRR}, vol. abs/2509.00576, 2025.

\bibitem{DBLP:conf/iclr/SudhakarNRLRC25}
H.~N. Arjun Vaithilingam Sudhakar~and, M.~Reymond, M.~Liu, J.~Rajendran, and
  S.~Chandar, ``A generalist hanabi agent,'' in \emph{{ICLR}}.\hskip 1em plus
  0.5em minus 0.4em\relax OpenReview.net, 2025.

\bibitem{vla_gen0_2025}
G.~A. Team, ``Gen-0: Embodied foundation models that scale with physical
  interaction,'' \emph{arXiv preprint}, 2025.

\bibitem{DBLP:journals/corr/abs-2509-14117}
M.~M. Ali Abouzeid~and, Z.~Sun, and D.~Song, ``Geoaware-vla: Implicit geometry
  aware vision-language-action model,'' \emph{CoRR}, vol. abs/2509.14117, 2025.

\bibitem{DBLP:conf/iros/SongZDCLFW24}
H.~Z. Wenxuan Song~and, P.~Ding, C.~Cui, S.~Lyu, Y.~Fan, and D.~Wang, ``Germ:
  {A} generalist robotic model with mixture-of-experts for quadruped, robot,''
  in \emph{{IROS}}.\hskip 1em plus 0.5em minus 0.4em\relax {IEEE}, 2024, pp.
  11\,879--11\,886.

\bibitem{DBLP:conf/iclr/ZhangDLPW25}
P.~D. Hongyin Zhang~and, S.~Lyu, Y.~Peng, and D.~Wang, ``{GEVRM:}
  goal-expressive video generation model for robust visual, manipulation,'' in
  \emph{{ICLR}}.\hskip 1em plus 0.5em minus 0.4em\relax OpenReview.net, 2025.

\bibitem{DBLP:journals/corr/abs-2508-05342}
L.~G. Shunlei Li~and, J.~Wang, C.~Che, X.~Xiao, J.~Cao, Y.~Hu, and H.~R.
  Karimi, ``Information-theoretic graph fusion with vision-language-action
  model, for policy reasoning and dual robotic control,'' \emph{CoRR}, vol.
  abs/2508.05342, 2025.

\bibitem{DBLP:journals/corr/abs-2512-09619}
M.~C. Minghao Guo~and, J.~Tao, R.~Xu, Y.~Yan, X.~Liang, I.~Laptev, and
  X.~Chang, ``Glad: Geometric latent distillation for vision-language-action
  models,'' \emph{CoRR}, vol. abs/2512.09619, 2025.

\bibitem{vla_gluestick_2025}
J.~Jabbour, D.-K. Kim, M.~Smith, J.~Patrikar, R.~Ghosal, Y.~Wang, A.~Agha,
  V.~J. Reddi, and S.~Omidshafiei, ``Dont run with scissors,'' \emph{arXiv
  preprint}, 2025.

\bibitem{DBLP:journals/corr/abs-2510-03342}
A.~A. Gemini Robotics Team~and, S.~Abeyruwan, J.~Ainslie, J.~Alayrac, M.~G.
  Arenas, A.~Balakrishna, N.~Batchelor, A.~Bewley, J.~T. Bingham, M.~Bloesch,
  K.~Bousmalis, P.~Brakel, A.~Brohan, T.~Buschmann, A.~Byravan, S.~Cabi,
  K.~Caluwaerts, F.~Casarini, C.~Chan, O.~Chang, L.~Chappellet{-}Volpini, J.~E.
  Chen, X.~Chen, H.~L. Chiang, K.~Choromanski, A.~Collister, D.~B. D'Ambrosio,
  S.~Dasari, T.~Davchev, M.~K. Dave, C.~Devin, N.~D. Palo, T.~Ding, C.~Doersch,
  A.~Dostmohamed, Y.~Du, D.~Dwibedi, S.~T. Egambaram, M.~Elabd, T.~Erez,
  X.~Fang, C.~Fantacci, C.~Fong, E.~Frey, C.~Fu, R.~Gao, M.~Giustina,
  K.~Gopalakrishnan, L.~Graesser, O.~Groth, A.~Gupta, R.~Hafner, S.~Hansen,
  L.~Hasenclever, S.~Haves, N.~Heess, B.~Hernaez, A.~Hofer, J.~Hsu, L.~Huang,
  S.~H. Huang, A.~Iscen, M.~G. Jacob, D.~Jain, S.~Jesmonth, A.~Jindal,
  R.~Julian, D.~Kalashnikov, M.~E. Karagozler, S.~Karp, M.~Kecman, J.~C. Kew,
  D.~Kim, F.~Kim, J.~Kim, T.~Kipf, S.~Kirmani, K.~Konyushkova, L.~Y. Ku,
  Y.~Kuang, T.~Lampe, A.~Laurens, T.~A. Le, I.~Leal, A.~X. Lee, T.~E. Lee,
  G.~Lever, J.~Liang, L.~Lin, F.~Liu, S.~Long, C.~Lu, S.~Maddineni,
  A.~Majumdar, K.~Maninis, A.~Marmon, S.~Martinez, A.~H. Michaely, and
  N.~Milonopoulos, ``Gemini robotics 1.5: Pushing the frontier of generalist
  robots with, advanced embodied reasoning, thinking, and motion transfer,''
  \emph{CoRR}, vol. abs/2510.03342, 2025.

\bibitem{DBLP:journals/corr/abs-2411-19309}
K.~Z. Zijian Zhang~and, Z.~Chen, J.~Jang, Y.~Li, C.~Wang, M.~Ding, D.~Fox, and
  H.~Yao, ``{GRAPE:} generalizing robot policy via preference alignment,''
  \emph{CoRR}, vol. abs/2411.19309, 2024.

\bibitem{DBLP:journals/corr/abs-2508-07650}
M.~C. Helong Huang~and, K.~Tan, X.~Quan, G.~Huang, and H.~Zhang,
  ``Graphcot-vla: {A} 3d spatial-aware reasoning vision-language-action, model
  for robotic manipulation with ambiguous instructions,'' \emph{CoRR}, vol.
  abs/2508.07650, 2025.

\bibitem{DBLP:journals/corr/abs-2505-03233}
M.~Y. Shengliang Deng~and, S.~Wei, H.~Ma, Y.~Yang, J.~Chen, Z.~Zhang, T.~Yang,
  X.~Zhang, H.~Cui, Z.~Zhang, and H.~Wang, ``Graspvla: a grasping foundation
  model pre-trained on billion-scale, synthetic action data,'' \emph{CoRR},
  vol. abs/2505.03233, 2025.

\bibitem{DBLP:journals/corr/abs-2511-04357}
Z.~F. Ma{\"{e}}lic Neau~and, P.~E. Santos, A.~Bosser, and C.~Buche,
  ``Grasp-vla: Graph-based symbolic action representation for long-horizon,
  planning with {VLA} policies,'' \emph{CoRR}, vol. abs/2511.04357, 2025.

\bibitem{DBLP:journals/corr/abs-2512-24210}
G.~C. Ruoshi Wen~and, Z.~Cui, M.~Du, Y.~Gou, Z.~Han, L.~Huang, M.~Lei, Y.~Li,
  Z.~Li, W.~Liu, Y.~Liu, X.~Ma, H.~Niu, Y.~Ouyang, Z.~Ren, H.~Shi, W.~Xu,
  H.~Zhang, J.~Zhang, X.~Zhang, L.~Zheng, W.~Zhong, Y.~Zhou, Z.~Zhu, and H.~Li,
  ``Gr-dexter technical report,'' \emph{CoRR}, vol. abs/2512.24210, 2025.

\bibitem{DBLP:journals/corr/abs-2512-01801}
X.~M. Yunfei Li~and, J.~Xu, Y.~Cui, Z.~Cui, Z.~Han, L.~Huang, T.~Kong, Y.~Liu,
  H.~Niu, W.~Peng, J.~Qiao, Z.~Ren, H.~Shi, Z.~Su, J.~Tian, Y.~Xiao, S.~Zhang,
  L.~Zheng, H.~Li, and Y.~Wu, ``{GR-RL:} going dexterous and precise for
  long-horizon robotic manipulation,'' \emph{CoRR}, vol. abs/2512.01801, 2025.

\bibitem{DBLP:conf/iclr/0038DZJMG0F00025}
Y.~D. Yi~Li~and, J.~Zhang, J.~Jang, M.~Memmel, C.~R. Garrett, F.~Ramos, D.~Fox,
  A.~Li, A.~Gupta, and A.~Goyal, ``{HAMSTER:} hierarchical action models for
  open-world robot manipulation,'' in \emph{{ICLR}}.\hskip 1em plus 0.5em minus
  0.4em\relax OpenReview.net, 2025.

\bibitem{vla_helix_2025}
FIGURE, ``Helix: A vision-language-action model for generalist humanoid
  control,'' \emph{arXiv preprint}, 2025.

\bibitem{DBLP:journals/corr/abs-2512-09928}
P.~D. Minghui Lin~and, S.~Wang, Z.~Zhuang, Y.~Liu, X.~Tong, W.~Song, S.~Lyu,
  S.~Huang, and D.~Wang, ``Hif-vla: Hindsight, insight and foresight through
  motion representation, for vision-language-action models,'' \emph{CoRR}, vol.
  abs/2512.09928, 2025.

\bibitem{DBLP:journals/corr/abs-2512-05693}
B.~L. Zhiying Du~and, Y.~Liang, Y.~Shen, H.~Cao, X.~Zheng, Z.~Feng, Z.~Wu,
  J.~Yang, and Y.~Jiang, ``Himoe-vla: Hierarchical mixture-of-experts for
  generalist vision-language-action, policies,'' \emph{CoRR}, vol.
  abs/2512.05693, 2025.

\bibitem{DBLP:journals/corr/abs-2510-26406}
R.~Z. Guanxing Lu~and, H.~Lin, H.~Zhang, and Y.~Tang, ``Human-in-the-loop
  online rejection sampling for robotic manipulation,'' \emph{CoRR}, vol.
  abs/2510.26406, 2025.

\bibitem{DBLP:conf/icml/ShiIEKPVTWWFLDG25}
B.~I. Lucy Xiaoyang Shi~and, M.~R. Equi, L.~Ke, K.~Pertsch, Q.~Vuong,
  J.~Tanner, A.~Walling, H.~Wang, N.~Fusai, A.~Li{-}Bell, D.~Driess, L.~Groom,
  S.~Levine, and C.~Finn, ``Hi robot: Open-ended instruction following with
  hierarchical vision-language-action, models,'' in \emph{{ICML}}.\hskip 1em
  plus 0.5em minus 0.4em\relax OpenReview.net, 2025.

\bibitem{DBLP:conf/corl/ZhangGCWHSC24}
Y.~G. Jianke Zhang~and, X.~Chen, Y.~Wang, Y.~Hu, C.~Shi, and J.~Chen, ``Hirt:
  Enhancing robotic control with hierarchical robot transformers,'' in
  \emph{CoRL}, ser. Proceedings of Machine Learning Research, vol. 270.\hskip
  1em plus 0.5em minus 0.4em\relax {PMLR}, 2024, pp. 933--946.

\bibitem{DBLP:conf/nips/XuZ00L24}
Y.~Z. Xinyu Xu~and, Y.~Li, L.~Han, and C.~Lu, ``Humanvla: Towards
  vision-language directed object rearrangement by, physical humanoid,'' in
  \emph{NeurIPS}, 2024.

\bibitem{DBLP:journals/corr/abs-2511-01914}
C.~X. Yuan Zhang~and, W.~Xu, C.~Ji, J.~Wu, and J.~Pan, ``iflybot-vla technical
  report,'' \emph{CoRR}, vol. abs/2511.01914, 2025.

\bibitem{DBLP:journals/corr/abs-2509-20841}
W.~G. Dekun Lu~and and K.~Jia, ``Imaginationpolicy: Towards generalizable,
  precise and reliable end-to-end, policy for robotic manipulation,''
  \emph{CoRR}, vol. abs/2509.20841, 2025.

\bibitem{DBLP:journals/corr/abs-2505-23757}
H.~G. Haohan Chi~and, Z.~Liu, J.~Liu, C.~Liu, J.~Li, K.~Yang, Y.~Yu, Z.~Wang,
  W.~Li, L.~Wang, X.~Hu, H.~Sun, H.~Zhao, and H.~Zhao, ``Impromptu {VLA:} open
  weights and open data for driving vision-language-action, models,''
  \emph{CoRR}, vol. abs/2505.23757, 2025.

\bibitem{DBLP:journals/corr/abs-2510-01389}
Z.~S. Ulas Berk Karli~and and T.~Fitzgerald, ``{INSIGHT:} inference-time
  sequence introspection for generating help, triggers in
  vision-language-action models,'' \emph{CoRR}, vol. abs/2510.01389, 2025.

\bibitem{DBLP:journals/corr/abs-2507-17520}
H.~L. Shuai Yang~and, Y.~Chen, B.~Wang, Y.~Tian, T.~Wang, H.~Wang, F.~Zhao,
  Y.~Liao, and J.~Pang, ``Instructvla: Vision-language-action instruction
  tuning from understanding, to manipulation,'' \emph{CoRR}, vol.
  abs/2507.17520, 2025.

\bibitem{DBLP:journals/corr/abs-2510-07778}
K.~G. Yandu Chen~and, Y.~Wen, Y.~Zhao, T.~Wang, and L.~Nie, ``Intentionvla:
  Generalizable and efficient embodied intention reasoning, for human-robot
  interaction,'' \emph{CoRR}, vol. abs/2510.07778, 2025.

\bibitem{DBLP:journals/corr/abs-2512-00797}
B.~M. Nan Sun~and, Y.~Li, C.~Wang, D.~Guo, and H.~Liu, ``Transforming
  monolithic foundation models into embodied multi-agent, architectures for
  human-robot collaboration,'' \emph{CoRR}, vol. abs/2512.00797, 2025.

\bibitem{DBLP:journals/corr/abs-2510-13778}
Y.~C. Xinyi Chen~and, Y.~Fu, N.~Gao, J.~Jia, W.~Jin, H.~Li, Y.~Mu, J.~Pang,
  Y.~Qiao, Y.~Tian, B.~Wang, B.~Wang, F.~Wang, H.~Wang, T.~Wang, Z.~Wang,
  X.~Wei, C.~Wu, S.~Yang, J.~Ye, J.~Yu, J.~Zeng, J.~Zhang, J.~Zhang, S.~Zhang,
  F.~Zheng, B.~Zhou, and Y.~Zhu, ``Internvla-m1: {A} spatially guided
  vision-language-action framework, for generalist robot policy,'' \emph{CoRR},
  vol. abs/2510.13778, 2025.

\bibitem{DBLP:conf/icra/GuoZCJWHC25}
J.~Z. Yanjiang Guo~and, X.~Chen, X.~Ji, Y.~Wang, Y.~Hu, and J.~Chen,
  ``Improving vision-language-action model with online reinforcement
  learning,'' in \emph{{ICRA}}.\hskip 1em plus 0.5em minus 0.4em\relax {IEEE},
  2025, pp. 15\,665--15\,672.

\bibitem{DBLP:journals/corr/abs-2508-06571}
Y.~G. Anqing Jiang~and, Y.~Wang, Z.~Sun, S.~Wang, Y.~Heng, H.~Sun, S.~Tang,
  L.~Zhu, J.~Chai, J.~Wang, Z.~Gu, H.~Jiang, and L.~Sun, ``{IRL-VLA:} training
  an vision-language-action policy via reward world, model,'' \emph{CoRR}, vol.
  abs/2508.06571, 2025.

\bibitem{vla_kvefficientvla_2025}
W.~Xu, L.~Zhuang, and L.~Shan, ``Kv-efficient vla: A method to speed up vision
  language models with rnn-gated chunked kv cache,'' \emph{arXiv preprint},
  2025.

\bibitem{DBLP:journals/corr/abs-2511-23034}
X.~G. Zuolei Li~and, X.~Wang, and J.~Fu, ``Latbot: Distilling universal latent
  actions for vision-language-action, models,'' \emph{CoRR}, vol.
  abs/2511.23034, 2025.

\bibitem{DBLP:journals/corr/abs-2509-18428}
Y.~N. Bahey Tharwat~and, A.~Abouzeid, and I.~Reid, ``Latent action pretraining
  through world modeling,'' \emph{CoRR}, vol. abs/2509.18428, 2025.

\bibitem{DBLP:journals/corr/abs-2512-10226}
K.~C. Shuhan Tan~and, Y.~Chen, R.~Tian, Y.~You, Y.~Wang, W.~Luo, Y.~Cao,
  P.~Kr{\"{a}}henb{\"{u}}hl, M.~Pavone, and B.~Ivanovic, ``Latent
  chain-of-thought world modeling for end-to-end driving,'' \emph{CoRR}, vol.
  abs/2512.10226, 2025.

\bibitem{DBLP:journals/corr/abs-2510-19752}
W.~C. Ameesh Shah~and, A.~Godbole, F.~Mora, S.~A. Seshia, and S.~Levine,
  ``Learning affordances at inference-time for vision-language-action,
  models,'' \emph{CoRR}, vol. abs/2510.19752, 2025.

\bibitem{DBLP:journals/corr/abs-2511-05642}
K.~D.~G. Justin Williams~and, R.~George, and M.~Sarkar, ``Lite {VLA:} efficient
  vision-language-action control on cpu-bound, edge robots,'' \emph{CoRR}, vol.
  abs/2511.05642, 2025.

\bibitem{DBLP:journals/corr/abs-2506-00411}
J.~S. Yi~Yang~and, S.~Kou, Y.~Wang, and Z.~Deng, ``Lohovla: {A} unified
  vision-language-action model for long-horizon, embodied tasks,'' \emph{CoRR},
  vol. abs/2506.00411, 2025.

\bibitem{DBLP:journals/corr/abs-2512-20166}
X.~G. Xiaofan Wang~and, J.~Fu, Z.~Li, D.~Fortier, G.~Mullins, A.~Kolobov, and
  B.~Guo, ``Lola: Long horizon latent action learning for general robot
  manipulation,'' \emph{CoRR}, vol. abs/2512.20166, 2025.

\bibitem{DBLP:journals/corr/abs-2508-19958}
P.~D. Yiguo Fan~and, S.~Bai, X.~Tong, Y.~Zhu, H.~Lu, F.~Dai, W.~Zhao, Y.~Liu,
  S.~Huang, Z.~Fan, B.~Chen, and D.~Wang, ``Long-vla: Unleashing long-horizon
  capability of vision language action, model for robot manipulation,''
  \emph{CoRR}, vol. abs/2508.19958, 2025.

\bibitem{DBLP:journals/corr/abs-2512-15840}
T.~Z. Boyuan Chen~and, H.~Geng, K.~Song, C.~Zhang, P.~Li, W.~T. Freeman,
  J.~Malik, P.~Abbeel, R.~Tedrake, V.~Sitzmann, and Y.~Du, ``Large video
  planner enables generalizable robot control,'' \emph{CoRR}, vol.
  abs/2512.15840, 2025.

\bibitem{DBLP:journals/corr/abs-2510-11660}
K.~G. Yi~Yang~and, Y.~Wen, H.~Li, Y.~Zhao, T.~Wang, and X.~Liu, ``Maniagent: An
  agentic framework for general robotic manipulation,'' \emph{CoRR}, vol.
  abs/2510.11660, 2025.

\bibitem{DBLP:journals/corr/abs-2511-16175}
X.~L. Yi~Yang~and, Y.~Chen, J.~Song, Y.~Wang, Z.~Xiao, J.~Su, Y.~Qiaoben,
  P.~Liu, and Z.~Deng, ``Mantis: {A} versatile vision-language-action model
  with disentangled, visual foresight,'' \emph{CoRR}, vol. abs/2511.16175,
  2025.

\bibitem{DBLP:journals/corr/abs-2512-02013}
J.~L. Chenyang Gu~and, H.~Chen, R.~Huang, Q.~Wuwu, Z.~Liu, X.~Li, Y.~Li,
  R.~Zhang, P.~Jia, P.~Heng, and S.~Zhang, ``Manualvla: {A} unified {VLA} model
  for chain-of-thought manual generation, and robotic manipulation,''
  \emph{CoRR}, vol. abs/2512.02013, 2025.

\bibitem{DBLP:journals/corr/abs-2511-09516}
W.~G. Runhao Li~and, Z.~Wu, C.~Wang, H.~Deng, Z.~Weng, Y.~Tan, and Z.~Wang,
  ``{MAP-VLA:} memory-augmented prompting for vision-language-action model, in
  robotic manipulation,'' \emph{CoRR}, vol. abs/2511.09516, 2025.

\bibitem{DBLP:journals/corr/abs-2511-21542}
J.~Z. Zhihao Zhan~and, L.~Zhang, Q.~Lv, H.~Liu, J.~Zhang, W.~Li, Z.~Chen,
  T.~Chen, K.~Wang, L.~Lin, and G.~Wang, ``{\(\mathscr{E}\)}\({}_{\mbox{0}}\):
  Enhancing generalization and fine-grained, control in {VLA} models via
  continuized discrete diffusion,'' \emph{CoRR}, vol. abs/2511.21542, 2025.

\bibitem{DBLP:conf/rss/ReussYWL24}
{\"{O}}.~E.~Y. Moritz Reuss~and, F.~Wenzel, and R.~Lioutikov, ``Multimodal
  diffusion transformer: Learning versatile behavior from, multimodal goals,''
  in \emph{Robotics: Science and Systems}, 2024.

\bibitem{DBLP:journals/corr/abs-2510-20328}
J.~P. Ajay Sridhar~and, S.~Sharma, and C.~Finn, ``Memer: Scaling up memory for
  robot control via experience retrieval,'' \emph{CoRR}, vol. abs/2510.20328,
  2025.

\bibitem{DBLP:journals/corr/abs-2508-19236}
B.~X. Hao Shi~and, Y.~Liu, L.~Sun, F.~Liu, T.~Wang, E.~Zhou, H.~Fan, X.~Zhang,
  and G.~Huang, ``Memoryvla: Perceptual-cognitive memory in
  vision-language-action models, for robotic manipulation,'' \emph{CoRR}, vol.
  abs/2508.19236, 2025.

\bibitem{DBLP:journals/corr/abs-2510-05580}
Z.~Y. Chen Li~and, H.~Zhang, F.~Chen, C.~Zhu, A.~Bolimera, and M.~Savvides,
  ``Metavla: Unified meta co-training for efficient embodied adaption,''
  \emph{CoRR}, vol. abs/2510.05580, 2025.

\bibitem{DBLP:journals/corr/abs-2510-05681}
D.~K. Suhyeok Jang~and, C.~Kim, Y.~Kim, and J.~Shin, ``Verifier-free test-time
  sampling for vision language action models,'' \emph{CoRR}, vol.
  abs/2510.05681, 2025.

\bibitem{DBLP:journals/corr/abs-2509-22199}
I.~Z. Haoyun Li~and, R.~Ouyang, X.~Wang, Z.~Zhu, Z.~Yang, Z.~Zhang, B.~Wang,
  C.~Ni, W.~Qin, X.~Chen, Y.~Ye, G.~Huang, Z.~Song, and X.~Wang,
  ``Mimicdreamer: Aligning human and robot demonstrations for scalable, {VLA}
  training,'' \emph{CoRR}, vol. abs/2509.22199, 2025.

\bibitem{DBLP:journals/corr/abs-2512-13636}
D.~Z. Haoyu Fu~and, Z.~Zhao, J.~Cui, H.~Xie, B.~Wang, G.~Chen, D.~Liang, and
  X.~Bai, ``Minddrive: {A} vision-language-action model for autonomous driving,
  via online reinforcement learning,'' \emph{CoRR}, vol. abs/2512.13636, 2025.

\bibitem{DBLP:journals/corr/abs-2512-08580}
S.~X. Peijun Tang~and, B.~Sun, B.~Huang, K.~Luo, H.~Yang, W.~Jin, and J.~Wang,
  ``Mind to hand: Purposeful robotic control via embodied reasoning,''
  \emph{CoRR}, vol. abs/2512.08580, 2025.

\bibitem{DBLP:journals/corr/abs-2512-15411}
X.~W. Zhenhan Yin~and, J.~Jiang, K.~Deng, P.~Chen, S.~Li, C.~Liu, X.~Xu,
  J.~Song, L.~Gao, and H.~T. Shen, ``Mivla: Towards generalizable
  vision-language-action model with human-robot, mutual imitation
  pre-training,'' \emph{CoRR}, vol. abs/2512.15411, 2025.

\bibitem{DBLP:journals/corr/abs-2509-26642}
J.~L. Zhuoyang Liu~and, J.~Xu, N.~Han, C.~Gu, H.~Chen, K.~Zhou, R.~Zhang,
  K.~Hsieh, K.~Wu, Z.~Che, J.~Tang, and S.~Zhang, ``{MLA:} {A} multisensory
  language-action model for multimodal understanding, and forecasting in
  robotic manipulation,'' \emph{CoRR}, vol. abs/2509.26642, 2025.

\bibitem{DBLP:journals/corr/abs-2512-00975}
X.~C. Haotian Liang~and, B.~Wang, M.~Chen, Y.~Liu, Y.~Zhang, Z.~Chen, T.~Yang,
  Y.~Chen, J.~Pang, D.~Liu, X.~Yang, Y.~Mu, W.~Shao, and P.~Luo, ``{MM-ACT:}
  learn from multimodal parallel generation to act,'' \emph{CoRR}, vol.
  abs/2512.00975, 2025.

\bibitem{DBLP:journals/corr/abs-2511-19433}
G.~W. Dong Jing~and, J.~Liu, W.~Tang, Z.~Sun, Y.~Yao, Z.~Wei, Y.~Liu, Z.~Lu,
  and M.~Ding, ``Mixture of horizons in action chunking,'' \emph{CoRR}, vol.
  abs/2511.19433, 2025.

\bibitem{DBLP:journals/corr/abs-2507-01843}
N.~S. Dmytro Kuzmenko~and, ``Moira: Modular instruction routing architecture
  for multi-task robotics,'' \emph{CoRR}, vol. abs/2507.01843, 2025.

\bibitem{DBLP:journals/corr/abs-2510-18316}
M.~X. Chengshu Li~and, A.~Bahety, H.~Yin, Y.~Jiang, H.~Huang, J.~Wong,
  S.~Garlanka, C.~Gokmen, R.~Zhang, W.~Liu, J.~Wu,
  R.~Mart{\'{\i}}n{-}Mart{\'{\i}}n, and L.~Fei{-}Fei, ``Momagen: Generating
  demonstrations under soft and hard constraints, for multi-step bimanual
  mobile manipulation,'' \emph{CoRR}, vol. abs/2510.18316, 2025.

\bibitem{DBLP:conf/cvpr/WuZX0Y25}
Y.~Z. Zhenyu Wu~and, X.~Xu, Z.~Wang, and H.~Yan, ``Momanipvla: Transferring
  vision-language-action models for general, mobile manipulation,'' in
  \emph{{CVPR}}.\hskip 1em plus 0.5em minus 0.4em\relax Computer Vision
  Foundation / {IEEE}, 2025, pp. 1714--1723.

\bibitem{DBLP:journals/corr/abs-2508-02549}
Y.~W. Shuo Wang~and, W.~Li, Y.~Wang, M.~Chen, K.~Wang, Z.~Su, X.~Cai, Y.~Jin,
  D.~Li, and Z.~Fan, ``Monodream: Monocular vision-language navigation with
  panoramic dreaming,'' \emph{CoRR}, vol. abs/2508.02549, 2025.

\bibitem{DBLP:conf/corl/StoneXLGLVWKZXF23}
T.~X. Austin Stone~and, Y.~Lu, K.~Gopalakrishnan, K.~Lee, Q.~Vuong,
  P.~Wohlhart, S.~Kirmani, B.~Zitkovich, F.~Xia, C.~Finn, and K.~Hausman,
  ``Open-world object manipulation using pre-trained vision-language models,''
  in \emph{CoRL}, ser. Proceedings of Machine Learning Research, vol.
  229.\hskip 1em plus 0.5em minus 0.4em\relax {PMLR}, 2023, pp. 3397--3417.

\bibitem{DBLP:conf/icra/ZhaoSWTDCG25}
W.~S. Han Zhao~and, D.~Wang, X.~Tong, P.~Ding, X.~Cheng, and Z.~Ge, ``More:
  Unlocking scalability in reinforcement learning for quadruped,
  vision-language-action models,'' in \emph{{ICRA}}.\hskip 1em plus 0.5em minus
  0.4em\relax {IEEE}, 2025, pp. 11\,212--11\,218.

\bibitem{DBLP:journals/corr/abs-2509-17759}
R.~Z. Chengbo Yuan~and, M.~Liu, Y.~Hu, S.~Wang, L.~Yi, C.~Wen, S.~Zhang, and
  Y.~Gao, ``Motiontrans: Human {VR} data enable motion-level learning for
  robotic, manipulation policies,'' \emph{CoRR}, vol. abs/2509.17759, 2025.

\bibitem{DBLP:journals/corr/abs-2509-01658}
A.~M. Zhenyu Wu~and, X.~Xu, H.~Yin, Y.~Liang, Z.~Wang, J.~Lu, and H.~Yan,
  ``Moto: {A} zero-shot plug-in interaction-aware navigation for general,
  mobile manipulation,'' \emph{CoRR}, vol. abs/2509.01658, 2025.

\bibitem{DBLP:journals/corr/abs-2509-19958}
J.~Z. Alexander Spiridonov~and, N.~Nikolov, L.~V. Gool, and D.~P. Paudel,
  ``Generalist robot manipulation beyond action labeled data,'' \emph{CoRR},
  vol. abs/2509.19958, 2025.

\bibitem{DBLP:journals/corr/abs-2512-13030}
H.~T. Hongzhe Bi~and, S.~Xie, Z.~Wang, S.~Huang, H.~Liu, R.~Zhao, Y.~Feng,
  C.~Xiang, Y.~Rong, H.~Zhao, H.~Liu, Z.~Su, L.~Ma, H.~Su, and J.~Zhu, ``Motus:
  {A} unified latent action world model,'' \emph{CoRR}, vol. abs/2512.13030,
  2025.

\bibitem{DBLP:journals/corr/abs-2512-22208}
X.~S. Pu~Zhao~and, Z.~Kong, Y.~Shen, S.~Chang, A.~Akbari, T.~Rupprecht, L.~Lu,
  E.~Nan, C.~Yang, Y.~He, W.~Shi, X.~Xu, Y.~Huang, W.~Jiang, W.~Wang, Y.~Chen,
  Y.~He, and Y.~Wang, ``Open-source multimodal moxin models with moxin-vlm and
  moxin-vla,'' \emph{CoRR}, vol. abs/2512.22208, 2025.

\bibitem{DBLP:journals/corr/abs-2509-25966}
F.~J. Peilong Han~and, M.~Zhang, Y.~Qiu, H.~Tang, Y.~Zheng, T.~Wang, and
  J.~Hao, ``{MUVLA:} learning to explore object navigation via map
  understanding,'' \emph{CoRR}, vol. abs/2509.25966, 2025.

\bibitem{DBLP:journals/corr/abs-2412-04453}
Y.~J. An{-}Chieh Cheng~and, Z.~Yang, X.~Zou, J.~Kautz, E.~Biyik, H.~Yin,
  S.~Liu, and X.~Wang, ``Navila: Legged robot vision-language-action model for
  navigation,'' \emph{CoRR}, vol. abs/2412.04453, 2024.

\bibitem{DBLP:journals/corr/abs-2511-22777}
M.~P. Sajjad Pakdamansavoji~and, A.~Sigal, Z.~Li, R.~H. Yang, and A.~Rasouli,
  ``Improving robotic manipulation robustness via {NICE} scene surgery,''
  \emph{CoRR}, vol. abs/2511.22777, 2025.

\bibitem{vla_nitrogen_2026}
L.~Magne, A.~Awadalla, G.~Wang, Y.~Xu, J.~Belofsky, F.~Hu, J.~Kim, L.~Schmidt,
  G.~Gkioxari, J.~Kautz, Y.~Yue, Y.~Choi, Y.~Zhu, and L.~J. Fan, ``Nitrogen: An
  open foundation model for generalist gaming agents,'' \emph{arXiv preprint},
  2026.

\bibitem{DBLP:journals/corr/abs-2509-23655}
D.~D. Rokas Bendikas~and, M.~Peschl, S.~Haresh, and P.~Mazzaglia, ``Focusing on
  what matters: Object-agent-centric tokenization for vision, language action
  models,'' \emph{CoRR}, vol. abs/2509.23655, 2025.

\bibitem{DBLP:journals/corr/abs-2512-22519}
T.~H. Khoa Vo~and, Y.~Ikebe, T.~Pham, N.~Chung, M.~N. Vu, D.~N.~H. Minh,
  A.~Nguyen, A.~Gunderman, C.~Rainwater, and N.~Le, ``Clutter-resistant
  vision-language-action models through object-centric, and geometry
  grounding,'' \emph{CoRR}, vol. abs/2512.22519, 2025.

\bibitem{DBLP:journals/corr/abs-2409-03272}
S.~Y. Julong Wei~and, P.~Li, Q.~Hu, Z.~Gan, and W.~Ding, ``Occllama: An
  occupancy-language-action generative world model for, autonomous driving,''
  \emph{CoRR}, vol. abs/2409.03272, 2024.

\bibitem{DBLP:journals/corr/abs-2509-05578}
L.~K. Ruixun Liu~and, D.~Li, and H.~Zhao, ``Occvla: Vision-language-action
  model with implicit 3d occupancy supervision,'' \emph{CoRR}, vol.
  abs/2509.05578, 2025.

\bibitem{DBLP:journals/corr/abs-2508-13103}
H.~D. Tianyi Zhang~and, H.~Hao, Y.~Qiao, J.~Dai, and Z.~Hou, ``Grounding
  actions in camera space: Observation-centric vision-language-action,
  policy,'' \emph{CoRR}, vol. abs/2508.13103, 2025.

\bibitem{DBLP:conf/nips/WangCMLZL0LML24}
S.~C. Zihao Wang~and, Z.~Mu, H.~Lin, C.~Zhang, X.~Liu, Q.~Li, A.~Liu, X.~S. Ma,
  and Y.~Liang, ``Omnijarvis: Unified vision-language-action tokenization
  enables open-world, instruction following agents,'' in \emph{NeurIPS}, 2024.

\bibitem{DBLP:journals/corr/abs-2509-00789}
Q.~N. Pei Liu~and, X.~Lu, H.~Liu, W.~Ma, D.~She, P.~Jia, X.~Lang, and J.~Ma,
  ``Omnireason: {A} temporal-guided vision-language-action framework for,
  autonomous driving,'' \emph{CoRR}, vol. abs/2509.00789, 2025.

\bibitem{DBLP:journals/corr/abs-2510-09667}
C.~C. Huaihai Lyu~and, S.~Xie, P.~Wang, X.~Chen, S.~Zhang, and C.~Xu,
  ``Omnisat: Compact action token, faster auto regression,'' \emph{CoRR}, vol.
  abs/2510.09667, 2025.

\bibitem{DBLP:journals/corr/abs-2509-19480}
C.~G. Noriaki Hirose~and, D.~Shah, and S.~Levine, ``Omnivla: An omni-modal
  vision-language-action model for robot navigation,'' \emph{CoRR}, vol.
  abs/2509.19480, 2025.

\bibitem{DBLP:journals/corr/abs-2505-11917}
R.~N. Fanqi Lin~and, Y.~Hu, J.~You, J.~Zhao, and Y.~Gao, ``Onetwovla: {A}
  unified vision-language-action model with adaptive, reasoning,'' \emph{CoRR},
  vol. abs/2505.11917, 2025.

\bibitem{DBLP:journals/corr/abs-2509-13347}
M.~L. Zihao Wang~and, K.~He, X.~Wang, Z.~Mu, A.~Liu, and Y.~Liang, ``Openha:
  {A} series of open-source hierarchical agentic models in minecraft,''
  \emph{CoRR}, vol. abs/2509.13347, 2025.

\bibitem{DBLP:conf/corl/KimPKXB0RFSVKBT24}
K.~P. Moo Jin Kim~and, S.~Karamcheti, T.~Xiao, A.~Balakrishna, S.~Nair,
  R.~Rafailov, E.~P. Foster, P.~R. Sanketi, Q.~Vuong, T.~Kollar, B.~Burchfiel,
  R.~Tedrake, D.~Sadigh, S.~Levine, P.~Liang, and C.~Finn, ``Openvla: An
  open-source vision-language-action model,'' in \emph{CoRL}, ser. Proceedings
  of Machine Learning Research, vol. 270.\hskip 1em plus 0.5em minus
  0.4em\relax {PMLR}, 2024, pp. 2679--2713.

\bibitem{DBLP:conf/icml/HuangLFWMMGA25}
F.~L. Huang Huang~and, L.~Fu, T.~Wu, M.~Mukadam, J.~Malik, K.~Goldberg, and
  P.~Abbeel, ``{OTTER:} {A} vision-language-action model with text-aware visual
  feature, extraction,'' in \emph{{ICML}}.\hskip 1em plus 0.5em minus
  0.4em\relax OpenReview.net, 2025.

\bibitem{DBLP:journals/corr/abs-2512-16793}
S.~L. Xiaopeng Lin~and, B.~Yu, R.~Yang, C.~Wu, Y.~Miao, Y.~Jin, Y.~Shi,
  C.~Huang, B.~Cheng, and K.~Chen, ``Physbrain: Human egocentric data as a
  bridge from vision language, models to physical intelligence,'' \emph{CoRR},
  vol. abs/2512.16793, 2025.

\bibitem{DBLP:journals/corr/abs-2509-24524}
J.~L. Zhihao Wang~and, J.~Zheng, W.~Zhang, D.~Liu, Y.~Zheng, H.~Niu, J.~Yu, and
  X.~Zhan, ``Physiagent: An embodied agent framework in physical world,''
  \emph{CoRR}, vol. abs/2509.24524, 2025.

\bibitem{DBLP:journals/corr/abs-2510-25889}
K.~Chen, Z.~Liu, T.~Zhang, Z.~Guo, S.~Xu, H.~Lin, H.~Zang, Q.~Zhang, Z.~Yu,
  G.~Fan, T.~Huang, Y.~Wang, and C.~Yu, ``{\(\pi\)}\({}_{\mbox{rl}}\): Online
  {RL} fine-tuning for flow-based vision-language-action models,'' \emph{CoRR},
  vol. abs/2510.25889, 2025.

\bibitem{DBLP:journals/corr/abs-2511-01571}
G.~S. Wenqi Liang~and, Y.~He, J.~Dong, S.~Dai, I.~Laptev, S.~H. Khan, and
  Y.~Cong, ``Pixelvla: Advancing pixel-level understanding in
  vision-language-action, model,'' \emph{CoRR}, vol. abs/2511.01571, 2025.

\bibitem{DBLP:journals/corr/abs-2507-23540}
E.~L.~H. Yi~Zhang~and, K.~Chao, N.~Petrovic, Y.~Song, C.~Wu, and A.~Knoll, ``A
  unified perception-language-action framework for adaptive autonomous,
  driving,'' \emph{CoRR}, vol. abs/2507.23540, 2025.

\bibitem{DBLP:journals/corr/abs-2512-18933}
J.~Z. Hang Yu~and, Y.~Liu, K.~Li, C.~Ma, D.~Zhang, Y.~Hu, G.~Chen, J.~Xie,
  J.~Guo, J.~Zhao, and Y.~Gao, ``Point what you mean: Visually grounded
  instruction policy,'' \emph{CoRR}, vol. abs/2512.18933, 2025.

\bibitem{DBLP:journals/corr/abs-2512-03724}
X.~W. Ziwen Li~and, H.~Zhang, R.~Chen, R.~Lin, X.~He, H.~Huang, Y.~Guo,
  F.~Karray, T.~Liu, and M.~Gong, ``Posa-vla: Enhancing action generation via
  pose-conditioned anchor, attention,'' \emph{CoRR}, vol. abs/2512.03724, 2025.

\bibitem{DBLP:journals/corr/abs-2511-20633}
Z.~H. Jiahui Zhang~and, C.~Gu, Z.~Ma, and L.~Zhang, ``Reinforcing action
  policies by prophesying,'' \emph{CoRR}, vol. abs/2511.20633, 2025.

\bibitem{DBLP:journals/corr/abs-2412-01034}
H.~K. Seongmin Park~and, W.~Jeon, J.~Yang, B.~Jeon, Y.~Oh, and J.~Choi,
  ``Quantization-aware imitation-learning for resource-efficient robotic,
  control,'' \emph{CoRR}, vol. abs/2412.01034, 2024.

\bibitem{DBLP:journals/corr/abs-2510-14836}
Y.~C. Yixuan Li~and, M.~Zhou, and H.~Li, ``Qdepth-vla: Quantized depth
  prediction as auxiliary supervision for, vision-language-action models,''
  \emph{CoRR}, vol. abs/2510.14836, 2025.

\bibitem{DBLP:conf/corl/ChebotarVHXLIKY23}
Q.~V. Yevgen Chebotar~and, K.~Hausman, F.~Xia, Y.~Lu, A.~Irpan, A.~Kumar,
  T.~Yu, A.~Herzog, K.~Pertsch, K.~Gopalakrishnan, J.~Ibarz, O.~Nachum, S.~A.
  Sontakke, G.~Salazar, H.~T. Tran, J.~Peralta, C.~Tan, D.~Manjunath, J.~Singh,
  B.~Zitkovich, T.~Jackson, K.~Rao, C.~Finn, and S.~Levine, ``Q-transformer:
  Scalable offline reinforcement learning via autoregressive, q-functions,'' in
  \emph{CoRL}, ser. Proceedings of Machine Learning Research, vol. 229.\hskip
  1em plus 0.5em minus 0.4em\relax {PMLR}, 2023, pp. 3909--3928.

\bibitem{DBLP:conf/icra/TongDFWZCSZZDHL25}
P.~D. Xinyang Tong~and, Y.~Fan, D.~Wang, W.~Zhang, C.~Cui, M.~Sun, H.~Zhao,
  H.~Zhang, Y.~Dang, S.~Huang, and S.~Lyu, ``Quart-online: Latency-free
  multimodal large language model for quadruped, robot learning,'' in
  \emph{{ICRA}}.\hskip 1em plus 0.5em minus 0.4em\relax {IEEE}, 2025, pp.
  9533--9539.

\bibitem{DBLP:journals/corr/abs-2505-09601}
L.~F. Justin Yu~and, H.~Huang, K.~El{-}Refai, R.~A. Ambrus, R.~Cheng, M.~Z.
  Irshad, and K.~Goldberg, ``Real2render2real: Scaling robot data without
  dynamics simulation or, robot hardware,'' \emph{CoRR}, vol. abs/2505.09601,
  2025.

\bibitem{DBLP:conf/iclr/LiuWLTCWX0025}
L.~W. Songming Liu~and, B.~Li, H.~Tan, H.~Chen, Z.~Wang, K.~Xu, H.~Su, and
  J.~Zhu, ``{RDT-1B:} a diffusion foundation model for bimanual manipulation,''
  in \emph{{ICLR}}.\hskip 1em plus 0.5em minus 0.4em\relax OpenReview.net,
  2025.

\bibitem{DBLP:journals/corr/abs-2511-19912}
Z.~Y. Dapeng Zhang~and, Z.~Chen, C.~Liao, Y.~Chen, F.~Shen, Q.~Zhou, and
  T.~Chua, ``Reasoning-vla: {A} fast and general vision-language-action
  reasoning, model for autonomous driving,'' \emph{CoRR}, vol. abs/2511.19912,
  2025.

\bibitem{DBLP:journals/corr/abs-2509-20109}
Y.~Z. Pengxiang Li~and, Y.~Wang, H.~Wang, H.~Zhao, J.~Liu, X.~Zhan, K.~Zhan,
  and X.~Lang, ``Discrete diffusion for reflective vision-language-action
  models in, autonomous driving,'' \emph{CoRR}, vol. abs/2509.20109, 2025.

\bibitem{DBLP:conf/iclr/SridharDJ025}
S.~D. Kaustubh Sridhar~and, D.~Jayaraman, and I.~Lee, ``{REGENT:} {A}
  retrieval-augmented generalist agent that can act in-context, in new
  environments,'' in \emph{{ICLR}}.\hskip 1em plus 0.5em minus 0.4em\relax
  OpenReview.net, 2025.

\bibitem{DBLP:conf/corl/HuangWLZF24}
C.~W. Wenlong Huang~and, Y.~Li, R.~Zhang, and L.~Fei{-}Fei, ``Rekep:
  Spatio-temporal reasoning of relational keypoint constraints, for robotic
  manipulation,'' in \emph{CoRL}, ser. Proceedings of Machine Learning
  Research, vol. 270.\hskip 1em plus 0.5em minus 0.4em\relax {PMLR}, 2024, pp.
  4573--4602.

\bibitem{DBLP:journals/corr/abs-2512-08333}
Z.~Z. Yajat Yadav~and, A.~Wagenmaker, K.~Pertsch, and S.~Levine, ``Robust
  finetuning of vision-language-action robot policies via parameter, merging,''
  \emph{CoRR}, vol. abs/2512.08333, 2025.

\bibitem{DBLP:journals/corr/abs-2509-21243}
T.~C. Jiyeon Koo~and, H.~Kang, E.~Pyo, T.~G. Oh, T.~Kim, and A.~J. Choi,
  ``Retovla: Reusing register tokens for spatial reasoning in
  vision-language-action, models,'' \emph{CoRR}, vol. abs/2509.21243, 2025.

\bibitem{DBLP:conf/icra/DeyZNGP25}
J.~Z. Sombit Dey~and, N.~Nikolov, L.~V. Gool, and D.~P. Paudel, ``Revla:
  Reverting visual domain limitation of robotic foundation models,'' in
  \emph{{ICRA}}.\hskip 1em plus 0.5em minus 0.4em\relax {IEEE}, 2025, pp.
  8679--8686.

\bibitem{DBLP:journals/corr/abs-2508-02062}
S.~D. Kaustubh Sridhar~and, D.~Jayaraman, and I.~Lee, ``{RICL:} adding
  in-context adaptability to pre-trained vision-language-action, models,''
  \emph{CoRR}, vol. abs/2508.02062, 2025.

\bibitem{DBLP:journals/tmlr/BousmalisVRDLVD24}
G.~V. Konstantinos Bousmalis~and, D.~Rao, C.~M. Devin, A.~X. Lee, M.~B.
  Villalonga, T.~Davchev, Y.~Zhou, A.~Gupta, A.~Raju, A.~Laurens, C.~Fantacci,
  V.~Dalibard, M.~Zambelli, M.~F. Martins, R.~Pevceviciute, M.~Blokzijl,
  M.~Denil, N.~Batchelor, T.~Lampe, E.~Parisotto, K.~Zolna, S.~E. Reed, S.~G.
  Colmenarejo, J.~Scholz, A.~Abdolmaleki, O.~Groth, J.~Regli, O.~Sushkov,
  T.~Roth{\"{o}}rl, J.~E. Chen, Y.~Aytar, D.~Barker, J.~Ortiz, M.~A.
  Riedmiller, J.~T. Springenberg, R.~Hadsell, F.~Nori, and N.~Heess, ``Robocat:
  {A} self-improving generalist agent for robotic manipulation,'' \emph{Trans.
  Mach. Learn. Res.}, vol. 2024, 2024.

\bibitem{DBLP:journals/corr/abs-2509-08820}
C.~Y. Zongzheng Zhang~and, H.~Xu, M.~Liao, X.~Qi, H.~Gao, Z.~Wang, and H.~Zhao,
  ``Robochemist: Long-horizon and safety-compliant robotic chemical
  experimentation,'' \emph{CoRR}, vol. abs/2509.08820, 2025.

\bibitem{DBLP:journals/corr/abs-2410-08001}
H.~L. Qingwen Bu~and, L.~Chen, J.~Cai, J.~Zeng, H.~Cui, M.~Yao, and Y.~Qiao,
  ``Towards synergistic, generalized, and efficient dual-system for robotic,
  manipulation,'' \emph{CoRR}, vol. abs/2410.08001, 2024.

\bibitem{DBLP:conf/iclr/LiLZYXWCJ0LLK24}
M.~L. Xinghang Li~and, H.~Zhang, C.~Yu, J.~Xu, H.~Wu, C.~Cheang, Y.~Jing,
  W.~Zhang, H.~Liu, H.~Li, and T.~Kong, ``Vision-language foundation models as
  effective robot imitators,'' in \emph{{ICLR}}.\hskip 1em plus 0.5em minus
  0.4em\relax OpenReview.net, 2024.

\bibitem{DBLP:journals/corr/abs-2412-00171}
W.~Z. Weixin Mao~and, Z.~Jiang, D.~Fang, Z.~Zhang, Z.~Lan, F.~Jia, T.~Wang,
  H.~Fan, and O.~Yoshie, ``Robomatrix: {A} skill-centric hierarchical framework
  for scalable, robot task planning and execution in open-world,'' \emph{CoRR},
  vol. abs/2412.00171, 2024.

\bibitem{DBLP:journals/corr/abs-2512-10394}
H.~X. Weifan Guan~and, C.~Zhang, A.~Li, Q.~Hu, and J.~Cheng, ``Roboneuron: {A}
  modular framework linking foundation models and {ROS}, for embodied {AI},''
  \emph{CoRR}, vol. abs/2512.10394, 2025.

\bibitem{DBLP:conf/iros/LiWDMNHL25}
J.~W. Shunlei Li~and, R.~Dai, W.~Ma, W.~Y. Ng, Y.~Hu, and Z.~Li,
  ``Robonurse-vla: Robotic scrub nurse system based on vision-language-action,
  model,'' in \emph{{IROS}}.\hskip 1em plus 0.5em minus 0.4em\relax {IEEE},
  2025, pp. 3986--3993.

\bibitem{DBLP:journals/corr/abs-2510-23763}
J.~F. Siyin Wang~and, F.~Liu, X.~He, H.~Wu, J.~Shi, K.~Huang, Z.~Fei, J.~Gong,
  Z.~Wu, Y.~Jiang, S.~Ng, T.~Chua, and X.~Qiu, ``Roboomni: Proactive robot
  manipulation in omni-modal context,'' \emph{CoRR}, vol. abs/2510.23763, 2025.

\bibitem{DBLP:journals/corr/abs-2510-26536}
C.~C. Huajie Tan~and, X.~Chen, Y.~Ji, Z.~Zhao, X.~Hao, Y.~Lyu, M.~Cao, J.~Zhao,
  H.~Lyu, E.~Zhou, N.~Chen, Y.~Fu, C.~Peng, W.~Guo, D.~Liang, Z.~Chen, M.~Lyu,
  C.~He, Y.~Ao, Y.~Lin, P.~Wang, Z.~Wang, and S.~Zhang, ``Roboos-next: {A}
  unified memory-based framework for lifelong, scalable, and robust multi-robot
  collaboration,'' \emph{CoRR}, vol. abs/2510.26536, 2025.

\bibitem{DBLP:conf/corl/YuanDBPKMMF24}
J.~D. Wentao Yuan~and, V.~Blukis, W.~Pumacay, R.~Krishna, A.~Murali,
  A.~Mousavian, and D.~Fox, ``Robopoint: {A} vision-language model for spatial
  affordance prediction, in robotics,'' in \emph{CoRL}, ser. Proceedings of
  Machine Learning Research, vol. 270.\hskip 1em plus 0.5em minus 0.4em\relax
  {PMLR}, 2024, pp. 4005--4020.

\bibitem{DBLP:journals/corr/abs-2510-25713}
C.~Y. Boshi An~and and R.~K. Katzschmann, ``Robotic assistant: Completing
  collaborative tasks with dexterous vision-language-action, models,''
  \emph{CoRR}, vol. abs/2510.25713, 2025.

\bibitem{DBLP:journals/corr/abs-2412-14058}
P.~L. Xinghang Li~and, M.~Liu, D.~Wang, J.~Liu, B.~Kang, X.~Ma, T.~Kong,
  H.~Zhang, and H.~Liu, ``Towards generalist robot policies: What matters in
  building vision-language-action, models,'' \emph{CoRR}, vol. abs/2412.14058,
  2024.

\bibitem{DBLP:journals/corr/abs-2510-00037}
Z.~W. Jianing Guo~and, C.~Tu, Y.~Ma, X.~Kong, Z.~Liu, J.~Ji, S.~Zhang, Y.~Chen,
  K.~Chen, Q.~Dou, Y.~Yang, X.~Liu, H.~Zhao, W.~Lv, and S.~Li, ``On robustness
  of vision-language-action model against multi-modal, perturbations,''
  \emph{CoRR}, vol. abs/2510.00037, 2025.

\bibitem{DBLP:journals/corr/abs-2510-10975}
L.~L. Mingtong Dai~and, Y.~Bai, Y.~Liu, Z.~Wang, R.~SU, C.~Chen, L.~Lin, and
  X.~Wu, ``Rover: Robot reward model as test-time verifier for
  vision-language-action, model,'' \emph{CoRR}, vol. abs/2510.10975, 2025.

\bibitem{DBLP:conf/iros/JulgBW25}
W.~B. Tobias J{\"{u}}lg~and and F.~Walter, ``Refined policy distillation: From
  {VLA} generalists to {RL} experts,'' in \emph{{IROS}}.\hskip 1em plus 0.5em
  minus 0.4em\relax {IEEE}, 2025, pp. 11\,677--11\,684.

\bibitem{DBLP:journals/corr/abs-2510-01711}
J.~L. Taeyoung Kim~and, M.~Koo, D.~Kim, K.~Lee, C.~Kim, Y.~Seo, and J.~Shin,
  ``Contrastive representation regularization for vision-language-action,
  models,'' \emph{CoRR}, vol. abs/2510.01711, 2025.

\bibitem{DBLP:conf/corl/ZitkovichYXXXXW23}
T.~Y. Brianna Zitkovich~and, S.~Xu, P.~Xu, T.~Xiao, F.~Xia, J.~Wu, P.~Wohlhart,
  S.~Welker, A.~Wahid, Q.~Vuong, V.~Vanhoucke, H.~T. Tran, R.~Soricut,
  A.~Singh, J.~Singh, P.~Sermanet, P.~R. Sanketi, G.~Salazar, M.~S. Ryoo,
  K.~Reymann, K.~Rao, K.~Pertsch, I.~Mordatch, H.~Michalewski, Y.~Lu,
  S.~Levine, L.~Lee, T.~E. Lee, I.~Leal, Y.~Kuang, D.~Kalashnikov, R.~Julian,
  N.~J. Joshi, A.~Irpan, B.~Ichter, J.~Hsu, A.~Herzog, K.~Hausman,
  K.~Gopalakrishnan, C.~Fu, P.~Florence, C.~Finn, K.~A. Dubey, D.~Driess,
  T.~Ding, K.~M. Choromanski, X.~Chen, Y.~Chebotar, J.~Carbajal, N.~Brown,
  A.~Brohan, M.~G. Arenas, and K.~Han, ``{RT-2:} vision-language-action models
  transfer web knowledge to robotic, control,'' in \emph{CoRL}, ser.
  Proceedings of Machine Learning Research, vol. 229.\hskip 1em plus 0.5em
  minus 0.4em\relax {PMLR}, 2023, pp. 2165--2183.

\bibitem{DBLP:conf/icra/NasirianyKDSZDSX25}
S.~K. Soroush Nasiriany~and, T.~Ding, L.~Smith, Y.~Zhu, D.~Driess, D.~Sadigh,
  and T.~Xiao, ``Rt-affordance: Affordances are versatile intermediate
  representations, for robot manipulation,'' in \emph{{ICRA}}.\hskip 1em plus
  0.5em minus 0.4em\relax {IEEE}, 2025, pp. 8249--8257.

\bibitem{DBLP:conf/rss/BelkhaleDXSVTCD24}
T.~D. Suneel Belkhale~and, T.~Xiao, P.~Sermanet, Q.~Vuong, J.~Tompson,
  Y.~Chebotar, D.~Dwibedi, and D.~Sadigh, ``{RT-H:} action hierarchies using
  language,'' in \emph{Robotics: Science and Systems}, 2024.

\bibitem{DBLP:journals/natmi/SchmidgallKKGK24}
J.~W.~K. Samuel Schmidgall~and, A.~Kuntz, A.~E. Ghazi, and A.~Krieger,
  ``General-purpose foundation models for increased autonomy in robot-assisted,
  surgery,'' \emph{Nat. Mac. Intell.}, vol.~6, no.~11, pp. 1275--1283, 2024.

\bibitem{DBLP:conf/iclr/GuKW0AR0FGXSX0H24}
S.~K. Jiayuan Gu~and, P.~Wohlhart, Y.~Lu, M.~G. Arenas, K.~Rao, W.~Yu, C.~Fu,
  K.~Gopalakrishnan, Z.~Xu, P.~Sundaresan, P.~Xu, H.~Su, K.~Hausman, C.~Finn,
  Q.~Vuong, and T.~Xiao, ``Rt-trajectory: Robotic task generalization via
  hindsight trajectory, sketches,'' in \emph{{ICLR}}.\hskip 1em plus 0.5em
  minus 0.4em\relax OpenReview.net, 2024.

\bibitem{DBLP:conf/icra/ONeillRMGPLPGMJ24}
A.~R. Abby O'Neill~and, A.~Maddukuri, A.~Gupta, A.~Padalkar, A.~Lee, A.~Pooley,
  A.~Gupta, A.~Mandlekar, A.~Jain, A.~Tung, A.~Bewley, A.~Herzog, A.~Irpan,
  A.~Khazatsky, A.~Rai, A.~Gupta, A.~E. Wang, A.~Singh, A.~Garg, A.~Kembhavi,
  A.~Xie, A.~Brohan, A.~Raffin, A.~Sharma, A.~Yavary, A.~Jain, A.~Balakrishna,
  A.~Wahid, B.~Burgess{-}Limerick, B.~Kim, B.~Sch{\"{o}}lkopf, B.~Wulfe,
  B.~Ichter, C.~Lu, C.~Xu, C.~Le, C.~Finn, C.~Wang, C.~Xu, C.~Chi, C.~Huang,
  C.~Chan, C.~Agia, C.~Pan, C.~Fu, C.~Devin, D.~Xu, D.~Morton, D.~Driess,
  D.~Chen, D.~Pathak, D.~Shah, D.~B{\"{u}}chler, D.~Jayaraman, D.~Kalashnikov,
  D.~Sadigh, E.~Johns, E.~P. Foster, F.~Liu, F.~Ceola, F.~Xia, F.~Zhao,
  F.~Stulp, G.~Zhou, G.~S. Sukhatme, G.~Salhotra, G.~Yan, G.~Feng, G.~Schiavi,
  G.~Berseth, G.~Kahn, G.~Wang, H.~Su, H.~Fang, H.~Shi, H.~Bao, H.~B. Amor,
  H.~I. Christensen, H.~Furuta, H.~Walke, H.~Fang, H.~Ha, I.~Mordatch,
  I.~Radosavovic, I.~Leal, J.~Liang, J.~Abou{-}Chakra, J.~Kim, J.~Drake,
  J.~Peters, J.~Schneider, J.~Hsu, J.~Bohg, J.~T. Bingham, J.~Wu, J.~Gao,
  J.~Hu, J.~Wu, J.~Wu, J.~Sun, J.~Luo, J.~Gu, J.~Tan, J.~Oh, J.~Wu, J.~Lu,
  J.~Yang, J.~Malik, J.~Silv{\'{e}}rio, J.~Hejna, J.~Booher, J.~Tompson,
  J.~Yang, J.~Salvador, J.~J. Lim, J.~Han, K.~Wang, K.~Rao, K.~Pertsch,
  K.~Hausman, K.~Go, K.~Gopalakrishnan, K.~Goldberg, K.~Byrne, K.~Oslund,
  K.~Kawaharazuka, K.~Black, K.~Lin, K.~Zhang, K.~Ehsani, K.~Lekkala, K.~Ellis,
  K.~Rana, K.~Srinivasan, K.~Fang, K.~P. Singh, K.~Zeng, K.~Hatch, K.~Hsu,
  L.~Itti, L.~Y. Chen, L.~Pinto, L.~Fei{-}Fei, L.~Tan, L.~J. Fan, L.~Ott,
  L.~Lee, L.~Weihs, M.~Chen, M.~Lepert, M.~Memmel, M.~Tomizuka, M.~Itkina,
  M.~G. Castro, M.~Spero, M.~Du, M.~Ahn, M.~C. Yip, M.~Zhang, M.~Ding, M.~Heo,
  M.~K. Srirama, M.~Sharma, M.~J. Kim, N.~Kanazawa, N.~Hansen, N.~Heess, N.~J.
  Joshi, N.~S{\"{u}}nderhauf, N.~Liu, N.~D. Palo, N.~M.~M. Shafiullah, O.~Mees,
  O.~Kroemer, O.~Bastani, P.~R. Sanketi, P.~T. Miller, P.~Yin, P.~Wohlhart,
  P.~Xu, P.~D. Fagan, P.~Mitrano, P.~Sermanet, P.~Abbeel, P.~Sundaresan,
  Q.~Chen, Q.~Vuong, R.~Rafailov, R.~Tian, R.~Doshi,
  R.~Mart{\'{\i}}n{-}Mart{\'{\i}}n, R.~Baijal, R.~Scalise, R.~Hendrix, R.~Lin,
  R.~Qian, R.~Zhang, R.~Mendonca, R.~Shah, R.~Hoque, R.~Julian,
  S.~Bustamante{-}Gomez, S.~Kirmani, S.~Levine, S.~Lin, S.~Moore, S.~Bahl,
  S.~Dass, S.~D. Sonawani, S.~Song, S.~Xu, S.~Haldar, S.~Karamcheti,
  S.~Adebola, S.~Guist, S.~Nasiriany, S.~Schaal, S.~Welker, S.~Tian,
  S.~Ramamoorthy, S.~Dasari, S.~Belkhale, S.~Park, S.~Nair, S.~Mirchandani,
  T.~Osa, T.~Gupta, T.~Harada, T.~Matsushima, T.~Xiao, T.~Kollar, T.~Yu,
  T.~Ding, T.~Davchev, T.~Z. Zhao, T.~Armstrong, T.~Darrell, T.~Chung, V.~Jain,
  V.~Vanhoucke, W.~Zhan, W.~Zhou, W.~Burgard, X.~Chen, X.~Wang, X.~Zhu,
  X.~Geng, X.~Liu, L.~Xu, X.~Li, Y.~Lu, Y.~J. Ma, Y.~Kim, Y.~Chebotar, Y.~Zhou,
  Y.~Zhu, Y.~Wu, Y.~Xu, Y.~Wang, Y.~Bisk, Y.~Cho, Y.~Lee, Y.~Cui, Y.~Cao,
  Y.~Wu, Y.~Tang, Y.~Zhu, Y.~Zhang, Y.~Jiang, Y.~Li, Y.~Li, Y.~Iwasawa,
  Y.~Matsuo, Z.~Ma, Z.~Xu, Z.~J. Cui, Z.~Zhang, and Z.~Lin, ``Open
  x-embodiment: Robotic learning datasets and {RT-X} models : Open,
  x-embodiment collaboration,'' in \emph{{ICRA}}.\hskip 1em plus 0.5em minus
  0.4em\relax {IEEE}, 2024, pp. 6892--6903.

\bibitem{DBLP:conf/rss/0001B0GCF24}
V.~B. Ankit Goyal~and, J.~Xu, Y.~Guo, Y.~Chao, and D.~Fox, ``{RVT-2:} learning
  precise manipulation from few demonstrations,'' in \emph{Robotics: Science
  and Systems}, 2024.

\bibitem{DBLP:journals/corr/abs-2509-15212}
S.~H. Yuming Jiang~and, S.~Xue, Y.~Zhao, J.~Cen, S.~Leng, K.~Li, J.~Guo,
  K.~Wang, M.~Chen, F.~Wang, D.~Zhao, and X.~Li, ``Rynnvla-001: Using human
  demonstrations to improve robot manipulation,'' \emph{CoRR}, vol.
  abs/2509.15212, 2025.

\bibitem{DBLP:journals/corr/abs-2511-17502}
S.~H. Jun Cen~and, Y.~Yuan, K.~Li, H.~Yuan, C.~Yu, Y.~Jiang, J.~Guo, X.~Li,
  H.~Luo, F.~Wang, D.~Zhao, and H.~Chen, ``Rynnvla-002: {A} unified
  vision-language-action and world model,'' \emph{CoRR}, vol. abs/2511.17502,
  2025.

\bibitem{vla_safevla_2025}
B.~Zhang, Y.~Zhang, J.~Ji, Y.~Lei, J.~Dai, Y.~Chen, and Y.~Yang, ``Safevla:
  Towards safety alignment of vision-language-action model via constrained
  learning,'' \emph{arXiv preprint}, 2025.

\bibitem{DBLP:conf/icra/LealCJDVR0LSVSO24}
K.~C. Isabel Leal~and, D.~Jain, A.~Dubey, J.~Varley, M.~S. Ryoo, Y.~Lu, F.~Liu,
  V.~Sindhwani, Q.~Vuong, T.~Sarl{\'{o}}s, K.~Oslund, K.~Hausman, and K.~Rao,
  ``{SARA-RT:} scaling up robotics transformers with self-adaptive robust,
  attention,'' in \emph{{ICRA}}.\hskip 1em plus 0.5em minus 0.4em\relax {IEEE},
  2024, pp. 6920--6927.

\bibitem{DBLP:conf/icra/ZhuZLWXLCSPFT25}
Y.~Z. Minjie Zhu~and, J.~Li, J.~Wen, Z.~Xu, N.~Liu, R.~Cheng, C.~Shen, Y.~Peng,
  F.~Feng, and J.~Tang, ``Scaling diffusion policy in transformer to 1 billion
  parameters for, robotic manipulation,'' in \emph{{ICRA}}.\hskip 1em plus
  0.5em minus 0.4em\relax {IEEE}, 2025, pp. 10\,838--10\,845.

\bibitem{vla_scvla_2024}
C.~Li, J.~Liu, G.~Wang, X.~Li, S.~Chen, L.~Heng, C.~Xiong, J.~Ge, R.~Zhang,
  K.~Zhou, and S.~Zhang, ``A self-correcting vision-language-action model for
  fast and slow system manipulation,'' \emph{arXiv preprint}, 2024.

\bibitem{DBLP:journals/corr/abs-2510-16281}
A.~L. Yilin Wu~and, T.~Hermans, F.~Ramos, A.~Bajcsy, and
  C.~P{\'{e}}rez{-}D'Arpino, ``Do what you say: Steering vision-language-action
  models via runtime, reasoning-action alignment verification,'' \emph{CoRR},
  vol. abs/2510.16281, 2025.

\bibitem{DBLP:conf/iros/WangZDFF25}
J.~Z. Beichen Wang~and, S.~Dong, I.~Fang, and C.~Feng, ``{VLM} see, robot do:
  Human demo video to robot action plan via vision, language model,'' in
  \emph{{IROS}}.\hskip 1em plus 0.5em minus 0.4em\relax {IEEE}, 2025, pp.
  17\,215--17\,222.

\bibitem{DBLP:journals/corr/abs-2509-14138}
Z.~A. Ran Yang~and, L.~Zhou, and Y.~Feng, ``Seqvla: Sequential task execution
  for long-horizon manipulation with, completion-aware vision-language-action
  model,'' \emph{CoRR}, vol. abs/2509.14138, 2025.

\bibitem{DBLP:conf/cvpr/LinLGYWBLWS25}
L.~L. Kevin Qinghong Lin~and, D.~Gao, Z.~Yang, S.~Wu, Z.~Bai, S.~W. Lei,
  L.~Wang, and M.~Z. Shou, ``Showui: One vision-language-action model for {GUI}
  visual agent,'' in \emph{{CVPR}}.\hskip 1em plus 0.5em minus 0.4em\relax
  Computer Vision Foundation / {IEEE}, 2025, pp. 19\,498--19\,508.

\bibitem{DBLP:journals/corr/abs-2512-00783}
L.~Wang, ``Sigma: The key for vision-language-action models toward telepathic,
  alignment,'' \emph{CoRR}, vol. abs/2512.00783, 2025.

\bibitem{DBLP:journals/corr/abs-2509-09674}
Y.~Z. Haozhan Li~and, J.~Yu, Y.~Zhang, Z.~Yang, K.~Zhang, X.~Zhu, Y.~Zhang,
  T.~Chen, G.~Cui, D.~Wang, D.~Luo, Y.~Fan, Y.~Sun, J.~Zeng, J.~Pang, S.~Zhang,
  Y.~Wang, Y.~Mu, B.~Zhou, and N.~Ding, ``Simplevla-rl: Scaling {VLA} training
  via reinforcement learning,'' \emph{CoRR}, vol. abs/2509.09674, 2025.

\bibitem{DBLP:journals/corr/abs-2510-04041}
H.~S. Ayudh Saxena~and, S.~Routray, R.~R. Shah, and E.~Pahwa, ``{SITCOM:}
  scaling inference-time compute for vlas,'' \emph{CoRR}, vol. abs/2510.04041,
  2025.

\bibitem{DBLP:journals/corr/abs-2506-01844}
D.~A. Mustafa Shukor~and, F.~Capuano, P.~Kooijmans, S.~Palma, A.~Zouitine,
  M.~Aractingi, C.~Pascal, M.~Russi, A.~Marafioti, S.~Alibert, M.~Cord,
  T.~Wolf, and R.~Cad{\`{e}}ne, ``Smolvla: {A} vision-language-action model for
  affordable and efficient, robotics,'' \emph{CoRR}, vol. abs/2506.01844, 2025.

\bibitem{DBLP:conf/cvpr/JiangXLZRGLCY025}
W.~X. Jianping Jiang~and, Z.~Lin, H.~Zhang, T.~Ren, Y.~Gao, Z.~Lin, Z.~Cai,
  L.~Yang, and Z.~Liu, ``{SOLAMI:} social vision-language-action modeling for
  immersive interaction, with 3d autonomous characters,'' in
  \emph{{CVPR}}.\hskip 1em plus 0.5em minus 0.4em\relax Computer Vision
  Foundation / {IEEE}, 2025, pp. 26\,887--26\,898.

\bibitem{DBLP:journals/corr/abs-2510-12276}
W.~S. Fuhao Li~and, H.~Zhao, J.~Wang, P.~Ding, D.~Wang, L.~Zeng, and H.~Li,
  ``Spatial forcing: Implicit spatial representation alignment for
  vision-language-action, model,'' \emph{CoRR}, vol. abs/2510.12276, 2025.

\bibitem{DBLP:journals/corr/abs-2509-05614}
J.~X. Hanzhen Wang~and, J.~Pan, Y.~Zhou, and G.~Dai, ``Specprune-vla:
  Accelerating vision-language-action models via action-aware, self-speculative
  pruning,'' \emph{CoRR}, vol. abs/2509.05614, 2025.

\bibitem{DBLP:journals/corr/abs-2507-22424}
R.~Y. Songsheng Wang~and, Z.~Yuan, C.~Yu, F.~Gao, Y.~Wang, and D.~F. Wong,
  ``Spec-vla: Speculative decoding for vision-language-action models with,
  relaxed acceptance,'' \emph{CoRR}, vol. abs/2507.22424, 2025.

\bibitem{DBLP:journals/corr/abs-2509-09090}
Y.~L. Hengyu Fang~and, Y.~Du, L.~Du, and H.~Yang, ``{SQAP-VLA:} {A} synergistic
  quantization-aware pruning framework for, high-performance
  vision-language-action models,'' \emph{CoRR}, vol. abs/2509.09090, 2025.

\bibitem{DBLP:journals/corr/abs-2509-26251}
Y.~Y. Zhejia Cai~and, X.~Chang, S.~Liang, R.~Chen, F.~Xiong, M.~Xu, and
  R.~Huang, ``Seeing space and motion: Enhancing latent actions with spatial
  and, dynamic awareness for {VLA},'' \emph{CoRR}, vol. abs/2509.26251, 2025.

\bibitem{DBLP:journals/corr/abs-2512-05107}
G.~Z. Feng Xu~and, X.~Kong, T.~Fu, D.~F.~N. Gordon, X.~An, and B.~Busam,
  ``{STARE-VLA:} progressive stage-aware reinforcement for fine-tuning,
  vision-language-action models,'' \emph{CoRR}, vol. abs/2512.05107, 2025.

\bibitem{DBLP:journals/corr/abs-2512-21970}
M.~Y. Shengliang Deng~and, Y.~Zheng, J.~Su, W.~Zhang, X.~Zhao, H.~Cui,
  Z.~Zhang, and H.~Wang, ``Stereovla: Enhancing vision-language-action models
  with stereo vision,'' \emph{CoRR}, vol. abs/2512.21970, 2025.

\bibitem{DBLP:journals/corr/abs-2512-18477}
J.~Z. Wenjun Lin~and, K.~Cai, and K.~Wang, ``{STORM:} search-guided generative
  world models for robotic manipulation,'' \emph{CoRR}, vol. abs/2512.18477,
  2025.

\bibitem{DBLP:journals/corr/abs-2512-23162}
P.~G. Yufan He~and, M.~Xu, Z.~Li, A.~Myronenko, D.~Imans, B.~Liu, D.~Yang,
  M.~Gu, Y.~Ji, Y.~Jin, R.~Zhao, B.~Shen, and D.~Xu, ``Surgworld: Learning
  surgical robot policies from videos via world, modeling,'' \emph{CoRR}, vol.
  abs/2512.23162, 2025.

\bibitem{DBLP:journals/corr/abs-2512-00903}
C.~C. Chaojun Ni~and, X.~Wang, Z.~Zhu, W.~Zheng, B.~Wang, T.~Chen, G.~Zhao,
  H.~Li, Z.~Dong, Q.~Zhang, Y.~Ye, Y.~Wang, G.~Huang, and W.~Mei, ``Swiftvla:
  Unlocking spatiotemporal dynamics for lightweight {VLA}, models at minimal
  overhead,'' \emph{CoRR}, vol. abs/2512.00903, 2025.

\bibitem{DBLP:journals/corr/abs-2510-10932}
X.~Z. Zonghuan Xu~and, X.~Ma, and Y.~Jiang, ``Tabvla: Targeted backdoor attacks
  on vision-language-action models,'' \emph{CoRR}, vol. abs/2510.10932, 2025.

\bibitem{DBLP:journals/corr/abs-2502-05171}
S.~M. Jonas Geiping~and, N.~Jain, J.~Kirchenbauer, S.~Singh, B.~R. Bartoldson,
  B.~Kailkhura, A.~Bhatele, and T.~Goldstein, ``Scaling up test-time compute
  with latent reasoning: {A} recurrent, depth approach,'' \emph{CoRR}, vol.
  abs/2502.05171, 2025.

\bibitem{DBLP:journals/corr/abs-2509-25746}
H.~Z. Shuaijun Wang~and, D.~Xiang, and Y.~You, ``Tacrefinenet: Tactile-only
  grasp refinement between arbitrary in-hand, object poses,'' \emph{CoRR}, vol.
  abs/2509.25746, 2025.

\bibitem{DBLP:journals/corr/abs-2507-09160}
S.~W. Jialei Huang~and, F.~Lin, Y.~Hu, C.~Wen, and Y.~Gao, ``Tactile-vla:
  Unlocking vision-language-action model's physical knowledge, for tactile
  generalization,'' \emph{CoRR}, vol. abs/2507.09160, 2025.

\bibitem{DBLP:journals/corr/abs-2509-07962}
H.~X. Zongzheng Zhang~and, Z.~Yang, C.~Yue, Z.~Lin, H.~Gao, Z.~Wang, and
  H.~Zhao, ``{TA-VLA:} elucidating the design space of torque-aware
  vision-language-action, models,'' \emph{CoRR}, vol. abs/2509.07962, 2025.

\bibitem{DBLP:journals/corr/abs-2507-16815}
C.~Huang, Y.~Wu, M.~Chen, Y.~F. Wang, and F.~Yang, ``Thinkact:
  Vision-language-action reasoning via reinforced visual latent planning,''
  \emph{CoRR}, vol. abs/2507.16815, 2025.

\bibitem{DBLP:conf/iclr/LuWLLT25}
Z.~W. Guanxing Lu~and, C.~Liu, J.~Lu, and Y.~Tang, ``Thinkbot: Embodied
  instruction following with thought chain reasoning,'' in \emph{{ICLR}}.\hskip
  1em plus 0.5em minus 0.4em\relax OpenReview.net, 2025.

\bibitem{DBLP:journals/corr/abs-2505-23189}
J.~Z. Shaoan Wang~and, M.~Li, J.~Liu, A.~Li, K.~Wu, F.~Zhong, J.~Yu, Z.~Zhang,
  and H.~Wang, ``Trackvla: Embodied visual tracking in the wild,'' \emph{CoRR},
  vol. abs/2505.23189, 2025.

\bibitem{DBLP:journals/corr/abs-2509-11839}
P.~D. Jiacheng Liu~and, Q.~Zhou, Y.~Wu, D.~Huang, Z.~Peng, W.~Xiao, W.~Zhang,
  L.~Yang, C.~Lu, and D.~Wang, ``Trajbooster: Boosting humanoid whole-body
  manipulation via trajectory-centric, learning,'' \emph{CoRR}, vol.
  abs/2509.11839, 2025.

\bibitem{DBLP:journals/corr/abs-2507-01424}
Y.~G. Zhenyang Liu~and, S.~Zheng, X.~Xue, and Y.~Fu, ``Trivla: {A}
  triple-system-based unified vision-language-action model, for general robot
  control,'' \emph{CoRR}, vol. abs/2507.01424, 2025.

\bibitem{vla_tvve_2025}
Y.~Bai, Z.~Wang, Y.~Liu, K.~Luo, Y.~Wen, M.~Dai, W.~Chen, Z.~Chen, L.~Liu,
  G.~Li, and L.~Lin, ``Learning to see and act: Task-aware virtual view
  exploration for robotic manipulation,'' \emph{arXiv preprint}, 2025.

\bibitem{DBLP:journals/corr/abs-2511-01718}
W.~S. Jiayi Chen~and, P.~Ding, Z.~Zhou, H.~Zhao, F.~Tang, D.~Wang, and H.~Li,
  ``Unified diffusion {VLA:} vision-language-action model via joint discrete,
  denoising diffusion process,'' \emph{CoRR}, vol. abs/2511.01718, 2025.

\bibitem{DBLP:journals/corr/abs-2509-22441}
Y.~Z. Zhangyuan Wang~and, Y.~Yan, X.~Tian, X.~Shao, M.~Li, W.~Li, G.~Su,
  W.~Cui, and D.~Fan, ``Underwatervla: Dual-brain vision-language-action
  architecture for, autonomous underwater navigation,'' \emph{CoRR}, vol.
  abs/2509.22441, 2025.

\bibitem{DBLP:journals/corr/abs-2510-10642}
Y.~H. Jianke Zhang~and, Y.~Guo, X.~Chen, Y.~Liu, W.~Chen, C.~Lu, and J.~Chen,
  ``Unicod: Enhancing robot policy via unified continuous and discrete,
  representation learning,'' \emph{CoRR}, vol. abs/2510.10642, 2025.

\bibitem{DBLP:conf/cvpr/LuCL0KMHK24}
C.~C. Jiasen Lu~and, S.~Lee, Z.~Zhang, S.~Khosla, R.~Marten, D.~Hoiem, and
  A.~Kembhavi, ``Unified-io 2: Scaling autoregressive multimodal models with
  vision, language, audio, and action,'' in \emph{{CVPR}}.\hskip 1em plus 0.5em
  minus 0.4em\relax {IEEE}, 2024, pp. 26\,429--26\,445.

\bibitem{DBLP:journals/corr/abs-2412-06224}
K.~W. Jiazhao Zhang~and, S.~Wang, M.~Li, H.~Liu, S.~Wei, Z.~Wang, Z.~Zhang, and
  H.~Wang, ``Uni-navid: {A} video-based vision-language-action model for
  unifying, embodied navigation tasks,'' \emph{CoRR}, vol. abs/2412.06224,
  2024.

\bibitem{DBLP:journals/corr/abs-2512-09864}
Z.~L. Hao Lu~and, G.~Jiang, Y.~Luo, S.~Chen, Y.~Zhang, and Y.~Chen, ``Uniugp:
  Unifying understanding, generation, and planing for end-to-end, autonomous
  driving,'' \emph{CoRR}, vol. abs/2512.09864, 2025.

\bibitem{DBLP:journals/corr/abs-2511-21192}
Y.~Y. Hui Lu~and, Y.~Yang, C.~Yi, Q.~Zhang, B.~Shen, A.~C. Kot, and X.~Jiang,
  ``When robots obey the patch: Universal transferable patch attacks on,
  vision-language-action models,'' \emph{CoRR}, vol. abs/2511.21192, 2025.

\bibitem{DBLP:conf/icml/ZhangGHCZ025}
Y.~G. Jianke Zhang~and, Y.~Hu, X.~Chen, X.~Zhu, and J.~Chen, ``{UP-VLA:} {A}
  unified understanding and prediction model for embodied, agent,'' in
  \emph{{ICML}}.\hskip 1em plus 0.5em minus 0.4em\relax OpenReview.net, 2025.

\bibitem{DBLP:journals/corr/abs-2510-23576}
Z.~W. Anqi Li~and, J.~Zhang, M.~Li, Y.~Qi, Z.~Chen, Z.~Zhang, and H.~Wang,
  ``Urbanvla: {A} vision-language-action model for urban micromobility,''
  \emph{CoRR}, vol. abs/2510.23576, 2025.

\bibitem{DBLP:journals/corr/abs-2510-07869}
Z.~W. Junwen Gu~and, P.~Si, S.~Qiu, Y.~Feng, L.~Sun, L.~Luo, L.~Yu, J.~Wang,
  and Z.~Wu, ``{USIM} and {U0:} {A} vision-language-action dataset and model
  for, general underwater robots,'' \emph{CoRR}, vol. abs/2510.07869, 2025.

\bibitem{DBLP:journals/corr/abs-2504-02792}
R.~Y. Chuning Zhu~and, S.~Feng, B.~Burchfiel, P.~Shah, and A.~Gupta, ``Unified
  world models: Coupling video and action diffusion for pretraining, on large
  robotic datasets,'' \emph{CoRR}, vol. abs/2504.02792, 2025.

\bibitem{DBLP:journals/corr/abs-2512-15692}
L.~A. Jonas Pai~and, V.~Montesinos, B.~Forrai, O.~Mees, and E.~Nava,
  ``mimic-video: Video-action models for generalizable robot control beyond,
  vlas,'' \emph{CoRR}, vol. abs/2512.15692, 2025.

\bibitem{DBLP:journals/corr/abs-2512-03044}
J.~L. Yueru Jia~and, S.~Liu, R.~Zhou, W.~Yu, Y.~Yan, X.~Chi, Y.~Guo, B.~Shi,
  and S.~Zhang, ``Video2act: {A} dual-system video diffusion policy with
  robotic spatio-motional, modeling,'' \emph{CoRR}, vol. abs/2512.03044, 2025.

\bibitem{DBLP:journals/corr/abs-2512-06963}
F.~W. Yichao Shen~and, Z.~Du, Y.~Liang, Y.~Lu, J.~Yang, N.~Zheng, and B.~Guo,
  ``Videovla: Video generators can be generalizable robot manipulators,''
  \emph{CoRR}, vol. abs/2512.06963, 2025.

\bibitem{DBLP:journals/corr/abs-2512-03913}
J.~Y. Jeongeun Park~and, B.~Jeon, J.~Park, J.~Shin, N.~Cho, K.~Lee, S.~Yun, and
  S.~Choi, ``Hierarchical vision language action model using success and
  failure, demonstrations,'' \emph{CoRR}, vol. abs/2512.03913, 2025.

\bibitem{DBLP:journals/corr/abs-2512-07582}
M.~W. Guangyan Chen~and, Q.~Shao, Z.~Zhou, W.~Mao, T.~Cui, M.~Zhu, Y.~Deng,
  L.~Yang, Z.~Zhang, Y.~Yang, H.~Chen, and Y.~Yue, ``See once, then act:
  Vision-language-action model with task learning, from one-shot video
  demonstrations,'' \emph{CoRR}, vol. abs/2512.07582, 2025.

\bibitem{DBLP:journals/corr/abs-2510-13054}
H.~H. Ankit Goyal~and, X.~Yang, V.~Blukis, and F.~Ramos, ``{VLA-0:} building
  state-of-the-art vlas with zero modification,'' \emph{CoRR}, vol.
  abs/2510.13054, 2025.

\bibitem{DBLP:journals/corr/abs-2511-17199}
C.~M. Hanyu Zhou~and and G.~H. Lee, ``{VLA-4D:} embedding 4d awareness into
  vision-language-action models, for spatiotemporally coherent robotic
  manipulation,'' \emph{CoRR}, vol. abs/2511.17199, 2025.

\bibitem{DBLP:journals/corr/abs-2509-09372}
P.~D. Yihao Wang~and, L.~Li, C.~Cui, Z.~Ge, X.~Tong, W.~Song, H.~Zhao, W.~Zhao,
  P.~Hou, S.~Huang, Y.~Tang, W.~Wang, R.~Zhang, J.~Liu, and D.~Wang,
  ``Vla-adapter: An effective paradigm for tiny-scale vision-language-action,
  model,'' \emph{CoRR}, vol. abs/2509.09372, 2025.

\bibitem{DBLP:journals/corr/abs-2512-15258}
M.~Z. Yuze Wu~and, X.~Li, Y.~Du, Y.~Fan, W.~Li, Z.~Han, X.~Zhou, and F.~Gao,
  ``{VLA-AN:} an efficient and onboard vision-language-action framework, for
  aerial navigation in complex environments,'' \emph{CoRR}, vol.
  abs/2512.15258, 2025.

\bibitem{DBLP:journals/corr/abs-2502-02175}
Y.~W. Siyu Xu~and, C.~Xia, D.~Zhu, T.~Huang, and C.~Xu, ``Vla-cache: Towards
  efficient vision-language-action model via adaptive, token caching in robotic
  manipulation,'' \emph{CoRR}, vol. abs/2502.02175, 2025.

\bibitem{DBLP:journals/corr/abs-2511-16203}
Y.~X. Yuping Yan~and, Y.~Zhang, L.~Lyu, H.~Wang, and Y.~Jin, ``When alignment
  fails: Multimodal adversarial attacks on vision-language-action, models,''
  \emph{CoRR}, vol. abs/2511.16203, 2025.

\bibitem{DBLP:journals/corr/abs-2506-17561}
Z.~L. Chongkai Gao~and, Z.~Chi, J.~Huang, X.~Fei, Y.~Hou, Y.~Zhang, Y.~Lin,
  Z.~Fang, Z.~Jiang, and L.~Shao, ``{VLA-OS:} structuring and dissecting
  planning representations and, paradigms in vision-language-action models,''
  \emph{CoRR}, vol. abs/2506.17561, 2025.

\bibitem{DBLP:journals/corr/abs-2511-14178}
J.~L. Zhuo Li~and, Z.~Dong, T.~Teng, Q.~Rouxel, D.~G. Caldwell, and F.~Chen,
  ``Towards deploying {VLA} without fine-tuning: Plug-and-play inference-time,
  {VLA} policy steering via embodied evolutionary diffusion,'' \emph{CoRR},
  vol. abs/2511.14178, 2025.

\bibitem{DBLP:journals/corr/abs-2511-16449}
Y.~C. Ziyan Liu~and, H.~Cai, T.~Lin, S.~Yang, Z.~Liu, and B.~Zhao,
  ``Vla-pruner: Temporal-aware dual-level visual token pruning for efficient,
  vision-language-action inference,'' \emph{CoRR}, vol. abs/2511.16449, 2025.

\bibitem{DBLP:journals/corr/abs-2508-12211}
O.~G.~Y. Cyrus Neary~and, A.~Kuramshin, O.~Aslan, and G.~Berseth, ``Improving
  pre-trained vision-language-action policies with model-based, search,''
  \emph{CoRR}, vol. abs/2508.12211, 2025.

\bibitem{DBLP:journals/corr/abs-2511-12405}
\BIBentryALTinterwordspacing
S.~M. Hyunki Seong~and, H.~Ahn, J.~Kang, and D.~H. Shim, ``Vla-r:
  Vision-language action retrieval toward open-world end-to-end autonomous
  driving,'' 2025. [Online]. Available:
  \url{https://dblp.org/rec/journals/corr/abs-2511-12405}
\BIBentrySTDinterwordspacing

\bibitem{DBLP:journals/corr/abs-2510-01623}
\BIBentryALTinterwordspacing
Z.~Z. Angen Ye~and, B.~Wang, X.~Wang, D.~Zhang, and Z.~Zhu, ``Vla-r1: Enhancing
  reasoning in vision-language-action models,'' 2025. [Online]. Available:
  \url{https://dblp.org/rec/journals/corr/abs-2510-01623}
\BIBentrySTDinterwordspacing

\bibitem{DBLP:journals/corr/abs-2512-24673}
L.~Z. Yongsheng Zhao~and, B.~Cheng, G.~Yao, X.~Wen, and H.~Gao, ``{VLA-RAIL:}
  {A} real-time asynchronous inference linker for {VLA}, models and robots,''
  \emph{CoRR}, vol. abs/2512.24673, 2025.

\bibitem{DBLP:journals/corr/abs-2510-00406}
P.~D. Hengtao Li~and, R.~Suo, Y.~Wang, Z.~Ge, D.~Zang, K.~Yu, M.~Sun, H.~Zhang,
  D.~Wang, and W.~Su, ``{VLA-RFT:} vision-language-action reinforcement
  fine-tuning with verified, rewards in world simulators,'' \emph{CoRR}, vol.
  abs/2510.00406, 2025.

\bibitem{DBLP:conf/iclr/ZhaoD0GB0W25}
P.~D. Wei Zhao~and, M.~Zhang, Z.~Gong, S.~Bai, H.~Zhao, and D.~Wang, ``{VLAS:}
  vision-language-action model with speech instructions for, customized robot
  manipulation,'' in \emph{{ICLR}}.\hskip 1em plus 0.5em minus 0.4em\relax
  OpenReview.net, 2025.

\bibitem{DBLP:journals/corr/abs-2510-11027}
T.~Z. Ganlin Yang~and, H.~Hao, W.~Wang, Y.~Liu, D.~Wang, G.~Chen, Z.~Cai,
  J.~Chen, W.~Su, W.~Zhou, Y.~Qiao, J.~Dai, J.~Pang, G.~Luo, W.~Wang, Y.~Mu,
  and Z.~Hou, ``Vlaser: Vision-language-action model with synergistic embodied
  reasoning,'' \emph{CoRR}, vol. abs/2510.11027, 2025.

\bibitem{DBLP:journals/corr/abs-2512-01031}
Y.~S. Jiaming Tang~and, Y.~Zhao, S.~Yang, Y.~Lin, Z.~Zhang, J.~Hou, Y.~Lu,
  Z.~Liu, and S.~Han, ``{VLASH:} real-time vlas via future-state-aware
  asynchronous inference,'' \emph{CoRR}, vol. abs/2512.01031, 2025.

\bibitem{DBLP:journals/corr/abs-2509-22195}
X.~W. Asher J. Hancock~and, L.~Zha, O.~Russakovsky, and A.~Majumdar, ``Actions
  as language: Fine-tuning vlms into vlas without catastrophic, forgetting,''
  \emph{CoRR}, vol. abs/2509.22195, 2025.

\bibitem{DBLP:conf/nips/Chen0C0LZPH0YYY24}
M.~W. Guangyan Chen~and, T.~Cui, Y.~Mu, H.~Lu, T.~Zhou, Z.~Peng, M.~Hu, H.~Li,
  L.~Yuan, Y.~Yang, and Y.~Yue, ``Vlmimic: Vision language models are visual
  imitation learner for fine-grained, actions,'' in \emph{NeurIPS}, 2024.

\bibitem{DBLP:conf/corl/HuangWZL0023}
C.~W. Wenlong Huang~and, R.~Zhang, Y.~Li, J.~Wu, and L.~Fei{-}Fei, ``Voxposer:
  Composable 3d value maps for robotic manipulation with language, models,'' in
  \emph{CoRL}, ser. Proceedings of Machine Learning Research, vol. 229.\hskip
  1em plus 0.5em minus 0.4em\relax {PMLR}, 2023, pp. 540--562.

\bibitem{DBLP:conf/wacv/XuWZLO25}
M.~W. Mingjie Xu~and, Y.~Zhao, J.~C.~L. Li, and W.~Ou, ``Llava-spacesgg: Visual
  instruct tuning for open-vocabulary scene graph, generation with enhanced
  spatial relations,'' in \emph{{WACV}}.\hskip 1em plus 0.5em minus 0.4em\relax
  {IEEE}, 2025, pp. 6362--6372.

\bibitem{DBLP:journals/corr/abs-2512-06112}
J.~C. Yifang Xu~and, F.~Cai, Z.~Zhu, H.~Shang, S.~Luan, M.~Xu, N.~Zhang, Y.~Li,
  J.~Cai, and S.~Zhu, ``Wam-flow: Parallel coarse-to-fine motion planning via
  discrete flow, matching for autonomous driving,'' \emph{CoRR}, vol.
  abs/2512.06112, 2025.

\bibitem{DBLP:journals/corr/abs-2512-11047}
J.~C. Haoran Jiang~and, Q.~Bu, L.~Chen, M.~Shi, Y.~Zhang, D.~Li, C.~Suo,
  C.~Wang, Z.~Peng, and H.~Li, ``Wholebodyvla: Towards unified latent {VLA} for
  whole-body loco-manipulation, control,'' \emph{CoRR}, vol. abs/2512.11047,
  2025.

\bibitem{DBLP:journals/corr/abs-2511-09515}
Z.~Y. Fangqi Zhu~and, Z.~Hong, Q.~Shou, X.~Ma, and S.~Guo, ``{WMPO:} world
  model-based policy optimization for vision-language-action, models,''
  \emph{CoRR}, vol. abs/2511.09515, 2025.

\bibitem{DBLP:journals/corr/abs-2510-07313}
X.~C. Zezhong Qian~and, Y.~Li, S.~Wang, Z.~Qin, X.~Ju, S.~Han, and S.~Zhang,
  ``Wristworld: Generating wrist-views via 4d world models for robotic,
  manipulation,'' \emph{CoRR}, vol. abs/2510.07313, 2025.

\bibitem{DBLP:journals/corr/abs-2512-04537}
H.~C. Pei Yang~and, Y.~Song, and M.~Z. Shou, ``X-humanoid: Robotize human
  videos to generate humanoid videos at scale,'' \emph{CoRR}, vol.
  abs/2512.04537, 2025.

\bibitem{DBLP:journals/corr/abs-2511-02776}
K.~W. Shichao Fan~and, Z.~Che, X.~Wang, D.~Wu, F.~Liao, N.~Liu, Y.~Zhang,
  Z.~Zhao, Z.~Xu, M.~Li, Q.~Liu, S.~Zhang, M.~Wan, and J.~Tang, ``{XR-1:}
  towards versatile vision-language-action models via learning, unified
  vision-motion representations,'' \emph{CoRR}, vol. abs/2511.02776, 2025.

\bibitem{DBLP:journals/corr/abs-2510-10274}
J.~L. Jinliang Zheng~and, Z.~Wang, D.~Liu, X.~Kang, Y.~Feng, Y.~Zheng, J.~Zou,
  Y.~Chen, J.~Zeng, Y.~Zhang, J.~Pang, J.~Liu, T.~Wang, and X.~Zhan, ``{X-VLA:}
  soft-prompted transformer as scalable cross-embodiment
  vision-language-action, model,'' \emph{CoRR}, vol. abs/2510.10274, 2025.

\end{thebibliography}

\clearpage
\appendix
\subsection{Background}

\subsubsection{Unimodal Models}
\label{sec:unimodal}

Vision-language-action models integrate three modalities, often relying on existing unimodal models. The transition from convolutional neural networks (e.g., ResNet \cite{DBLP:conf/cvpr/HeZRS16}) to visual Transformers (e.g., ViT \cite{DBLP:conf/iclr/DosovitskiyB0WZ21}, SAM \cite{DBLP:journals/corr/abs-2304-02643}) in computer vision has facilitated the development of vision foundation models (VFMs). In natural language processing, the evolution from recurrent neural networks (e.g., LSTM \cite{DBLP:journals/neco/HochreiterS97}, GRU \cite{DBLP:conf/emnlp/ChoMGBBSB14}) to Transformers \cite{DBLP:conf/nips/VaswaniSPUJGKP17} initially led to the pretrain-finetune paradigm (e.g., BERT \cite{DBLP:conf/naacl/DevlinCLT19}, ChatGPT \cite{chatgpt-openai}), followed by the recent success of prompt tuning driven by large language models. Reinforcement learning (e.g., DQN \cite{DBLP:journals/nature/MnihKSRVBGRFOPB15}, AlphaGo \cite{DBLP:journals/nature/SilverHMGSDSAPL16}, PPO \cite{DBLP:journals/corr/SchulmanWDRK17}, Dactyl \cite{DBLP:journals/corr/abs-1808-00177}) has also witnessed a shift towards employing Transformers to model the Markov Decision Process as autoregressive sequential data. DPO \cite{DBLP:conf/nips/RafailovSMMEF23} directly trains LLMs on human preferences, simplifying RLHF.

\subsubsection{Vision-Language Models}
\label{sec:vlm}

Vision-language tasks, encompassing image captioning \cite{DBLP:journals/corr/ChenFLVGDZ15}, visual question answering \cite{DBLP:conf/iccv/AntolALMBZP15}, visual grounding \cite{DBLP:conf/eccv/YuPYBB16}, require the fusion of computer vision and natural language processing models. Early efforts, such as Show and Tell \cite{DBLP:conf/cvpr/VinyalsTBE15}, leveraged the success of early CNNs and RNNs. The advent of high-capacity language models, including BERT \cite{DBLP:conf/naacl/DevlinCLT19} and GPT \cite{radford2018improving}, ushered in a new era of Transformer-based VLMs. One of the pioneering self-supervised pretraining methods is ViLBERT \cite{DBLP:conf/nips/LuBPL19}, while CLIP \cite{DBLP:conf/icml/RadfordKHRGASAM21} popularized contrastive pretraining approaches for multimodal alignment. The recent emergence of large language models has driven the development of multimodal LLMs (MLLMs) or large multimodal models (LMMs), which achieve state-of-the-art performance on multimodal instruction-following tasks. Representative MLLMs include Flamingo \cite{DBLP:conf/nips/AlayracDLMBHLMM22}, BLIP-2 \cite{DBLP:conf/icml/0008LSH23}, and LLaVA \cite{DBLP:journals/corr/abs-2304-08485}. VLMs share a close relationship with VLAs, as the multimodal architectures of VLMs can be readily adopted for VLAs. Additionally, VLMs can function as high-level task planners if they possess sufficient reasoning capabilities. 

\subsection{Background (Extended Version): Unimodal Models}
\label{sec:unimodal_models_all}

\begin{figure*}
    \centering
    \includegraphics[width=\textwidth, trim={0cm 875pt 425pt 30pt},clip]{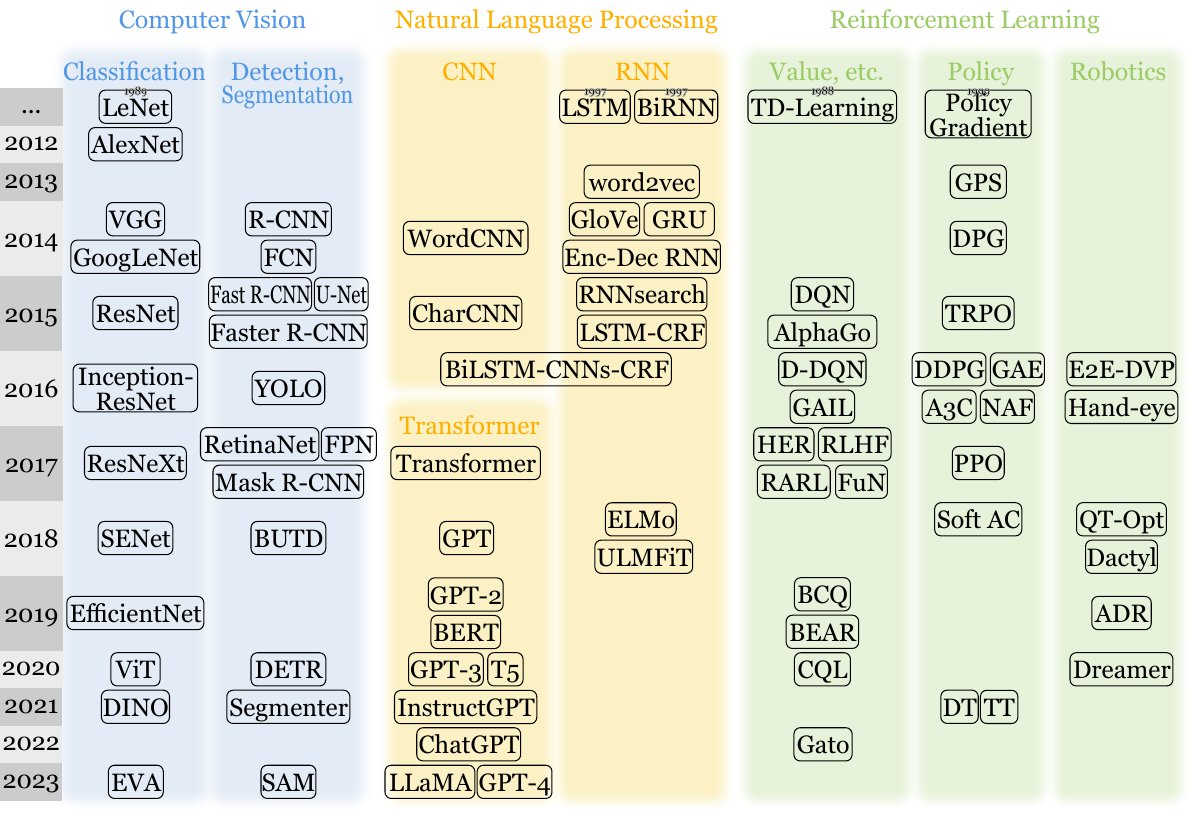}
    \caption{A brief timeline of pivotal unimodal models leading to the development of vision-language-action models, organized by their publication years. Details can be found in Appendix:\ref{sec:unimodal_models_all}. }
    \label{fig:timeline}
\end{figure*}

\begin{figure*}
    \centering
    \includegraphics[width=\textwidth, trim={2cm 0.5cm 2cm 1cm},clip]{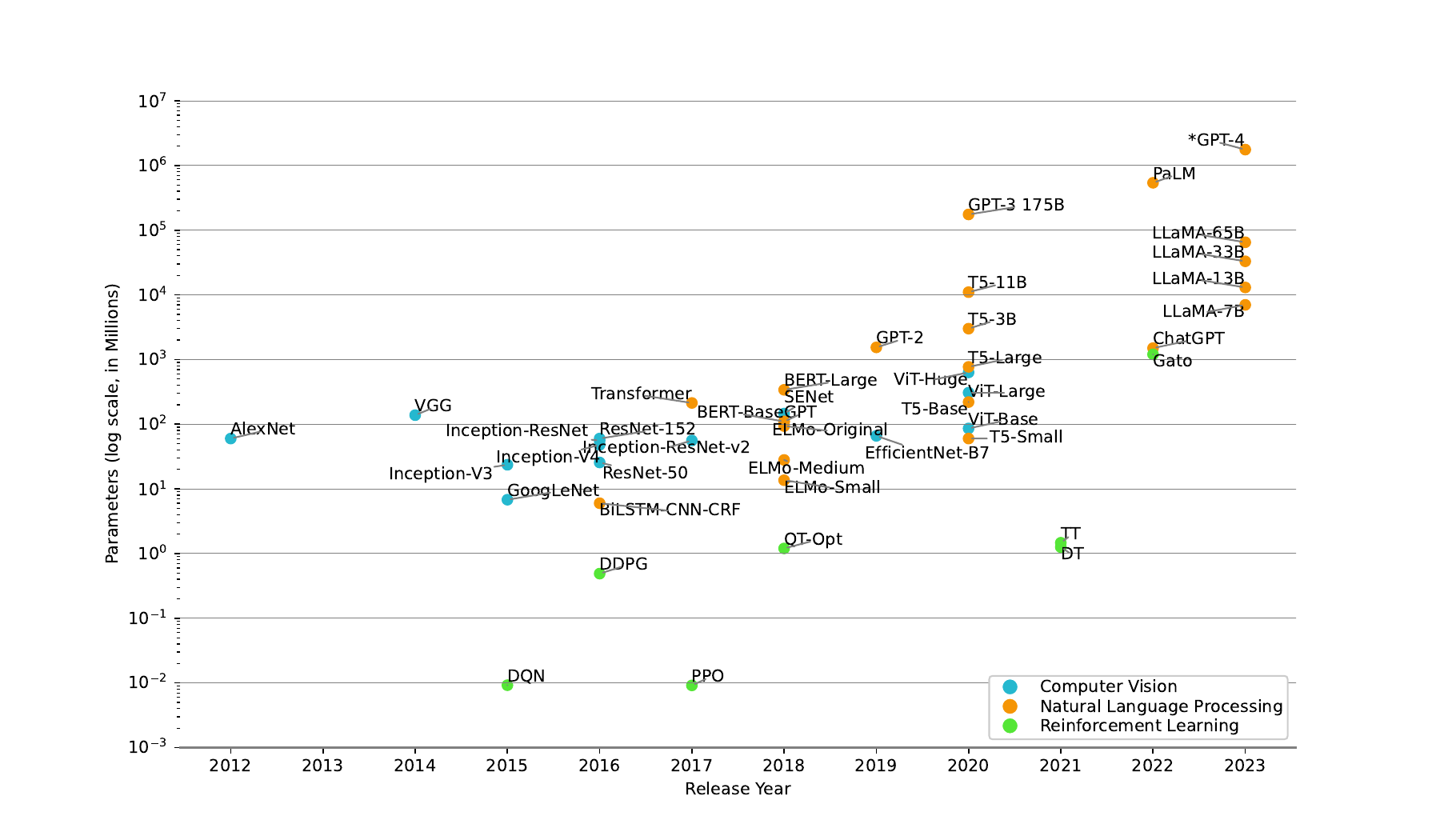}
    \caption{The growing scale of unimodal models over the years. $\ast$GPT-4's model size is estimated since it has not been officially disclosed.}
    \label{fig:model_size}
\end{figure*}

Vision-language-action models involve three modalities, and consequently, many VLAs depend on existing unimodal models for processing inputs from different modalities. Therefore, it is crucial to summarize representative developments in unimodal models, as they often serve as integral components in VLAs. Specifically, for the vision modality, we collect models designed for image classification, object detection, and image segmentation, as these tasks are particularly relevant for robotic learning. Natural language processing models play a crucial role in enabling VLAs to understand language instructions or generate language responses. Reinforcement learning is a foundational component for obtaining optimal policies, facilitating the generation of appropriate actions in a given environment and condition. A brief timeline of the development of unimodal models is depicted in Figure~\ref{fig:timeline}. Additionally, Figure~\ref{fig:model_size} highlights the progressive increase in model size within these fields.

\subsubsection{Computer Vision}
\label{sec:cv}

Computer vision witnessed the inception of modern neural networks. In robotics, object classification models can be used to inform a policy about which objects are of interest, and models for object detection or image segmentation can help precisely locate objects. Therefore, we mainly summarize approaches for these tasks, but numerous excellent surveys on visual models, ranging from convolutional neural networks (CNNs) \cite{DBLP:journals/air/KhanSZQ20} to Transformers \cite{DBLP:journals/csur/KhanNHZKS22}, offer more detailed insights. Interested readers are directed to these surveys for a more comprehensive introduction. Here, we will briefly touch upon some of the key developments in the field of computer vision. 

\paragraph{Convolutional Neural Network} Early developments in computer vision (CV) were primarily focused on the image classification task. LeNet \cite{DBLP:journals/neco/LeCunBDHHHJ89} was among the first convolutional neural networks, designed for identifying handwritten digits in zip codes. In 2012, AlexNet \cite{DBLP:conf/nips/KrizhevskySH12} emerged as a breakthrough by winning the ImageNet challenge, showcasing the potential of neural networks. VGG \cite{DBLP:journals/corr/SimonyanZ14a} demonstrated the benefits of increasing the depth of CNNs. GoogLeNet \cite{DBLP:conf/cvpr/SzegedyLJSRAEVR15}, also known as Inception-V1, introduced the concept of blocks. ResNet \cite{DBLP:conf/cvpr/HeZRS16} introduced skip connections or residual connections. Inception-ResNet \cite{DBLP:conf/aaai/SzegedyIVA17}, as the name suggests, combines residual connects and inception blocks. ResNeXt \cite{DBLP:conf/cvpr/XieGDTH17} explored the concept of split, transform, and merge. SENet \cite{DBLP:conf/cvpr/HuSS18} introduces the squeeze-and-excitation blocks, utilizing a type of attention mechanism. EfficientNet \cite{DBLP:conf/icml/TanL19} studied the width, depth, and resolution of CNN models with ``compound scaling,'' highlighting the trade-off between efficiency and performance. 

Alongside image classification, object detection became an integral component in many applications. Building upon the success of image classification backbone networks, a series of works optimized region-based methods: R-CNN \cite{DBLP:conf/cvpr/GirshickDDM14}, Fast R-CNN \cite{DBLP:conf/iccv/Girshick15}, Faster R-CNN \cite{DBLP:conf/nips/RenHGS15}, and Mask R-CNN \cite{DBLP:conf/iccv/HeGDG17}. Grid-based methods like YOLO \cite{DBLP:conf/cvpr/RedmonDGF16} are also widely adopted. Bottom-up, top-down is also a popular strategy, employed by FPN \cite{DBLP:conf/cvpr/LinDGHHB17}, RetinaNet \cite{DBLP:conf/iccv/LinGGHD17}, BUTD \cite{DBLP:conf/cvpr/00010BT0GZ18}, etc. In scenarios requiring more detailed and precise object detection, image segmentation aims to determine the exact outline of objects. Many popular models adopt an ``encoder-decoder'' architecture, where the encoder understands both the global and local context of the image, and the decoder produces a segmentation map based on this context information. Representative works following this idea include FCN \cite{DBLP:conf/cvpr/LongSD15}, SegNet \cite{DBLP:journals/pami/BadrinarayananK17}, Mask R-CNN \cite{DBLP:conf/iccv/HeGDG17}, and U-Net \cite{DBLP:conf/miccai/RonnebergerFB15}.

\paragraph{Vision Transformer} Convolutional neural networks (CNNs) have historically been the foundation of computer vision models. However, the landscape shifted with the introduction of the Transformer architecture in the seminal work by \cite{DBLP:conf/nips/VaswaniSPUJGKP17}. This paradigm shift was initiated by ViT \cite{DBLP:conf/iclr/DosovitskiyB0WZ21}. It revolutionizes image processing by breaking down images into 16-by-16 pixel patches, treating each as a token akin to those in NLP; leveraging a BERT-like model, ViT encodes these patches and has exhibited superior performance over many traditional CNN models in image classification tasks. 

The transformative power of the Transformer extends beyond classification. DETR \cite{DBLP:conf/eccv/CarionMSUKZ20} employs an encoder-decoder Transformer architecture to tackle object detection. The encoder processes the input image, and its output embeddings are fed into the decoder through cross-attention. Notably, DETR introduces learnable object queries to the decoder, facilitating the extraction of crucial object-wise information from the encoder's output. Venturing into image segmentation, Segmenter \cite{DBLP:conf/iccv/StrudelPLS21} was the first to utilize Transformer on this task. The Segment Anything model (SAM) \cite{DBLP:journals/corr/abs-2304-02643} achieves remarkable milestones in promptable segmentation, zero-shot performance, and versatile architecture, further underlining the transformative impact of Vision Transformer models in various computer vision domains.

\paragraph{Vision in 3D} Aside from the most common RGB data, other types of visual inputs are widely used \cite{DBLP:journals/csur/IoannidouCNK17, DBLP:journals/corr/abs-1808-01462}. In robotics, depth maps are useful since they provide essential 3D information that is not explicitly stored in RGB images. Depth maps can be captured with Microsoft Kinect \footnote{\url{https://azure.microsoft.com/en-us/products/kinect-dk/}} or Intel RealSense \footnote{\url{https://www.intelrealsense.com}} or recovered from pure RGB images. Point clouds \cite{DBLP:journals/pami/GuoWHLLB21} are also popular visual input types due to the widespread adoption of LiDARs and 3D scanners; depth maps can be easily converted to point clouds. Volumetric data \cite{DBLP:conf/cvpr/QiSNDYG16}, such as voxels or octrees, is usually more information-rich than depth maps and is suitable for representing rigid objects. Despite the widespread use of 3D meshes as the default data format in computer graphics, their irregular nature poses challenges for neural networks \cite{DBLP:conf/aaai/FengFYZG19}.

\subsubsection{Natural Language Processing}
\label{sec:nlp}

Natural Language Processing (NLP) plays a pivotal role in VLA, serving as a vital component for understanding user instructions or even generating appropriate textual responses. The recent surge in NLP owes much to the success of Transformer models \cite{DBLP:conf/nips/VaswaniSPUJGKP17}. In the landscape of contemporary NLP, there is a noticeable shift towards implicit learning of language syntax and semantics, a departure from the previous paradigms. To provide context, this subsection will commence with a concise overview of fundamental yet enduring concepts before delving into the noteworthy advancements in contemporary NLP. For an in-depth exploration of the progress in the NLP domain, readers are directed to comprehensive surveys by \cite{DBLP:journals/tnn/OtterMK21} and \cite{DBLP:journals/csur/LiuYFJHN23}, which meticulously review the trajectory of advancements in NLP.

\paragraph{Early Developments} The field of NLP, which was more frequently referred to as Computational Linguistics (CL) in the early days, tries to solve various tasks regarding to natural language. In CL, natural languages used to be processed in a hierarchical way: word, syntax, and semantics. Firstly, on the word level, many aspects need to be accounted for, including morphology, lexicology, phonology, etc. This leads to problems like tokenization, lemmatization, stemming, semantic relations, word sense disambiguation, and Zipf's Law problem. Then, in terms of syntax, natural language, in contrast to formal language, has much less restricted grammar and thus is more challenging to parse: in the Chomsky hierarchy, natural language is generally considered to follow context-sensitive grammar, while programming languages are covered under context-free grammar. Syntactic parsing includes tasks such as part-of-speech tagging, constituency parsing, dependency parsing, and named entity recognition. Finally, to understand the semantics of a sentence in written language or an utterance in spoken language, the following tasks were studied: semantic role labeling, frame semantic parsing, abstract meaning representation, logical form parsing, etc.

\paragraph{Recurrent Neural Network \& Convolutional Neural Network} In the initial stages of NLP, rudimentary models relied on simple feed-forward neural networks to tackle various tasks \cite{DBLP:conf/nips/BengioDV00}. After the introduction of word embeddings like word2vec \cite{DBLP:conf/nips/MikolovSCCD13, DBLP:journals/corr/abs-1301-3781} and GloVe \cite{DBLP:conf/emnlp/PenningtonSM14}, NLP techniques embraced recurrent neural networks (RNNs) \cite{DBLP:journals/cogsci/Elman90}, such as LSTM \cite{DBLP:journals/neco/HochreiterS97}, GRU \cite{DBLP:conf/emnlp/ChoMGBBSB14}, RNNsearch \cite{DBLP:journals/corr/BahdanauCB14}, LSTM-CRF \cite{DBLP:journals/corr/HuangXY15}, etc. Examples of representative RNNs in NLP include ELMo \cite{DBLP:conf/naacl/PetersNIGCLZ18} and ULMFiT \cite{DBLP:conf/acl/RuderH18}. While RNNs played a significant role, alternative models utilizing convolutional neural networks also emerged. WordCNN \cite{DBLP:conf/emnlp/Kim14} employed CNNs at the word level, with word2vec features serving as input to the CNNs. Another approach, CharCNN \cite{DBLP:conf/nips/ZhangZL15}, focused on modeling language at the character level with CNN. Subsequent research \cite{DBLP:conf/acl/MaH16} highlighted that character-level CNNs excel at capturing word morphology, and their combination with a word-level LSTM backbone can significantly enhance performance.

\paragraph{Transformer \& Large Language Model} The groundbreaking Transformer model, introduced by \cite{DBLP:conf/nips/VaswaniSPUJGKP17}, revolutionized natural language processing through the introduction of the self-attention mechanism, inspiring a cascade of subsequent works. BERT \cite{DBLP:conf/naacl/DevlinCLT19} leverages the Transformer encoder stack, excelling in natural language understanding. On the other hand, the GPT family \cite{radford2018improving, radford2019language, brown2020language, DBLP:journals/corr/abs-2303-08774} is built upon Transformer decoder blocks, showcasing prowess in natural language generation tasks. A line of work strives to refine the original BERT, including RoBERTa \cite{DBLP:journals/corr/abs-1907-11692}, ALBERT \cite{DBLP:conf/iclr/LanCGGSS20}, ELECTRA \cite{DBLP:conf/iclr/ClarkLLM20}. Simultaneously, a parallel line of research following the GPT paradigm has given rise to models like XLNet \cite{DBLP:conf/nips/YangDYCSL19}, OPT \cite{DBLP:journals/corr/abs-2205-01068}. BART \cite{DBLP:conf/acl/LewisLGGMLSZ20}, an encoder-decoder Transformer, distinguishes itself through pretraining using the denoising sequence-to-sequence task. Meanwhile, T5 \cite{DBLP:journals/jmlr/RaffelSRLNMZLL20} introduces modifications to the original Transformer, maintaining the encoder-decoder architecture. T5 unifies various NLP tasks through a shared text-to-text format, exhibiting enhanced performance in transfer learning. These diverse models collectively showcase the versatility and ongoing evolution within the NLP landscape.

Over the past few years, there has been a remarkable expansion in the size of language models, driven by the scalability of the Transformer architecture. This trend has given rise to a series of large language models (LLMs) that have demonstrated breakthrough performance and capabilities not achievable with smaller models. A landmark model in this evolution is ChatGPT \cite{chatgpt-openai}, which has sparked considerable interest and inspired a series of works in this domain, such as GPT-4 \cite{DBLP:journals/corr/abs-2303-08774}, PaLM \cite{DBLP:journals/jmlr/ChowdheryNDBMRBCSGSSTMRBTSPRDHPBAI23}, PaLM-2 \cite{DBLP:journals/corr/abs-2305-10403}, LLaMA \cite{DBLP:journals/corr/abs-2302-13971}, LLaMA 2 \cite{DBLP:journals/corr/abs-2307-09288}, ERNIE 3.5 \cite{ernie35-baidu}. Notably, LLaMA stands out as one of the few open-source LLMs, fostering interesting developments. The introduction of  ``instruction-tuning'' has allowed efficient finetuning of a pretrained LLM to become an instruction-following model. This technique is popularized by InstructGPT \cite{DBLP:conf/nips/Ouyang0JAWMZASR22} and FLAN \cite{DBLP:conf/iclr/WeiBZGYLDDL22, DBLP:journals/corr/abs-2210-11416}. Recent advancements in instruction-following models include Alpaca \cite{alpaca}, employing self-instruction, and Vicuna \cite{vicuna2023}, leveraging conversations from \href{https://sharegpt.com}{ShareGPT}. As LLMs grow in scale and power, there is a shift away from the need for finetuning on downstream tasks. With appropriate prompts, LLMs can produce accurate outputs without task-specific training, a paradigm known as \textit{prompt engineering}. This approach differs from the traditional \textit{pretrain-finetune} paradigm. DPO \cite{DBLP:conf/nips/RafailovSMMEF23} introduces a method to train LLMs directly on human preferences without requiring explicit reward modeling or reinforcement learning, simplifying upon previous RLHF methods.

\subsubsection{Reinforcement Learning}
\label{sec:rl}

Reinforcement learning (RL) seeks to acquire a policy capable of taking optimal actions based on observations of the environment. Numerous vision-language-action models are constructed based on paradigms such as imitation learning or Temporal Difference (TD) learning within RL. Many challenges faced in the development of robotic policies can be effectively addressed through insights gained from the field of RL. Consequently, delving into RL methods presents a valuable avenue for enhancing robotic learning. For a deeper exploration of RL methods, readers can refer to more comprehensive reviews provided by \cite{DBLP:journals/corr/Li17b}, \cite{DBLP:journals/corr/abs-1708-05866}, and \cite{DBLP:journals/corr/abs-2301-03044}.


\paragraph{Deep Reinforcement Learning} The advent of deep reinforcement learning can be attributed to the success of pioneering models, Deep Q-Network \cite{DBLP:journals/nature/MnihKSRVBGRFOPB15} and AlphaGo \cite{DBLP:journals/nature/SilverHMGSDSAPL16}. Deep learning, with its ability to provide low-dimensional representations, proved instrumental in overcoming traditional computational and memory complexity challenges in reinforcement learning. In recent years, a multitude of value-function-based approaches have surfaced. Double DQN (D-DQN) \cite{DBLP:conf/aaai/HasseltGS16} addresses the action value overestimation issue of DQN. Hindsight experience replay (HER) \cite{DBLP:conf/nips/AndrychowiczCRS17} focuses on the sparse reward issue. Batch-Constrained deep Q-learning (BCQ) \cite{DBLP:conf/icml/FujimotoMP19} presents an approach aimed at enhancing off-policy learning by constraining the action space. BEAR \cite{DBLP:conf/nips/KumarFSTL19} endeavors to alleviate instability arising from bootstrapping errors in off-policy RL. \cite{DBLP:conf/nips/KumarZTL20} introduces conservative Q-learning (CQL) to address the overestimation of values by standard off-policy RL methods.

Another paradigm within reinforcement learning is policy search, encompassing methods such as policy gradient and actor-critic techniques. These approaches aim to combat persistent challenges, including instability, slow convergence, and data inefficiency. Guided policy search (GPS) \cite{DBLP:conf/icml/LevineK13} learns a policy with importance sampling guided by another controller toward a local optimum. Deterministic policy gradient (DPG) \cite{DBLP:conf/icml/SilverLHDWR14}, deep deterministic policy gradient (DDPG) \cite{DBLP:journals/corr/LillicrapHPHETS15}, and asynchronous advantage actor-critic (A3C) \cite{DBLP:conf/icml/MnihBMGLHSK16} improve efficiency without compromising stability. Normalized advantage functions (NAF) \cite{DBLP:conf/icml/GuLSL16} are a continuous variant of Q-learning, allowing for Q-learning with experience replay. Soft actor-critic (Soft AC) \cite{DBLP:conf/icml/HaarnojaZAL18} takes advantage of the maximum entropy RL framework to lower sample complexity and improve convergence properties. Trust region policy optimization (TRPO) \cite{DBLP:conf/icml/SchulmanLAJM15} and proximal policy optimization (PPO) \cite{DBLP:journals/corr/SchulmanWDRK17} utilize trust region methods to stabilize policy gradients, with PPO additionally incorporating a truncated generalized advantage estimation (GAE) \cite{DBLP:journals/corr/SchulmanMLJA15}.

Beyond these, various other RL methodologies exist, such as imitation learning and hierarchical reinforcement learning. Generative adversarial imitation learning (GAIL) \cite{DBLP:conf/nips/HoE16} is an imitation learning method that uses a generative adversarial framework to discriminate expert trajectories against generated trajectories. Robust adversarial reinforcement learning (RARL) \cite{DBLP:conf/icml/PintoDSG17} incorporates adversarial agents to enhance generalization. RLHF \cite{DBLP:conf/nips/ChristianoLBMLA17} utilizes human preferences without access to the reward function. FeUdal Networks (FuN) \cite{DBLP:conf/icml/VezhnevetsOSHJS17} introduce a hierarchical reinforcement learning architecture featuring a Manager module and a Worker module.


\paragraph{Robotics} In the field of RL, robotics stands out as one of the most prevalent and impactful applications. A noteworthy contribution to this field is E2E-DVP \cite{DBLP:journals/jmlr/LevineFDA16}. This pioneering model represents one of the first end-to-end solutions for robotic control. Its neural network is designed to take raw images as input and generate motor torques as output. \cite{DBLP:conf/iser/LevinePKQ16} curated a substantial real-world dataset and developed a CNN that predicts grasps based on monocular input images. Building upon these foundations, QT-Opt \cite{DBLP:journals/corr/abs-1806-10293} further expands the dataset and model scale, introducing closed-loop control capabilities to enhance robotic control systems. Then Dreamer \cite{DBLP:conf/iclr/HafnerLB020} addresses long-horizon tasks. OpenAI also developed a dexterous robot hand that can solve the Rubik's cube \cite{DBLP:journals/corr/abs-1808-00177, DBLP:journals/corr/abs-1910-07113}. 


\subsubsection{Graph}
Graph is ubiquitous in many scenarios, such as social networks, molecule structures, 3D object meshes, etc. Even images and text can be modeled as a 2D grid and a linear graph (path graph), respectively. To process graph-structured data, recurrent graph neural networks \cite{DBLP:journals/tnn/ScarselliGTHM09} were first introduced, which were later optimized by convolutional graph neural networks. The review of graph neural networks (GNN) \cite{DBLP:journals/tnn/WuPCLZY21} can be referred to for more in-depth details.

Convolutional graph neural networks can be generally divided into two categories: spectral-based and spatial-based. Spectral-based convolutional GNNs draw inspiration from graph signal processing, which provides theoretical support for the design of the networks. However, spatial-based convolutional GNNs have an advantage in terms of their efficiency and flexibility. Spectral CNN \cite{DBLP:journals/corr/BrunaZSL13} is one of the first convolutional GNNs, but it is not robust to changes in graph structure and has a high computational cost. ChebNet \cite{DBLP:conf/nips/DefferrardBV16} and GCN \cite{DBLP:conf/iclr/KipfW17} significantly reduced the cost: ChebNet used an approximation based on Chebyshev polynomials, and GCN is its first-order approximation. Neural Network for Graphs (NN4G) \cite{DBLP:journals/tnn/Micheli09} is the first spatial network. MPNN \cite{DBLP:conf/icml/GilmerSRVD17} introduced a general framework of spatial-based networks under which most existing GNNs can be covered. But the drawback of MPNN is that it does not embed graph structure information, which is later solved by GIN \cite{DBLP:conf/iclr/XuHLJ19}. GraphSage \cite{DBLP:conf/nips/HamiltonYL17} improves efficiency by sampling a fixed number of neighbors. Graph Attention Network (GAT) \cite{DBLP:conf/iclr/VelickovicCCRLB18} incorporates the attention mechanism. 

Besides recurrent and convolutional graph neural networks, there are graph autoencoders \cite{DBLP:conf/aaai/CaoLX16, DBLP:journals/corr/KipfW16a} and spatial-temporal graph neural networks \cite{DBLP:conf/iconip/SeoDVB18}. Equivariant message passing networks are recently introduced to handle 3D graphs, including E(n)- and SE(3)- Equivariant GNN \cite{DBLP:conf/icml/DuZDMC0SL22, DBLP:conf/icml/SatorrasHW21}. Graph Transformer models make use of the power of Transformers to process graph data. There are already over 20 such graph Transformer models, such as GROVER \cite{DBLP:conf/nips/RongBXX0HH20} and SE(3)-Transformers \cite{DBLP:conf/nips/FuchsW0W20}.

\paragraph{Graph and Vision} Graph structures also exist in some computer vision tasks. Scene graph \cite{DBLP:journals/ijcv/KrishnaZGJHKCKL17} can be used to express object relationships in most visual inputs. In addition to detecting objects in an image, scene graph generation necessitates the understanding of the relationships between detected objects. For example, a model needs to detect a person and a cup, and then understand that the person is holding the cup, which is the relationship between the two objects. Knowledge graphs often contain visual illustrations, such as WikiData. Those graphs can be helpful in downstream computer vision tasks.

\paragraph{Graph and Language} Graph structures are ubiquitous in language data \cite{DBLP:journals/ftml/WuCSGGLPL23}. Word-level graphs include dependency graphs, constituency graphs, AMR graphs, etc. Word-level means each node of those graphs corresponds to a word in the original text. These graphs can be used to explicitly represent the syntax or semantics of the raw sentence. Sentence-level graphs can be useful in dialog tracking \cite{DBLP:conf/emnlp/GhosalMPCG19}, fact checking \cite{DBLP:conf/acl/ZhongXTXDZWY20}, etc. Document-level graphs include knowledge graphs \cite{DBLP:journals/corr/abs-2107-09556}, citation graphs \cite{DBLP:conf/kdd/TangZYLZS08}, etc. They can be used in document-level tasks, such as document retrieval, document clustering, etc. Different types of language graphs are often processed using the aforementioned GNNs to facilitate downstream tasks.

\subsection{Background (Extended Version): Vision-Language Models}
\label{sec:vision-language}

\begin{table*}
    \centering
    \caption{Vision-language models. In the ``Objective'' column: ``MLM'': masked language modeling. ``MVM'': masked vision modeling, reconstructing masked image regions. ``VLM'': binary classification of whether vision and language inputs match. ``LM'': autoregressive language modeling. ``VLCL'': vision-language contrastive learning. We only include representative multimodal datasets due to limited space. ``MM'' includes multimodal tasks such as visual question answering, image captioning, and vision-language retrieval. ``Vision'' represents computer vision tasks, such as image classification. ``Language'' represents natural language processing tasks}
    \resizebox{\textwidth}{!}{%
    \begin{tabular}{>{\RaggedRight}p{0.3cm} 
                    >{\RaggedRight}p{2.5cm} 
                    >{\RaggedRight}p{3cm} 
                    >{\RaggedRight}p{1cm} 
                    >{\RaggedRight}p{2cm} 
                    >{\RaggedRight}p{1cm} 
                    >{\RaggedRight}p{2cm} 
                    >{\RaggedRight}p{1.75cm} 
                    >{\RaggedRight}p{4cm} 
                    >{\RaggedRight}p{2cm} 
                    @{} 
                    }
        \toprule
                   & & \multicolumn{2}{c}{\textbf{Vision Encoder}} & \multicolumn{2}{c}{\textbf{Language Encoder}} & & & & \\
        & \textbf{Model} & \textbf{Name} & \textbf{Params} & \textbf{Name} & \textbf{Params} & \textbf{VL-Fusion} &  \textbf{Objectives} & \textbf{Datasets} & \textbf{Tasks}\\
        \midrule
        \multirow{20}{*}{\rotatebox[origin=c]{90}{Self-supervised}}
        & ViLBERT \cite{DBLP:conf/nips/LuBPL19} &
            Faster R-CNN \cite{DBLP:conf/nips/RenHGS15, DBLP:conf/cvpr/00010BT0GZ18} & 44M &
            Dual-stream BERT-base \cite{DBLP:conf/naacl/DevlinCLT19} & 221M & 
            Dual-stream &
            MLM, MVM, VLM &
            COCO, VG &
            MM
            \\ \cmidrule(lr){2-2} \cmidrule(lr){3-4} \cmidrule(lr){5-6} \cmidrule(lr){7-7} \cmidrule(lr){8-10}
        & LXMERT \cite{DBLP:conf/emnlp/TanB19} & 
            Faster R-CNN \cite{DBLP:conf/nips/RenHGS15, DBLP:conf/cvpr/00010BT0GZ18} & 44M &
            Dual-stream BERT-base \cite{DBLP:conf/naacl/DevlinCLT19} & 183M & 
            Dual-stream &
            MLM, MVM, VLM, VQA & 
            COCO, VG, VQA, GQA, VGQA & 
            MM
            \\ \cmidrule(lr){2-2} \cmidrule(lr){3-4} \cmidrule(lr){5-6} \cmidrule(lr){7-7} \cmidrule(lr){8-10}
        & VisualBERT \cite{DBLP:journals/corr/abs-1908-03557} &
            Faster R-CNN \cite{DBLP:conf/nips/RenHGS15, DBLP:conf/cvpr/00010BT0GZ18} & 60M &
            BERT-base \cite{DBLP:conf/naacl/DevlinCLT19} & 110M & 
            Single-stream &
            MLM, VLM &
            COCO & 
            MM
            \\ \cmidrule(lr){2-2} \cmidrule(lr){3-4} \cmidrule(lr){5-6} \cmidrule(lr){7-7} \cmidrule(lr){8-10}
        & VL-BERT \cite{DBLP:conf/iclr/SuZCLLWD20} &
            Faster R-CNN \cite{DBLP:conf/nips/RenHGS15, DBLP:conf/cvpr/00010BT0GZ18} & 44M &
            BERT-base \cite{DBLP:conf/naacl/DevlinCLT19} & 110M & 
            Single-stream &
            MLM, MVM & 
            CC &
            MM
            \\ \cmidrule(lr){2-2} \cmidrule(lr){3-4} \cmidrule(lr){5-6} \cmidrule(lr){7-7} \cmidrule(lr){8-10}
        & UNITER \cite{DBLP:conf/eccv/ChenLYK0G0020} & 
            Faster R-CNN \cite{DBLP:conf/nips/RenHGS15, DBLP:conf/cvpr/00010BT0GZ18} & 44M &
            BERT-base/ \newline BERT-large \cite{DBLP:conf/naacl/DevlinCLT19} & 86M/ \newline 303M & 
            Single-stream &
            MLM, VLM, MVM, WRA & 
            COCO, VG, SBU, CC &
            MM
            \\ \cmidrule(lr){2-2} \cmidrule(lr){3-4} \cmidrule(lr){5-6} \cmidrule(lr){7-7} \cmidrule(lr){8-10}
        & ViLT \cite{DBLP:conf/icml/KimSK21} & 
            Linear projection \cite{DBLP:conf/iclr/DosovitskiyB0WZ21} & 2.4M &
            BERT-base \cite{DBLP:conf/naacl/DevlinCLT19} & 85M &
            Single-stream &
            MLM, VLM & 
            COCO, VG, SBU, CC &
            MM
            \\ \cmidrule(lr){2-2} \cmidrule(lr){3-4} \cmidrule(lr){5-6} \cmidrule(lr){7-7} \cmidrule(lr){8-10}
        & SimVLM \cite{DBLP:conf/iclr/WangYYDT022} &
            ViT/CoAtNet-huge (shared encoder) \cite{DBLP:conf/nips/DaiLLT21} & 632M &
            Shared encoder & 632M & 
            Single-stream &
            PrefixLM & 
            ALIGN dataset & 
            MM
            \\ \cmidrule(lr){2-2} \cmidrule(lr){3-4} \cmidrule(lr){5-6} \cmidrule(lr){7-7} \cmidrule(lr){8-10}
        & GIT \cite{DBLP:journals/tmlr/WangYHLLGLLW22} &
            Florence \cite{DBLP:journals/corr/abs-2111-11432} & 637M & 
            Transformer \cite{DBLP:conf/nips/VaswaniSPUJGKP17} & 60M &
            Single-stream &
            LM & 
            COCO, VG, SBU, CC, etc. & 
            MM
            \\ \cmidrule(lr){2-2} \cmidrule(lr){3-4} \cmidrule(lr){5-6} \cmidrule(lr){7-7} \cmidrule(lr){8-10}
        & BEiT-3 \cite{DBLP:conf/cvpr/WangBDBPLAMSSW23} &
            V-FFN \newline (+Shared Attn) & 692M (+317M) &
            L-FFN \newline (+Shared Attn) & 692M (+317M) & 
            Modality experts & 
            MLM, VLM & 
            COCO, VG, SBU, CC &
            MM, Vision
        \\ \midrule

        \multirow{10}{*}{\rotatebox[origin=c]{90}{Contrastive}}
        & CLIP \cite{DBLP:conf/icml/RadfordKHRGASAM21} &
            ViT \cite{DBLP:conf/iclr/DosovitskiyB0WZ21} & 428M & 
            GPT-2 \cite{radford2019language} & 63M & 
            Two-tower &
            VLCL & 
            WTI & 
            Vision
            \\ \cmidrule(lr){2-2} \cmidrule(lr){3-4} \cmidrule(lr){5-6} \cmidrule(lr){7-7} \cmidrule(lr){8-10}
        & FILIP \cite{DBLP:conf/iclr/YaoHHLNXLLJX22} &
            ViT-L/14 \cite{DBLP:conf/iclr/DosovitskiyB0WZ21} & 428M & 
            GPT \cite{radford2019language} & 117M & 
            Two-tower &
            VLCL & 
            FILIP300M (Self-collect) &
            MM, Vision
            \\ \cmidrule(lr){2-2} \cmidrule(lr){3-4} \cmidrule(lr){5-6} \cmidrule(lr){7-7} \cmidrule(lr){8-10}
        & ALIGN \cite{DBLP:conf/icml/JiaYXCPPLSLD21} &
            EfficientNet-L2 \cite{DBLP:conf/cvpr/XieLHL20} & 480M &
            BERT-large \cite{DBLP:conf/naacl/DevlinCLT19} & 336M &
            Two-tower &
            VLCL & 
            ALIGN dataset (Self-collect) & 
            MM, Vision
            \\ \cmidrule(lr){2-2} \cmidrule(lr){3-4} \cmidrule(lr){5-6} \cmidrule(lr){7-7} \cmidrule(lr){8-10}
        & ALBEF \cite{DBLP:conf/nips/LiSGJXH21} &
            ViT-B/16 \cite{DBLP:conf/iclr/DosovitskiyB0WZ21} & 87M &
            BERT-base \cite{DBLP:conf/naacl/DevlinCLT19} & 85M & 
            Dual-stream &
            MLM, VLM, VLCL & 
            COCO, VG, CC, SBU & 
            MM
            \\ \cmidrule(lr){2-2} \cmidrule(lr){3-4} \cmidrule(lr){5-6} \cmidrule(lr){7-7} \cmidrule(lr){8-10}
        & FLAVA \cite{DBLP:conf/cvpr/SinghHGCGRK22} &
            ViT-B/16 \cite{DBLP:conf/iclr/DosovitskiyB0WZ21} & 87M & 
            RoBERTa-base \cite{DBLP:journals/corr/abs-1907-11692} & 125M & 
            Dual-stream &
            MLM, MVM, VLM, VLCL & 
            COCO, VG, CC, SBU, etc. (PMD) & 
            MM, Vision, Language
            \\ \cmidrule(lr){2-2} \cmidrule(lr){3-4} \cmidrule(lr){5-6} \cmidrule(lr){7-7} \cmidrule(lr){8-10}
        & Florence \cite{DBLP:journals/corr/abs-2111-11432} &
            Hierarchical Vision Transformers \cite{DBLP:conf/iccv/LiuL00W0LG21, DBLP:conf/iccv/WuXCLDY021} & 637M & 
            RoBERTa \cite{DBLP:journals/corr/abs-1907-11692} & 125M & 
            Two-tower &
            VLCL & 
            FLD-900M (Self-collect) &
            Vision
        \\ \midrule

        \multirow{40}{*}{\rotatebox[origin=c]{90}{Large Multimodal Model}}
        & Flamingo \cite{DBLP:conf/nips/AlayracDLMBHLMM22} &
            NFNet-F6 \cite{DBLP:conf/icml/BrockDSS21} & 438M &
            Chinchilla \cite{DBLP:journals/corr/abs-2203-15556} & 70B &
            Dual-stream &
            LM &
            M3W, ALIGN dataset, LTIP, VTP &
            MM
            \\ \cmidrule(lr){2-2} \cmidrule(lr){3-4} \cmidrule(lr){5-6} \cmidrule(lr){7-7} \cmidrule(lr){8-10}
        & BLIP-2 \cite{DBLP:conf/icml/0008LSH23} &
            CLIP ViT-L/14 \cite{DBLP:conf/icml/RadfordKHRGASAM21}, EVA ViT-G/14 \cite{DBLP:conf/cvpr/FangWXSWW0WC23} \newline + Q-Former & 428M, \newline 1B & 
            OPT \cite{DBLP:journals/corr/abs-2205-01068} \newline Flan-T5 \cite{DBLP:journals/corr/abs-2210-11416} & 6.7B \newline 3B/11B &
            Single-stream &
            BLIP, LM &
            COCO, VG, CC, SBU, LAION & 
            MM
            \\ \cmidrule(lr){2-2} \cmidrule(lr){3-4} \cmidrule(lr){5-6} \cmidrule(lr){7-7} \cmidrule(lr){8-10}
        & PaLI \cite{DBLP:conf/iclr/Chen0CPPSGGMB0P23} &
            ViT-e \cite{DBLP:conf/iclr/Chen0CPPSGGMB0P23} & 4B &
            mT5-XXL \cite{DBLP:conf/naacl/XueCRKASBR21} & 13B &
            Single-stream &
            Mixed &
            WebLI, etc &
            MM
            \\ \cmidrule(lr){2-2} \cmidrule(lr){3-4} \cmidrule(lr){5-6} \cmidrule(lr){7-7} \cmidrule(lr){8-10}
        & PaLI-X \cite{DBLP:journals/corr/abs-2305-18565} &
            ViT-22B \cite{DBLP:journals/corr/abs-2302-05442} & 22B &
            UL2 \cite{DBLP:conf/iclr/Tay00GW0CBSZZHM23} & 32B &
            Single-stream &
            Mixed &
            WebLI, etc &
            MM
            \\ \cmidrule(lr){2-2} \cmidrule(lr){3-4} \cmidrule(lr){5-6} \cmidrule(lr){7-7} \cmidrule(lr){8-10}
        & LLaMA-Adapter \cite{DBLP:journals/corr/abs-2303-16199} & 
            CLIP ViT-B/16 \cite{DBLP:conf/icml/RadfordKHRGASAM21} & 87M & 
            LLaMA \cite{DBLP:journals/corr/abs-2302-13971} & 7B &
            Single-stream &
            LM &
            Self-instruct & 
            Instruction-following
            \\ \cmidrule(lr){2-2} \cmidrule(lr){3-4} \cmidrule(lr){5-6} \cmidrule(lr){7-7} \cmidrule(lr){8-10}
        & LLaMA-Adapter-V2 \cite{DBLP:journals/corr/abs-2304-15010} & 
            CLIP ViT-L/14 \cite{DBLP:conf/icml/RadfordKHRGASAM21} & 428M & 
            LLaMA \cite{DBLP:journals/corr/abs-2302-13971} & 7B &
            Single-stream &
            LM &
            GPT-4-LLM, COCO, ShareGPT &
            Instruction-following
            \\ \cmidrule(lr){2-2} \cmidrule(lr){3-4} \cmidrule(lr){5-6} \cmidrule(lr){7-7} \cmidrule(lr){8-10}
        & Kosmos-1 \cite{DBLP:journals/corr/abs-2302-14045}, \newline Kosmos-2 \cite{DBLP:journals/corr/abs-2306-14824} & 
            CLIP ViT-L/14 \cite{DBLP:conf/icml/RadfordKHRGASAM21} & 428M & 
            Magneto \cite{DBLP:journals/corr/abs-2210-06423} & 1.3B & 
            Single-stream &
            LM & 
            LAION, COYO, CC; Unnatural Instructions, FLANv2 &
            Instruction-following (Kosmos-2 w/ grounding, referring)
            \\ \cmidrule(lr){2-2} \cmidrule(lr){3-4} \cmidrule(lr){5-6} \cmidrule(lr){7-7} \cmidrule(lr){8-10}
        & InstructBLIP \cite{DBLP:journals/corr/abs-2305-06500} & 
            EVA ViT-G/14 \cite{DBLP:conf/cvpr/FangWXSWW0WC23} & 1B & 
            Flan-T5 \cite{DBLP:journals/corr/abs-2210-11416} \newline Vicuna \cite{vicuna2023} & 3B/11B \newline 7B/13B &
            Single-stream &
            BLIP, LM & 
            COCO, VQA, LLaVA-Instruct-150K, etc. (26 datasets) & 
            Instruction-following
            \\ \cmidrule(lr){2-2} \cmidrule(lr){3-4} \cmidrule(lr){5-6} \cmidrule(lr){7-7} \cmidrule(lr){8-10}
        & LLaVA \cite{DBLP:journals/corr/abs-2304-08485} &
            CLIP ViT-L/14 \cite{DBLP:conf/icml/RadfordKHRGASAM21} & 428M & 
            LLaMA \cite{DBLP:journals/corr/abs-2302-13971} & 13B &
            Single-stream &
            LM & 
            CC, (FT: GPT-assisted Visual Instruction Data Generation) & 
            Instruction-following
            \\ \cmidrule(lr){2-2} \cmidrule(lr){3-4} \cmidrule(lr){5-6} \cmidrule(lr){7-7} \cmidrule(lr){8-10}
        & MiniGPT-4 \cite{DBLP:journals/corr/abs-2304-10592} &
            EVA ViT-G/14 \cite{DBLP:conf/cvpr/FangWXSWW0WC23} \newline  + Q-Former \cite{DBLP:conf/icml/0008LSH23} & 1B & 
            Vicuna \cite{vicuna2023} \newline LLaMA2 \cite{DBLP:journals/corr/abs-2307-09288} & 7B/13B \newline 7B &
            Single-stream &
            LM &
            CC, SBU, LAION (FT: SC) &
            Instruction-following
            \\ \cmidrule(lr){2-2} \cmidrule(lr){3-4} \cmidrule(lr){5-6} \cmidrule(lr){7-7} \cmidrule(lr){8-10}
        & Video-LLaMA \cite{DBLP:journals/corr/abs-2306-02858} &
            EVA ViT-G/14 \cite{DBLP:conf/cvpr/FangWXSWW0WC23} \newline  + Q-Former \cite{DBLP:conf/icml/0008LSH23} & 1B & 
            LLaMA \cite{DBLP:journals/corr/abs-2302-13971} & 7B/13B & 
            Single-stream &
            BLIP, LM & 
            CC595k &
            Instruction-following
            \\ \cmidrule(lr){2-2} \cmidrule(lr){3-4} \cmidrule(lr){5-6} \cmidrule(lr){7-7} \cmidrule(lr){8-10}
        & PandaGPT \cite{DBLP:journals/corr/abs-2305-16355} &
            ImageBind ViT-H \cite{DBLP:conf/cvpr/GirdharELSAJM23} & 632M & 
            Vicuna \cite{vicuna2023} & 13B &
            Single-stream &
            LM & 
            (FT: LLaVA data, MiniGPT-4 data) & 
            Instruction-following
            \\ \cmidrule(lr){2-2} \cmidrule(lr){3-4} \cmidrule(lr){5-6} \cmidrule(lr){7-7} \cmidrule(lr){8-10}
        & VideoChat \cite{DBLP:journals/corr/abs-2305-06355} &
            EVA ViT-G/14 \cite{DBLP:conf/cvpr/FangWXSWW0WC23} \newline  + Q-Former \cite{DBLP:conf/icml/0008LSH23} & 1B & 
            StableVicuna \cite{stablelm} & 13B &
            Single-stream &
            LM &
            COCO, VG, CC, SBU (FT: SC, MiniGPT-4, LLaVA data) &
            Instruction-following
            \\ \cmidrule(lr){2-2} \cmidrule(lr){3-4} \cmidrule(lr){5-6} \cmidrule(lr){7-7} \cmidrule(lr){8-10}
        & ChatSpot \cite{DBLP:journals/corr/abs-2307-09474} & 
            CLIP ViT-L/14 \cite{DBLP:conf/icml/RadfordKHRGASAM21} & 428M & 
            Vicuna \cite{vicuna2023} & 7B & 
            Single-stream &
            LM & 
            MGVLID, RegionChat &
            Instruction-following, Vision
            \\ \cmidrule(lr){2-2} \cmidrule(lr){3-4} \cmidrule(lr){5-6} \cmidrule(lr){7-7} \cmidrule(lr){8-10}
        & mPLUG-Owl \cite{DBLP:journals/corr/abs-2304-14178}, mPLUG-Owl2 \cite{DBLP:journals/corr/abs-2311-04257} &
            CLIP ViT-L/14 \cite{DBLP:conf/icml/RadfordKHRGASAM21} \newline + Visual Abstractor & 428M & 
            LLaMA \cite{DBLP:journals/corr/abs-2302-13971} & 7B &
            Single-stream &
            LM &
            LAION, COYO, CC, COCO (FT: Alpaca, Vicuna, Baize \cite{DBLP:conf/emnlp/XuGDM23} data) &
            Instruction-following
            \\ \cmidrule(lr){2-2} \cmidrule(lr){3-4} \cmidrule(lr){5-6} \cmidrule(lr){7-7} \cmidrule(lr){8-10}
        & Visual ChatGPT \cite{DBLP:journals/corr/abs-2303-04671} &
            (22 different models) & - &
            ChatGPT \cite{chatgpt-openai} & - &
            Prompt Manager &
            - &
            \multicolumn{2}{l}{Add image understanding and generation to ChatGPT}
            \\ \cmidrule(lr){2-2} \cmidrule(lr){3-4} \cmidrule(lr){5-6} \cmidrule(lr){7-7} \cmidrule(lr){8-10}
        & X-LLM \cite{DBLP:journals/corr/abs-2305-04160} &
            ViT-G \cite{DBLP:conf/cvpr/Zhai0HB22} + Q-Former \cite{DBLP:conf/icml/0008LSH23} + Adapter & 1.8B &
            ChatGLM \cite{DBLP:conf/acl/DuQLDQY022} & 6B &
            Single-stream &
            Three-stage training &
            MiniGPT-4 data, AISHELL-2, ActivityNet, VSDial-CN (SC) &
            Instruction-following
            \\ 
        \bottomrule
    \end{tabular}
    }
    \label{tb:vlm}
\end{table*}

Comprehensive surveys on VLMs exist, covering early BERT-based VLMs \cite{DBLP:journals/ijautcomp/ChenZHCSXX23, DBLP:journals/pami/XuZC23} (Section~\ref{sec:ss_pretrain}), as well as more recent VLMs with contrastive pretraining \cite{DBLP:journals/ijautcomp/WangCQGWWTG23, DBLP:journals/corr/abs-2304-00685} (Section~\ref{sec:ctrs_pretrain}). Given the rapid evolution of this field and the emergence of new VLMs based on large language models, commonly known as large multimodal models (LMMs), we also compile the latest developments of LMMs (Section~\ref{sec:LMM}). To compare the most representative VLMs, we include their specifications in Table~\ref{tb:vlm}.


\subsubsection{Self-Supervised Pretraining} 
\label{sec:ss_pretrain}

The evolution of the Transformer architecture to accommodate various modalities has given rise to robust multimodal Transformer models. Initial VLMs based on BERT can be broadly categorized into two types: single-stream and multi-stream \cite{DBLP:journals/ijautcomp/ChenZHCSXX23}. Single-stream models employ a single stack of Transformer blocks to process both visual and linguistic inputs, whereas multi-stream models utilize a separate Transformer stack for each modality, with Transformer cross-attention layers exchanging multimodal information. To enhance alignment among modalities, these models incorporate various pretraining tasks aimed at absorbing knowledge from out-of-domain data. ViLBERT \cite{DBLP:conf/nips/LuBPL19} stands as the pioneer in this line of work, featuring a multi-stream Transformer architecture. Text input undergoes standard processing in the language Transformer; image input is first processed using Faster R-CNN, and the output embeddings of all objects are then passed into the vision stream. The two Transformer outputs---language embeddings and vision embeddings---are combined using a novel co-attention Transformer layer. VL-BERT \cite{DBLP:conf/iclr/SuZCLLWD20} adopts a single-stream multimodal Transformer, simply concatenating vision and language tokens into a single input sequence. VideoBERT \cite{DBLP:conf/iccv/SunMV0S19} adapts mulitmodal Transformer models to video inputs. UNITER \cite{DBLP:conf/eccv/ChenLYK0G0020} proposes the word-region alignment loss to explicitly align word with image regions. ViLT \cite{DBLP:conf/icml/KimSK21} uses ViT-style \cite{DBLP:conf/iclr/DosovitskiyB0WZ21} image patch projection to embed images, deviating from previous region or grid features. 

SimVLM \cite{DBLP:conf/iclr/WangYYDT022} opts for a streamlined approach, relying solely on a single prefix language modeling objective to reduce training costs. VLMo \cite{DBLP:conf/nips/BaoW0LMASPW22} and BEiT-3 \cite{DBLP:conf/cvpr/WangBDBPLAMSSW23} both introduce mixture-of-modality-experts Transformers to effectively handle multimodal inputs.

\subsubsection{Contrastive Pretraining}
\label{sec:ctrs_pretrain}

Vision-language pretraining in the initial series of BERT-based VLMs has evolved, with refinements such as curating larger-scale pretraining datasets, leveraging multimodal contrastive learning, and exploring specialized multimodal architectures. CLIP \cite{DBLP:conf/icml/RadfordKHRGASAM21} is one of the earliest attempts in vision-language contrastive pretraining. By contrastive pretraining on a large-scale image-text pair dataset, CLIP exhibits the capability to be transferred to downstream tasks in a zero-shot fashion.  In the same line of work, other few-/zero-shot learners have emerged. FILIP \cite{DBLP:conf/iclr/YaoHHLNXLLJX22} concentrates on finer-grained multimodal interactions with a token-wise contrastive objective. ALIGN \cite{DBLP:conf/icml/JiaYXCPPLSLD21} focuses on learning from noisy datasets collected without filtering and post-processing. ``Locked-image Tuning'' (LiT) \cite{DBLP:conf/cvpr/ZhaiWMSK0B22} posits that only training the text model while freezing the image model yields the best results on new tasks. In Frozen \cite{DBLP:conf/nips/TsimpoukelliMCE21}, the pretrained language model is frozen and a vision encoder is trained to produce image embeddings as a part of language model prompts, exemplifying an instance of prompt tuning. 

Unlike the two-tower frameworks of CLIP, FILIP, and ALIGN, which solely train unimodal encoders (image encoder and text encoder), ALBEF \cite{DBLP:conf/nips/LiSGJXH21} additionally incorporates training a multimodal encoder on top of the unimodal encoders, with FLAVA \cite{DBLP:conf/cvpr/SinghHGCGRK22} sharing a similar idea. In contrast to contrastive pretraining methods, CoCa \cite{DBLP:journals/tmlr/YuWVYSW22} seeks to amalgamate the strengths of CLIP's contrastive learning and SimVLM's generative objective. Florence \cite{DBLP:journals/corr/abs-2111-11432} generalizes representations from coarse, scene-level to fine, object-level, expands from images to videos, and encompasses modalities beyond RGB channels. OFA \cite{DBLP:conf/icml/WangYMLBLMZZY22} draws inspiration from T5 \cite{DBLP:journals/jmlr/RaffelSRLNMZLL20} and proposes unifying diverse unimodal and multimodal tasks under a sequence-to-sequence learning framework.

\subsubsection{Large Multimodal Model} 
\label{sec:LMM}

Large language models (LLMs) encapsulate extensive knowledge, and efforts have been made to transfer this knowledge to multimodal tasks. Given the resource-intensive nature of finetuning entire LLMs on multimodal tasks due to their large size, various techniques have been explored to effectively connect frozen LLMs with vision encoders, enabling the combined model to acquire multimodal capabilities. Flamingo \cite{DBLP:conf/nips/AlayracDLMBHLMM22} connects a pretrained vision encoder, NFNet \cite{DBLP:conf/icml/BrockDSS21}, and a large language model, Chinchilla \cite{DBLP:journals/corr/abs-2203-15556}, by inserting trainable gated cross-attention layers while keeping the rest of the model frozen. BLIP-2 \cite{DBLP:conf/icml/0008LSH23} introduces Q-Former, bootstrapping vision-language representation learning first from a frozen CLIP ViT \cite{DBLP:conf/icml/RadfordKHRGASAM21} and then from a frozen LLM, OPT \cite{DBLP:journals/corr/abs-2205-01068} or Flan-T5 \cite{DBLP:journals/corr/abs-2210-11416}. PaLI \cite{DBLP:conf/iclr/Chen0CPPSGGMB0P23} and PaLI-X \cite{DBLP:journals/corr/abs-2305-18565} investigate the advantages of jointly scaling up the vision and language components using large-scale multilingual image-text data.

Similar to developments in NLP, instruction-following has become a crucial aspect of VLMs, prompting the exploration of various multimodal instruction-tuning methods. LLaMA-Adapter \cite{DBLP:journals/corr/abs-2303-16199, DBLP:journals/corr/abs-2304-15010} employs a parameter-efficient finetuning (PEFT) technique, enabling LLaMA \cite{DBLP:journals/corr/abs-2302-13971} to process visual inputs. Kosmos-1 \cite{DBLP:journals/corr/abs-2302-14045} introduces a less restrictive input format that accommodates interleaved image and text. Its Magneto LLM \cite{DBLP:journals/corr/abs-2210-06423} serves as a ``general-purpose interface'' for docking with perception modules \cite{DBLP:conf/icml/RadfordKHRGASAM21}. Kosmos-2 \cite{DBLP:journals/corr/abs-2306-14824} adds additional grounding and referring capabilities. InstructBLIP \cite{DBLP:journals/corr/abs-2305-06500} achieves instruction-following using an instruction-aware Q-Former based on BLIP-2's Q-Former \cite{DBLP:conf/icml/0008LSH23}. Comparable to Kosmos-2, ChatSpot \cite{DBLP:journals/corr/abs-2307-09474} excels at following precise referring instructions, utilizing CLIP ViT \cite{DBLP:conf/icml/RadfordKHRGASAM21} and Vicuna \cite{vicuna2023}. X-LLM \cite{DBLP:journals/corr/abs-2305-04160} converts multimodality data into LLM inputs using X2L interfaces and treats them as foreign languages, where the X2L interface is inspired by the Q-Former from BLIP-2 \cite{DBLP:conf/icml/0008LSH23}. mPLUG-Owl \cite{DBLP:journals/corr/abs-2304-14178, DBLP:journals/corr/abs-2311-04257} introduces a two-stage training paradigm that establishes a connection between a pretrained LLM with a visual encoder and a visual abstractor, thereby endowing LLMs with multimodality abilities. Visual ChatGPT \cite{DBLP:journals/corr/abs-2303-04671} proposes a prompt manager that manages the interaction between ChatGPT and 22 visual foundation models, with the goal of equipping ChatGPT with the capability to understand and generate images. 

Rather than employing intricate mechanisms to connect components for different modalities, both LLaVA \cite{DBLP:journals/corr/abs-2304-08485} and MiniGPT-4 \cite{DBLP:journals/corr/abs-2304-10592} propose connecting vision encoders with LLMs through a single linear layer. LLaVA adopts a two-stage instruction-tuning approach, pretraining the CLIP ViT vision encoder \cite{DBLP:conf/icml/RadfordKHRGASAM21} in the first stage and finetuning the linear layer and the LLaMA LLM \cite{DBLP:journals/corr/abs-2302-13971} in the second stage. In contrast, MiniGPT-4 freezes both the vision encoder (BLIP-2's ViT + Q-Former \cite{DBLP:conf/icml/0008LSH23}) and the Vicuna LLM \cite{vicuna2023}, only training the linear layer. Following MiniGPT-4, Video-LLaMA \cite{DBLP:journals/corr/abs-2306-02858} handles videos by incorporating two branches for video and audio, each comprising a video/audio encoder and a BLIP-2-style Q-Former \cite{DBLP:conf/icml/0008LSH23}. PandaGPT \cite{DBLP:journals/corr/abs-2305-16355} leverages ImageBind \cite{DBLP:conf/cvpr/GirdharELSAJM23} to encode vision/text/audio/depth/thermal/IMU data, feeding them to the Vicuna model \cite{vicuna2023} also through a linear layer. PandaGPT diverges from MiniGPT-4 by using LoRA \cite{DBLP:conf/iclr/HuSWALWWC22} to train Vicuna alongside the linear layer.

\subsection{Supplementary Related Work}
Due to page limitations, we were unable to provide citations for all the mentioned models in the main text. Here is a list of supplementary related works:\\

\noindent Computer Vision:
\begin{itemize}
    \item Mask R-CNN \cite{DBLP:conf/iccv/HeGDG17}
    \item YOLOv8 \cite{DBLP:conf/cvpr/RedmonDGF16}
    \item EfficientNet \cite{DBLP:conf/icml/TanL19}
    \item UNet \cite{DBLP:conf/miccai/RonnebergerFB15}
    \item VQ-GAN \cite{DBLP:conf/cvpr/EsserRO21}
    \item MoVQ \cite{DBLP:conf/nips/ZhengVCP22}
    \item SigLIP \cite{DBLP:conf/iccv/ZhaiM0B23}
    \item ViLD \cite{DBLP:conf/iclr/GuLKC22}
    \item MDETR \cite{DBLP:conf/iccv/KamathSLSMC21}
    \item HLSM object detector \cite{DBLP:conf/corl/BlukisPFGA21}
    \item ResNet \cite{DBLP:conf/cvpr/HeZRS16}
    \item ViT, ViT-B, ViT-L \cite{DBLP:conf/iclr/DosovitskiyB0WZ21, DBLP:conf/corl/RadosavovicXJAM22}
    \item MAE \cite{DBLP:conf/cvpr/HeCXLDG22}
    \item ViT-22B \cite{DBLP:journals/corr/abs-2302-05442}
    \item EVA ViT-G/14 \cite{DBLP:conf/cvpr/FangWXSWW0WC23}
    \item OWL-ViT \cite{DBLP:journals/corr/abs-2205-06230}
    \item ConvNeXt \cite{DBLP:conf/cvpr/0003MWFDX22}
    \item SAM \cite{DBLP:journals/corr/abs-2304-02643}
    \item Imagen video \cite{DBLP:journals/corr/abs-2210-02303}
    \item C-ViViT \cite{DBLP:conf/iclr/VillegasBKM0SCK23}
    \item OSRT \cite{DBLP:conf/nips/SajjadiDMSPLGGK22}
    \item PointNet \cite{DBLP:conf/cvpr/QiSMG17}
    \item PointNet++ \cite{DBLP:conf/nips/QiYSG17}
    \item 3D-CLR \cite{DBLP:conf/cvpr/HongLDCTG23}
    \item ConceptFusion \cite{DBLP:conf/rss/JatavallabhulaK23}
\end{itemize}

\vspace{1em}
\noindent Natural Language Processing, LLM, VLM:
\begin{itemize}
    \item Long Short-Term Memory (LSTM) \cite{DBLP:journals/neco/HochreiterS97}
    \item Transformer \cite{DBLP:conf/nips/VaswaniSPUJGKP17}
    \item Sentence-BERT / Sent.-BERT \cite{DBLP:conf/emnlp/ReimersG19}
    \item DistilBERT \cite{DBLP:journals/corr/abs-1910-01108}
    \item T5-small, T5-base, T5-XXL \cite{DBLP:journals/jmlr/RaffelSRLNMZLL20}
    \item LLaMA \cite{DBLP:journals/corr/abs-2302-13971}
    \item LLaMA 2 \cite{DBLP:journals/corr/abs-2307-09288}
    \item Vicuna-V1.5 13B \cite{vicuna2023}
    \item MPT \cite{MosaicML2023Introducing}
    \item FLAN \cite{DBLP:conf/iclr/WeiBZGYLDDL22}
    \item Flamingo \cite{DBLP:conf/nips/AlayracDLMBHLMM22, DBLP:journals/corr/abs-2308-01390}
    \item Pythia \cite{DBLP:conf/icml/BidermanSABOHKP23}
    \item Phi-2 \cite{javaheripi2023phi}
    \item GPT-NeoX \cite{DBLP:journals/corr/abs-2204-06745}
    \item Codex \cite{DBLP:journals/corr/abs-2107-03374}
    \item InstructGPT \cite{DBLP:journals/corr/abs-2305-16355}
    \item ChatGPT \cite{chatgpt-openai}
    \item GPT-2 \cite{radford2019language} 
    \item GPT-3 \cite{brown2020language}
    \item GPT-4 \cite{DBLP:journals/corr/abs-2303-08774}
    \item PaLI-X \cite{DBLP:journals/corr/abs-2305-18565}
    \item Prismatic-7B VLM \cite{DBLP:conf/icml/Karamcheti0BLKS24}
    \item PaliGemma \cite{DBLP:journals/corr/abs-2407-07726}
    \item PaliGemma 2 \cite{DBLP:journals/corr/abs-2412-03555}
    \item Chameleon \cite{DBLP:journals/corr/abs-2405-09818}
    \item Emu3 \cite{DBLP:journals/corr/abs-2409-18869}
    \item LWM-Chat-1M \cite{DBLP:conf/iclr/0055YZA25}
    \item LLaMA-Adapter \cite{DBLP:journals/corr/abs-2303-16199}
\end{itemize}

\vspace{1em}
\noindent Action Modules:
\begin{itemize}
    \item LingUNet \cite{DBLP:conf/emnlp/MisraBBNSA18}
    \item Latent Motor Plans (LMP) \cite{DBLP:conf/corl/LynchKXKTLS19}
    \item STORM \cite{DBLP:conf/corl/BhardwajSMRFRB21}
    \item PerceiverIO \cite{DBLP:conf/iclr/JaegleBADIDKZBS22}
    \item DD-PPO \cite{DBLP:conf/iclr/WijmansKMLEPSB20}
    \item ScaleDP \cite{DBLP:journals/corr/abs-2409-14411}
    \item Symbol-tuning \cite{DBLP:journals/corr/abs-2305-08298}
    \item DiT \cite{DBLP:conf/iccv/PeeblesX23}
\end{itemize}

\noindent Others:
\begin{itemize}
    \item DDPM \cite{DBLP:conf/nips/HoJA20}
    \item DDIM \cite{DBLP:conf/iclr/0011SKKEP21}
    \item Ravens \cite{DBLP:conf/corl/ZengFTWCAAKDSL20}
    \item RLBench \cite{DBLP:journals/ral/JamesMAD20}
    \item Something V2: Something-something V2 \cite{DBLP:conf/iccv/GoyalKMMWKHFYMH17}
\end{itemize}

\vspace{1em}
\noindent Abbreviations:
\begin{itemize}
    \item TFM: Transformer
    \item MCIL: multicontext imitation learning
    \item MLE: maximum likelihood estimation
    \item MPC: model predictive control
    \item CVAE: conditional variational autoencoder
    \item EDR: Everyday Robots
\end{itemize}

\subsection{Additional VLA-Related Work}

We include additional VLA-related work that could not be included in the main text due to the page limit.

\subsubsection{Components of VLA: Pretraining}

Value-Implicit Pre-training (VIP) \cite{DBLP:conf/iclr/MaSJBK023} capitalizes on the temporal sequences present in videos, distinguishing itself from R3M \cite{DBLP:conf/corl/NairRKF022} by attracting both the initial and target frames while simultaneously repelling successive frames. This objective aims to capture long-range temporal relationships and uphold local smoothness. While their own experiments demonstrate superior performance compared to R3M in specific tasks, subsequent research presents conflicting findings through more comprehensive evaluations \cite{DBLP:conf/nips/MajumdarYAMCSJB23}.

SpawnNet \cite{DBLP:journals/corr/abs-2307-03567} utilizes a two-stream architecture incorporating adapter layers to fuse features from both a pretrained vision encoder and features learned from scratch. This innovative approach eliminates the necessity of training the pretrained vision encoders while surpassing the performance of parameter-efficient finetuning (PEFT) methods, as evidenced by experimental results in robot manipulation tasks.

Holo-Dex \cite{DBLP:conf/icra/ArunachalamGCP23} proposes learning low-dimensional representations for visual policies using self-supervised learning (SSL). In Dobb$\cdot$E \cite{DBLP:journals/corr/abs-2311-16098}, SSL is also explored for 3D visual inputs, through a method termed ``Home Pretrained Representations,'' using data collected from 22 homes in NYC. The authors extend SSL-pretrained representations beyond visual inputs. AuRL \cite{DBLP:conf/corl/ThankarajP23} introduces learning dynamic behaviors from audio, an often-overlooked yet important information source. Additionally, visual inputs were found to be insufficient for multi-fingered robotic hands, leading to T-Dex \cite{DBLP:conf/corl/GuzeyECP23}, a tactile-based dexterous policy.

\subsubsection{Components of VLA: World Models}

Masked World Model (MWM) \cite{DBLP:conf/corl/SeoHLLJLA22} innovates by modifying the vision encoder of DreamerV2 to a hybrid composition of convolutional neural network and vision Transformer. In training this novel Convolutional-ViT vision encoder, MWM draws inspiration from the approach proposed in MAE \cite{DBLP:conf/cvpr/HeCXLDG22}. Through the incorporation of an auxiliary reward prediction loss, the resulting learned latent dynamics model exhibits a notable performance improvement across various visual robotic tasks.

SWIM \cite{DBLP:conf/rss/MendoncaBP23} advocates for the use of human videos in training a world model due to the availability of large-scale human-centric data. However, acknowledging the substantial gap between human and robot data, SWIM addresses this disparity by grounding actions in visual affordance maps. This approach involves inferring target poses based on the affordance maps, facilitating an effective transfer of knowledge from human data to enhance robot control.

Iso-Dream \cite{DBLP:conf/nips/Pan0WY22} introduces two key enhancements to the Dreamer framework. The first enhancement focuses on optimizing inverse dynamics by decoupling controllable and noncontrollable dynamics. The second enhancement involves optimizing agent behavior based on the decoupled latent imaginations. This approach is particularly advantageous, as the noncontrollable state transition branch can be rolled out independently of actions, yielding benefits for long-horizon decision-making processes.

Transformer-based World Model (TWM) \cite{DBLP:conf/iclr/RobineHUH23} shares the same motivation as Dreamer but adopts a different approach by constructing a world model based on the Transformer-XL architecture \cite{DBLP:conf/acl/DaiYYCLS19}. This Transformer-based world model trains a model-free agent in latent imagination by generating new trajectories. Additionally, TWM suggests a modification to the balanced KL divergence loss from DreamerV2 and introduces a novel thresholded entropy loss tailored for the advantage actor-critic framework.

SceneScript \cite{DBLP:journals/corr/abs-2403-13064} introduces a set of structured language commands that can define an entire scene, specifying the layout and objects. This language-based scene representation distinguishes itself from previous methods---such as meshes, voxel grids, point clouds, or radiance fields---by generating scenes in an autoregressive, token-based fashion. It achieves competitive results against state-of-the-art methods in layout estimation and object detection.

\subsubsection{Components of VLA: Imitation Learning}

BeT \cite{DBLP:conf/nips/ShafiullahCAP22}, an imitation learning variant of the Trajectory Transformer, addresses the noisy and multimodal nature of non-expert human demonstration data by using k-means-based action discretization coupled with continuous action correction. C-BeT \cite{DBLP:conf/iclr/CuiWSP23} extends BeT by adding a target frame or demonstration as the goal specification. VQ-BeT \cite{DBLP:conf/icml/00010EKSP24} further improves BeT by incorporating vector quantization instead of k-means, enabling better modeling of long-range actions.

\subsubsection{Subsequent Surveys on VLA}

Since the initial release of this survey was made public, the field of embodied AI has experienced a proliferation of VLA models, leading to the emergence of several related surveys \cite{DBLP:journals/corr/abs-2505-04769, DBLP:journals/corr/abs-2507-01925, DBLP:journals/corr/abs-2507-10672, DBLP:journals/corr/abs-2509-19012, DBLP:journals/corr/abs-2509-23121, DBLP:journals/access/KawaharazukaOYPZ25, DBLP:journals/corr/abs-2510-24795, DBLP:journals/corr/abs-2511-05936, DBLP:journals/corr/abs-2512-11362, pine2025rlvla, DBLP:journals/corr/abs-2512-16760}.

\subsubsection{Latest Developments of VLA}

LLaRA \cite{DBLP:conf/iclr/0109M0KJSRBCLR25} introduces a data-centric method that uses a Vision-Language Model to automatically relabel existing robot demonstration videos with rich, descriptive language, significantly boosting the performance of downstream vision-language policies.

$\pi_0$ \cite{DBLP:journals/corr/abs-2410-24164} undergoes further enhancements, including FAST \cite{DBLP:journals/corr/abs-2501-09747} and $\pi_{0.5}$ \cite{DBLP:journals/corr/abs-2504-16054}. FAST is a novel robot action tokenization method designed to prioritize dexterous skills. $\pi_{0.5}$ pushes the boundaries of in-the-wild generalization by incorporating co-training with diverse data sources.

Mobility VLA \cite{DBLP:conf/corl/XuCFJZL0SRS0HHF24} proposes a hierarchical policy that integrates the commonsense reasoning capabilities of VLMs with a topological-graph-based low-level navigation policy.

While many VLAs focus primarily on robotic arms, recent advancements have extended their application to more complex embodiments, including humanoid and quadruped robots.

\vspace{1em}
\noindent Humanoid Robot:
\begin{itemize}
    \item GR00T N1 \cite{DBLP:journals/corr/abs-2503-14734}
    \item Humanoid-VLA \cite{DBLP:journals/corr/abs-2502-14795}
\end{itemize}

\noindent Quadruped Robot:
\begin{itemize}
    \item QUAR-VLA \cite{DBLP:conf/eccv/DingZZSZHYW24}
    \item QUART-Online \cite{DBLP:journals/corr/abs-2412-15576}
    \item MoRE \cite{DBLP:journals/corr/abs-2503-08007}
\end{itemize}    

\noindent Dexterous Hand:
\begin{itemize}
    \item DexGraspVLA \cite{DBLP:journals/corr/abs-2502-20900}
    \item DexGrasp-VLA \cite{DBLP:journals/corr/abs-2511-00139}
\end{itemize}

\noindent Autonomous Vehicle:
\begin{itemize}
    \item Alpamayo-R1 \cite{DBLP:journals/corr/abs-2511-00088}
    \item CoVLA \cite{DBLP:conf/wacv/AraiMSWY0Y25}
    \item DriveVLA-W0 \cite{DBLP:journals/corr/abs-2510-12796}
    \item EMMA \cite{DBLP:journals/tmlr/HwangXLHJCHHCSZGAT25}
\end{itemize}

\vspace{1em}
Additionally, some VLAs are designed to function as virtual assistants capable of taking actions within operating systems, video games, and other environments. Notable examples include:
\begin{itemize}
    \item CombatVLA \cite{DBLP:journals/corr/abs-2503-09527}
    \item JARVIS-VLA \cite{DBLP:conf/acl/LiWH0L25}
\end{itemize}

\begin{figure*}[t]
    \centering
    \includegraphics[width=1.0\textwidth, trim=0 0.4in 0 0.3in, clip]{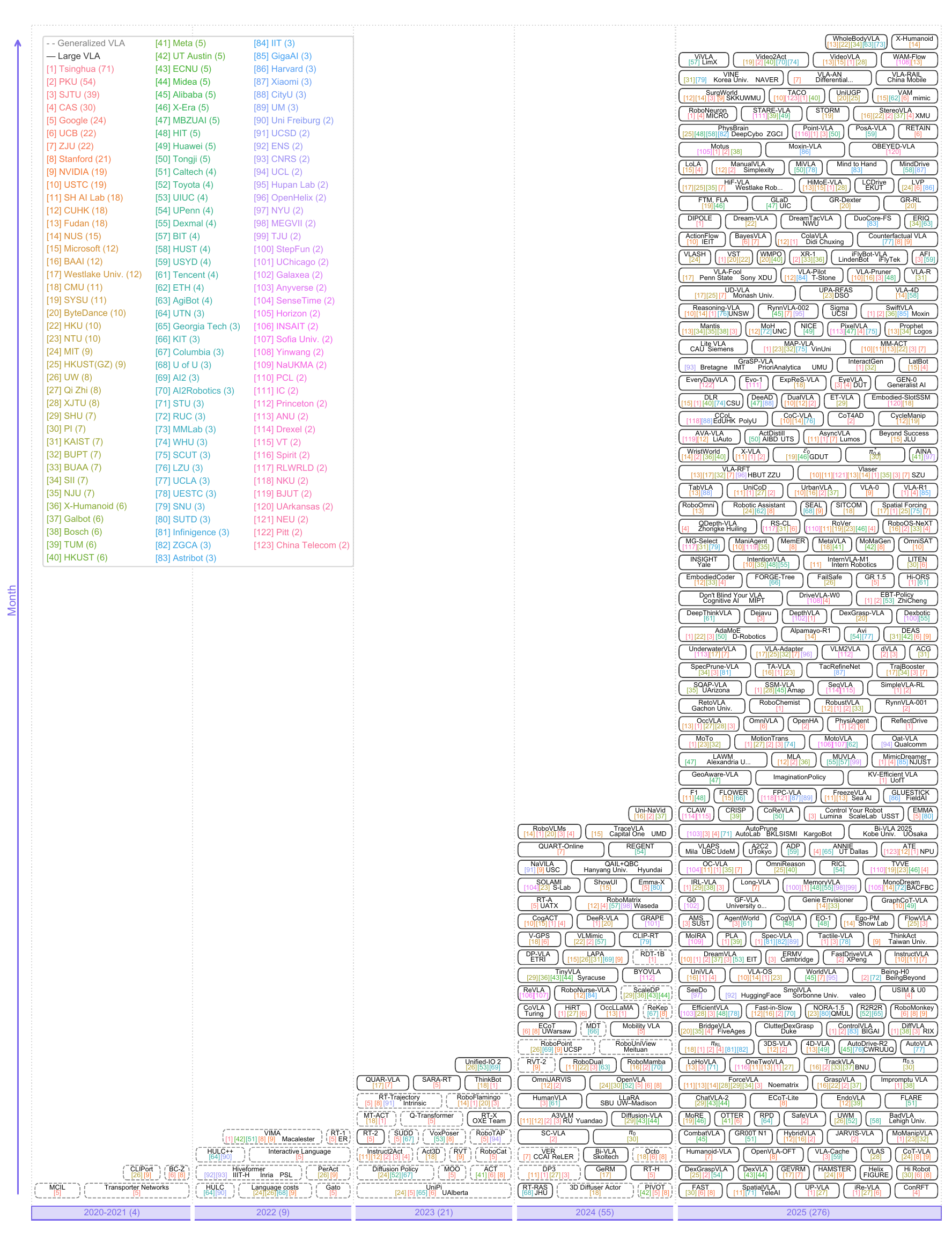}%
    \caption{Timeline of VLA models from 2020 to 2025. Bracketed numbers indicate the publication count for the corresponding year or institute.}
    \label{fig:vla_timeline}
\end{figure*}

\begin{figure*}[t]
    \centering
    \includegraphics[width=\textwidth, trim=2.25in 0.5in 1.5in 0, clip]{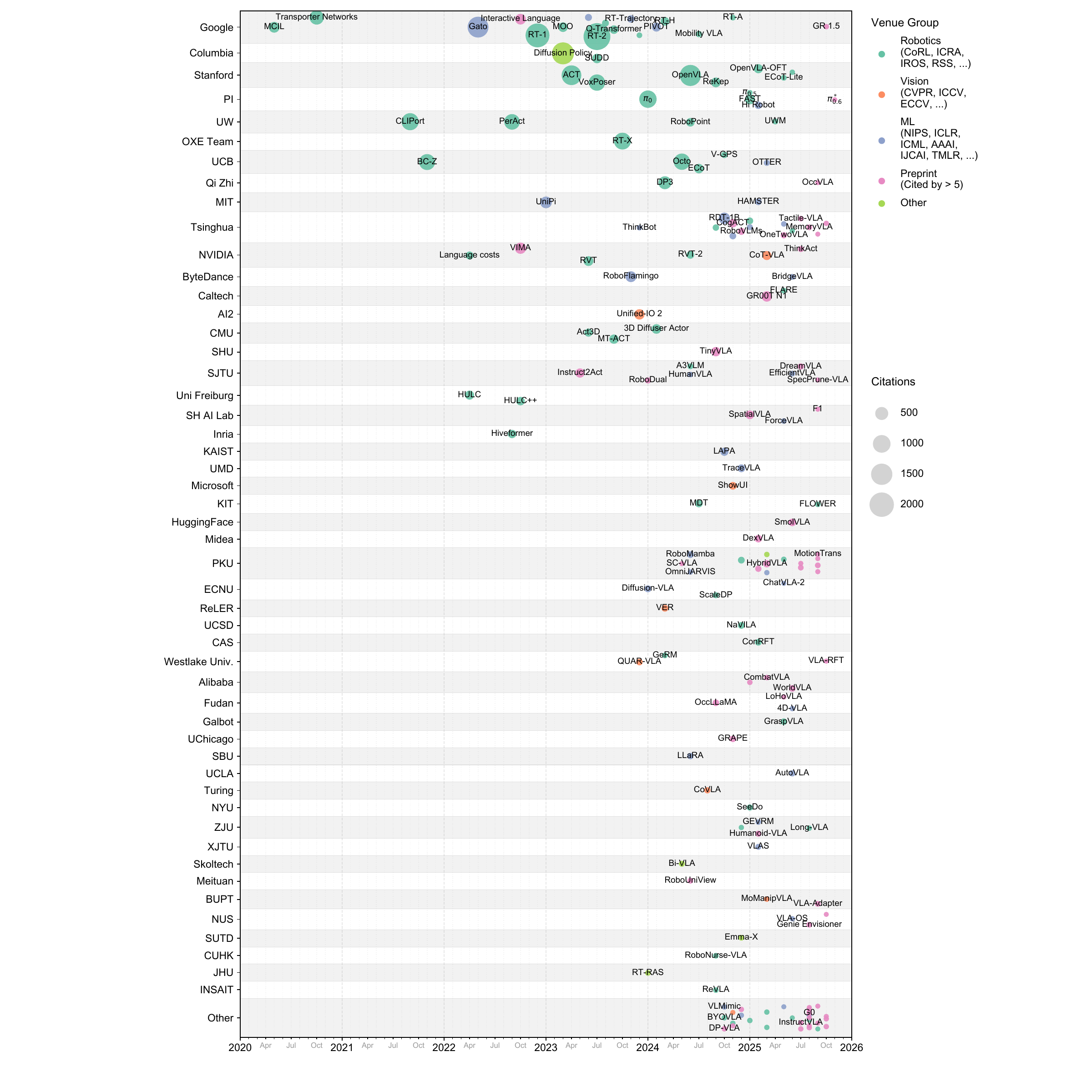}%
    \caption{Visualization of the VLA research landscape.}
    \label{fig:vla_bubble_landscape}
\end{figure*}

\begin{figure*}[t]
    \centering
    \subfloat[Papers per Year\label{fig:years_papers}]{%
        \includegraphics[width=0.49\textwidth, trim=0.6in 0.5in 0 0.25in, clip]{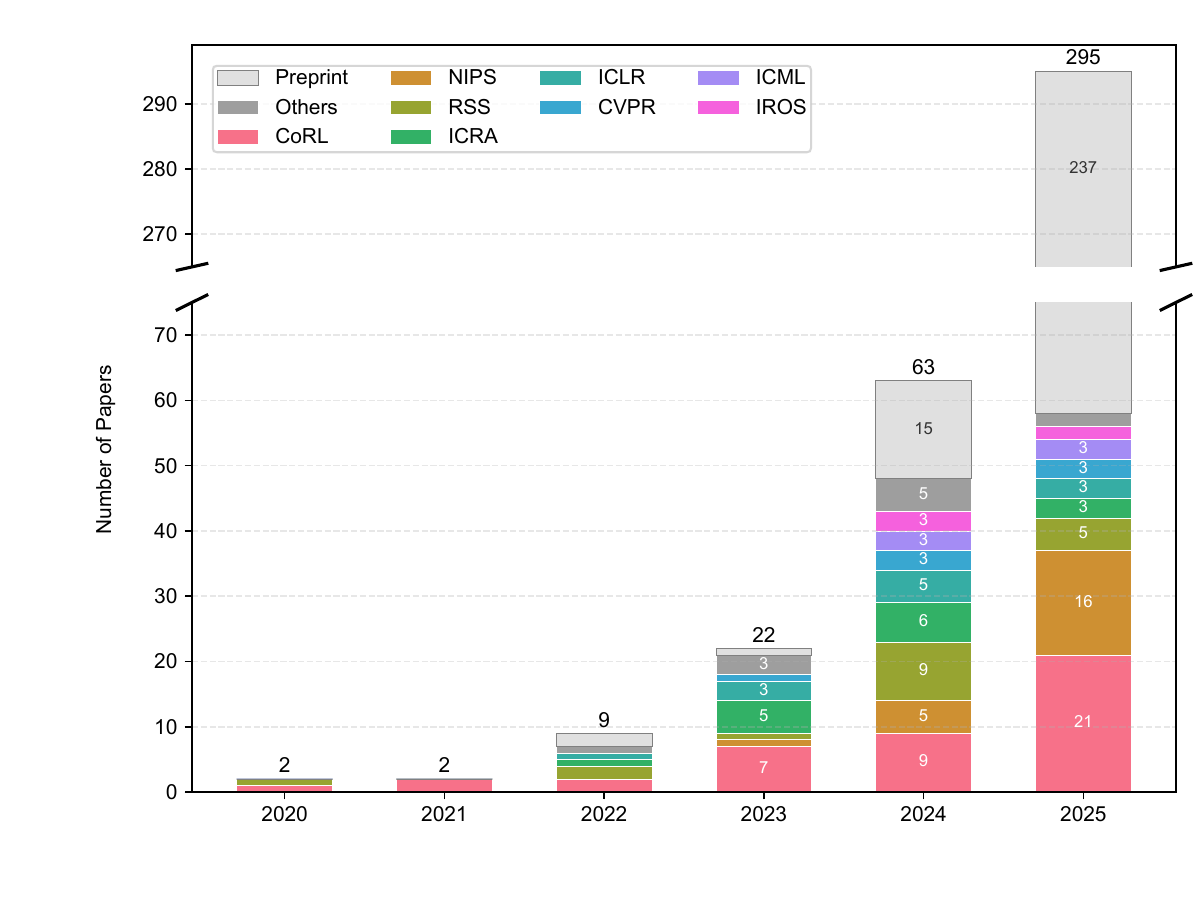}%
    }
    \hfill
    \subfloat[Individual Paper Citations per Year\label{fig:years_citations_dist}]{%
        \includegraphics[width=0.49\textwidth, trim=0.6in 0.5in 0 0.25in, clip]{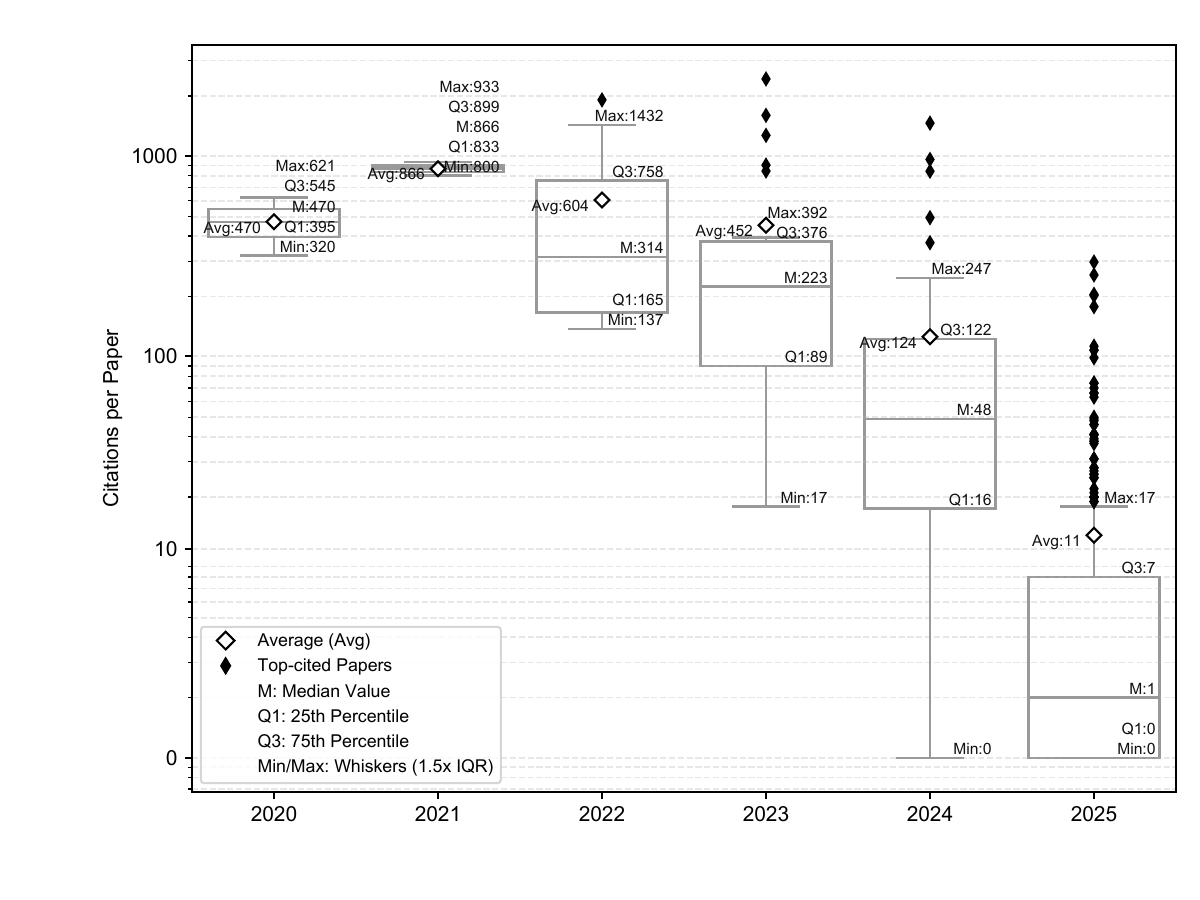}%
    }
    \hfill
    \subfloat[Institutes per Year\label{fig:years_institutes}]{%
        \includegraphics[width=0.49\textwidth, trim=0.6in 0.5in 0 0.25in, clip]{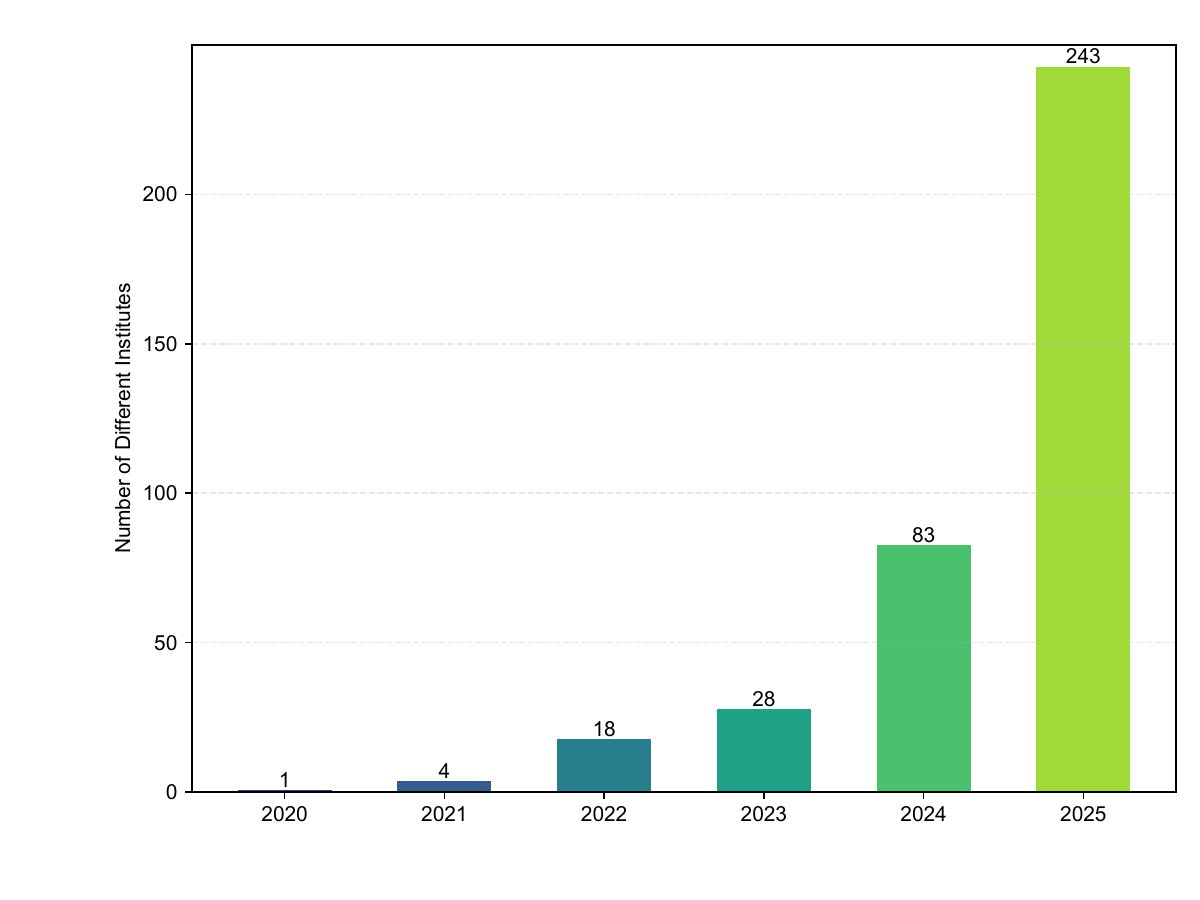}%
    }
    \hfill
    \subfloat[Total Citations per Year\label{fig:years_citations}]{%
        \includegraphics[width=0.49\textwidth, trim=0.6in 0.5in 0 0.25in, clip]{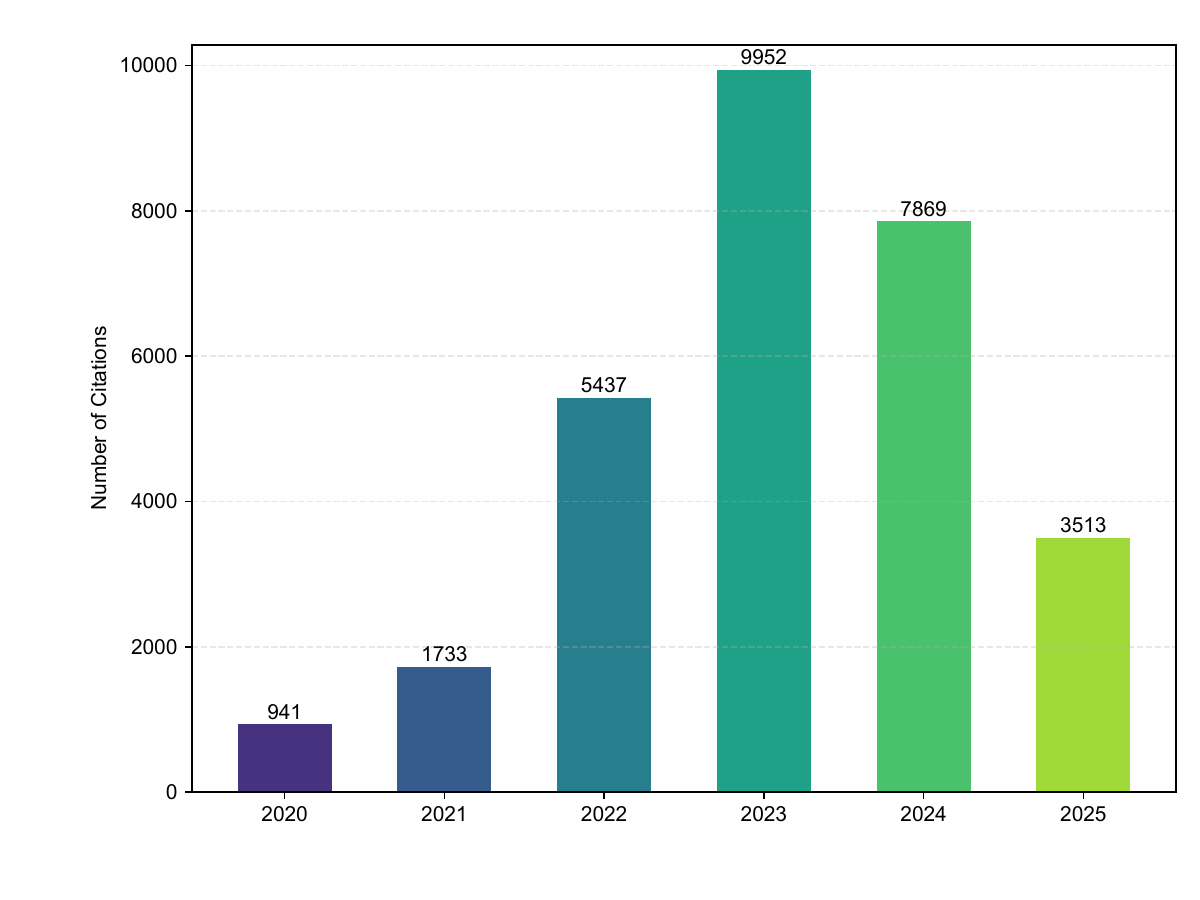}%
    }
    \hfill
    \subfloat[Collaboration Heatmap\label{fig:collaboration_heatmap}]{%
        \includegraphics[width=0.7\textwidth, trim=0 1.15in 0 0.5in, clip]{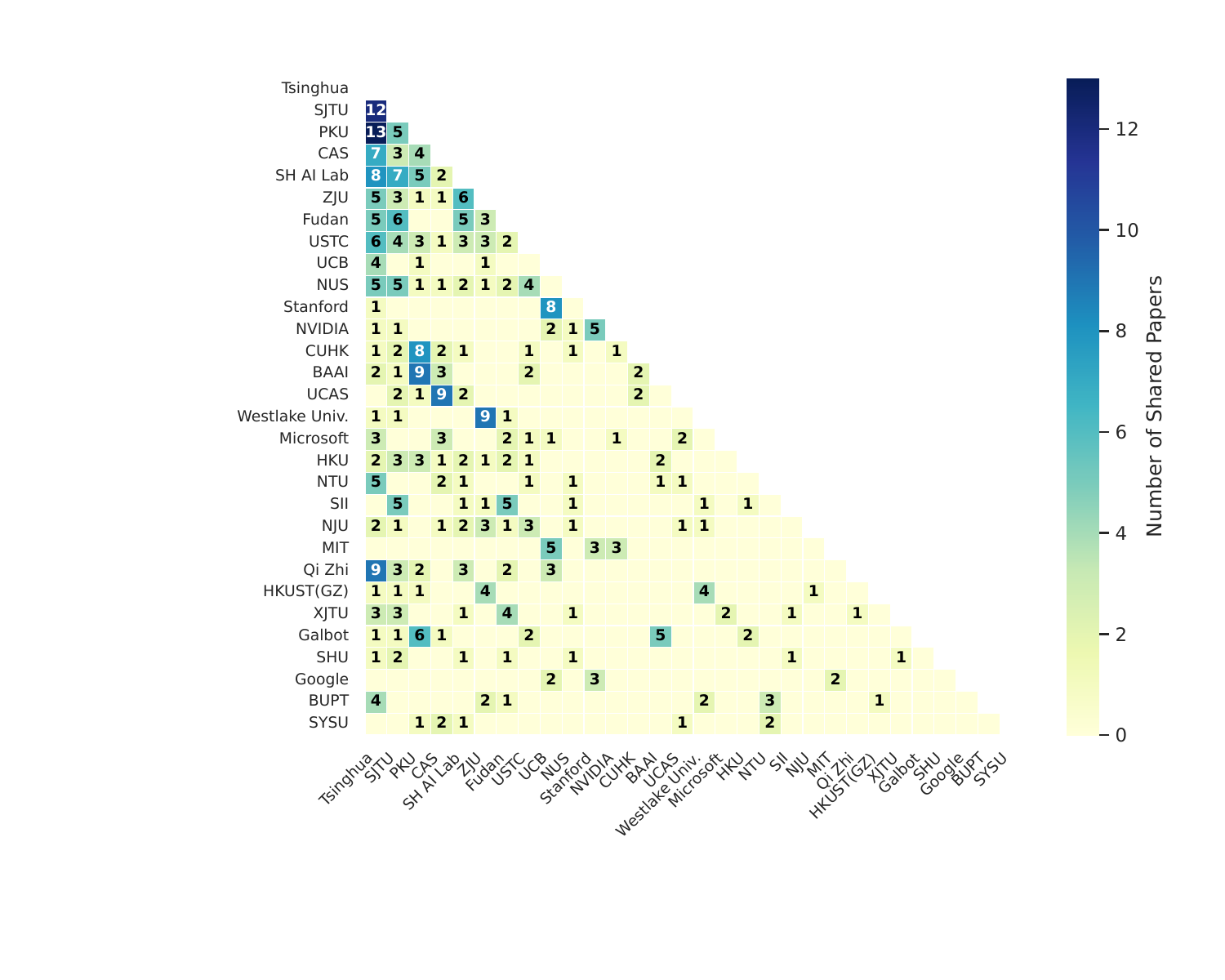}%
    }

    \caption{Quantitative analysis of VLA development trends.}
    \label{fig:vla_years_comparison}
\end{figure*}

\begin{figure*}[t]
    \centering
    \subfloat[Number of Papers per Institute\label{fig:inst_papers}]{%
        \includegraphics[width=1\textwidth, trim=0.9in 0.45in 0 0.25in, clip]{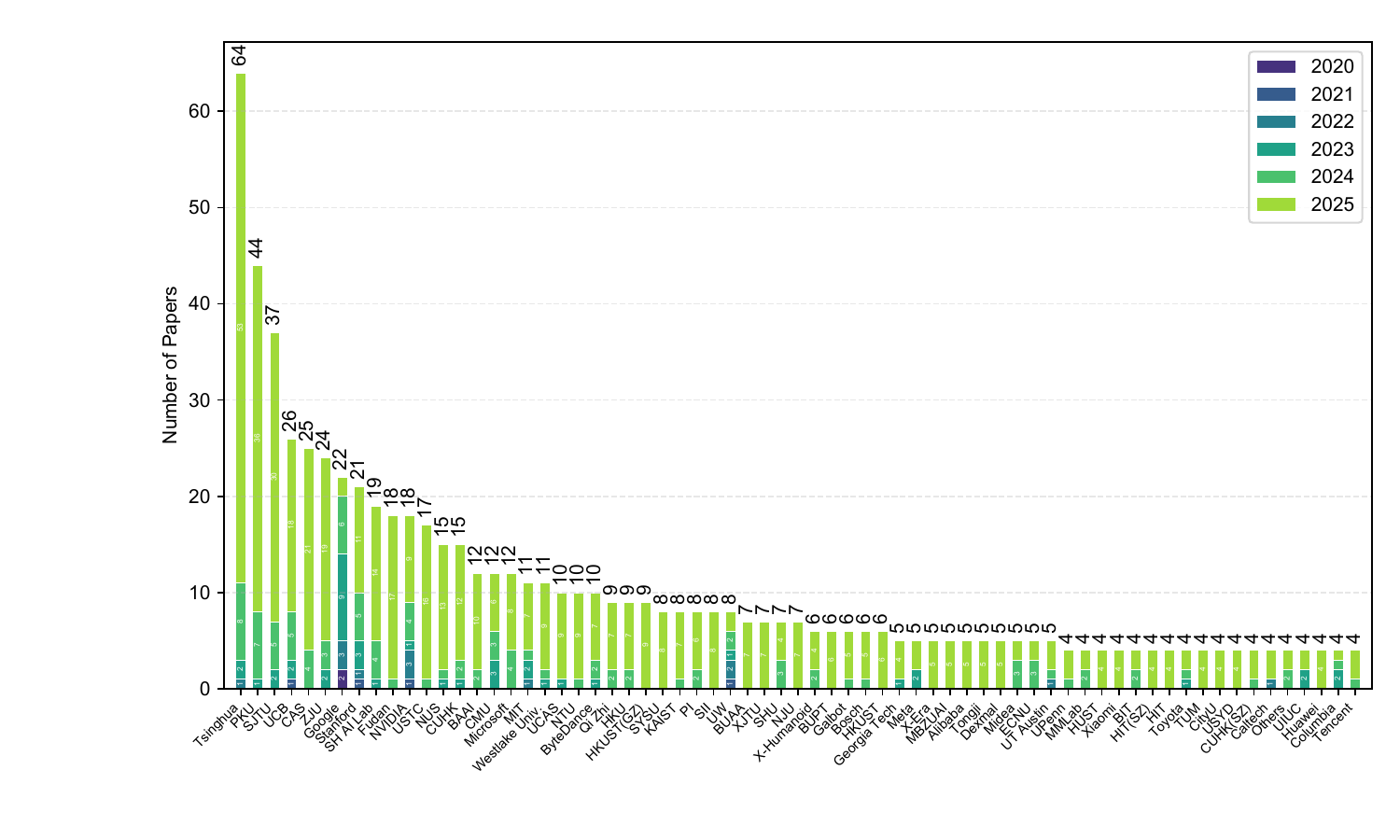}%
    }
    \
    \subfloat[Number of Citations per Institute\label{fig:inst_citations}]{%
        \includegraphics[width=1\textwidth, trim=0.9in 0.45in 0 0.25in, clip]{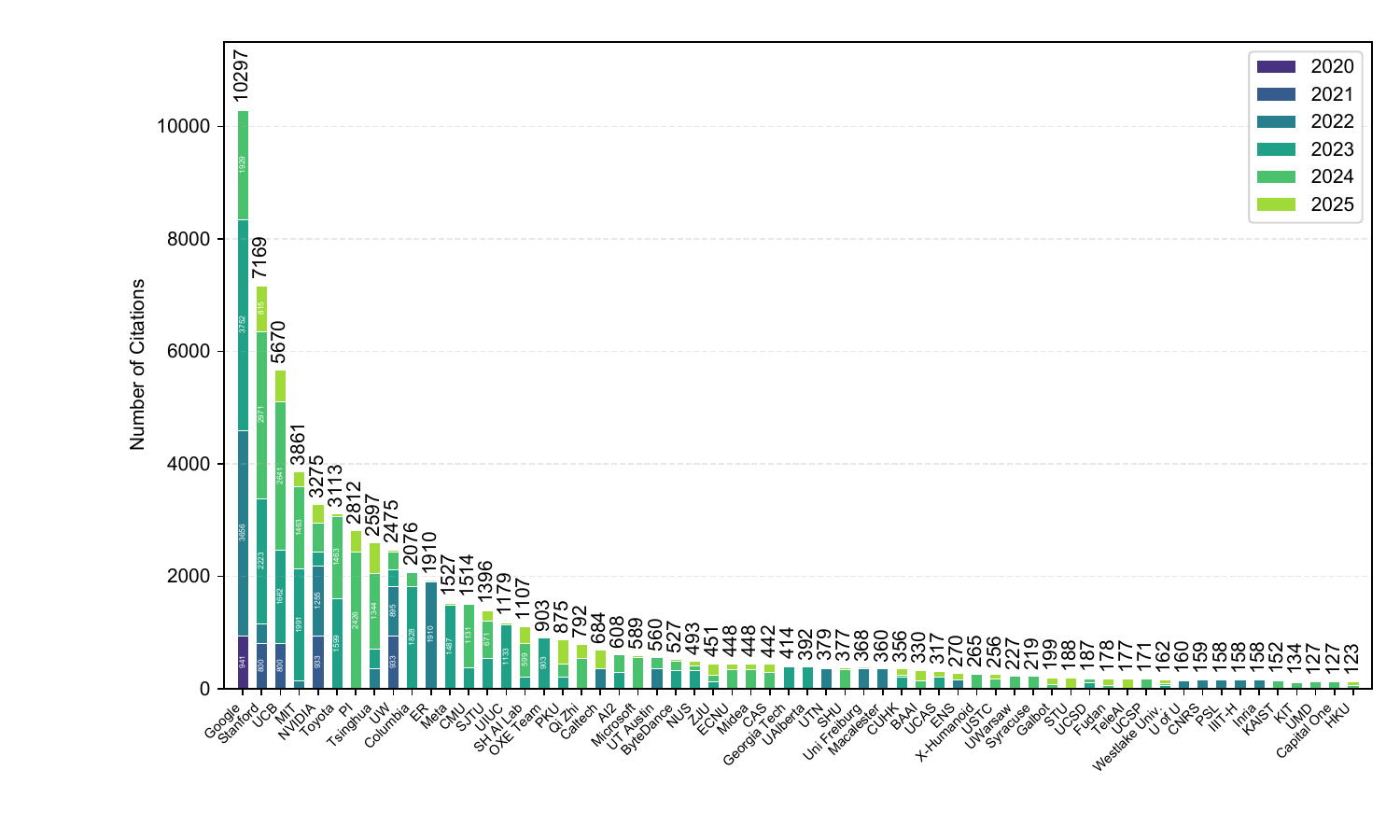}%
    }
    
    \caption{VLA research output and impact by institution.}
    \label{fig:vla_institutes_comparison}
\end{figure*}

\vspace{1em}
We utilize various charts to visualize key aspects of VLA developments from 2020 to 2025, including Figure~\ref{fig:vla_timeline}, Figure~\ref{fig:vla_bubble_landscape}, Figure~\ref{fig:vla_years_comparison}, and Figure~\ref{fig:vla_institutes_comparison}. To supplement the VLAs discussed in the main text, we employed a hybrid approach combining automated scripting and manual searching to retrieve VLA-related papers published between January 2020 and December 2025. We queried the keywords ``VLA'', ``Vision-language-action'', and ``Vision language action'', filtering false positives based on their relevance to ``embodied AI'' and ``robotics''. This pipeline yielded approximately 400 VLA-related papers. Acknowledging the potential for automated errors, we welcome feedback and requests for corrections regarding the included data.

\vspace{1em}
\noindent Abbreviations of institutes appearing in Figure~\ref{fig:vla_timeline}:
\begin{itemize}
    \item \textbf{AI2}: Allen Institute for AI
    \item \textbf{AIBD}: Advanced Institute of Big Data
    \item \textbf{ANU}: Australian National University
    \item \textbf{AgiBot}: AgiBot Research
    \item \textbf{Alexandria Univ.}: Alexandria University
    \item \textbf{Alibaba}: Alibaba Group, DAMO Academy
    \item \textbf{Anyverse}: Anyverse Intelligence
    \item \textbf{BAAI}: Beijing Academy of Artificial Intelligence
    \item \textbf{BACFBC}: Innovation Center for Future Blockchain and Privacy Computing, Beijing
    \item \textbf{BIGAI}: State Key Lab of General Artiﬁcial Intelligence
    \item \textbf{BIT}: Beijing Institute of Technology
    \item \textbf{BJUT}: Beijing University of Technology
    \item \textbf{BKLSISMI}: Beijing Key Laboratory of Super Intelligent Security of Multi-Modal Information
    \item \textbf{BNU}: Beijing Normal University
    \item \textbf{BUAA}: Beihang University
    \item \textbf{BUPT}: Beijing University of Posts and Telecommunications
    \item \textbf{Bosch}: Bosch Center for Artificial Intelligence, Bosch Corporate Research, Bosch Mobility Solutions, Bosch Research
    \item \textbf{Bretagne}: Bretagne INP - ENIB
    \item \textbf{ByteDance}: ByteDance Research, ByteDance Seed
    \item \textbf{CAS}: Chinese Academy of Sciences
    \item \textbf{CAU}: Clark Atlanta University
    \item \textbf{CMU}: Carnegie Mellon University
    \item \textbf{CNRS}: CNRS IRL 2010 CROSSING
    \item \textbf{CSU}: Central South University
    \item \textbf{CUHK}: The Chinese University of Hong Kong
    \item \textbf{CUHK(SZ)}: The Chinese University of Hong Kong (Shenzhen)
    \item \textbf{CWRU}: Case Western Reserve University
    \item \textbf{Caltech}: California Institute of Technology
    \item \textbf{Cambridge}: Cambridge University
    \item \textbf{China Mobile}: China Mobile Information Technology Co., Ltd.,
    \item \textbf{China Telecom}: China Telecom
    \item \textbf{CityU}: City University of Hong Kong
    \item \textbf{Cognitive AI}: Cognitive AI Lab
    \item \textbf{Columbia}: Columbia University
    \item \textbf{DSO}: DSO National Laboratories
    \item \textbf{DUT}: Dalian University of Technology
    \item \textbf{Didi Chuxing}: Didi Chuxing
    \item \textbf{Drexel}: Drexel University
    \item \textbf{Duke}: Duke University
    \item \textbf{ECNU}: East China Normal University
    \item \textbf{EIT}: Eastern Institute of Technology
    \item \textbf{EKUT}: University of Tübingen
    \item \textbf{ENS}: École Normale Supérieure
    \item \textbf{ER}: Everyday Robots
    \item \textbf{ETH}: ETH Zurich
    \item \textbf{EdUHK}: The Education University of Hong Kong
    \item \textbf{Fudan}: Fudan University
    \item \textbf{GDUT}: Guangdong University of Technology
    \item \textbf{Gachon Univ.}: Gachon University
    \item \textbf{Galaxea}: Galaxea AI
    \item \textbf{Georgia Tech}: Georgia Institute of Technology
    \item \textbf{Google}: Google, Google DeepMind, Robotics at Google, Google Research, Brain Team, Gemini Robotics Team
    \item \textbf{HBUT}: Hebei University of Technology
    \item \textbf{HIT}: Harbin Institute of Technology
    \item \textbf{HIT(SZ)}: Harbin Institute of Technology (Shenzhen)
    \item \textbf{HKU}: The University of Hong Kong
    \item \textbf{HKUST}: Hong Kong University of Science and Technology
    \item \textbf{HKUST(GZ)}: Hong Kong University of Science and Technology (Guangzhou)
    \item \textbf{HUST}: Huazhong University of Science and Technology
    \item \textbf{Hanyang Univ.}: Hanyang University
    \item \textbf{Harvard}: Harvard University
    \item \textbf{Horizon}: Horizon Robotics
    \item \textbf{Huawei}: Huawei Technologies, Huawei Noah's Ark Lab
    \item \textbf{HuggingFace}: Hugging Face
    \item \textbf{Hyundai}: Hyundai Motor Company
    \item \textbf{IC}: Imperial College London
    \item \textbf{IEIT}: IEIT SYSTEMS Co., Ltd.
    \item \textbf{IIIT-H}: IIIT Hyderabad
    \item \textbf{IIT}: Istituto Italiano di Tecnologia
    \item \textbf{IMT}: IMT Atlantique
    \item \textbf{Infinigence}: Infinigence AI
    \item \textbf{JHU}: Johns Hopkins University
    \item \textbf{JLU}: Jilin University
    \item \textbf{KAIST}: KAIST AI
    \item \textbf{KIT}: Karlsruhe Institute of Technology
    \item \textbf{Kobe Univ.}: Kobe University
    \item \textbf{Korea Univ.}: Korea University
    \item \textbf{LZU}: Lanzhou University
    \item \textbf{Lehigh Univ.}: Lehigh University
    \item \textbf{LiAuto}: LiAuto Inc.
    \item \textbf{LimX}: LimX Dynamics
    \item \textbf{Logos}: Logos Robotics
    \item \textbf{Lumina}: Lumina Group
    \item \textbf{Lumos}: Lumos Robotics
    \item \textbf{MBZUAI}: Mohamed bin Zayed University of Artificial Intelligence
    \item \textbf{MEGVII}: MEGVII Technology
    \item \textbf{MIPT}: IAI MIPT
    \item \textbf{MIT}: Massachusetts Institute of Technology
    \item \textbf{MMLab}: MMLab @ CUHK, MMLab @ HKU
    \item \textbf{Macalester}: Macalester College
    \item \textbf{Meta}: Meta AI, FAIR-MetaAI, Meta Reality Labs
    \item \textbf{Microsoft}: Microsoft Research, Microsoft Research Asia, Microsoft Zurich
    \item \textbf{Midea}: Midea Group
    \item \textbf{Mila}: Mila — Quebec AI Institute
    \item \textbf{Monash Univ.}: Monash University
    \item \textbf{Moxin}: Moxin (Huzhou) Technology Co., Ltd.
    \item \textbf{NAVER}: NAVER AI Lab
    \item \textbf{NEU}: Northeastern University
    \item \textbf{NJU}: Nanjing University
    \item \textbf{NJUPT}: Nanjing University of Posts and Telecommunications
    \item \textbf{NKU}: Nankai University
    \item \textbf{NPU}: Northwestern Polytechnical University
    \item \textbf{NTU}: Nanyang Technological University
    \item \textbf{NUS}: National University of Singapore
    \item \textbf{NVIDIA}: NVIDIA Research
    \item \textbf{NWU}: Northwestern University
    \item \textbf{NYU}: New York University
    \item \textbf{NaUKMA}: National University of Kyiv-Mohyla Academy
    \item \textbf{Noematrix}: Noematrix Intelligence
    \item \textbf{OXE Team}: Open X-Embodiment Collaboration
    \item \textbf{OpenHelix}: OpenHelix Team
    \item \textbf{PCL}: Peng Cheng Laboratory
    \item \textbf{PI}: Physical Intelligence
    \item \textbf{PKU}: Peking University
    \item \textbf{PSL}: PSL Research University
    \item \textbf{Penn State}: Pennsylvania State University
    \item \textbf{Pitt}: University of Pittsburgh
    \item \textbf{PolyU}: The Hong Kong Polytechnic University
    \item \textbf{Princeton}: Princeton University
    \item \textbf{QMUL}: Queen Mary University of London
    \item \textbf{Qi Zhi}: Shanghai Qi Zhi Institute, Shanghai Qizhi Institute
    \item \textbf{Qualcomm}: Qualcomm AI Research
    \item \textbf{RU}: Rutgers University
    \item \textbf{RUC}: Renmin University of China
    \item \textbf{SBU}: Stony Brook University
    \item \textbf{SCUT}: South China University of Technology
    \item \textbf{SH AI Lab}: Shanghai AI Laboratory
    \item \textbf{SHU}: Shanghai University
    \item \textbf{SII}: Shanghai Innovation Institute
    \item \textbf{SJTU}: Shanghai Jiao Tong University
    \item \textbf{SKKU}: Sungkyunkwan University
    \item \textbf{SNU}: Seoul National University
    \item \textbf{STU}: ShanghaiTech University
    \item \textbf{SUST}: Southern University of Science and Technology
    \item \textbf{SUTD}: Singapore University of Technology and Design
    \item \textbf{SYSU}: Sun Yat-sen University
    \item \textbf{SZU}: Shenzhen University
    \item \textbf{Sea AI}: Sea AI Lab
    \item \textbf{SenseTime}: SenseTime Research
    \item \textbf{Siemens}: Siemens Corporation
    \item \textbf{Simplexity}: Simplexity Robotics
    \item \textbf{Sofia Univ.}: Sofia University ``St. Kliment Ohridski''
    \item \textbf{Sony}: Sony Interactive Entertainment, Sony Research
    \item \textbf{Sorbonne Univ.}: Sorbonne University
    \item \textbf{Spirit}: Spirit AI
    \item \textbf{Stanford}: Stanford University
    \item \textbf{Syracuse}: Syracuse University
    \item \textbf{T-Stone}: T-Stone Robotics Institute
    \item \textbf{TJU}: Tianjin University
    \item \textbf{TUM}: Technical University of Munich
    \item \textbf{Taiwan Univ.}: National Taiwan University
    \item \textbf{Tencent}: Tencent Inc., Tencent Robotics X, Wechat AI
    \item \textbf{Tongji}: Tongji University
    \item \textbf{Toyota}: Toyota Research Institute
    \item \textbf{Tsinghua}: Tsinghua University
    \item \textbf{Turing}: Turing Inc.
    \item \textbf{U of U}: University of Utah
    \item \textbf{UATX}: The University of Austin at Texas
    \item \textbf{UAlberta}: University of Alberta
    \item \textbf{UArizona}: University of Arizona
    \item \textbf{UArkansas}: University of Arkansas
    \item \textbf{UBC}: The University of British Columbia
    \item \textbf{UCAS}: University of Chinese Academy of Sciences
    \item \textbf{UCB}: University of California, Berkeley
    \item \textbf{UCL}: University College London
    \item \textbf{UCLA}: University of California, Los Angeles
    \item \textbf{UCSD}: University of California San Diego
    \item \textbf{UCSI}: UCSI University
    \item \textbf{UCSP}: Universidad Católica San Pablo
    \item \textbf{UChicago}: University of Chicago
    \item \textbf{UESTC}: University of Electronic Science and Technology of China
    \item \textbf{UIC}: University of Illinois Chicago
    \item \textbf{UIUC}: University of Illinois Urbana-Champaign
    \item \textbf{UM}: University of Macau
    \item \textbf{UMD}: University of Maryland, College Park
    \item \textbf{UMU}: Umeå University
    \item \textbf{UNSW}: University of New South Wales
    \item \textbf{UOsaka}: The University of Osaka
    \item \textbf{UPenn}: University of Pennsylvania
    \item \textbf{UQ}: University of Queensland
    \item \textbf{USC}: University of Southern California
    \item \textbf{USST}: University of Shanghai for Science and Technology
    \item \textbf{USTC}: University of Science and Technology of China
    \item \textbf{USYD}: The University of Sydney
    \item \textbf{UT Austin}: The University of Texas at Austin
    \item \textbf{UT Dallas}: University of Texas at Dallas
    \item \textbf{UTN}: University of Technology Nuremberg
    \item \textbf{UTS}: University of Technology Sydney
    \item \textbf{UTokyo}: The University of Tokyo
    \item \textbf{UW}: University of Washington
    \item \textbf{UWarsaw}: University of Warsaw
    \item \textbf{UW–Madison}: University of Wisconsin-Madison
    \item \textbf{UdeM}: Universit\'e de Montr\'eal
    \item \textbf{Uni Freiburg}: University of Freiburg
    \item \textbf{UofT}: University of Toronto
    \item \textbf{VT}: Virginia Tech
    \item \textbf{VinUni}: VinUniversity
    \item \textbf{WHU}: Wuhan University
    \item \textbf{WMU}: Wenzhou Medical University
    \item \textbf{Waseda}: Waseda University
    \item \textbf{Westlake Univ.}: Westlake University
    \item \textbf{X-Era}: X-Era AI Lab
    \item \textbf{X-Humanoid}: Beijing Innovation Center of Humanoid Robotics
    \item \textbf{XDU}: Xidian University
    \item \textbf{XJTU}: Xi'an Jiaotong University
    \item \textbf{XMU}: Xiamen University Malaysia
    \item \textbf{XPeng}: XPeng Motors
    \item \textbf{Xiaomi}: Beijing Xiaomi Robot Technology Co., Ltd., Xiaomi EV, Xiaomi Robotics
    \item \textbf{Yale}: Yale University
    \item \textbf{Yinwang}: Yinwang Intelligent Technology Co., Ltd.
    \item \textbf{Yuandao}: Yuandao AI
    \item \textbf{ZGCA}: Zhongguancun Academy
    \item \textbf{ZGCI}: Zhongguancun Institute of Artificial Intelligence
    \item \textbf{ZJU}: Zhejiang University
    \item \textbf{ZZU}: Zhengzhou University
    \item \textbf{ZhiCheng}: ZhiCheng AI
    \item \textbf{Zhongke Huiling}: Beijing Zhongke Huiling Robot Technology Co
    \item \textbf{iFlyTek}: iFlyTek Research and Development Group
    \item \textbf{mimic}: mimic robotics
    \item \textbf{valeo}: valeo.ai
\end{itemize}

\vspace{1em}
\noindent List of VLAs appearing in Figure~\ref{fig:vla_timeline}:
\begin{itemize}
    \item 3D Diffuser Actor \cite{DBLP:conf/corl/KeGF24}
    \item 3DS-VLA \cite{vla_3dsvla_2025}
    \item 4D-VLA \cite{DBLP:journals/corr/abs-2506-22242}
    \item A2C2 \cite{DBLP:journals/corr/abs-2509-23224}
    \item A3VLM \cite{DBLP:conf/corl/0004CLZ0B0024}
    \item ACG \cite{DBLP:journals/corr/abs-2510-22201}
    \item ACT \cite{DBLP:conf/rss/ZhaoKLF23}
    \item Act3D \cite{DBLP:conf/corl/GervetXGF23}
    \item ActDistill \cite{DBLP:journals/corr/abs-2511-18082}
    \item ActionFlow \cite{DBLP:journals/corr/abs-2512-20276}
    \item AdaMoE \cite{DBLP:journals/corr/abs-2510-14300}
    \item ADP \cite{DBLP:journals/corr/abs-2509-22093}
    \item AFI \cite{DBLP:journals/corr/abs-2512-07472}
    \item AgentWorld \cite{DBLP:journals/corr/abs-2508-07770}
    \item AINA \cite{DBLP:journals/corr/abs-2511-16661}
    \item Alpamayo-R1 \cite{DBLP:journals/corr/abs-2511-00088}
    \item AMS \cite{DBLP:journals/corr/abs-2508-10259}
    \item ANNIE \cite{DBLP:journals/corr/abs-2509-03383}
    \item AsyncVLA \cite{DBLP:journals/corr/abs-2511-14148}
    \item ATE \cite{DBLP:journals/corr/abs-2509-02055}
    \item AutoDrive-R2 \cite{DBLP:journals/corr/abs-2509-01944}
    \item AutoPrune \cite{DBLP:journals/corr/abs-2509-23931}
    \item AutoVLA \cite{DBLP:journals/corr/abs-2506-13757}
    \item AVA-VLA \cite{DBLP:journals/corr/abs-2511-18960}
    \item Avi \cite{DBLP:journals/corr/abs-2510-21746}
    \item BadVLA \cite{DBLP:journals/corr/abs-2505-16640}
    \item BayesVLA \cite{DBLP:journals/corr/abs-2512-11218}
    \item BC-Z \cite{DBLP:conf/corl/JangIKKELLF21}
    \item Being-H0 \cite{DBLP:journals/corr/abs-2507-15597}
    \item Beyond Success \cite{DBLP:journals/corr/abs-2511-22555}
    \item Bi-VLA \cite{DBLP:conf/smc/GbagbeCAALT24}
    \item Bi-VLA 2025 \cite{DBLP:journals/corr/abs-2509-18865}
    \item BridgeVLA \cite{DBLP:journals/corr/abs-2506-07961}
    \item BYOVLA \cite{DBLP:conf/icra/HancockRM25}
    \item CCoL \cite{DBLP:journals/corr/abs-2511-14396}
    \item ChatVLA-2 \cite{DBLP:journals/corr/abs-2505-21906}
    \item CLAW \cite{DBLP:journals/corr/abs-2509-14143}
    \item CLIPort \cite{DBLP:conf/corl/ShridharMF21}
    \item CLIP-RT \cite{DBLP:journals/corr/abs-2411-00508}
    \item ClutterDexGrasp \cite{DBLP:journals/corr/abs-2506-14317}
    \item CoC-VLA \cite{DBLP:journals/corr/abs-2511-19914}
    \item CogACT \cite{DBLP:journals/corr/abs-2411-19650}
    \item CogVLA \cite{DBLP:journals/corr/abs-2508-21046}
    \item ColaVLA \cite{DBLP:journals/corr/abs-2512-22939}
    \item CombatVLA \cite{DBLP:journals/corr/abs-2503-09527}
    \item ConRFT \cite{DBLP:journals/corr/abs-2502-05450}
    \item ControlVLA \cite{DBLP:journals/corr/abs-2506-16211}
    \item Control Your Robot \cite{DBLP:journals/corr/abs-2509-23823}
    \item CoReVLA \cite{DBLP:journals/corr/abs-2509-15968}
    \item CoT4AD \cite{DBLP:journals/corr/abs-2511-22532}
    \item CoT-VLA \cite{DBLP:conf/cvpr/ZhaoLKFZWLMHFHL25}
    \item Counterfactual VLA \cite{DBLP:journals/corr/abs-2512-24426}
    \item CoVLA \cite{DBLP:conf/wacv/AraiMSWY0Y25}
    \item CRISP \cite{DBLP:journals/corr/abs-2509-06819}
    \item CycleManip \cite{DBLP:journals/corr/abs-2512-01022}
    \item DEAS \cite{DBLP:journals/corr/abs-2510-07730}
    \item DeeAD \cite{DBLP:journals/corr/abs-2511-20720}
    \item DeepThinkVLA \cite{DBLP:journals/corr/abs-2511-15669}
    \item DeeR-VLA \cite{DBLP:conf/nips/YueWKHWSF024}
    \item Dejavu \cite{vla_dejavu_2025}
    \item DepthVLA \cite{DBLP:journals/corr/abs-2510-13375}
    \item Dexbotic \cite{DBLP:journals/corr/abs-2510-23511}
    \item DexGrasp-VLA \cite{DBLP:journals/corr/abs-2511-00139}
    \item DexGraspVLA \cite{DBLP:journals/corr/abs-2502-20900}
    \item DexVLA \cite{DBLP:journals/corr/abs-2502-05855}
    \item Diffusion Policy \cite{DBLP:journals/ijrr/ChiXFCDBTS25}
    \item Diffusion-VLA \cite{DBLP:journals/corr/abs-2412-03293}
    \item DiffVLA \cite{DBLP:journals/corr/abs-2510-17148}
    \item DIPOLE \cite{vla_dipole_2025}
    \item DLR \cite{DBLP:journals/corr/abs-2511-19528}
    \item Don't Blind Your VLA \cite{DBLP:journals/corr/abs-2510-25616}
    \item DP3 \cite{DBLP:conf/rss/ZeZZHWX24}
    \item DP-VLA \cite{DBLP:journals/corr/abs-2410-15549}
    \item DreamTacVLA \cite{DBLP:journals/corr/abs-2512-23864}
    \item DreamVLA \cite{DBLP:journals/corr/abs-2507-04447}
    \item Dream-VLA \cite{DBLP:journals/corr/abs-2512-22615}
    \item DriveVLA-W0 \cite{DBLP:journals/corr/abs-2510-12796}
    \item DualVLA \cite{DBLP:journals/corr/abs-2511-22134}
    \item DuoCore-FS \cite{DBLP:journals/corr/abs-2512-20188}
    \item dVLA \cite{DBLP:journals/corr/abs-2509-25681}
    \item EBT-Policy \cite{DBLP:journals/corr/abs-2510-27545}
    \item ECoT \cite{DBLP:conf/corl/ZawalskiCPMFL24}
    \item ECoT-Lite \cite{DBLP:journals/corr/abs-2505-08243}
    \item EfficientVLA \cite{DBLP:journals/corr/abs-2506-10100}
    \item Ego-PM \cite{DBLP:journals/corr/abs-2508-19852}
    \item EmbodiedCoder \cite{DBLP:journals/corr/abs-2510-06207}
    \item Embodied-SlotSSM \cite{DBLP:journals/corr/abs-2511-11478}
    \item EMMA \cite{DBLP:journals/corr/abs-2509-22407}
    \item Emma-X \cite{DBLP:conf/acl/SunHDTTGP25}
    \item EndoVLA \cite{DBLP:journals/corr/abs-2505-15206}
    \item EO-1 \cite{DBLP:journals/corr/abs-2508-21112}
    \item ERIQ \cite{DBLP:journals/corr/abs-2512-24125}
    \item ERMV \cite{DBLP:journals/corr/abs-2507-17462}
    \item ET-VLA \cite{DBLP:journals/corr/abs-2511-01224}
    \item EveryDayVLA \cite{DBLP:journals/corr/abs-2511-05397}
    \item Evo-1 \cite{DBLP:journals/corr/abs-2511-04555}
    \item ExpReS-VLA \cite{DBLP:journals/corr/abs-2511-06202}
    \item EyeVLA \cite{DBLP:journals/corr/abs-2511-15279}
    \item F1 \cite{DBLP:journals/corr/abs-2509-06951}
    \item FailSafe \cite{DBLP:journals/corr/abs-2510-01642}
    \item FAST \cite{DBLP:journals/corr/abs-2501-09747}
    \item FastDriveVLA \cite{DBLP:journals/corr/abs-2507-23318}
    \item Fast-in-Slow \cite{vla_fastinslow_2025}
    \item FLARE \cite{DBLP:journals/corr/abs-2505-15659}
    \item FLOWER \cite{DBLP:journals/corr/abs-2509-04996}
    \item FlowVLA \cite{vla_flowvla_2025}
    \item ForceVLA \cite{DBLP:journals/corr/abs-2505-22159}
    \item FORGE-Tree \cite{DBLP:journals/corr/abs-2510-21744}
    \item FPC-VLA \cite{DBLP:journals/corr/abs-2509-04018}
    \item FreezeVLA \cite{DBLP:journals/corr/abs-2509-19870}
    \item FTM, FLA \cite{DBLP:journals/corr/abs-2512-02902}
    \item G0 \cite{DBLP:journals/corr/abs-2509-00576}
    \item Gato \cite{DBLP:conf/iclr/SudhakarNRLRC25}
    \item GEN-0 \cite{vla_gen0_2025}
    \item Genie Envisioner \cite{DBLP:journals/corr/abs-2508-05635}
    \item GeoAware-VLA \cite{DBLP:journals/corr/abs-2509-14117}
    \item GeRM \cite{DBLP:conf/iros/SongZDCLFW24}
    \item GEVRM \cite{DBLP:conf/iclr/ZhangDLPW25}
    \item GF-VLA \cite{DBLP:journals/corr/abs-2508-05342}
    \item GLaD \cite{DBLP:journals/corr/abs-2512-09619}
    \item GLUESTICK \cite{vla_gluestick_2025}
    \item GR00T N1 \cite{DBLP:journals/corr/abs-2503-14734}
    \item GR 1.5 \cite{DBLP:journals/corr/abs-2510-03342}
    \item GRAPE \cite{DBLP:journals/corr/abs-2411-19309}
    \item GraphCoT-VLA \cite{DBLP:journals/corr/abs-2508-07650}
    \item GraspVLA \cite{DBLP:journals/corr/abs-2505-03233}
    \item GraSP-VLA \cite{DBLP:journals/corr/abs-2511-04357}
    \item GR-Dexter \cite{DBLP:journals/corr/abs-2512-24210}
    \item GR-RL \cite{DBLP:journals/corr/abs-2512-01801}
    \item HAMSTER \cite{DBLP:conf/iclr/0038DZJMG0F00025}
    \item Helix \cite{vla_helix_2025}
    \item HiF-VLA \cite{DBLP:journals/corr/abs-2512-09928}
    \item HiMoE-VLA \cite{DBLP:journals/corr/abs-2512-05693}
    \item Hi-ORS \cite{DBLP:journals/corr/abs-2510-26406}
    \item Hi Robot \cite{DBLP:conf/icml/ShiIEKPVTWWFLDG25}
    \item HiRT \cite{DBLP:conf/corl/ZhangGCWHSC24}
    \item Hiveformer \cite{DBLP:conf/corl/GuhurCPTLS22}
    \item HULC \cite{DBLP:journals/ral/MeesHB22}
    \item HULC++ \cite{DBLP:conf/icra/MeesBB23}
    \item Humanoid-VLA \cite{DBLP:journals/corr/abs-2502-14795}
    \item HumanVLA \cite{DBLP:conf/nips/XuZ00L24}
    \item HybridVLA \cite{DBLP:journals/corr/abs-2503-10631}
    \item iFlyBot-VLA \cite{DBLP:journals/corr/abs-2511-01914}
    \item ImaginationPolicy \cite{DBLP:journals/corr/abs-2509-20841}
    \item Impromptu VLA \cite{DBLP:journals/corr/abs-2505-23757}
    \item INSIGHT \cite{DBLP:journals/corr/abs-2510-01389}
    \item Instruct2Act \cite{DBLP:journals/corr/abs-2305-11176}
    \item InstructVLA \cite{DBLP:journals/corr/abs-2507-17520}
    \item IntentionVLA \cite{DBLP:journals/corr/abs-2510-07778}
    \item InteractGen \cite{DBLP:journals/corr/abs-2512-00797}
    \item Interactive Language \cite{DBLP:journals/corr/abs-2210-06407}
    \item InternVLA-M1 \cite{DBLP:journals/corr/abs-2510-13778}
    \item iRe-VLA \cite{DBLP:conf/icra/GuoZCJWHC25}
    \item IRL-VLA \cite{DBLP:journals/corr/abs-2508-06571}
    \item JARVIS-VLA \cite{DBLP:conf/acl/LiWH0L25}
    \item KV-Efficient VLA \cite{vla_kvefficientvla_2025}
    \item Language costs \cite{DBLP:conf/rss/SharmaSBPH0AF22}
    \item LAPA \cite{DBLP:conf/iclr/YeJJJYPMTCLLL0Z25}
    \item LatBot \cite{DBLP:journals/corr/abs-2511-23034}
    \item LAWM \cite{DBLP:journals/corr/abs-2509-18428}
    \item LCDrive \cite{DBLP:journals/corr/abs-2512-10226}
    \item LITEN \cite{DBLP:journals/corr/abs-2510-19752}
    \item Lite VLA \cite{DBLP:journals/corr/abs-2511-05642}
    \item LLaRA \cite{DBLP:conf/iclr/0109M0KJSRBCLR25}
    \item LoHoVLA \cite{DBLP:journals/corr/abs-2506-00411}
    \item LoLA \cite{DBLP:journals/corr/abs-2512-20166}
    \item Long-VLA \cite{DBLP:journals/corr/abs-2508-19958}
    \item LVP \cite{DBLP:journals/corr/abs-2512-15840}
    \item ManiAgent \cite{DBLP:journals/corr/abs-2510-11660}
    \item Mantis \cite{DBLP:journals/corr/abs-2511-16175}
    \item ManualVLA \cite{DBLP:journals/corr/abs-2512-02013}
    \item MAP-VLA \cite{DBLP:journals/corr/abs-2511-09516}
    \item $\mathcal{E}_0$ \cite{DBLP:journals/corr/abs-2511-21542}
    \item MCIL \cite{DBLP:journals/ral/MeesHB22}
    \item MDT \cite{DBLP:conf/rss/ReussYWL24}
    \item MemER \cite{DBLP:journals/corr/abs-2510-20328}
    \item MemoryVLA \cite{DBLP:journals/corr/abs-2508-19236}
    \item MetaVLA \cite{DBLP:journals/corr/abs-2510-05580}
    \item MG-Select \cite{DBLP:journals/corr/abs-2510-05681}
    \item MimicDreamer \cite{DBLP:journals/corr/abs-2509-22199}
    \item MindDrive \cite{DBLP:journals/corr/abs-2512-13636}
    \item Mind to Hand \cite{DBLP:journals/corr/abs-2512-08580}
    \item MiVLA \cite{DBLP:journals/corr/abs-2512-15411}
    \item MLA \cite{DBLP:journals/corr/abs-2509-26642}
    \item MM-ACT \cite{DBLP:journals/corr/abs-2512-00975}
    \item Mobility VLA \cite{DBLP:conf/corl/XuCFJZL0SRS0HHF24}
    \item MoH \cite{DBLP:journals/corr/abs-2511-19433}
    \item MoIRA \cite{DBLP:journals/corr/abs-2507-01843}
    \item MoMaGen \cite{DBLP:journals/corr/abs-2510-18316}
    \item MoManipVLA \cite{DBLP:conf/cvpr/WuZX0Y25}
    \item MonoDream \cite{DBLP:journals/corr/abs-2508-02549}
    \item MOO \cite{DBLP:conf/corl/StoneXLGLVWKZXF23}
    \item MoRE \cite{DBLP:conf/icra/ZhaoSWTDCG25}
    \item MotionTrans \cite{DBLP:journals/corr/abs-2509-17759}
    \item MoTo \cite{DBLP:journals/corr/abs-2509-01658}
    \item MotoVLA \cite{DBLP:journals/corr/abs-2509-19958}
    \item Motus \cite{DBLP:journals/corr/abs-2512-13030}
    \item Moxin-VLA \cite{DBLP:journals/corr/abs-2512-22208}
    \item MT-ACT \cite{DBLP:conf/icra/BharadhwajVSGTK24}
    \item MUVLA \cite{DBLP:journals/corr/abs-2509-25966}
    \item NaVILA \cite{DBLP:journals/corr/abs-2412-04453}
    \item NICE \cite{DBLP:journals/corr/abs-2511-22777}
    \item NitroGen \cite{vla_nitrogen_2026}
    \item NORA-1.5 \cite{DBLP:journals/corr/abs-2511-14659}
    \item Oat-VLA \cite{DBLP:journals/corr/abs-2509-23655}
    \item OBEYED-VLA \cite{DBLP:journals/corr/abs-2512-22519}
    \item OccLLaMA \cite{DBLP:journals/corr/abs-2409-03272}
    \item OccVLA \cite{DBLP:journals/corr/abs-2509-05578}
    \item Octo \cite{DBLP:conf/rss/GhoshWPBMDHK0LT24}
    \item OC-VLA \cite{DBLP:journals/corr/abs-2508-13103}
    \item OmniJARVIS \cite{DBLP:conf/nips/WangCMLZL0LML24}
    \item OmniReason \cite{DBLP:journals/corr/abs-2509-00789}
    \item OmniSAT \cite{DBLP:journals/corr/abs-2510-09667}
    \item OmniVLA \cite{DBLP:journals/corr/abs-2509-19480}
    \item OneTwoVLA \cite{DBLP:journals/corr/abs-2505-11917}
    \item OpenHA \cite{DBLP:journals/corr/abs-2509-13347}
    \item OpenVLA \cite{DBLP:conf/corl/KimPKXB0RFSVKBT24}
    \item OpenVLA-OFT \cite{DBLP:journals/corr/abs-2502-19645}
    \item OTTER \cite{DBLP:conf/icml/HuangLFWMMGA25}
    \item PerAct \cite{DBLP:conf/corl/ShridharMF22}
    \item PhysBrain \cite{DBLP:journals/corr/abs-2512-16793}
    \item PhysiAgent \cite{DBLP:journals/corr/abs-2509-24524}
    \item $\pi_{0}$ \cite{DBLP:journals/corr/abs-2410-24164}
    \item $\pi_{0.5}$ \cite{DBLP:journals/corr/abs-2504-16054}
    \item $\pi_{0.6}^*$ \cite{DBLP:journals/corr/abs-2511-14759}
    \item $\pi_{\text{RL}}$ \cite{DBLP:journals/corr/abs-2510-25889}
    \item PIVOT \cite{DBLP:conf/icml/NasirianyX0XL0X24}
    \item PixelVLA \cite{DBLP:journals/corr/abs-2511-01571}
    \item PLA \cite{DBLP:journals/corr/abs-2507-23540}
    \item Point-VLA \cite{DBLP:journals/corr/abs-2512-18933}
    \item PosA-VLA \cite{DBLP:journals/corr/abs-2512-03724}
    \item Prophet \cite{DBLP:journals/corr/abs-2511-20633}
    \item QAIL+QBC \cite{DBLP:journals/corr/abs-2412-01034}
    \item QDepth-VLA \cite{DBLP:journals/corr/abs-2510-14836}
    \item Q-Transformer \cite{DBLP:conf/corl/ChebotarVHXLIKY23}
    \item QUART-Online \cite{DBLP:conf/icra/TongDFWZCSZZDHL25}
    \item QUAR-VLA \cite{DBLP:conf/eccv/DingZZSZHYW24}
    \item R2R2R \cite{DBLP:journals/corr/abs-2505-09601}
    \item RDT-1B \cite{DBLP:conf/iclr/LiuWLTCWX0025}
    \item Reasoning-VLA \cite{DBLP:journals/corr/abs-2511-19912}
    \item ReflectDrive \cite{DBLP:journals/corr/abs-2509-20109}
    \item REGENT \cite{DBLP:conf/iclr/SridharDJ025}
    \item ReKep \cite{DBLP:conf/corl/HuangWLZF24}
    \item RETAIN \cite{DBLP:journals/corr/abs-2512-08333}
    \item RetoVLA \cite{DBLP:journals/corr/abs-2509-21243}
    \item ReVLA \cite{DBLP:conf/icra/DeyZNGP25}
    \item RICL \cite{DBLP:journals/corr/abs-2508-02062}
    \item RoboCat \cite{DBLP:journals/tmlr/BousmalisVRDLVD24}
    \item RoboChemist \cite{DBLP:journals/corr/abs-2509-08820}
    \item RoboDual \cite{DBLP:journals/corr/abs-2410-08001}
    \item RoboFlamingo \cite{DBLP:conf/iclr/LiLZYXWCJ0LLK24}
    \item RoboMamba \cite{DBLP:conf/nips/LiuLWALZYZGZ24}
    \item RoboMatrix \cite{DBLP:journals/corr/abs-2412-00171}
    \item RoboMonkey \cite{DBLP:journals/corr/abs-2506-17811}
    \item RoboNeuron \cite{DBLP:journals/corr/abs-2512-10394}
    \item RoboNurse-VLA \cite{DBLP:conf/iros/LiWDMNHL25}
    \item RoboOmni \cite{DBLP:journals/corr/abs-2510-23763}
    \item RoboOS-NeXT \cite{DBLP:journals/corr/abs-2510-26536}
    \item RoboPoint \cite{DBLP:conf/corl/YuanDBPKMMF24}
    \item RoboTAP \cite{DBLP:conf/icra/VecerikD0DAZHAS24}
    \item Robotic Assistant \cite{DBLP:journals/corr/abs-2510-25713}
    \item RoboUniView \cite{DBLP:journals/corr/abs-2406-18977}
    \item RoboVLMs \cite{DBLP:journals/corr/abs-2412-14058}
    \item RobustVLA \cite{DBLP:journals/corr/abs-2510-00037}
    \item RoVer \cite{DBLP:journals/corr/abs-2510-10975}
    \item RPD \cite{DBLP:conf/iros/JulgBW25}
    \item RS-CL \cite{DBLP:journals/corr/abs-2510-01711}
    \item RT-1 \cite{DBLP:conf/rss/BrohanBCCDFGHHH23}
    \item RT-2 \cite{DBLP:conf/corl/ZitkovichYXXXXW23}
    \item RT-A \cite{DBLP:conf/icra/NasirianyKDSZDSX25}
    \item RT-H \cite{DBLP:conf/rss/BelkhaleDXSVTCD24}
    \item RT-RAS \cite{DBLP:journals/natmi/SchmidgallKKGK24}
    \item RT-Trajectory \cite{DBLP:conf/iclr/GuKW0AR0FGXSX0H24}
    \item RT-X \cite{DBLP:conf/icra/ONeillRMGPLPGMJ24}
    \item RVT \cite{DBLP:conf/corl/GoyalXGBCF23}
    \item RVT-2 \cite{DBLP:conf/rss/0001B0GCF24}
    \item RynnVLA-001 \cite{DBLP:journals/corr/abs-2509-15212}
    \item RynnVLA-002 \cite{DBLP:journals/corr/abs-2511-17502}
    \item SafeVLA \cite{vla_safevla_2025}
    \item SARA-RT \cite{DBLP:conf/icra/LealCJDVR0LSVSO24}
    \item ScaleDP \cite{DBLP:conf/icra/ZhuZLWXLCSPFT25}
    \item SC-VLA \cite{vla_scvla_2024}
    \item SEAL \cite{DBLP:journals/corr/abs-2510-16281}
    \item SeeDo \cite{DBLP:conf/iros/WangZDFF25}
    \item SeqVLA \cite{DBLP:journals/corr/abs-2509-14138}
    \item ShowUI \cite{DBLP:conf/cvpr/LinLGYWBLWS25}
    \item Sigma \cite{DBLP:journals/corr/abs-2512-00783}
    \item SimpleVLA-RL \cite{DBLP:journals/corr/abs-2509-09674}
    \item SITCOM \cite{DBLP:journals/corr/abs-2510-04041}
    \item SmolVLA \cite{DBLP:journals/corr/abs-2506-01844}
    \item SOLAMI \cite{DBLP:conf/cvpr/JiangXLZRGLCY025}
    \item Spatial Forcing \cite{DBLP:journals/corr/abs-2510-12276}
    \item SpatialVLA \cite{DBLP:journals/corr/abs-2501-15830}
    \item SpecPrune-VLA \cite{DBLP:journals/corr/abs-2509-05614}
    \item Spec-VLA \cite{DBLP:journals/corr/abs-2507-22424}
    \item SQAP-VLA \cite{DBLP:journals/corr/abs-2509-09090}
    \item SSM-VLA \cite{DBLP:journals/corr/abs-2509-26251}
    \item STARE-VLA \cite{DBLP:journals/corr/abs-2512-05107}
    \item StereoVLA \cite{DBLP:journals/corr/abs-2512-21970}
    \item STORM \cite{DBLP:journals/corr/abs-2512-18477}
    \item SUDD \cite{DBLP:conf/corl/HaFS23}
    \item SurgWorld \cite{DBLP:journals/corr/abs-2512-23162}
    \item SwiftVLA \cite{DBLP:journals/corr/abs-2512-00903}
    \item TabVLA \cite{DBLP:journals/corr/abs-2510-10932}
    \item TACO \cite{DBLP:journals/corr/abs-2502-05171}
    \item TacRefineNet \cite{DBLP:journals/corr/abs-2509-25746}
    \item Tactile-VLA \cite{DBLP:journals/corr/abs-2507-09160}
    \item TA-VLA \cite{DBLP:journals/corr/abs-2509-07962}
    \item ThinkAct \cite{DBLP:journals/corr/abs-2507-16815}
    \item ThinkBot \cite{DBLP:conf/iclr/LuWLLT25}
    \item TinyVLA \cite{DBLP:journals/corr/abs-2409-12514}
    \item TraceVLA \cite{DBLP:conf/iclr/ZhengLH0DKHY25}
    \item TrackVLA \cite{DBLP:journals/corr/abs-2505-23189}
    \item TrajBooster \cite{DBLP:journals/corr/abs-2509-11839}
    \item Transporter Networks \cite{DBLP:conf/corl/ZengFTWCAAKDSL20}
    \item TriVLA \cite{DBLP:journals/corr/abs-2507-01424}
    \item TVVE \cite{vla_tvve_2025}
    \item UD-VLA \cite{DBLP:journals/corr/abs-2511-01718}
    \item UnderwaterVLA \cite{DBLP:journals/corr/abs-2509-22441}
    \item UniCoD \cite{DBLP:journals/corr/abs-2510-10642}
    \item Unified-IO 2 \cite{DBLP:conf/cvpr/LuCL0KMHK24}
    \item Uni-NaVid \cite{DBLP:journals/corr/abs-2412-06224}
    \item UniPi \cite{DBLP:conf/nips/DuY0DN0SA23}
    \item UniUGP \cite{DBLP:journals/corr/abs-2512-09864}
    \item UniVLA \cite{DBLP:journals/corr/abs-2506-19850}
    \item UPA-RFAS \cite{DBLP:journals/corr/abs-2511-21192}
    \item UP-VLA \cite{DBLP:conf/icml/ZhangGHCZ025}
    \item UrbanVLA \cite{DBLP:journals/corr/abs-2510-23576}
    \item USIM \& U0 \cite{DBLP:journals/corr/abs-2510-07869}
    \item UWM \cite{DBLP:journals/corr/abs-2504-02792}
    \item VAM \cite{DBLP:journals/corr/abs-2512-15692}
    \item VER \cite{DBLP:conf/cvpr/LiuWY24}
    \item V-GPS \cite{DBLP:conf/corl/NakamotoMKL24}
    \item Video2Act \cite{DBLP:journals/corr/abs-2512-03044}
    \item VideoVLA \cite{DBLP:journals/corr/abs-2512-06963}
    \item VIMA \cite{DBLP:journals/corr/abs-2210-03094}
    \item VINE \cite{DBLP:journals/corr/abs-2512-03913}
    \item ViVLA \cite{DBLP:journals/corr/abs-2512-07582}
    \item VLA-0 \cite{DBLP:journals/corr/abs-2510-13054}
    \item VLA-4D \cite{DBLP:journals/corr/abs-2511-17199}
    \item VLA-Adapter \cite{DBLP:journals/corr/abs-2509-09372}
    \item VLA-AN \cite{DBLP:journals/corr/abs-2512-15258}
    \item VLA-Cache \cite{DBLP:journals/corr/abs-2502-02175}
    \item VLA-Fool \cite{DBLP:journals/corr/abs-2511-16203}
    \item VLA-OS \cite{DBLP:journals/corr/abs-2506-17561}
    \item VLA-Pilot \cite{DBLP:journals/corr/abs-2511-14178}
    \item VLA-Pruner \cite{DBLP:journals/corr/abs-2511-16449}
    \item VLAPS \cite{DBLP:journals/corr/abs-2508-12211}
    \item VLA-R \cite{DBLP:journals/corr/abs-2511-12405}
    \item VLA-R1 \cite{DBLP:journals/corr/abs-2510-01623}
    \item VLA-RAIL \cite{DBLP:journals/corr/abs-2512-24673}
    \item VLA-RFT \cite{DBLP:journals/corr/abs-2510-00406}
    \item VLAS \cite{DBLP:conf/iclr/ZhaoD0GB0W25}
    \item Vlaser \cite{DBLP:journals/corr/abs-2510-11027}
    \item VLASH \cite{DBLP:journals/corr/abs-2512-01031}
    \item VLM2VLA \cite{DBLP:journals/corr/abs-2509-22195}
    \item VLMimic \cite{DBLP:conf/nips/Chen0C0LZPH0YYY24}
    \item VoxPoser \cite{DBLP:conf/corl/HuangWZL0023}
    \item VST \cite{DBLP:conf/wacv/XuWZLO25}
    \item WAM-Flow \cite{DBLP:journals/corr/abs-2512-06112}
    \item WholeBodyVLA \cite{DBLP:journals/corr/abs-2512-11047}
    \item WMPO \cite{DBLP:journals/corr/abs-2511-09515}
    \item WorldVLA \cite{DBLP:journals/corr/abs-2506-21539}
    \item WristWorld \cite{DBLP:journals/corr/abs-2510-07313}
    \item X-Humanoid \cite{DBLP:journals/corr/abs-2512-04537}
    \item XR-1 \cite{DBLP:journals/corr/abs-2511-02776}
    \item X-VLA \cite{DBLP:journals/corr/abs-2510-10274}
\end{itemize}

\vspace{1em}
\subsubsection{Beyond VLA}

Recent research has expanded beyond standard VLA architectures in several key directions:
\begin{itemize}
    \item \textbf{VLA + World Model:} Architectures that unify VLAs with world models.
    \begin{itemize}
        \item WorldVLA \cite{DBLP:journals/corr/abs-2506-21539}
        \item UniVLA \cite{DBLP:journals/corr/abs-2506-19850}
        \item NORA-1.5 \cite{DBLP:journals/corr/abs-2511-14659}
        \item RynnVLA-002 \cite{DBLP:journals/corr/abs-2511-17502}
        \item Motus \cite{DBLP:journals/corr/abs-2512-13030}
    \end{itemize}

    \item \textbf{PLA, MLA:} Models incorporating perception modalities beyond vision.
    \begin{itemize}
        \item Perception-Language-Action framework (PLA) \cite{DBLP:journals/corr/abs-2507-23540}
        \item Multisensory Language–Action model (MLA) \cite{DBLP:journals/corr/abs-2509-26642}
    \end{itemize}

    \item \textbf{VAM:} Video-Action Models that jointly capture semantics and dynamics during video pretraining, enabling more efficient post-training for low-level robot control.
    \begin{itemize}
        \item mimic-video (VAM) \cite{DBLP:journals/corr/abs-2512-15692}
    \end{itemize}
\end{itemize}

\end{document}